\documentclass[10pt,journal,cspaper,compsoc]{IEEEtran}
\usepackage{amsmath}
\usepackage{rotating}
\usepackage{amssymb}
\usepackage{graphicx}
\usepackage{hyperref}
\usepackage{color}

\newcommand{\boldhead}[1]{\vspace{0.05in}\noindent\textbf{#1.}}

\newcommand{\ea}[0]{et al.}
\newcommand{\ve}[1]{{\bf #1}}
\newcommand{\afterfigure}{\vspace{-1.2em}}
\newcommand{\comment}[1]{}

\newcommand{\newtext}[1]{{\color{black}#1}} %Change this when editing!! (back to \color{blue})

\newcommand{\newtextRev}[1]{{\color{black}#1}}

\renewcommand{\baselinestretch}{1}

\begin{document}
\title{DepthTransfer: Depth Extraction from Video Using Non-parametric Sampling}

\author{Kevin~Karsch,~\IEEEmembership{Student Member,~IEEE,}
        Ce~Liu,~\IEEEmembership{Member,~IEEE,}
        and~Sing~Bing~Kang,~\IEEEmembership{Fellow,~IEEE}% <-this % stops a space
\IEEEcompsocitemizethanks{
\IEEEcompsocthanksitem Kevin Karsch is a graduate student in the Department of Computer Science, University of Illinois, Urbana, IL 61801. \protect\\ E-mail: karsch1@illinois.edu
\IEEEcompsocthanksitem Ce Liu is with Microsoft Research New England, Cambridge, MA 02142. \protect E-mail: celiu@microsoft.com
\IEEEcompsocthanksitem Sing Bing Kang is with Microsoft Research, Redmond, WA 98052. \protect E-mail: sbkang@microsoft.com}% <-this % stops a space
\thanks{}}


\markboth{Journal of \LaTeX\ Class Files,~Vol.~6, No.~1, January~2007}%
{Shell \MakeLowercase{\textit{et al.}}: Bare Demo of IEEEtran.cls for Computer Society Journals}

%\IEEEpubid{\makebox[\columnwidth]{\hfill 0000--0000/00/\$00.00~\copyright~2007 IEEE}%

% PRIOR to the title within the \IEEEcompsoctitleabstractindextext IEEEtran
\IEEEcompsoctitleabstractindextext{%
\begin{abstract}
We describe a technique that automatically generates plausible depth maps from videos using non-parametric depth sampling. We demonstrate our technique in cases where past methods fail (non-translating cameras and dynamic scenes). Our technique is applicable to single images as well as videos. For videos, we use local motion cues to improve the inferred depth maps, while optical flow is used to ensure temporal depth consistency. For training and evaluation, we use a Kinect-based system to collect a large dataset containing stereoscopic videos with known depths. We show that our depth estimation technique outperforms the state-of-the-art on benchmark databases. Our technique can be used to automatically convert a monoscopic video into stereo for 3D visualization, and we demonstrate this through a variety of visually pleasing results for indoor and outdoor scenes, including results from the feature film {\it Charade}.
\end{abstract}
% not want either math or citations to appear in the abstract.

% Note that keywords are not normally used for peer review papers.
}

\maketitle

% is not selected <OR> if conference mode is selected - because compsoc
% conference papers position the abstract like regular (non-compsoc)
\IEEEdisplaynotcompsoctitleabstractindextext

% \fi
\IEEEpeerreviewmaketitle

%%%%%%%%% BODY

%%%%%%%%%%%%%%%%%%%%%%%%%%%%%%%%%%%%%%
\section{Introduction}
\label{sec:intro}
% Computer Society journal papers do something a tad strange with the very
% first section heading (almost always called "Introduction"). They place it
% ABOVE the main text! IEEEtran.cls currently does not do this for you.
% However, You can achieve this effect by making LaTeX jump through some
% hoops via something like:
%
%\ifCLASSOPTIONcompsoc
%  \noindent\raisebox{2\baselineskip}[0pt][0pt]%
%  {\parbox{\columnwidth}{\section{Introduction}\label{sec:introduction}%
%  \global\everypar=\everypar}}%
%  \vspace{-1\baselineskip}\vspace{-\parskip}\par
%\else
%  \section{Introduction}\label{sec:introduction}\par
%\fi
%
% Admittedly, this is a hack and may well be fragile, but seems to do the
% trick for me. Note the need to keep any \label that may be used right
% after \section in the above as the hack puts \section within a raised box.


%%%%%%%% Teaser
\begin{figure*}[t]
%\vspace{-3.0in}
\begin{minipage}{\textwidth}
\begin{minipage}{0.485\linewidth}
\begin{center}
\includegraphics[width=.22\linewidth, trim=0pt 25pt 0pt 52pt, clip=true]{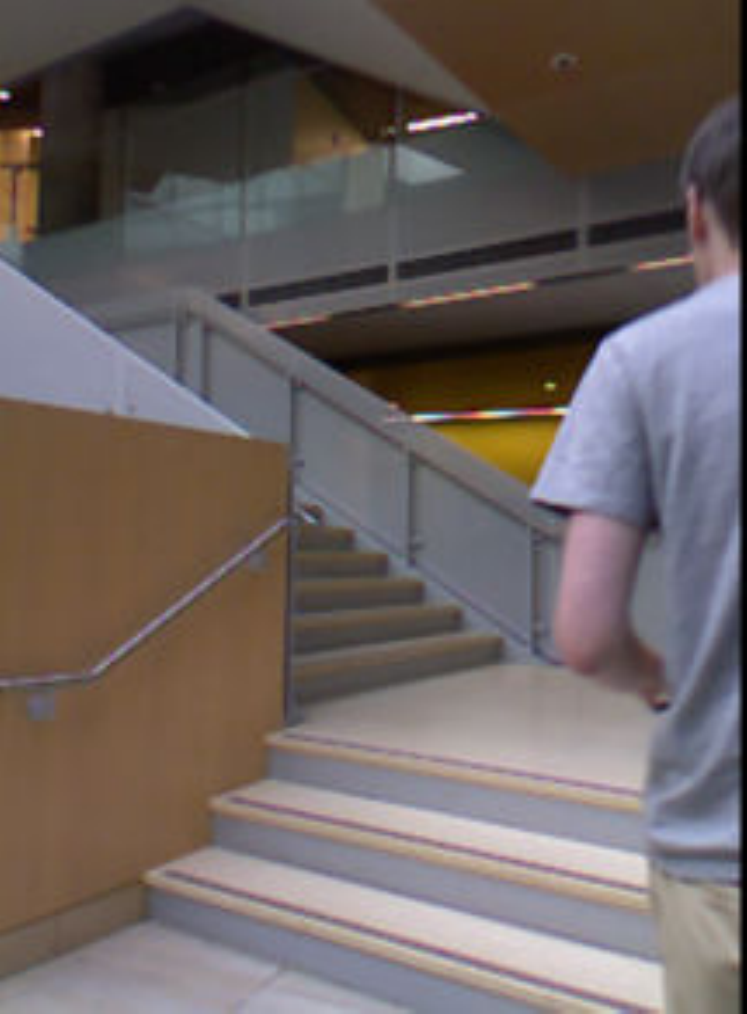}
\includegraphics[width=.22\linewidth, trim=0pt 25pt 0pt 52pt, clip=true]{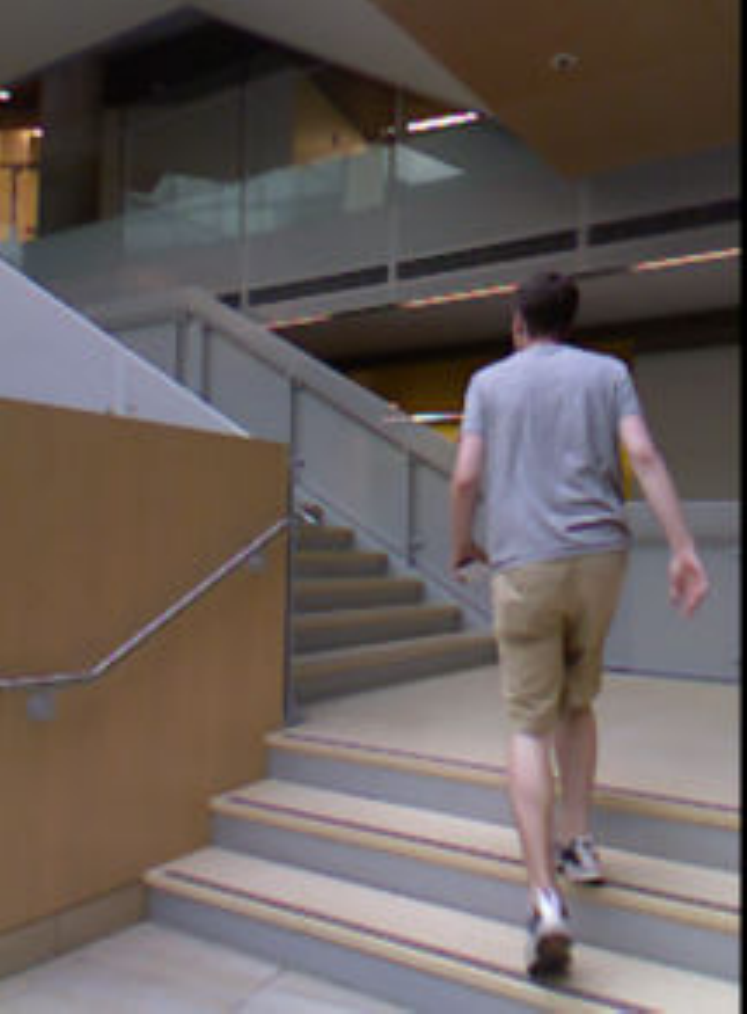}
\begin{sideways} \hspace{5mm} \tiny $\bullet$ \end{sideways}
\begin{sideways} \hspace{5mm} \tiny $\bullet$ \end{sideways}
%\begin{sideways} \hspace{9mm} \small $\bullet$ \end{sideways}
\includegraphics[width=.22\linewidth, trim=0pt 25pt 0pt 52pt, clip=true]{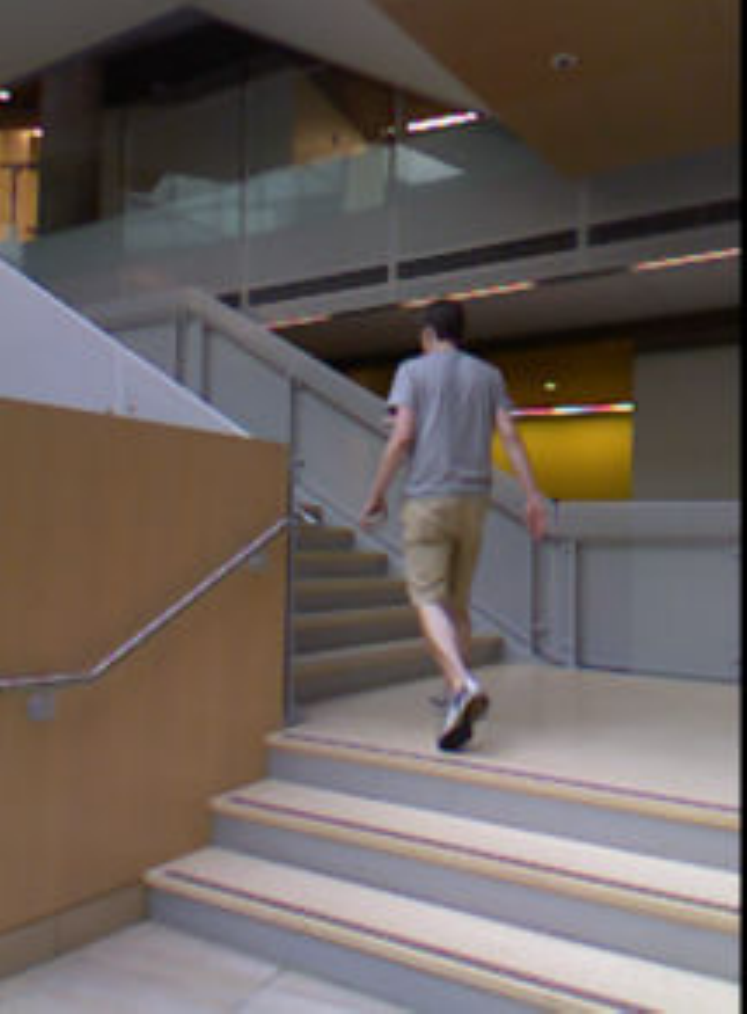}
\includegraphics[width=.22\linewidth, trim=0pt 25pt 0pt 52pt, clip=true]{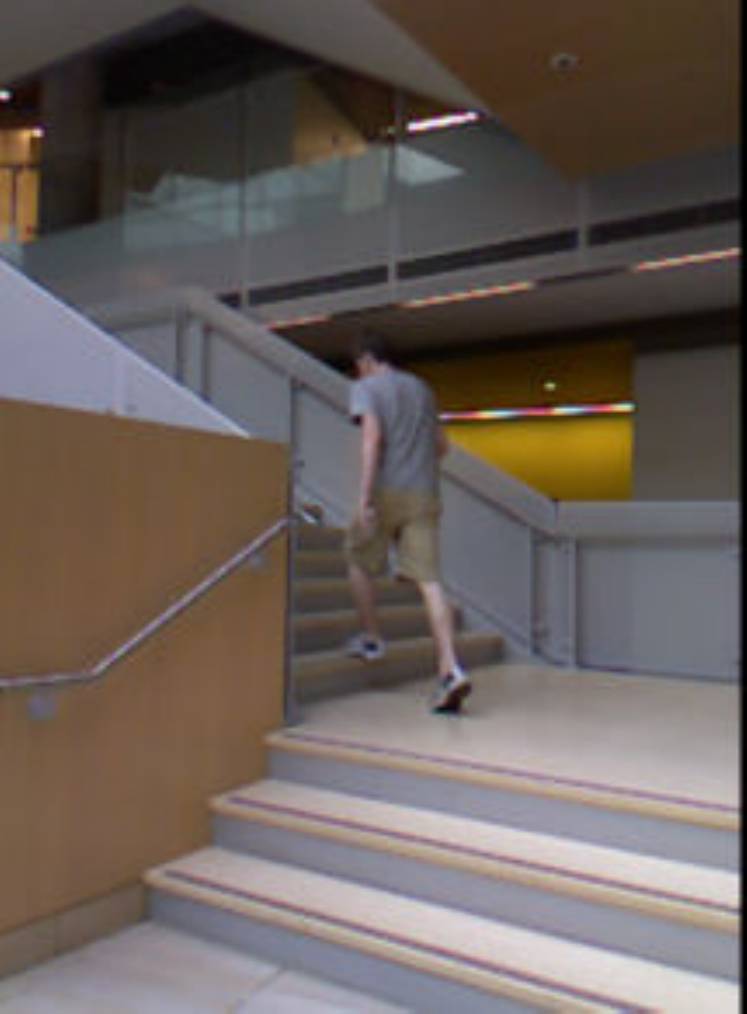}
\\
\includegraphics[width=.22\linewidth, trim=0pt 25pt 0pt 52pt, clip=true]{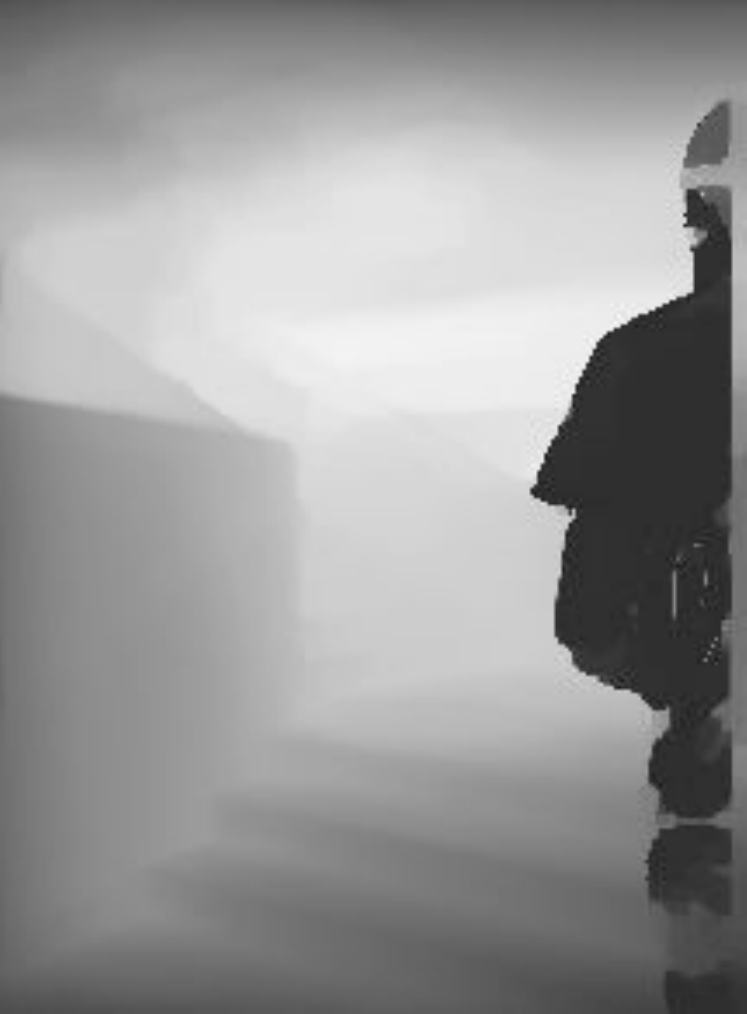}
\includegraphics[width=.22\linewidth, trim=0pt 25pt 0pt 52pt, clip=true]{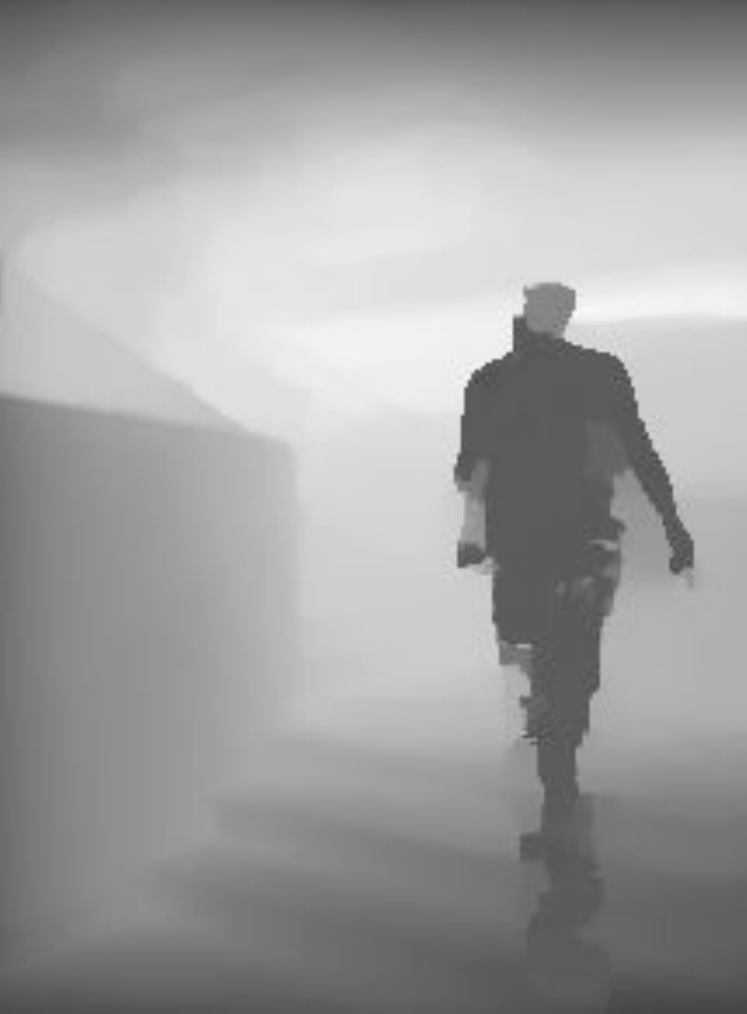}
\begin{sideways} \hspace{5mm} \tiny $\bullet$ \end{sideways}
\begin{sideways} \hspace{5mm} \tiny $\bullet$ \end{sideways}
%\begin{sideways} \hspace{9mm} \small $\bullet$ \end{sideways}
\includegraphics[width=.22\linewidth, trim=0pt 25pt 0pt 52pt, clip=true]{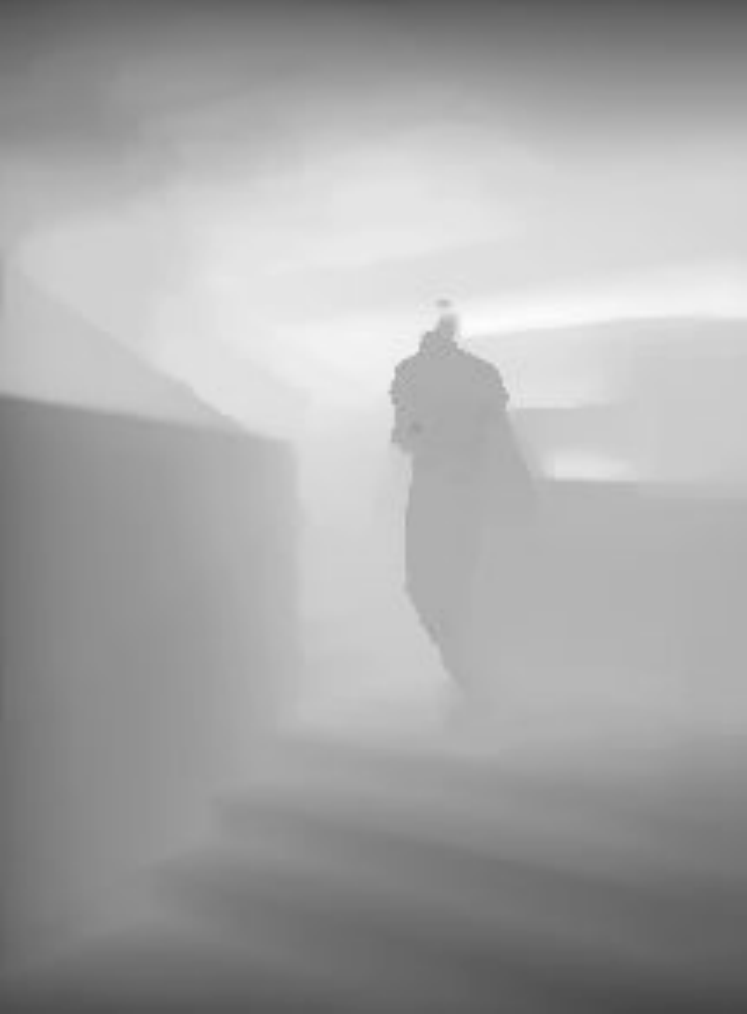}
\includegraphics[width=.22\linewidth, trim=0pt 25pt 0pt 52pt, clip=true]{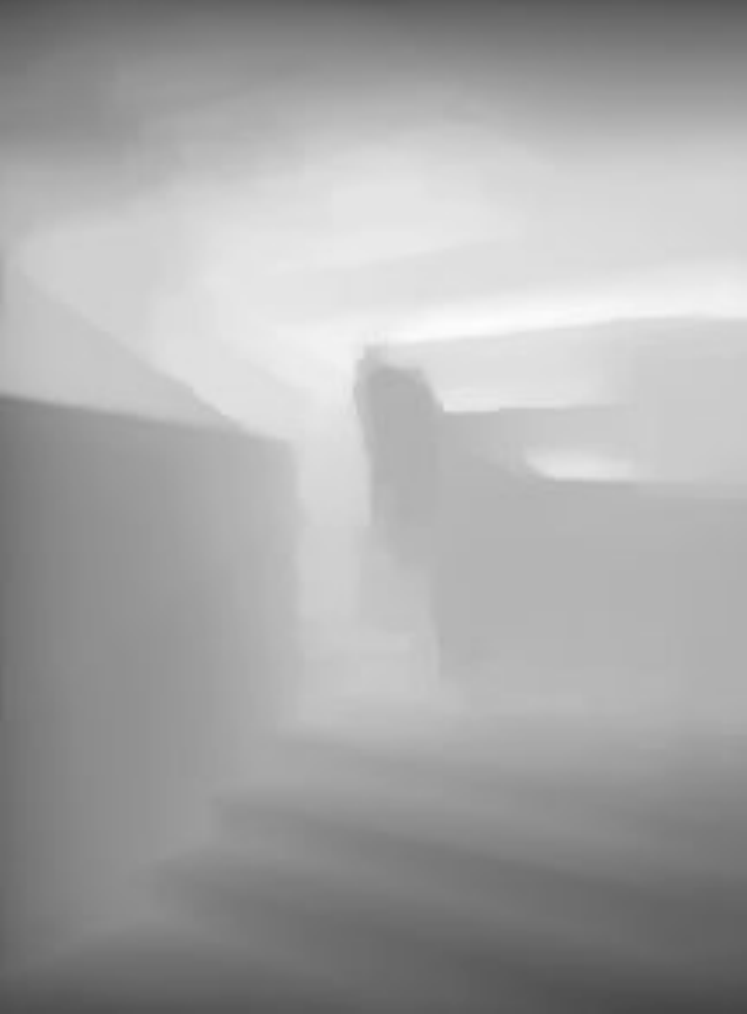}
\end{center}
\end{minipage}
\hfill
\begin{minipage}{0.485\linewidth}
\begin{center}
\includegraphics[width=.22\linewidth, trim=0pt 30pt 3pt 50pt, clip=true]{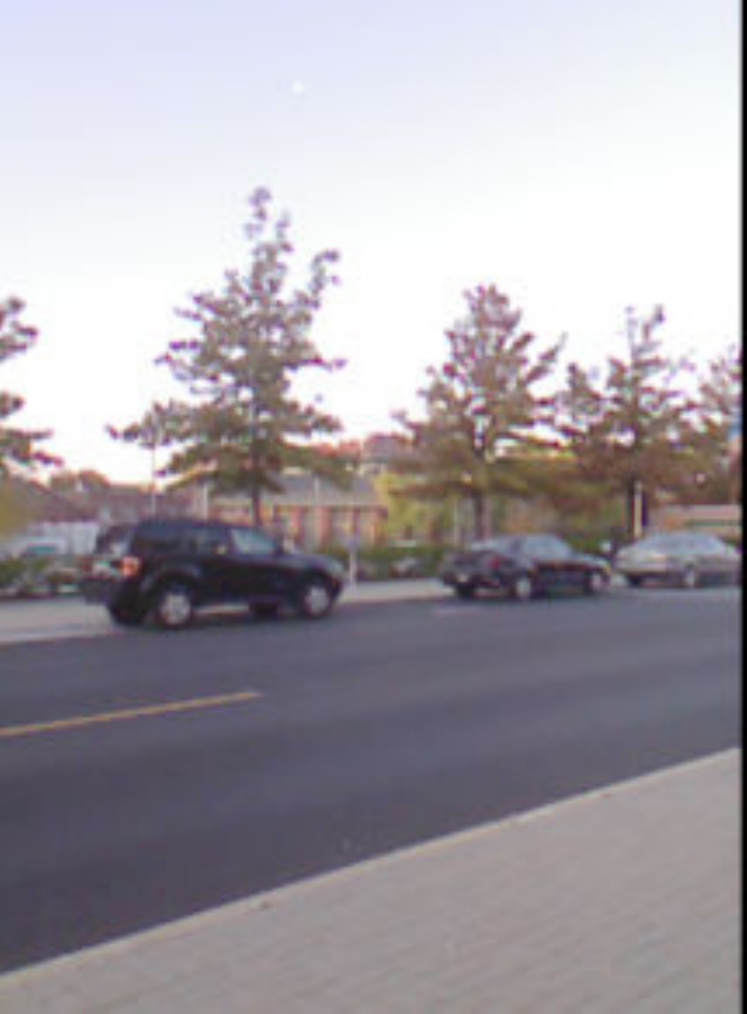}
\includegraphics[width=.22\linewidth, trim=0pt 30pt 3pt 50pt, clip=true]{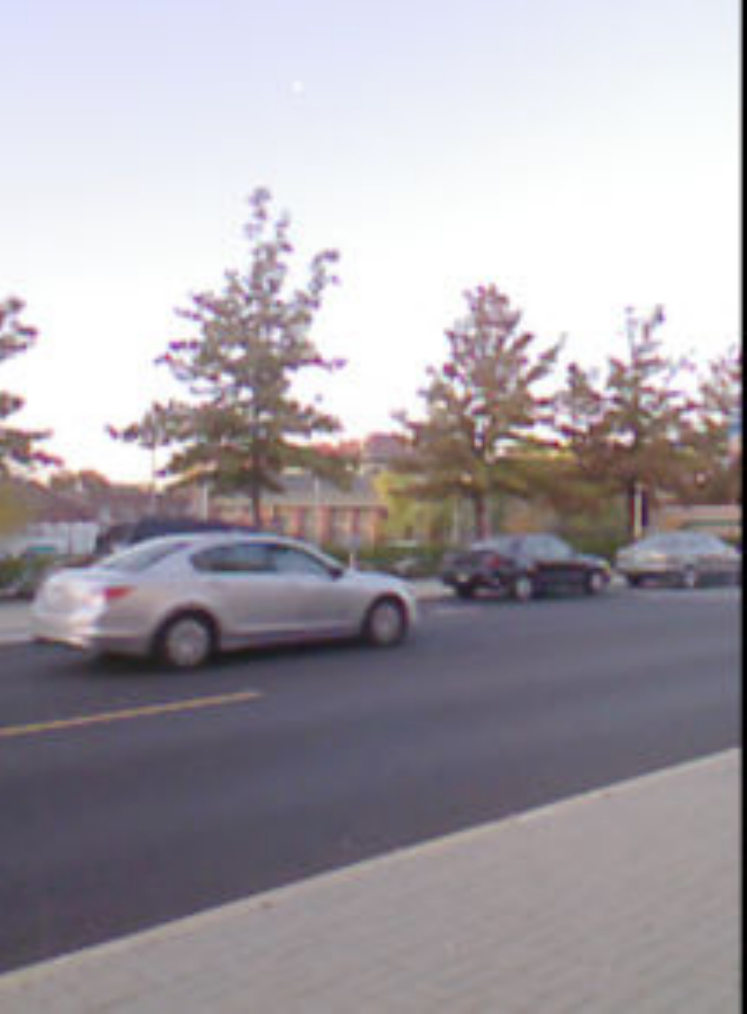}
\begin{sideways} \hspace{5mm} \tiny $\bullet$ \end{sideways}
\begin{sideways} \hspace{5mm} \tiny $\bullet$ \end{sideways}
%\begin{sideways} \hspace{9mm} \small $\bullet$ \end{sideways}
\includegraphics[width=.22\linewidth, trim=0pt 30pt 3pt 50pt, clip=true]{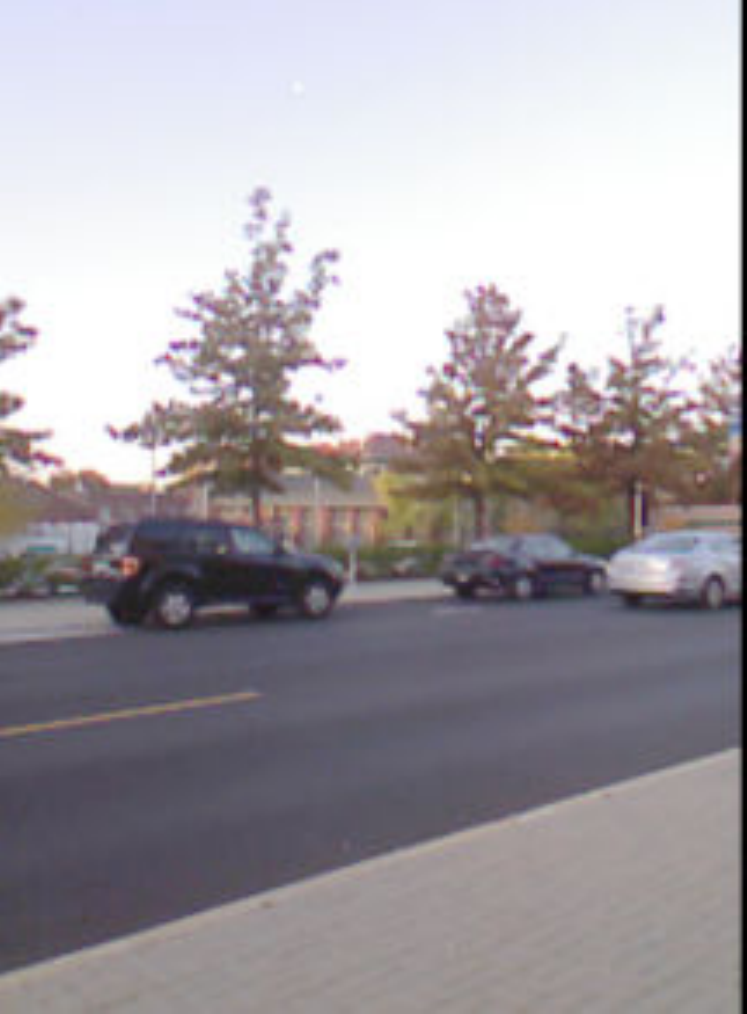}
\includegraphics[width=.22\linewidth, trim=0pt 30pt 3pt 50pt, clip=true]{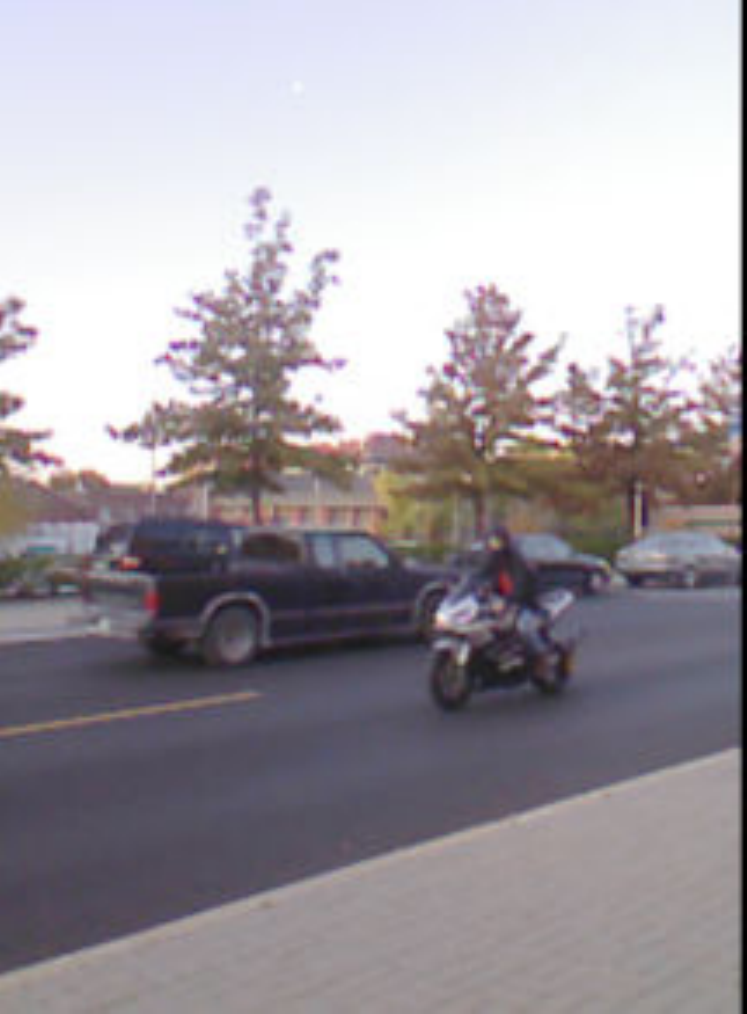}
\\
\includegraphics[width=.22\linewidth, trim=0pt 30pt 3pt 50pt, clip=true]{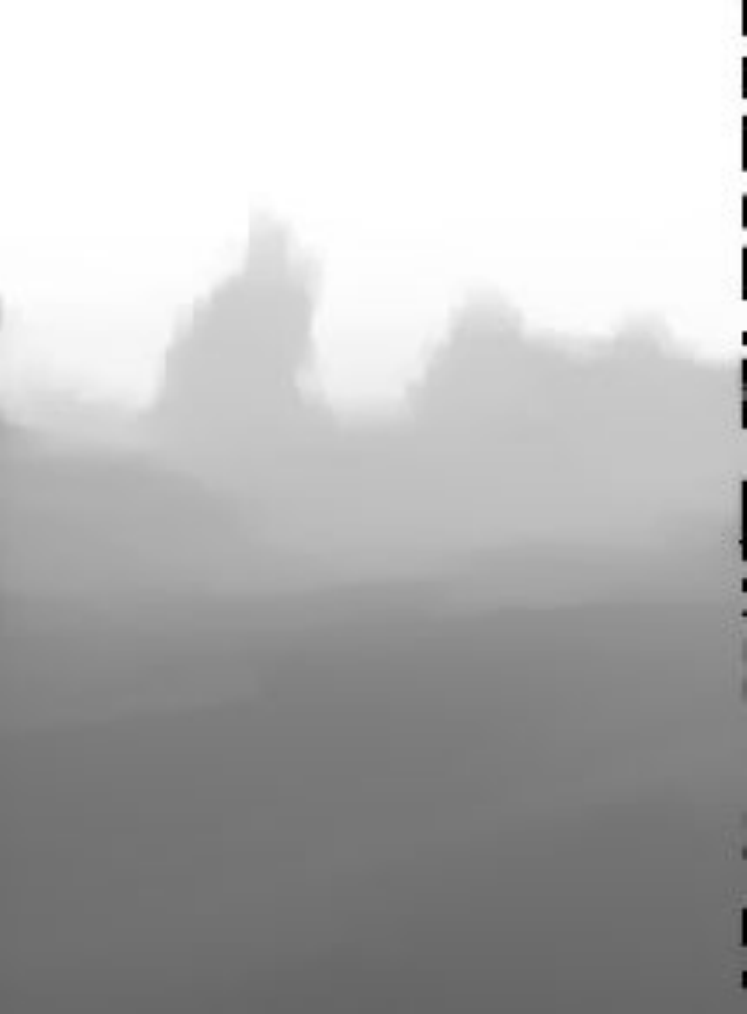}
\includegraphics[width=.22\linewidth, trim=0pt 30pt 3pt 50pt, clip=true]{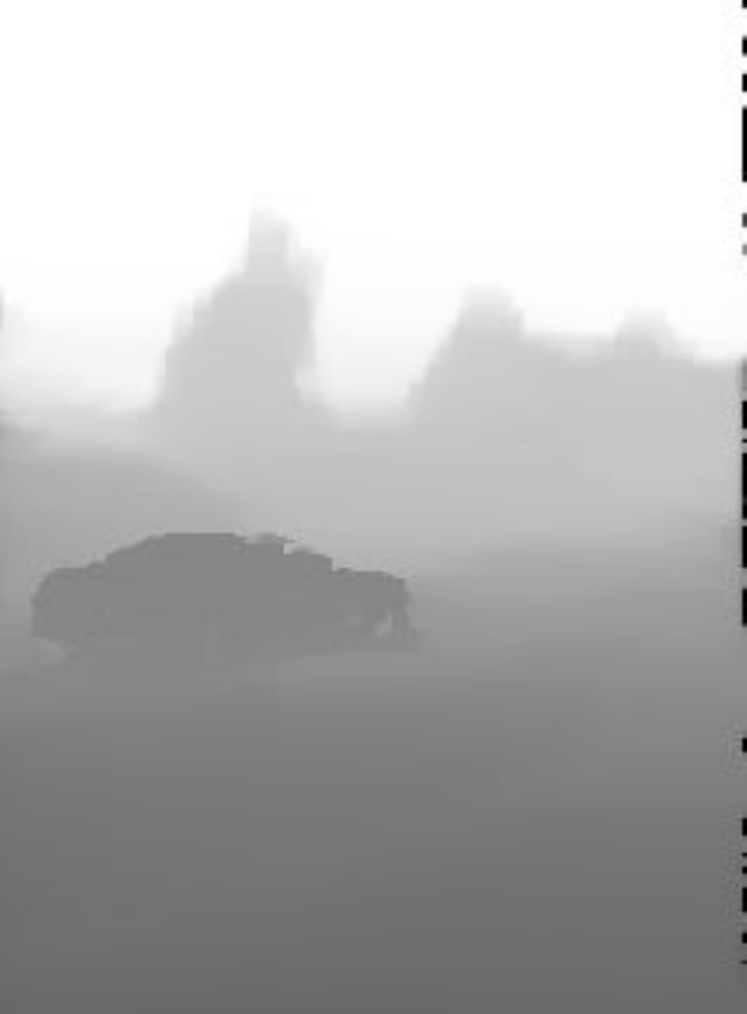}
\begin{sideways} \hspace{5mm} \tiny $\bullet$ \end{sideways}
\begin{sideways} \hspace{5mm} \tiny $\bullet$ \end{sideways}
%\begin{sideways} \hspace{9mm} \small $\bullet$ \end{sideways}
\includegraphics[width=.22\linewidth, trim=0pt 30pt 3pt 50pt, clip=true]{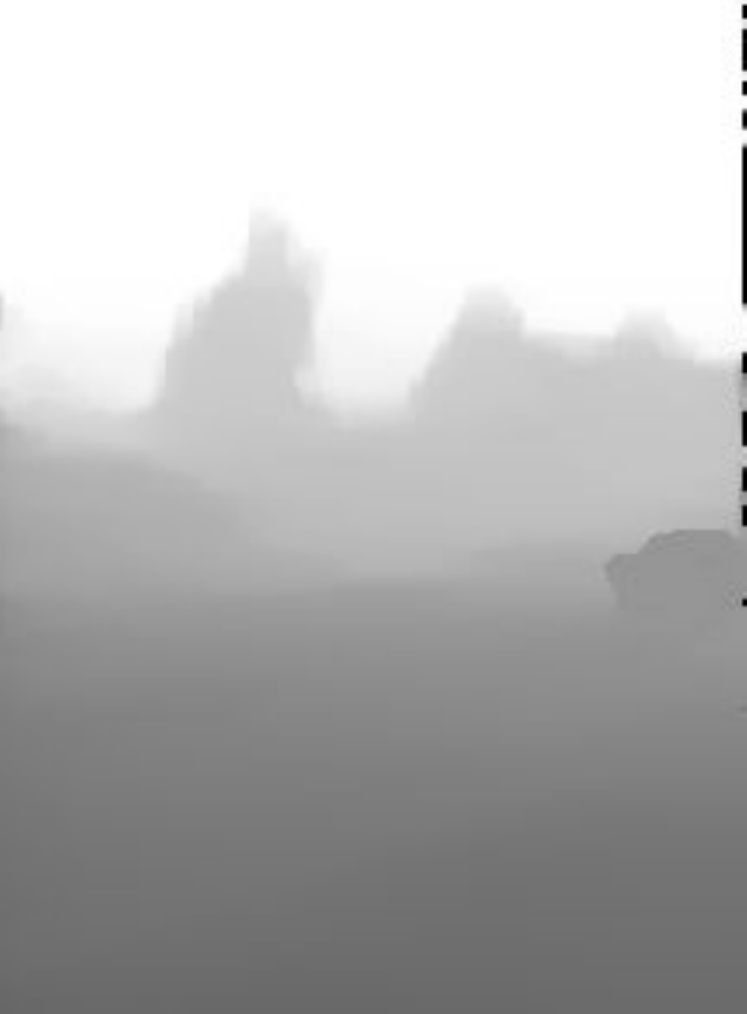}
\includegraphics[width=.22\linewidth, trim=0pt 30pt 3pt 50pt, clip=true]{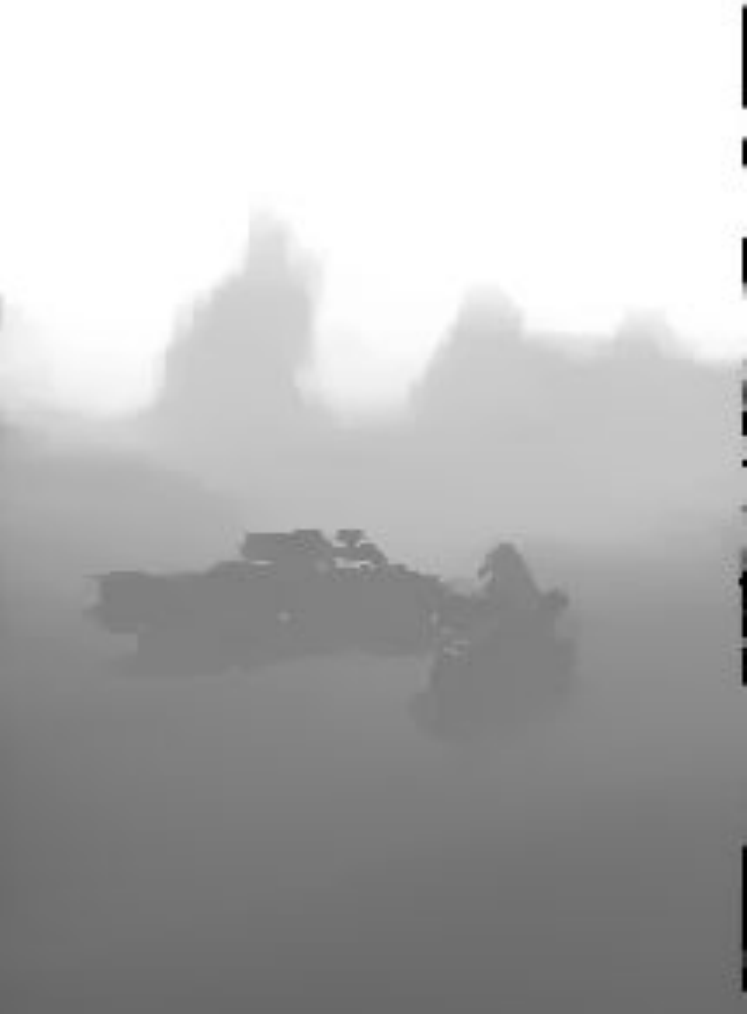}
\end{center}
\end{minipage}
\vspace{0mm}
\caption{Our technique takes a video sequence (\emph{top row}) and automatically estimates per-pixel depth (\emph{bottom row}). Our method does not require any cues from motion parallax or static scene elements; these videos were captured using a stationary camera with multiple moving objects.
%\vspace{-4mm}
}
\label{fig:teaser}
\end{minipage}
\end{figure*}
%%%%%%%%%%%%%%%%%

Scene depth is useful for a variety of tasks, ranging from 3D modeling and visualization to robot navigation. It also facilitates spatial reasoning about objects in the scene, in the context of scene understanding. In the growing 3D movie industry, knowing the scene depth greatly simplifies the process of converting 2D movies to their stereoscopic counterparts. The problem we are tackling in this paper is: given an arbitrary 2D video, how can we automatically extract plausible depth maps at every frame? At a deeper level, we investigate how we can reasonably extract depth maps in cases where conventional structure-from-motion and motion stereo fail.

While many reconstruction techniques for extracting depth from video sequences exist, they typically assume moving cameras and static scenes. They do not work for dynamic scenes or for stationary, rotating, or variable focal length sequences. There are some exceptions, e.g., \cite{Zhang:2011pami}, which can handle some moving objects, but they still require camera motion to induce parallax and allow depth estimation.

In this paper, we present a novel solution to generate depth maps from ordinary 2D videos; our solution also applies to single images. This technique is applicable to arbitrary videos, and works in cases where conventional depth recovery methods fail (static/rotating camera; change in focal length; dynamic scenes). Our primary contribution is the use of a non-parametric ``depth transfer'' approach for inferring temporally consistent depth maps without imposing requirements on the video (Sec~\ref{sec:depthest} and~\ref{sec:depthest:video}), including a method for improving the depth estimates of moving objects (Sec~\ref{sec:motion_est}). In addition, we introduce a new, ground truth stereo RGBD (RGB+depth) video dataset\footnote{Our dataset and code are publicly available at \url{http://kevinkarsch.com/depthtransfer}} (Sec~\ref{sec:dataset}). We also describe how we synthesize stereo videos from ordinary 2D videos using the results of our technique (Sec~\ref{sec:stereoest}). Exemplar results are shown in Fig~\ref{fig:teaser}.

\newtext{

\section{Related work}

In this section, we briefly survey techniques related to our work, namely 2D-to-3D conversion techniques (single image and video, automatic and manual) and non-parametric learning.

\subsection{Single image depth estimation and 2D-to-3D}

Early techniques for single image 2D-to-3D are semi-automatic; probably the most well-known of them is the ``tour-into-picture'' work of Horry~\ea~\cite{Horry:SIGGRAPH97}. Here, the user interactively adds planes to the single image for virtual view manipulation. Two other representative examples of interactive 2D-to-3D conversion systems are those of Oh~\ea~\cite{Oh:SIGGRAPH01} (where a painting metaphor is used to assign depths and extract layers) and Zhang~\ea~\cite{Zhang:JVCA02} (where the user adds surface normals, silhouettes, and creases as constraints for depth reconstruction).

One of the earliest automatic methods for single image 2D-to-3D was proposed by Hoiem~\ea~\cite{Hoiem:2005:APP:1186822.1073232}. They created convincing-looking reconstructions of outdoor images by assuming an image could be broken into a few planar surfaces; similarly, Delage~\ea~developed a Bayesian framework for reconstructing indoor scenes~\cite{Delage:06}. Saxena \ea~devised a supervised learning strategy for predicting depth from a single image~\cite{Saxena05learningdepth}, which was further improved to create realistic reconstructions for general scenes~\cite{Saxena:09}, and efficient learning strategies have since been proposed~\cite{Batra:CVPR:12}. Better depth estimates have been achieved by incorporating semantic labels~\cite{Liu:cvpr10}, or more sophisticated models~\cite{Li:2010}.

Apart from learning-based techniques for extracting depths, more conventional techniques have been used based on image content. For example, repetitive structures have been used for stereo reconstruction from a single image~\cite{Wu:2011wf}. \newtextRev{The dark channel prior has also proven effective for estimating depth from images containing haze~\cite{darkchannel}.} In addition, single-image shape from shading is also possible for known (a priori) object classes~\cite{Han:2003,Hassner:06}.

Compared to techniques here, we not only focus on depth from a {\it single image}, but also present a framework for using temporal information for enhanced and time-coherent depth when multiple frames are available.

%When repetitive structures exist, Wu~\ea~demonstrated how to acquire dense depth from a single image~\cite{Wu:2011wf}. For known classes of objects, it has been shown how to estimate depth~\cite{Hassner:06} and geometry~\cite{Han:2003}. In this paper, we are interested in not only single image depth estimation, but also inferring temporally coherent depth for videos of dynamic scenes.

The approach closest to ours is the contemporaneous work of Konrad et al.~\cite{Konrad:SPIE:12,Konrad:3DCINE:12}, which also uses some form of non-parametric depth sampling to automatically convert monocular images into stereoscopic images. They make similar assumptions to ours (e.g. appearance and depth are correlated), but use a simpler optimization scheme, and argue that the use SIFT flow only provides marginal improvements (opposed to not warping candidates via SIFT flow). Our improvements are two-fold: their depth maps are computed using the median of the candidate disparity fields and smoothed with a cross bilateral filter, while we consider the candidate depths (and depth gradients) on a per-pixel basis. Furthermore, we propose novel solutions to incorporate temporal information from videos, whereas the method of Konrad et al. works on single images. We also show a favorable comparison in Table~\ref{tab:nyu_results}.

%Their database is the YouTube3D dataset, used as the dictionary for learning the depth of a 2D image; since their dataset consists of stereo pairs, they train on extracting disparity rather than depth. They compute the median of the disparity fields of the best photometric matches and post-process using cross-bilateral filtering. While the search for the closest candidates is similar to ours, the subsequent operations differ significantly. For one, they compute depth per frame independently. In addition, their initial depth map is computed using the median of the candidate disparity fields, while we consider the candidate depths (and depth gradients) on a per-pixel basis and using temporal information for video. Finally, we do not post-process our depth maps as was done in \cite{Konrad:3DCINE:12}. \singbing{I'm a bit worried that reviewers will ask us to compare numbers with those in their paper.}

\subsection{Video depth estimation and 2D-to-3D}

A number of video 2D-to-3D techniques exist, but many of them are interactive. Examples of interactive system include those of Guttman~\ea~\cite{Guttmann:ICCV:09} (scribbles with depth properties are added to frames for video cube propagation), Ward~\ea~\cite{Ward:11} (``depth templates'' for primitive shapes are specified by the user, which the system propagates over the video), and Liao~\ea~\cite{Liao:TVCG:12} (user interaction to propagate structure-from-motion information with aid of optical flow).

There are a few commercial solutions that are automatic, e.g., Tri-Def DDD, but anecdotal tests using the demo version revealed room for improvement. There is even hardware available for real-time 2D-to-3D video conversion, such as the DA8223 chip by Dialog Semiconductor. It is not clear, however, how well the conversion works, since typically simple assumptions are made on what constitite foreground and background areas based on motion estimation. There are also a number of production houses specializing in 2D-to-3D conversions (e.g., In-Three~\cite{VanPernis:SPIE08} and Identity FX, Inc.), but their solutions are customized and likely to be manual-intensive. Furthermore, their tools are not publicly available.

If the video is mostly amenable to structure-from-motion and motion stereo, such technologies can be used to compute dense depth maps at every frame of the video. One system that does this is that of Zhang~\ea~\cite{Zhang:TVCG09}, which also showed how depth-dependent effects such as synthetic fog and depth-of-focus can be achieved. While this system (and others~\cite{Zhang:2011pami,Zhang:09}) can handle some moving objects, there is significant reliance on camera motion that induces parallax.

\subsection{Non-parametric learning}

%\singbing{Ce should probably add other representative techniques on non-parametric learning.}

As the conventional notion of image correspondences was extended from different views of the same 3D scene to semantically similar, but different, 3D scenes \cite{Liu:11a}, the information such as labels and motion can be transferred from a large database to parse an input image \cite{Liu:11b}. Given an unlabeled input image and a database with known per-pixel labels (e.g., sky, car, tree, window), their method works by transferring the labels from the database to the input image based on SIFT flow, which is estimated by matching pixel-wise, dense SIFT features. This simple methodology has been widely used in many computer vision applications such as image super resolution \cite{Tappen:ECCV12}, image tagging \cite{Rubinstein:ECCV12} and object discovery \cite{Rubinstein:CVPR13}. 

We build on this work by transferring  depth instead of semantic labels. Furthermore, we show that this ``transfer'' approach can be applied in a continuous optimization framework (Sec~\ref{sec:depthest}), whereas their method used a discrete MRFs.

}

%%%%%%%%%%%%%%%%%%%%%%%%%%%%%%%%%%%%%%
\section{Non-parametric depth estimation by continuous label transfer}
\label{sec:depthest}

We leverage recent work on non-parametric learning~\cite{Liu:09}, which avoids explicitly defining a parametric model and requires fewer assumptions as in past methods (e.g.,~\cite{Saxena05learningdepth,Saxena:09,Liu:cvpr10}). This approach also scales better with respect to the training data size, requiring virtually no training time. Our technique imposes no requirements on the video, such as motion parallax or sequence length, and can even be applied to a single image. We first describe our depth estimation technique as it applies to single images below, and in Sec~\ref{sec:depthest:video} we discuss novel additions that allow for improved depth estimation in videos.

%%%%%%%%%%%%%%%%%
\begin{figure*}[t]
\centerline{\includegraphics[width=\linewidth]{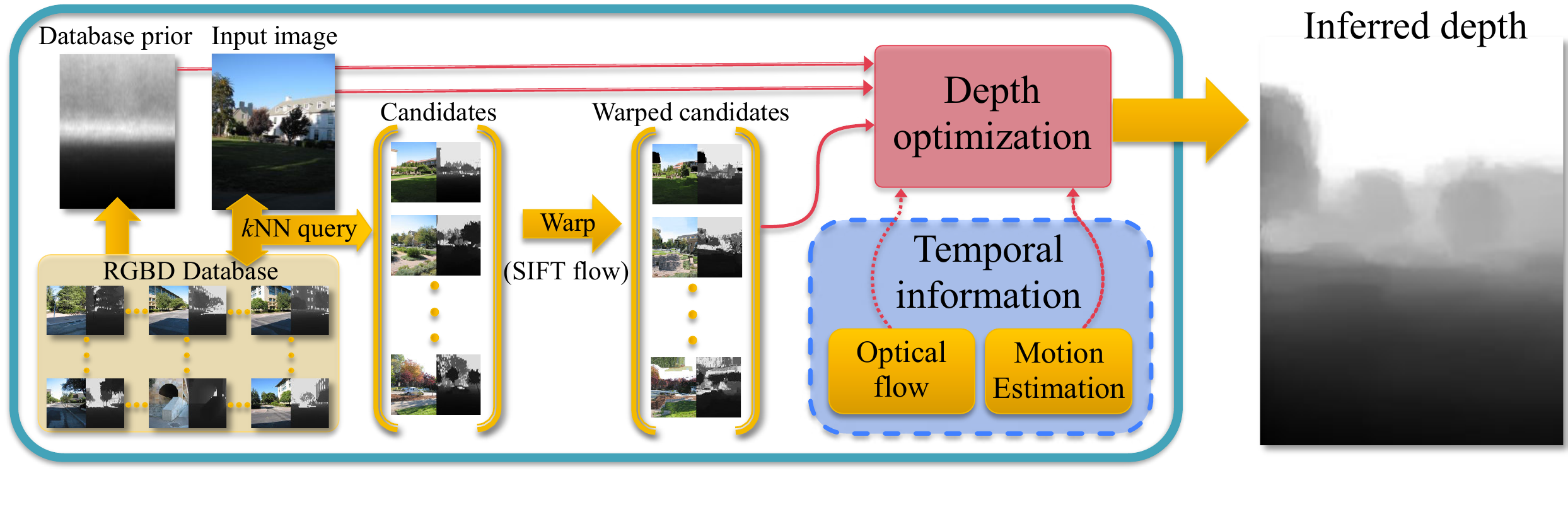}}
\vspace{-5mm}
\caption{Our pipeline for estimating depth. Given an input image, we find matching candidates in our database, and warp the candidates to match the structure of the input image. We then use a global optimization procedure to interpolate the warped candidates (Eq.~\ref{eq:depth_single}), producing per-pixel depth estimates for the input image. With temporal information (e.g., extracted from a video), our algorithm can achieve more accurate, temporally coherent depth.
}
\label{fig:depthoverview}\afterfigure
\end{figure*}
%%%%%%%%%%%%%%%%%

Our depth transfer approach, outlined in Fig~\ref{fig:depthoverview}, has three stages. First, given a database RGBD images, we find candidate images in the database that are ``similar'' to the input image in RGB space. Then, a warping procedure (SIFT Flow~\cite{Liu:11a}) is applied to the candidate images and depths to align them with the input. Finally, an optimization procedure is used to interpolate and smooth the warped candidate depth values; this results in the inferred depth.

Our core idea is that scenes with similar semantics should have roughly similar depth distributions when densely aligned. In other words, images of semantically alike scenes are expected to have similar depth values in regions with similar appearance. Of course, not all of these estimates will be correct, which is why we find several candidate images and refine and interpolate these estimates using a global optimization technique that considers factors other than just absolute depth values.

\subsection{RGBD database}

Our system requires a database of RGBD images and/or videos.  We have collected our own RGBD video dataset, as described in Sec~\ref{sec:dataset}; a few already exist online, though they are for single images only.\footnote{Examples: Make3D range image dataset (\url{http://make3d.cs.cornell.edu/data.html}),  B3DO dataset (\url{http://kinectdata.com/}),  NYU depth datasets (\url{http://cs.nyu.edu/~silberman/datasets/}), RGB-D dataset (\url{http://www.cs.washington.edu/rgbd-dataset/}), and our own (\url{http://kevinkarsch.com/depthtransfer}).}

\subsection{Candidate matching and warping}

Given a database and an input image, we compute high-level image features (we use GIST~\cite{Oliva:ijcv01} and optical flow features) for each image or frame of video in the database as well as the input image. We then select the top $K$ ($=7$ in our work, unless explicitly stated) matching frames from the database, but ensure that each video in the database contributes no more than one matching frame. This forces matching images to be from differing viewpoints, allowing for greater variety among matches. We call these matching images {\it candidate images}, and their corresponding depths {\it candidate depths}.

%%%%%%%%%%%%%%%%%
\begin{figure}[t]
\centerline{
\hfill
(a)\includegraphics[width=0.44\columnwidth,page=1]{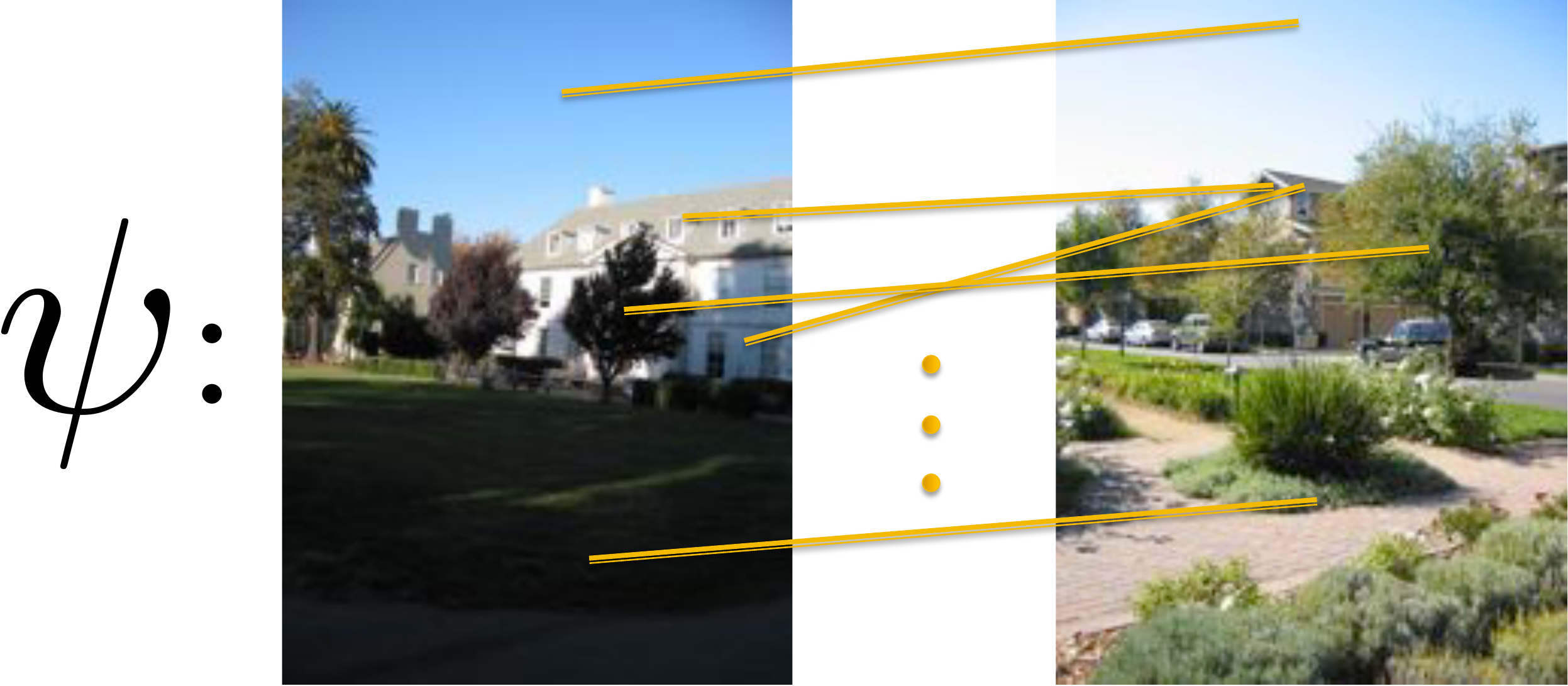}
\hfill
(b)\includegraphics[width=0.44\columnwidth,page=2]{fig/siftflow}
\hfill
}
\caption{SIFT flow warping. (a) SIFT features are calculated and matched in a one-to-many fashion, which defines $\psi$. (b) $\psi$ is applied to achieve dense scene alignment.
}
\label{fig:siftflow}\afterfigure
\end{figure}
%%%%%%%%%%%%%%%%%

Because the candidate images match the input closely in feature space, it is expected that the overall semantics of the scene are roughly similar. We also make the critical assumption that the distribution of depth is comparable among the input and candidates. However, we want pixel-to-pixel correspondences between the input and all candidates, as to limit the search space when inferring depth from the candidates.

We achieve this pixel-to-pixel correspondence through SIFT flow~\cite{Liu:11a}, which matches per-pixel SIFT features to estimate dense scene alignment. Using SIFT flow, we estimate warping functions $\psi_i, i \in \{1,\ldots,K\}$ for each candidate image; this process is illustrated in Fig~\ref{fig:siftflow}. These warping functions map pixel locations from a given candidate's domain to pixel locations in the input's domain. The warping functions can be one-to-many.

\newtext{

\subsection{Features for candidate image matching}

In order to find candidate images which match the input image/sequence semantically and in terms of depth distribution, we use a combination of GIST features~\cite{Oliva:ijcv01} and features derived from optical flow (used only in the case of videos, as in Sec~\ref{sec:depthest:video}). To create flow features for a video, we first compute optical flow (using Liu's implementation~\cite{Liu:thesis09}) for each pair of consecutive frames, which defines a warping from frame $i$ to frame $i+1$. If the input is a single image, or the last image in a video sequence, we consider this warping to be the identity warp. Then, we segment the image into $b \times b$ uniform blocks, and compute the mean and standard deviation over the flow field in each block, for both components of the flow (horizontal and vertical warpings), and for the second moments as well (each component squared). This leads to eight features per block, for a total of $8b^2$ features per image. We use $b=4$ for our results.

To determine the matching score between two images, we take a linear combination of the difference in GIST and optical flow features described above. Denoting $G_1, G_2$ and $F_1,F_2$ as the GIST and flow feature vectors for two images respectively, we define the matching score as
\begin{equation}
\label{eq:matchingscore}
 (1- \omega)||G_1-G_2|| + \omega ||F_1-F_2||,
\end{equation}
where $\omega =0.5 $ in our implementation.

}

\subsection{Depth optimization}

Each warped candidate depth is deemed to be a rough approximation of the input's depth map. Unfortunately, such candidate depths may still contain inaccuracies and are often not spatially smooth. Instead, we generate the most likely depth map by considering all of the warped candidates, optimizing with spatial regularization in mind.

Let $\ve{L}$ be the input image and $\ve{D}$ the depth map we wish to infer. We minimize
\begin{align}
\label{eq:depth_single}
-\log( P(\ve{D} | \ve{L}) ) &= E(\ve{D}) = \\  \sum_{i\in\text{pixels}} E_{\text{t}}(\ve{D}_i) &+ \alpha E_{\text{s}}(\ve{D}_i) + \beta E_{\text{p}}(\ve{D}_i) + \log(Z), \nonumber
\end{align}
where $Z$ is the normalization constant of the probability, and $\alpha$ and $\beta$ are parameters ($\alpha=10, \beta=0.5$). For a single image, our objective contains three terms: data ($E_{\text{t}}$), spatial smoothness ($E_{\text{s}}$), and database prior ($E_{\text{p}}$).

\boldhead{Data cost}
The data term measures how close the inferred depth map $\ve{D}$ is to each of the warped candidate depths, $\psi_j(C^{(j)})$. This distance measure is defined by $\phi$, a robust error norm (we use an approximation to the $L1$ norm, $\phi(x) = \sqrt{x^2+\epsilon}$, with $\epsilon = \text{10}^{-4}$). We define the data term as
\begin{equation}
\label{eq:transfer}
%\begin{split}
%& E_{\text{t}}(\ve{D}_i) = \sum_{j=1}^K w^{(j)}_i [\phi(\ve{D}_i-\psi_j(C^{(j)}_i)) + \\
%&\gamma (\phi(\nabla_x  \ve{D}_i-\psi_j(\nabla_x C^{(j)}_i)) + \phi(\nabla_y \ve{D}_i-\psi_j(\nabla_y C^{(j)}_i))) ],
%\end{split}
\begin{split}
E_{\text{t}}(\ve{D}_i) =  \sum_{j=1}^K w^{(j)}_i \Bigl[ & \phi(\ve{D}_i-\psi_j(C^{(j)}_i))\ +  \\
& \gamma \bigl[ \phi(\nabla_x  \ve{D}_i-\psi_j(\nabla_x C^{(j)}_i))\ + \\ &\ \ \ \phi(\nabla_y \ve{D}_i-\psi_j(\nabla_y C^{(j)}_i)) \bigr]  \Bigr],
\end{split}
\end{equation}
where $w_i^{(j)}$ is a confidence measure of the accuracy of the $j^{th}$ candidate's warped depth at pixel $i$, and $K$ ($=7$) is the total number of candidates. We measure not only absolute differences, but also relative depth changes, i.e., depth gradients ($\nabla_x$, $\nabla_y$ are spatial derivatives). The latter terms of Eq~\ref{eq:transfer} enforce similarity among candidate depth gradients and inferred depth gradients, weighted by $\gamma$ ($=10$).
%These terms are important, as they allow for more accurate intra-object depth inference.

\newtext{

Note that we warp the candidate's gradients in depth, rather than warping depth and then differentiating. Warping depth induces many new discontinuities, which would result in large gradients solely due to the warping, rather than actual depth discontinuities in the data. The downside of this is that the warped gradients are not integrable, but this does not signify as we optimize over depth anyhow (ensuring the resulting depth map is integrable).

%During optimization, we ensure that the inferred depth is similar to each of the $K$ warped candidate depths. However,
Some of the candidate depth values will be more reliable than others, and we model this reliability with a confidence weighting for each pixel in each candidate image (e.g. $w_i^{(j)}$ is the weight of the $i^{th}$ pixel from the $j^{th}$ candidate image). We compute these weights by comparing per-pixel SIFT descriptors, obtained during the SIFT flow computation, of both the input image and the candidate images:

\begin{equation}
\label{eq:smoothweights}
w_i^{(j)} = (1+e^{(||\ve{S}_i - \psi_j(\mathcal{S}_i^{(j)})|| - \mu_s)/\sigma_s})^{-1},
\end{equation}
where $\ve{S}_i$ and $\mathcal{S}_i^{(j)}$ are the SIFT feature vectors at pixel $i$ in candidate image $j$. We set $\mu_s = 0.5$ and $\sigma_s = 0.01$. Notice that the candidate image's SIFT features are computed first, and then warped using the warping function ($\psi_j$) calculated with SIFT flow.

}

\boldhead{Spatial smoothness}
\newtext{
While we encourage spatial smoothness, we do not want the smoothness applied uniformly to the inferred depth, since there is typically some correlation between image appearance and depth. Therefore, we assume that regions in the image with similar texture are likely to have similar, smooth depth transitions, and that discontinuities in the image are likely to correspond to discontinuities in depth.

We enforce appearance-dependent smoothness with a per-pixel weighting of the spatial regularization term such that this weight is large where the image gradients are small, and vice-versa. We determine this weighting by applying a sigmoidal function to the gradients, which we found to produce more pleasing inferred depth maps than using other boundary detecting schemes such as~\cite{Hoiem:2007,Maire:cvpr08}.

The smoothness term is specified as:\vspace{-.5em}
\begin{equation}
\label{eq:smooth}
E_{\text{s}}(\ve{D}_i) = s_{x,i} \phi(\nabla_x \ve{D}_i) + s_{y,i} \phi(\nabla_y \ve{D}_i).
\end{equation}
The depth gradients along x and y ($\nabla_x \ve{D}, \nabla_y \ve{D}$) are modulated by soft thresholds (sigmoidal functions) of image gradients in the same directions ($\nabla_x \ve{L}, \nabla_y \ve{L}$), namely, $s_{x,i} = (1+e^{(||\nabla_x \ve{L}_i|| - \mu_L)/\sigma_L})^{-1}$ and $s_{y,i} = (1+e^{(||\nabla_y \ve{L}_i|| - \mu_L)/\sigma_L})^{-1}$. We set $\mu_L = 0.05$ and $\sigma_L = 0.01$.

%Similar reasoning is applied for temporal smoothness. We can measure the confidence of the  optical flow by computing the reprojection error of the flow, which we use to weight the temporal smoothness terms in the coherence term of our video objective function.

}

\boldhead{Prior}
We also incorporate assumptions from our database that will guide the inference when pixels have little or no influence from other terms (due to weights $w$ and $s$):
\begin{equation}
\label{eq:prior}
E_{\text{p}}(\ve{D}_i)= \phi(\ve{D}_i - \mathcal{P}_i).
\end{equation}
We compute the prior $\mathcal{P}$ by averaging all depth images in our database.

\newtext{

\subsection{Numerical optimization details}

Equation~\ref{eq:depth_single} requires an unconstrained, non-linear optimization, and we use iteratively reweighted least squares to minimize our objective function. We choose IRLS because it is a fast alternative for solving unconstrained, nonlinear minimization problems such as ours. IRLS works by approximating the objective by a linear function of the parameters, and solving the system by minimizing the squared residual (e.g. with least squares); it is repeated until convergence.

As an example, consider a sub-portion of our objective, the second term in Eq~\ref{eq:transfer} for candidate \#1: $\sum_{i\in \text{pixels}} \phi(\nabla_x \ve{D}_i -\psi_1(\nabla_x C^{(1)}_i)$. To minimize, we differentiate with respect to depth and set equal to zero (letting $b = [\psi_1(\nabla_x C^{(1)}_1), \ \ldots, \ \psi_1(\nabla_x C^{(1)}_N)]^T$, keeping in mind that $\phi(x) = \sqrt{x^2+\epsilon}$, and $\nabla_x$ is the horizontal derivative):
\begin{equation}
\sum_{i\in \text{pixels}} \frac{d}{d\ve{D}} \phi(\nabla_x \ve{D}_i - b) \ \nabla_x (\nabla_x \ve{D}_i - b) = 0.
\end{equation}
We rewrite this using matrix notation as $G_x^T W (G_x \ve{D}-b) = 0$, where $G_x$ is the $N\times N$ linear operator corresponding to a horizontal gradient (e.g., $[G_x \ve{D}]_i = \nabla_x \ve{D}_i$), and $W$ is a diagonal matrix of ``weights'' computed from the non-linear portion of the derivative of $\phi$. By fixing $W$ (computed for a given $\ve{D}$), we arrive at the following IRLS solution:
\begin{align}
W &= diag\left(  \frac{d}{d\ve{D}} \phi(\nabla_x \ve{D}^{(t)} - b) \right) \nonumber \\
\ve{D}^{(t+1)} &= (G_x^T W G_x) ^ {+} (G_x^T W b),
\end{align}
where $()^+$ is the pseudoinverse and $D^{(t)}$ is the inferred depth at iteration $t$. We have found that thirty iterations typically approximates convergence. Extending IRLS to our full objective follows the same logic.\footnote{For further details and discussion of IRLS, see the appendix of Liu's thesis~\cite{Liu:thesis09}.}

%%%%%%%%%%%%%%%%%
\begin{figure}[t]
{\small \hspace{10mm} Initial \hspace{7mm} Iteration 5 \hspace{4mm} Iteration 10 \hspace{3mm} Iteration 30}  \vspace{1mm} \\
{\small \begin{sideways} \hspace{3mm} Random \hspace{4mm} Dataset avg \hspace{6mm} Mean \hspace{8mm} Median \end{sideways} }
\includegraphics[width=.95\linewidth]{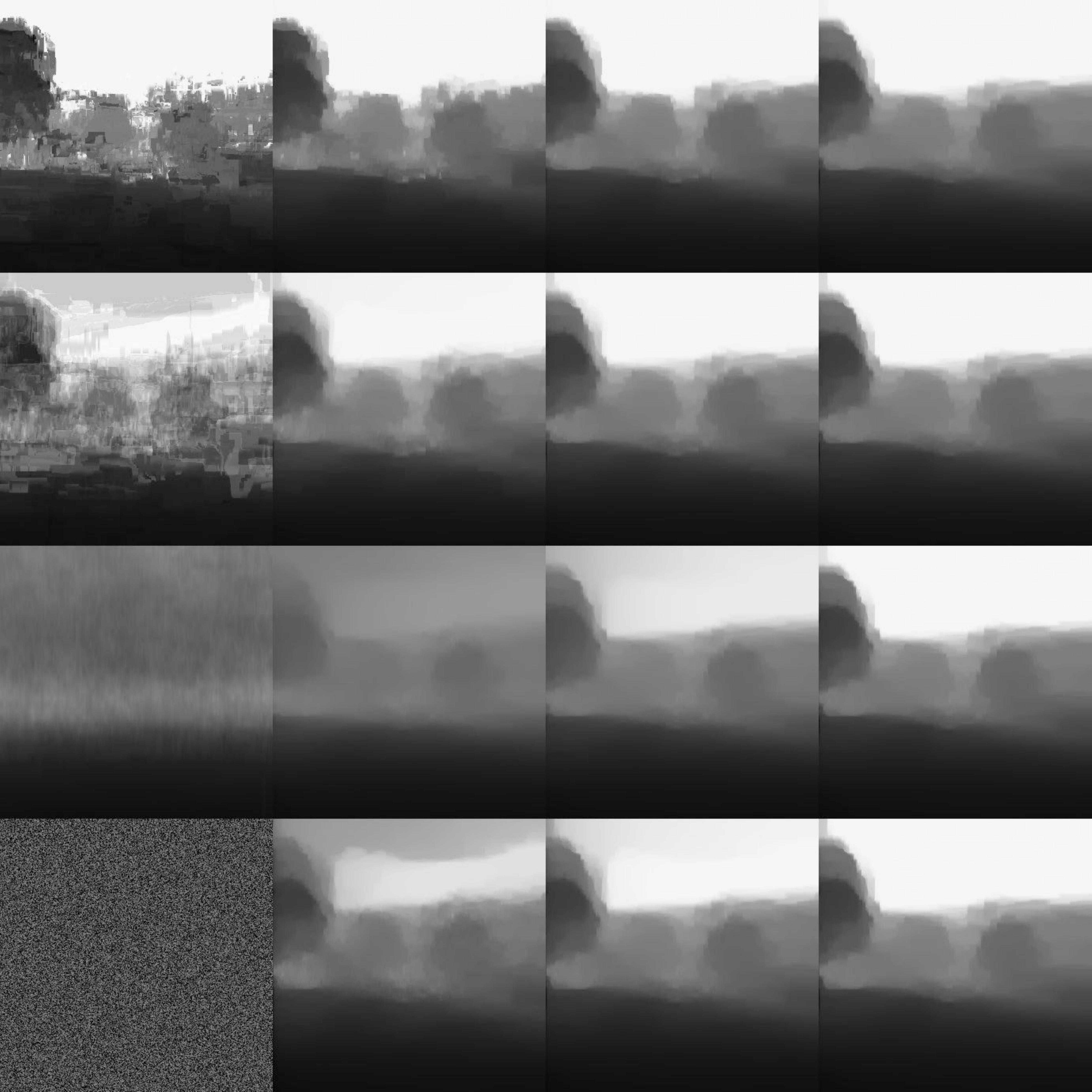}
\caption{Result of our optimization from different starting points (initialization methods on left). Our method typically converges to the same point given any initialization, but the median method (see text) is usually the most efficient. All depths are displayed at the same scale. The input image and depth can be seen in Fig~\ref{fig:depthoverview}.
}
\label{fig:startpoint}
\end{figure}
%%%%%%%%%%%%%%%%%

In the general case of videos, the size of this system can be very large (number of pixels $\times$ number of frames squared), although it will be sparse because of the limited number of pairwise interactions in the optimization. Still, given modern hardware limitations, we cannot solve this system directly, so we also must use an iterative method to solve the least squares system at each iteration of our IRLS procedure; we use preconditioned conjugate gradient and construct a preconditioner using incomplete Cholesky factorization.

Because we use iterative optimization, starting from a good initial estimate is helpful for quick convergence with fewer iterations. We have found that initializing with some function of the warped candidate depths provide a decent starting point, and we use the median value (per-pixel) of all candidate depths in our implementation (i.e., $\ve{D}^{(0)}_i = \text{median}\{\phi_1(C_i^{(1)}), \ \ldots,\  \phi_K(C_i^{(K)}) \}$). 
\newtextRev{
The method of Konrad et al.~\cite{Konrad:SPIE:12,Konrad:3DCINE:12} also uses the median of retrieved depth maps, but this becomes their final depth estimate. Our approach uses the median only for initialization and allows any of the warped candidate depths to contribute and influence the final estimate (dictated by the objective function).
} 
Figure~\ref{fig:startpoint} shows the optimization process for different initializations.

One issue is that this optimization can require a great deal of data to be stored concurrently in memory (several GBs for a standard definition clip of a few seconds). Solving this optimization efficiently, both in terms of time and space, is beyond the scope of this paper.

}

\section{{\kern-1pt Improved depth estimation for video}}
\label{sec:depthest:video}

Generating depth maps frame-by-frame without incorporating temporal information often leads to temporal discontinuities; past methods that ensure temporal coherence rely on a translating camera and static scene objects.  Here, we present a framework that improves depth estimates and enforces temporal coherence for {\it arbitrary video sequences}. That is, our algorithm is suitable for videos with moving scene objects and rotating/zooming views where conventional SfM and stereo techniques fail. (Here, we assume that zooming induces little or no parallax.)

%%%%%%%%%%%%%%%%%
\begin{figure}[t]
\begin{center}
\begin{minipage}{.23\linewidth}
%\centerline{\footnotesize{Input frames}}
%\vspace{.75mm}
\includegraphics[width=\linewidth, trim=0pt 30pt 1pt 48pt, clip=true]{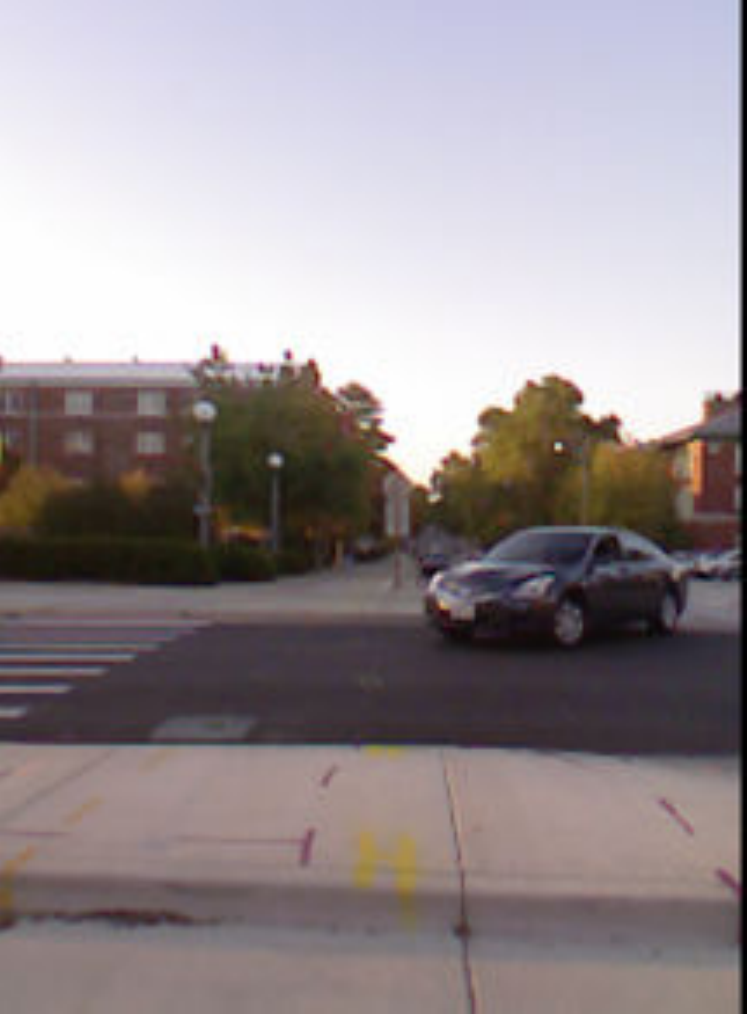}
\includegraphics[width=\linewidth, trim=0pt 30pt 1pt 48pt, clip=true]{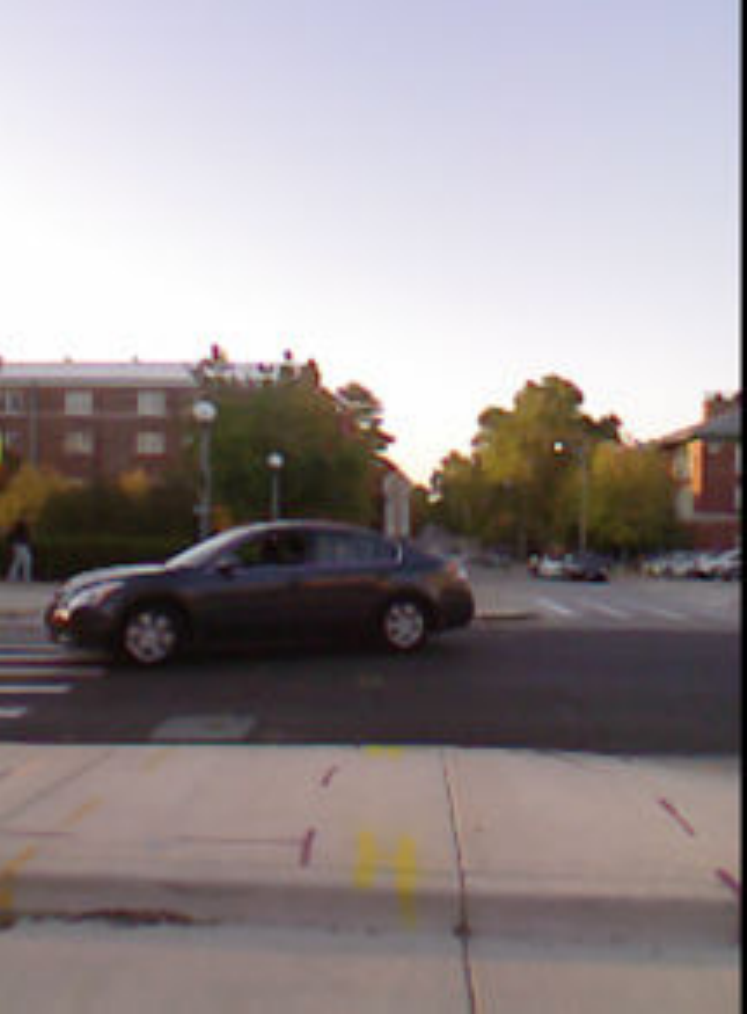}
\includegraphics[width=\linewidth, trim=0pt 30pt 1pt 48pt, clip=true]{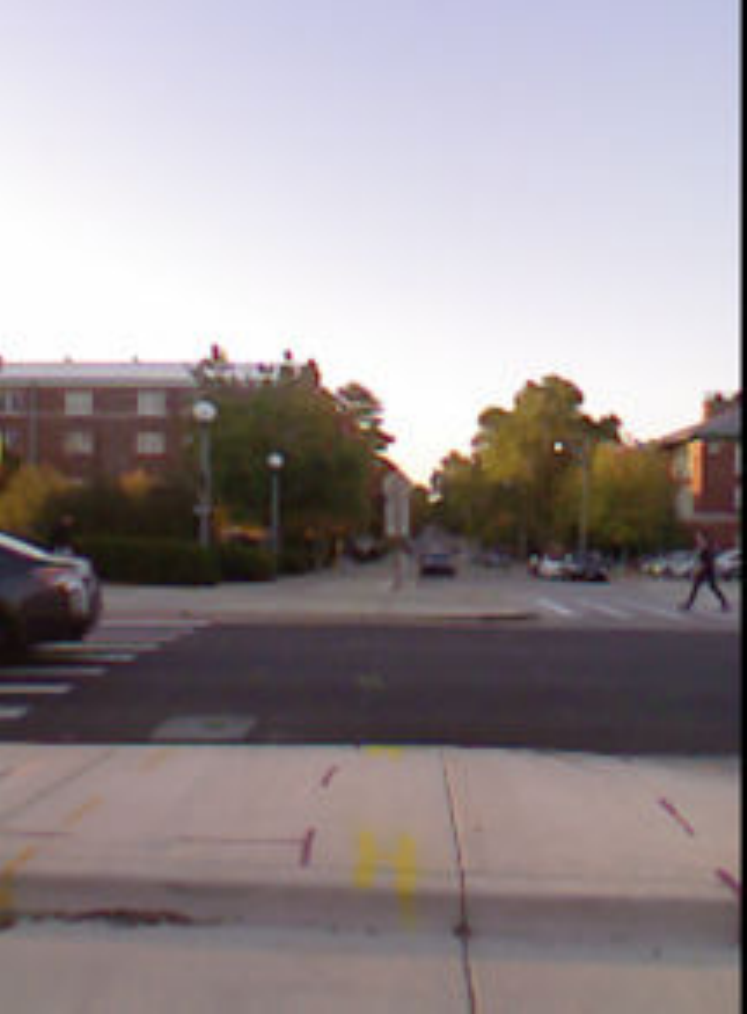}
\end{minipage}
\begin{minipage}{.23\linewidth}
%\centerline{\footnotesize{Image inference}}
%\vspace{.75mm}
\includegraphics[width=\linewidth, trim=0pt 30pt 1pt 48pt, clip=true]{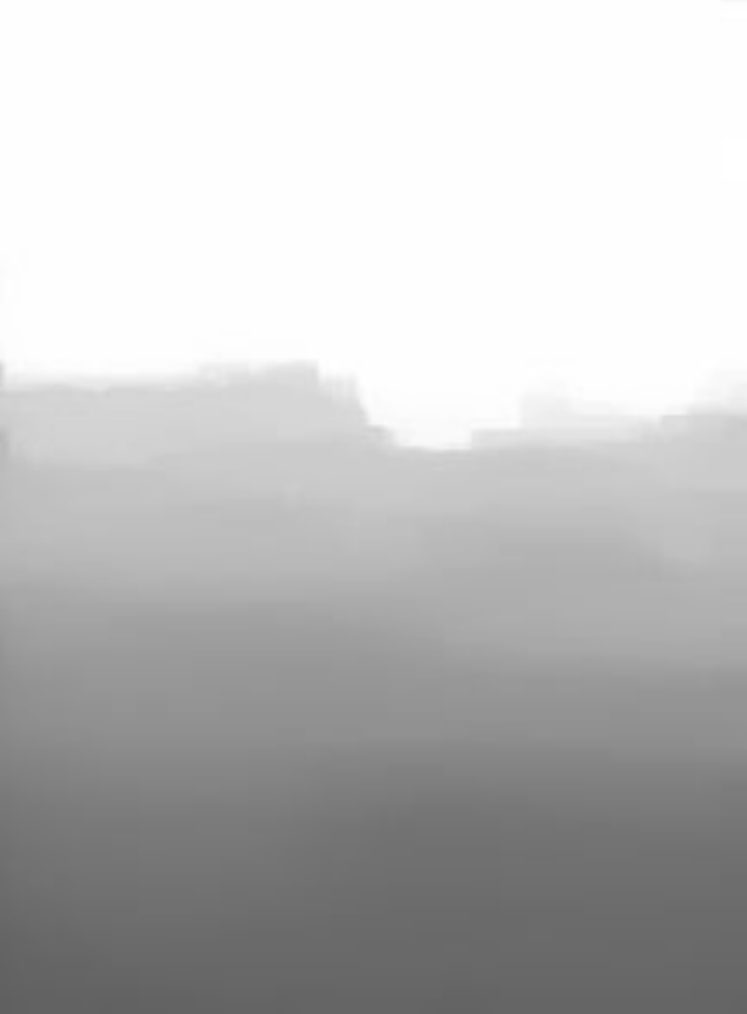}
\includegraphics[width=\linewidth, trim=0pt 30pt 1pt 48pt, clip=true]{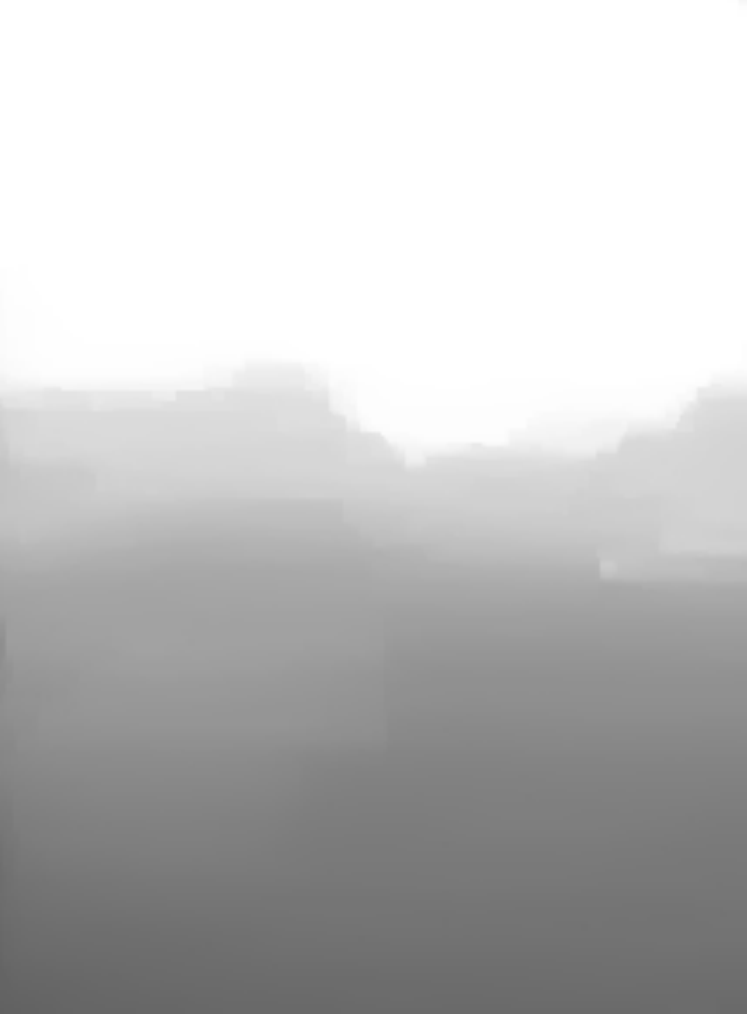}
\includegraphics[width=\linewidth, trim=0pt 30pt 1pt 48pt, clip=true]{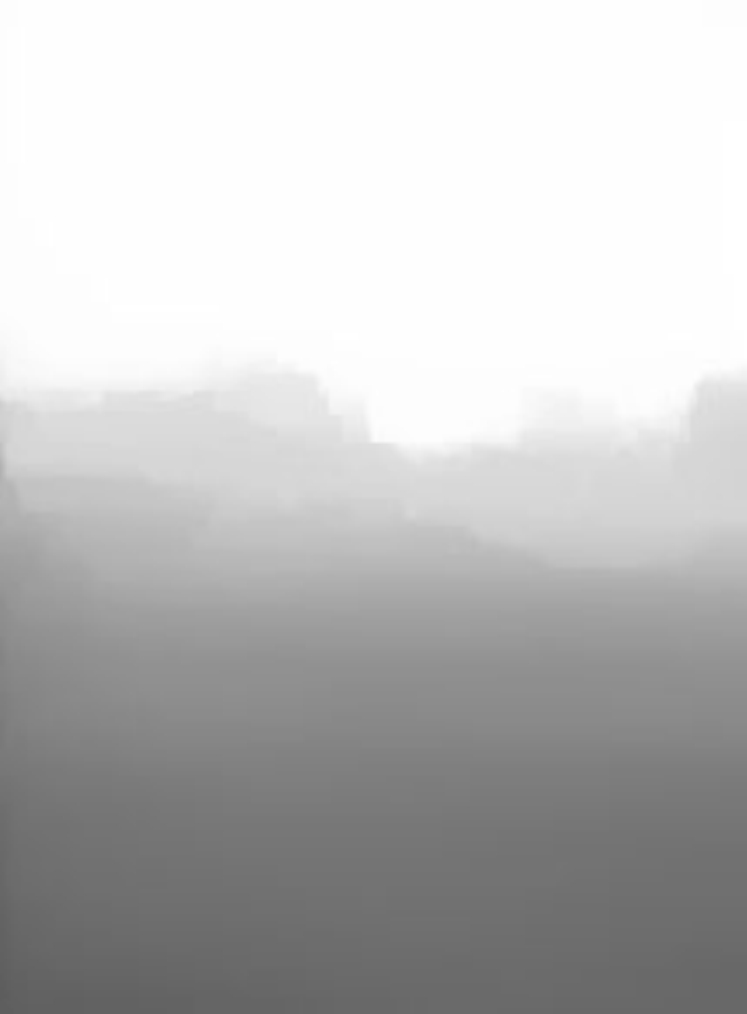}
\end{minipage}
\begin{minipage}{.23\linewidth}
%\centerline{\footnotesize{Video inference}}
\vspace{.5mm}
\includegraphics[width=\linewidth, trim=0pt 30pt 1pt 48pt, clip=true]{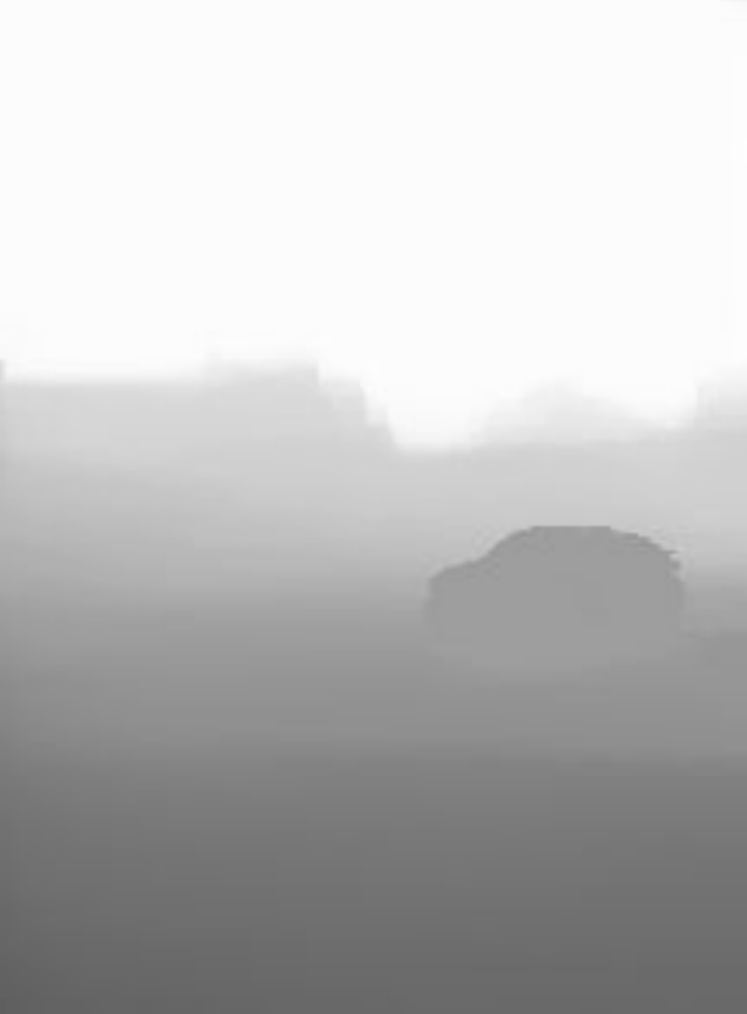}
\includegraphics[width=\linewidth, trim=0pt 30pt 1pt 48pt, clip=true]{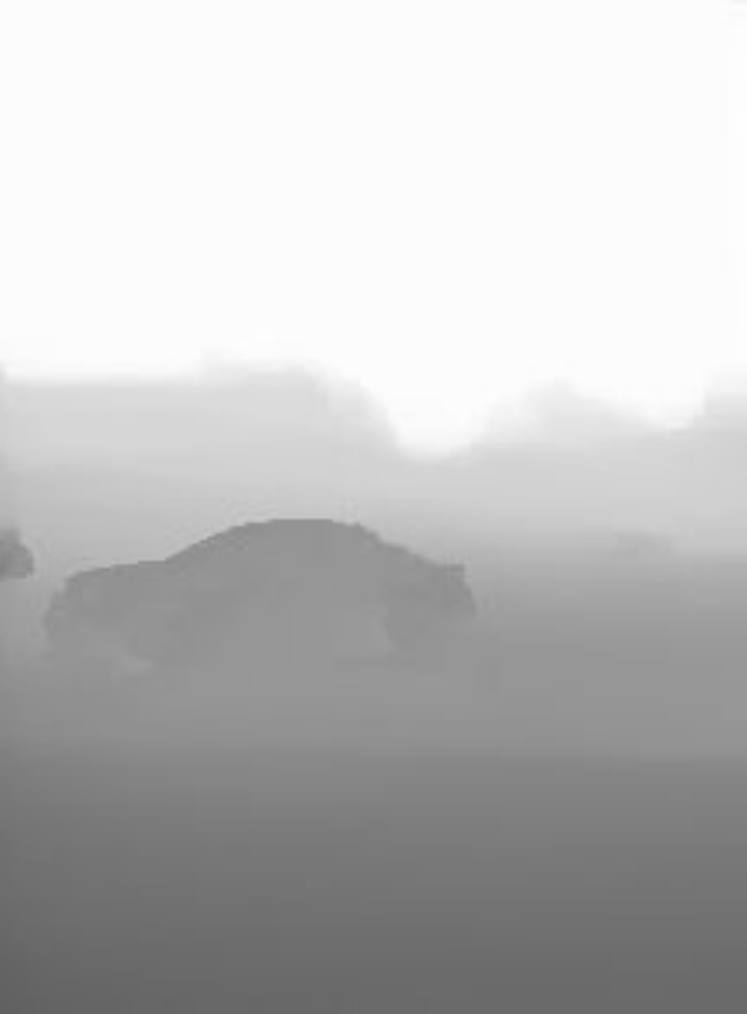}
\includegraphics[width=\linewidth, trim=0pt 30pt 1pt 48pt, clip=true]{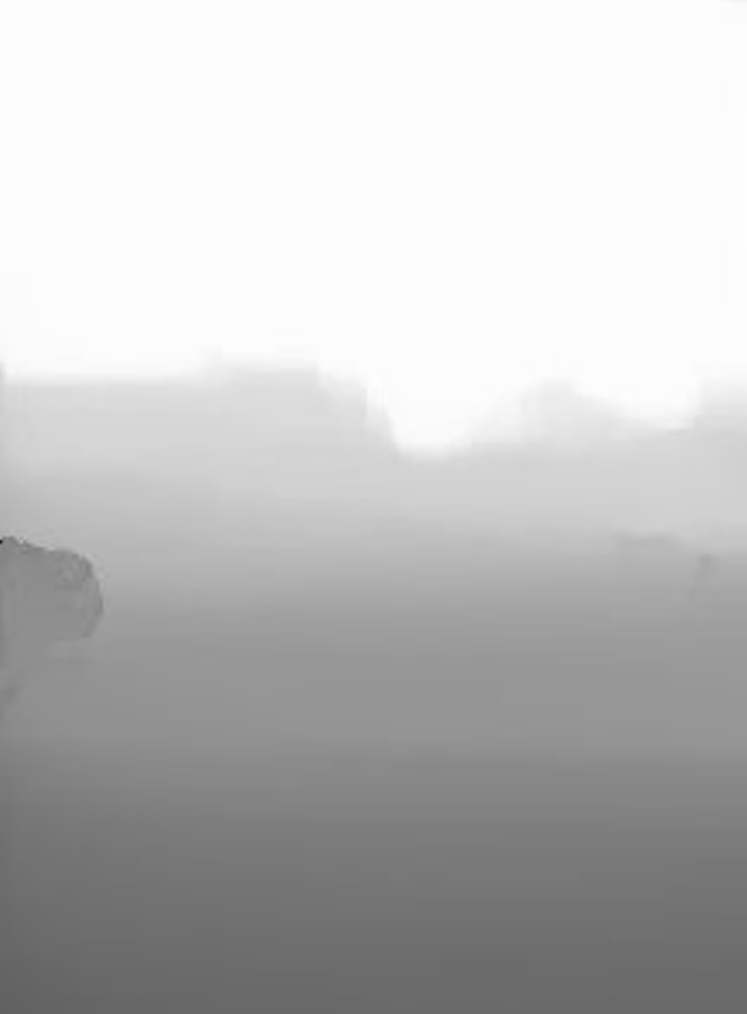}
\end{minipage}
\begin{minipage}{.23\linewidth}
%\centerline{\footnotesize {Segmentation}}
\includegraphics[width=\linewidth, trim=0pt 30pt 1pt 48pt, clip=true]{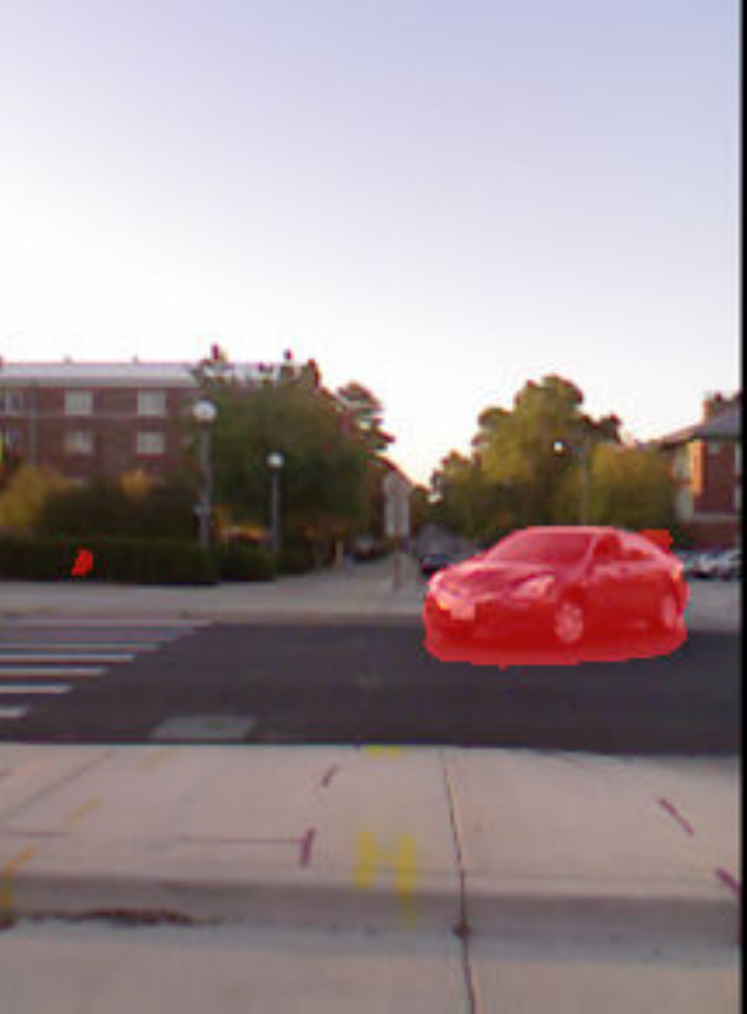}
\includegraphics[width=\linewidth, trim=0pt 30pt 1pt 48pt, clip=true]{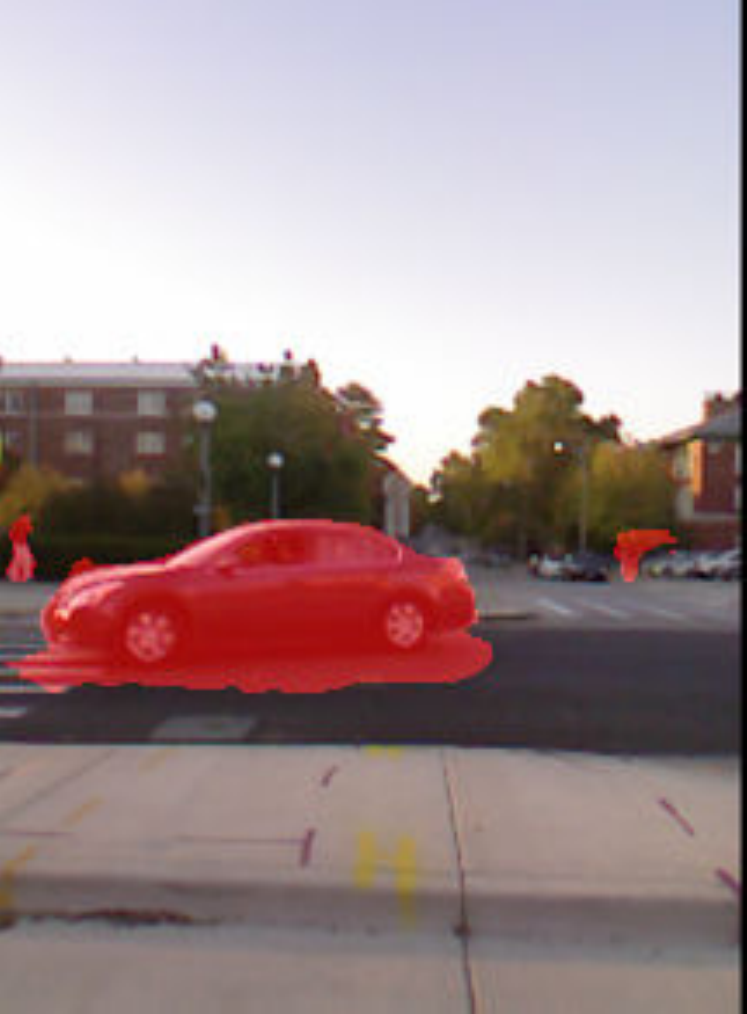}
\includegraphics[width=\linewidth, trim=0pt 30pt 1pt 48pt, clip=true]{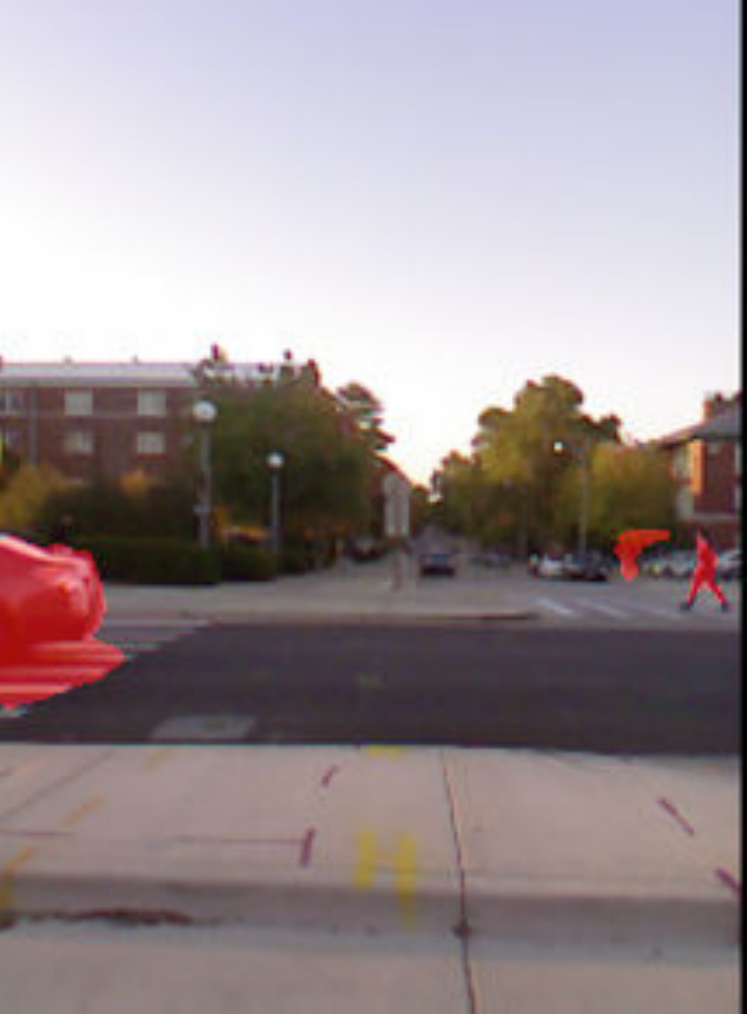}
\end{minipage}
\end{center}
\caption{Importance of temporal information. Left: input frames. Mid-left: predicted depth without temporal information. Note that the car is practically ignored here. Mid-right: predicted depth with temporal information, with the depth of the moving car recovered. Right: detected moving object.
}
\label{fig:temporal_benefit}
\end{figure}
%%%%%%%%%%%%%%%%%

Our idea is to incorporate temporal information through additional terms in the optimization that ensure (a) depth estimates are consistent over time and (b) that moving objects have depth similar to their contact point with the ground. Each frame is processed the same as in the single image case (candidate matching and warping), except that now we employ a global optimization (described below) that infers depth for the entire sequence at once, incorporating temporal information from all frames in the video.
%Our idea is to perform the same candidate matching and warping scheme as in the single image case for each frame of the video sequence, and also incorporate temporal information as an additional term in the optimization to improve depth estimation and maintain coherence throughout the sequence.
Fig~\ref{fig:temporal_benefit} illustrates the importance of these additional terms in our optimization.

We formulate the objective for handling video by adding two terms to the single-image objective function:
\begin{equation}
\label{eq:depth_video}
E_{\text{video}}(\ve{D}) = E(\ve{D}) +  \sum_{i\in\text{pixels}} \nu  E_{\text{c}}(\ve{D}_i) + \eta E_{\text{m}}(\ve{D}_i),
\end{equation}
where $E_{\text{c}}$ encourages temporal coherence while $E_{\text{m}}$ uses motion cues to improve the depth of moving objects. The weights $\nu$ and $\eta$ balance the relative influence of each term ($\nu = 100, \eta = 5$).

We model temporal coherence first by computing per-pixel optical flow for each pair of consecutive frames in the video (using Liu's publicly available code~\cite{Liu:thesis09}). We define the {\it flow difference}, $\nabla_{flow}$, as a linear operator which returns the change in the flow across two corresponding pixels, and model the coherence term as
\begin{equation}
\label{eq:coherence}
E_{\text{c}}(\ve{D}_i) =   s_{t,i} \phi(\nabla_{flow}\ve{D}_i).
\end{equation}
We weight each term by a measure of flow confidence, \\$s_{t,i} = (1+e^{-(||\nabla_{flow} \ve{L}_i|| - \mu_L)/\sigma_L})^{-1}$, which intuitively is a soft threshold on the reprojection error ($\mu_L = 0.05$, $\sigma_L = 0.01$). Minimizing the weighted flow differences has the effect of temporally smoothing inferred depth in regions where optical flow estimates are accurate.

To handle motion, we detect moving objects in the video (Sec~\ref{sec:motion_est}) and constrain their depth such that these objects touch the floor. Let $m$ be the binary motion segmentation mask and $\mathcal{M}$ the depth in which connected components in the segmentation mask contact the floor. We define the motion term as
\begin{equation}
\label{eq:motion}
E_{\text{m}}(\ve{D}_i) = m_i \phi(\ve{D}_i - \mathcal{M}_i).
\end{equation}

%% Old motion segmentation figure
%%%%%%%%%%%%%%%%%%
%\begin{figure}[t]
%\includegraphics[width=1\columnwidth,page=1]{fig/motionseg/motionseg_eccv.pdf}
%\caption{We apply median filtering on the stabilized images to extract the background image. A pixel is deemed to be in motion if there is sufficient relative difference from the background image.}
%\label{fig:motionseg}
%\end{figure}
%%%%%%%%%%%%%%%%%%

%%%%%%%%%%%%%%%%%
\begin{figure*}[t]
\includegraphics[width=\linewidth]{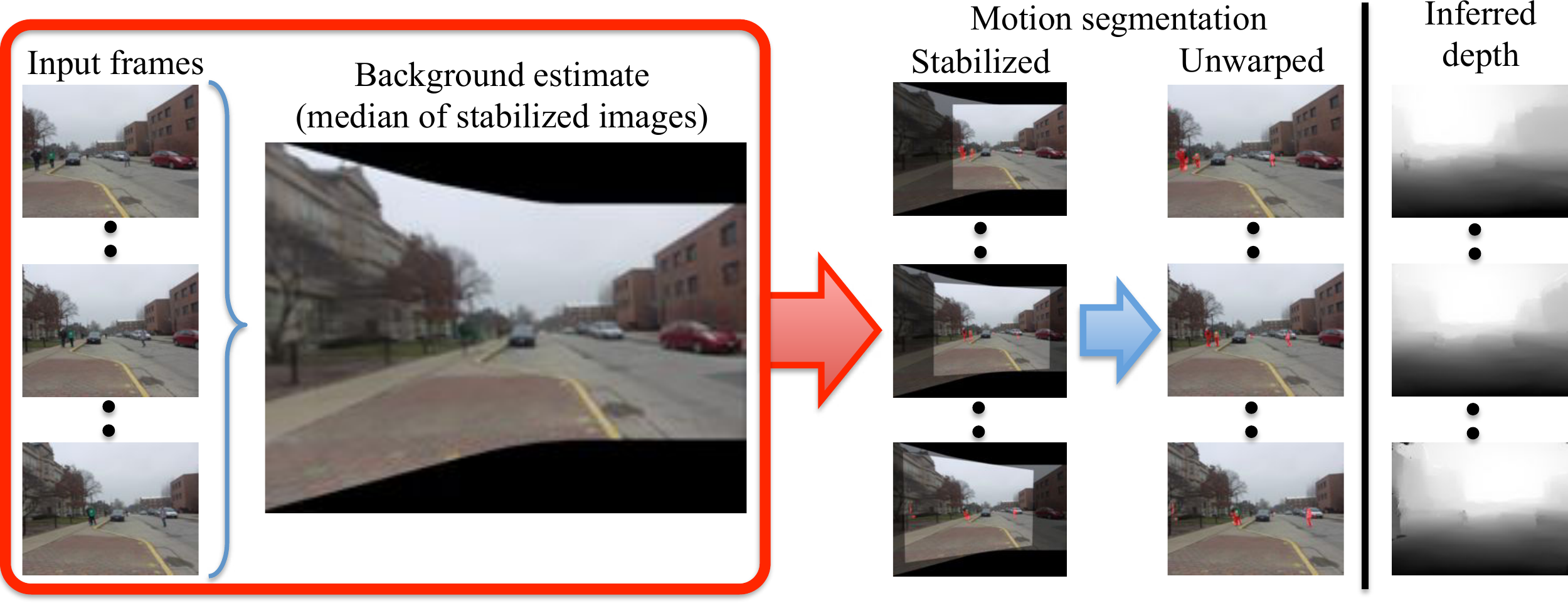}
\caption{Example of our motion segmentation applied to a rotating sequence. We first estimate homographies to stabilize the video frames and create a clean background image using a temporal median filter. We then evaluate our estimate of motion thresholding metric on the stabilized sequences, and unwarp the result (via the corresponding inverse homography) to segment the motion in the original sequence. We can then improve our inferred depth using this segmentation. Note that this technique is applicable to all video sequences that do not contain parallax induced from camera motion.}
\label{fig:motionseg_homography}
\end{figure*}
%%%%%%%%%%%%%%%%%

\subsection{Detecting moving objects}
\label{sec:motion_est}

Differentiating moving and stationary objects in the scene can be a useful cue when estimating depth. Here we describe our algorithm for detecting objects in motion in non-translating movies (i.e., static, rotational, and variable focal length videos)\footnote{In all other types of videos (e.g. those with parallax or fast moving objects/pose), we do not employ this algorithm; equivalently we set the motion segmentation weight to zero ($\eta=0$).}. Note that there are many existing techniques for detecting moving objects (e.g., \cite{Heikkila:PAMI06,Sheikh:ICCV09}); we use what we consider to be easy to implement and effective for our purpose of depth extraction.

First, to account for dynamic exposure changes throughout the video, we find the image with the lowest overall intensity in the sequence and perform histogram equalization on all other frames in the video. We use this image as to not propagate spurious noise found in brighter images. Next, we use RANSAC on point correspondences to compute the dominant camera motion (modeled using homography) to align neighboring frames in the video.  Median filtering is then used on the stabilized images to extract the background $B$ (ideally, without all the moving objects).

%In this paper, we handle videos from a stationary camera; the stationary background is used to detect and segment moving objects. Next, we compute optical flow for each pair of consecutive frames, and estimate the background image ($B$) with the flow-weighted average of the input sequence:
%\begin{equation}
%\label{eq:background}
%B_i = \frac{\sum_{k=1}^n L_{i,k} f_{i,k} }{\sum_{k=1}^n f_{i,k}},
%\end{equation}
%where $L_{i,k}$ and $f_{i,k}$ are the $i^{th}$ pixel from the $k^{th}$ video frame of the input image and flow weights respectively. Pixels with high motion (large flow magnitude) have less weight: $f_{i,k} =  (1+e^{-(||flow_{i,k}||-1)/.01})^{-1}$, where $||flow_{i,k}||$ is the flow magnitude at pixel $i$ of frame $k$. (Median filtering could also be used to obtain the background, but we found that this technique works better in our case.)

In our method, the likelihood of a pixel being in motion depends on how different it is from the background, weighted by the optical flow magnitude which is computed between stabilized frames (rather than between the original frames). We use relative differencing (relative to background pixels) to reduce reliance on absolute intensity values, and then threshold to produce a mask:%\vspace{-.5em}
\begin{equation}
\label{eq:moseg}
m_{i,k} = \left\{ \begin{array}{ll} 1 & \text{if } ||flow_{i,k}|| \frac{||W_{i,k}-B_i||^2}{B_i} > \tau \\ 0 & \text{otherwise,} \end{array} \right.
\end{equation}
%\vspace{-1.em}\\
where $\tau = 0.01$ is the threshold, and $W_{i,k}$ is the $k^{th}$ pixel of the $i^{th}$ stabilized frame (i.e., warped according to the homography that aligns $W$ with $B$). This produces a motion estimate in the background's coordinate system, so we apply the corresponding inverse homography to each warped frame to find the motion relative to each frame of the video. This segmentation mask is used (as in Eq~\ref{eq:motion}) to improve depth estimates for moving objects in our optimization. Fig~\ref{fig:motionseg_homography} illustrates this technique.

%%%%%%%%%%%%%%%%%%%%%%%%%%%%%%%%%%%%%%
\section{MSR-V3D Dataset}
\label{sec:dataset}

%%%%%%%%%%%%%%%%%
\begin{figure}[t]
\centerline{\includegraphics[width=\columnwidth]{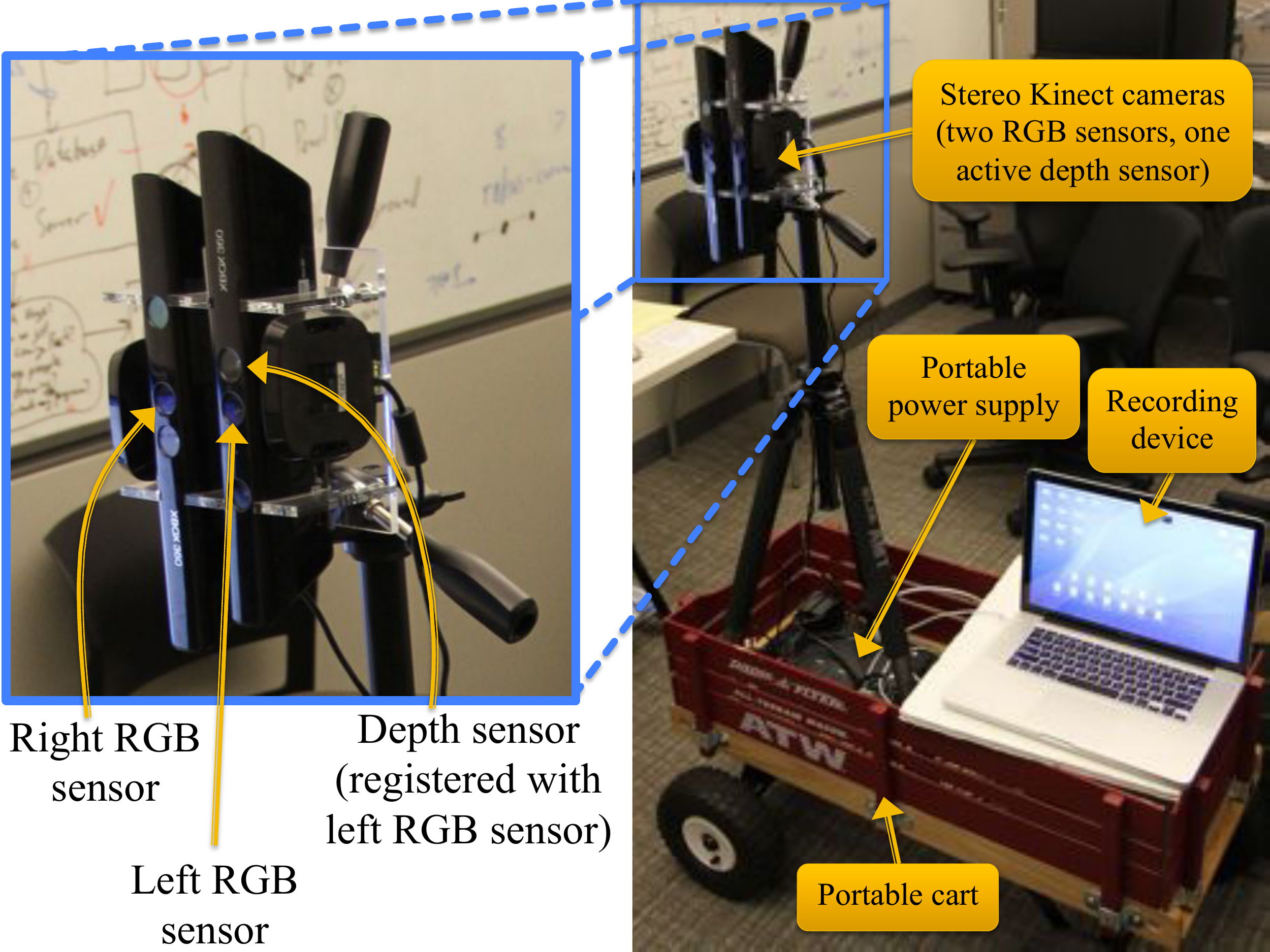}}
\caption{Our stereo-RGBD collection rig consists of two side-by-side Microsoft Kinects. The rig is mobile through the use of an uninterruptible power supply, laptop, and rolling mount.
}\afterfigure
\label{fig:kinectrig}
\end{figure}
%%%%%%%%%%%%%%%%%

\newtext{

In order to train and test our technique on image/video input, we collected a dataset containing stereo video clips for a variety of scenes and viewpoints (116 indoor, 61 outdoor). The dataset primarily contains videos recorded from a static viewpoint with moving objects in the scene (people, cars, etc.). There are also 100 indoor frames (single images) in addition to the 116 indoor video clips within the database (stereo RGB, depth for the left frame). These sequences come from four different buildings in two cities and contain substantial scene variation (e.g., hallways, rooms, foyers). Each clip is filmed with camera viewpoints that are either static or slowly rotated. We have entitled our dataset the Microsoft Research Stereo Video + Depth (MSR-V3D), and it is available online at \url{http://kevinkarsch.com/depthtransfer}.

We captured the MSR-V3D dataset with two side-by-side, vertically mounted Microsoft Kinects shown in Fig~\ref{fig:kinectrig} (positioned about 5cm apart). We collected the color images from both Kinects and only the depth map from the left Kinect. For each indoor clip, the left stereo view also contains view-aligned depth from the Kinect. Due to IR interference, depth for the right view was not captured indoors, and depth was totally disregarded outdoors due to limitations of the Kinect.

We also collected outdoor data with our stereo device. However, because the Kinect cannot produce depth maps outdoors due to IR interference from the sunlight, we could not use these sequences for training. \newtextRev{We attempted to extract ground truth disparity between stereo pairs, but the quality/resolution of Kinect images were too low to get decent estimates.} We did, however, use this data for testing and evaluation purposes.

Because the Kinect estimates depth by triangulating a pattern of projected infrared (IR) dots, multiple Kinects can interfere with each other, causing noisy and inaccurate depth. Thus, we mask out the depth sensor/IR projector from the rightmost Kinect, and only collect depth corresponding to the left views. This is suitable for our needs, as we only need to estimate depth for the input (left) sequence.

Kinect depth is also susceptible to ``holes'' in the data caused typically by surface properties, disoccluded/interfered IR pattern, or because objects are simply too far away from the device. For training, we disregard all pixels of the videos which contain holes, and for visualization, we fill the holes in using a na\"{\i}ve horizontal dilation approach.

%The  data we have collected is typically on the scale of rooms, and most objects are approximately 1-8 meters from the cameras. Sunlight also causes IR interference, thus we have only collected indoor scenes in our dataset.

%Kinects can capture RGB and depth data at up to 30Hz; however, this puts a significant strain on the USB bus that communicates the Kinect data to the computer. Furthermore, when two Kinects transfer data simultaneously (as in our setup), the bandwidth requirement is too great too handle with one USB bus at 30Hz, which corrupts the data stream. Thus, we only collect data at 20Hz, which we found empirically to be the highest data rate that avoids this corruption issue and is still suitable for our purposes. Ideally, the data collection rig would include a computer with multiple USB busses, but this would likely require a desktop PC rather than a laptop.

}

%%%%%%%%%%%%%%%%%%%%%%%%%%%%%%%%%%%%%%
\section{Experiments}
%\vspace{-0.03mm}
\label{sec:results}

In this section, we show results of experiments involving single image and video depth extraction. We also discuss the importance of scale associated with training and the effect of the number of candidates used for depth hypothesis.

\subsection{Single image results}
We evaluate our technique on two single image RGBD datasets: the Make3D Range Image Dataset~\cite{Saxena:09}, and the NYU Depth Dataset~\cite{Silberman:ECCV12}.

\subsubsection{Make3D Range Image Dataset}
Of the 534 images in the Make3D dataset, we use 400 for testing and 134 for training (the same as was done before, e.g.,~\cite{Saxena05learningdepth,Saxena:09,Li:2010,Liu:cvpr10}). We report error for three common metrics in Table~\ref{tab:make3d_results}. Denoting $\ve{D}$ as estimated depth and $\ve{D}^*$ as ground truth depth, we compute {\it relative} ({\bf rel}) {\it error} $\frac{|\ve{D}-\ve{D}^*|}{\ve{D}^*}$, {\bf log$_{10}$} {\it error} $|\log_{10}(\ve{D})-\log_{10}(\ve{D}^*)|$, and {\it root mean squared} ({\bf RMS}) {\it error} $\sqrt{ \sum_{i=1}^N (\ve{D}_i-\ve{D}_i^*)^2/N}$. Error measures are averaged over all pixels/images in the test set. Our estimated depth maps are computed at 345$\times$460 pixels (maintaining the aspect ratio of the Make3D dataset input images).

\begin{figure*}[t]
\begin{center}
\begin{minipage}[b]{0.49\linewidth}
\includegraphics[width=.24\linewidth]{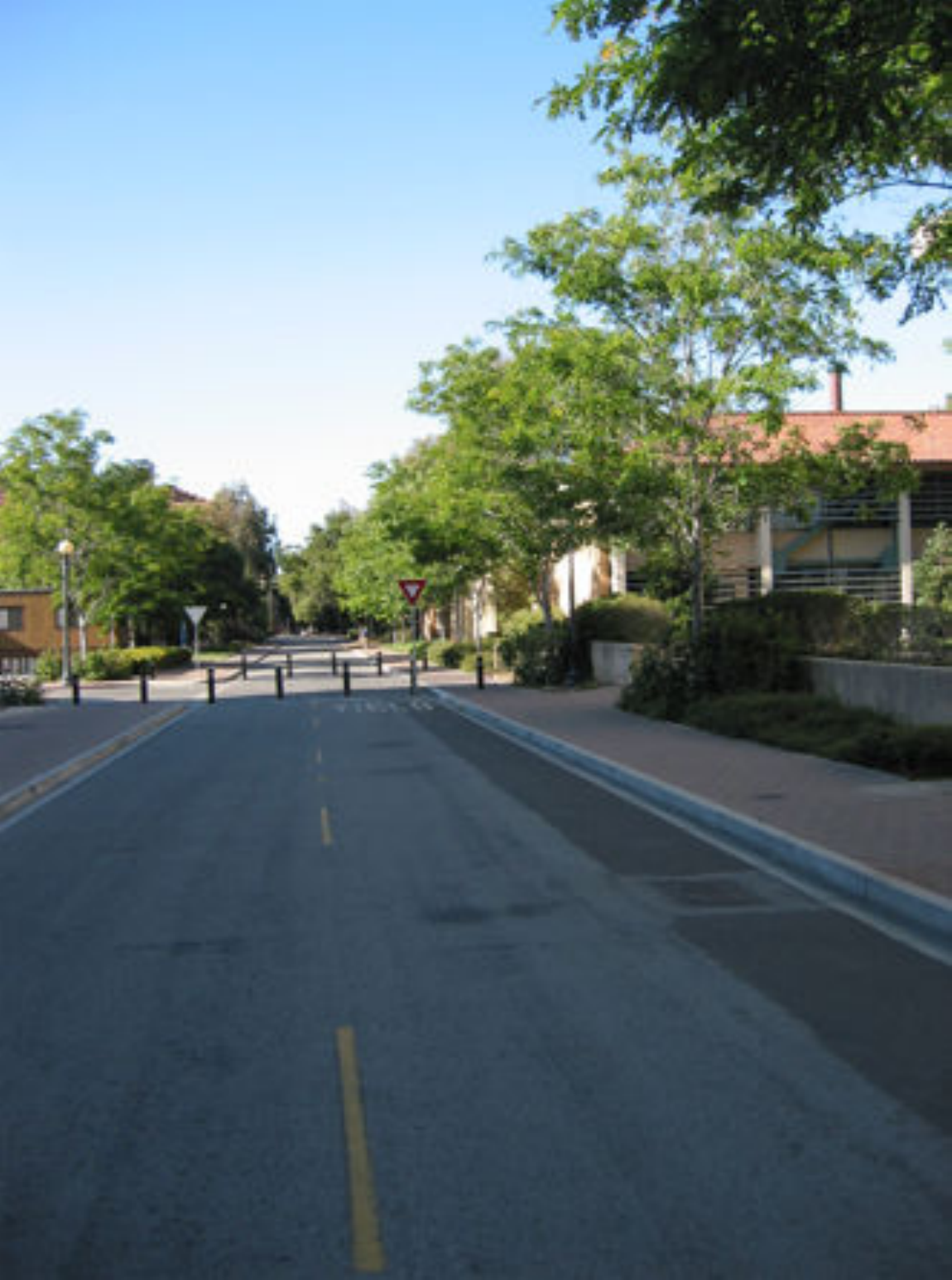}
\includegraphics[width=.24\linewidth]{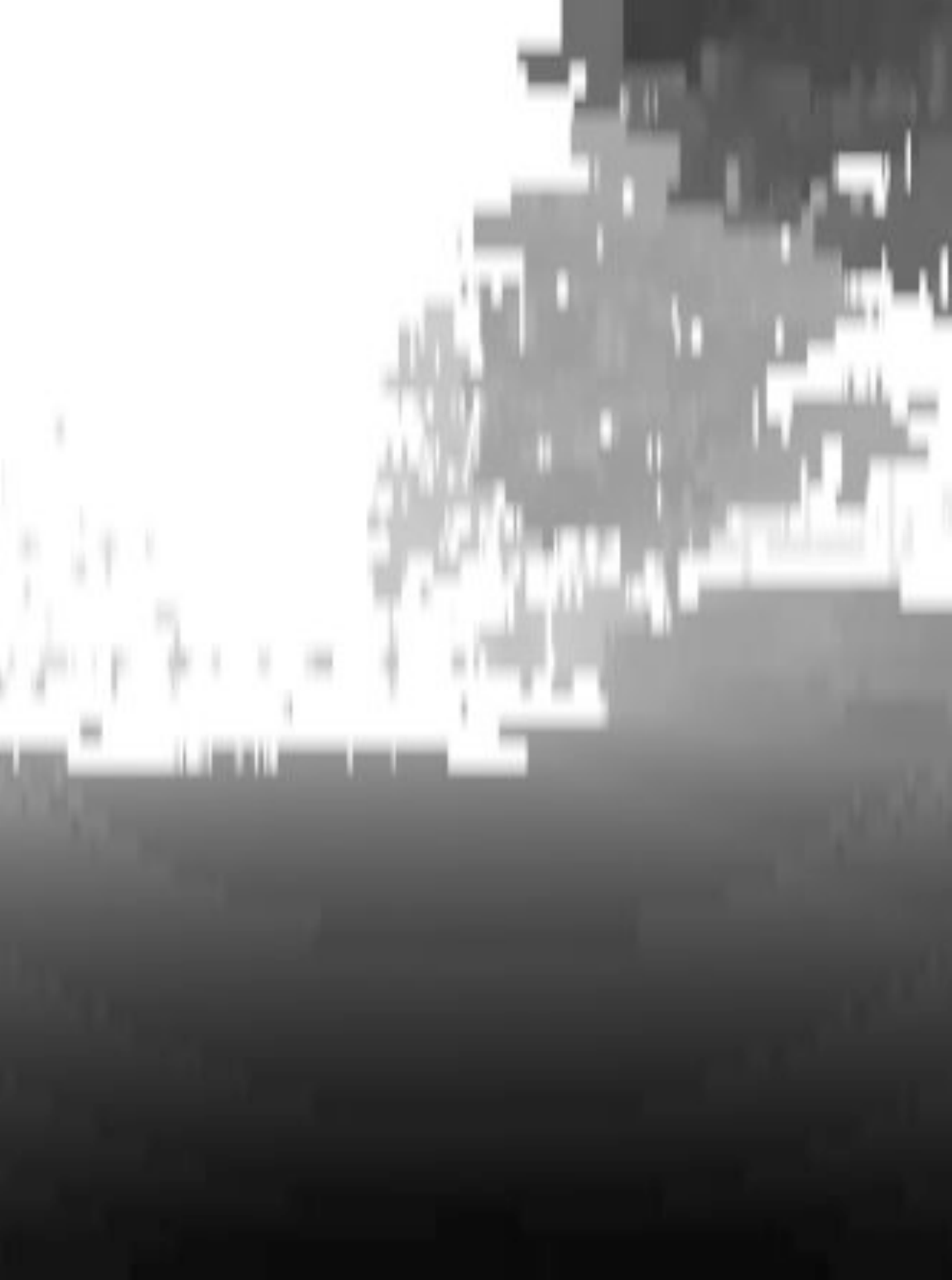}
\includegraphics[width=.24\linewidth]{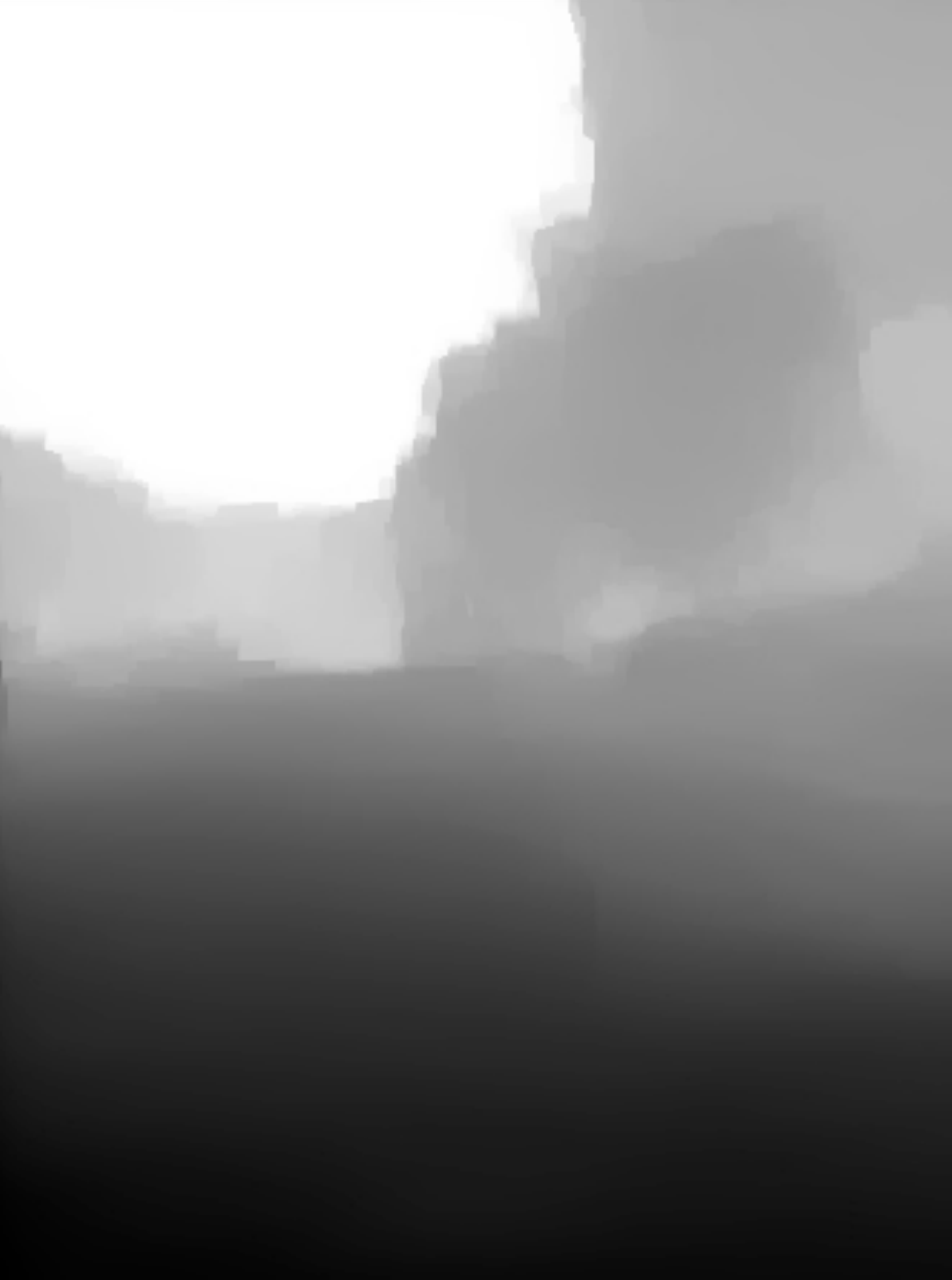}
\includegraphics[height=.323\linewidth]{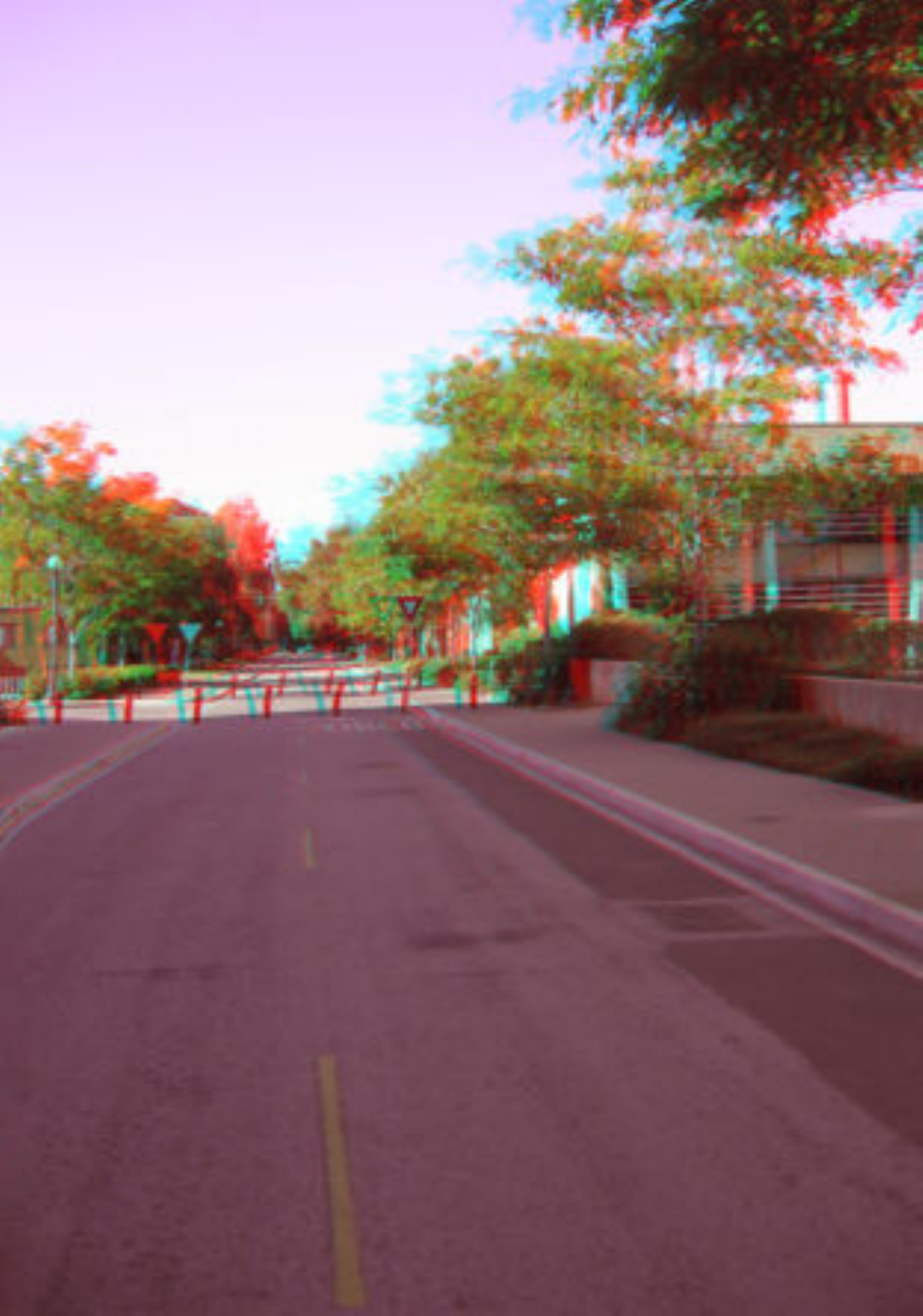}
\end{minipage}
\hfill
%\rule[0mm]{.15mm}{.165\linewidth}
\hfill
\begin{minipage}[b]{0.49\linewidth}
\includegraphics[width=.235\linewidth]{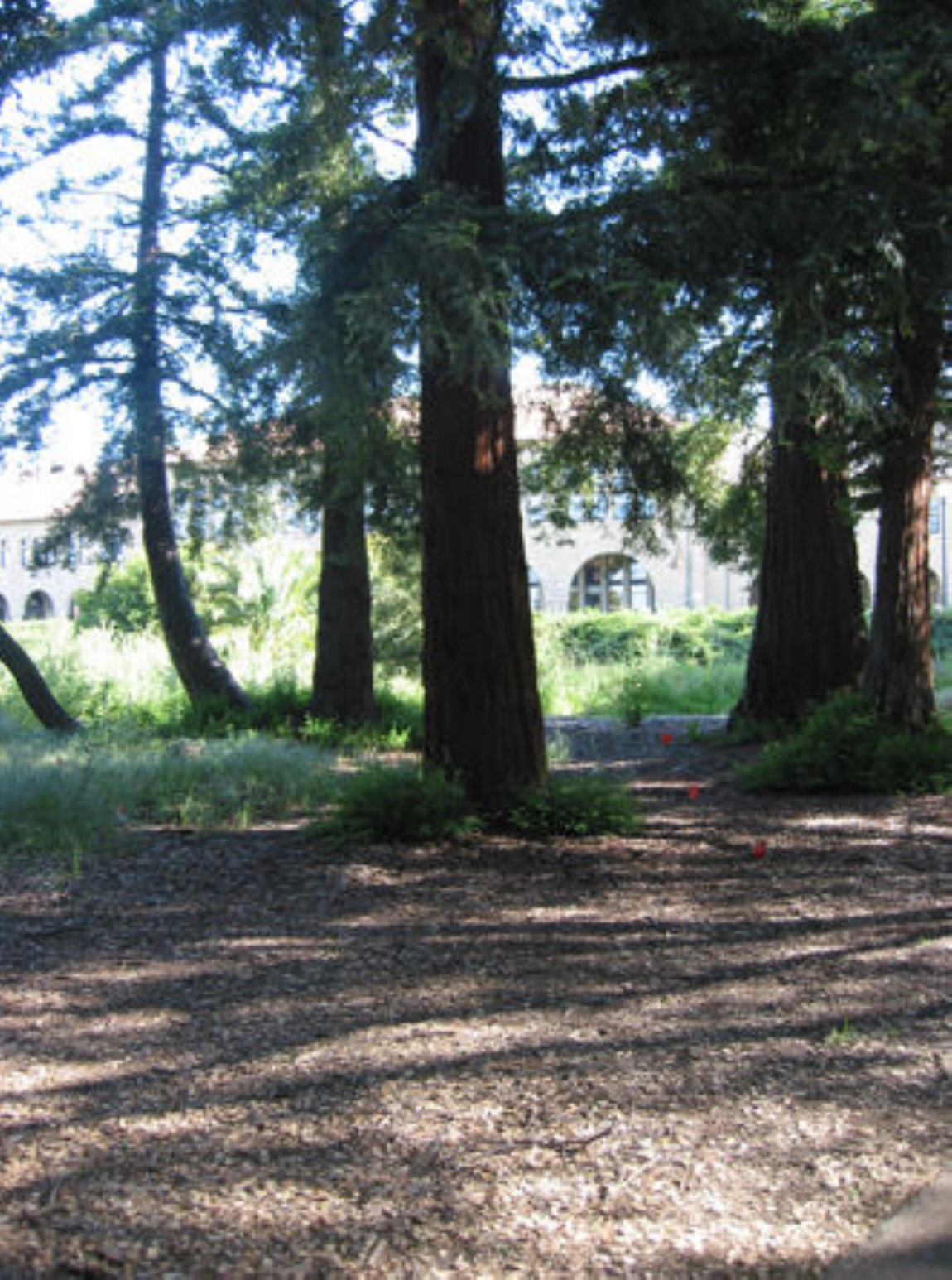}
\includegraphics[width=.235\linewidth]{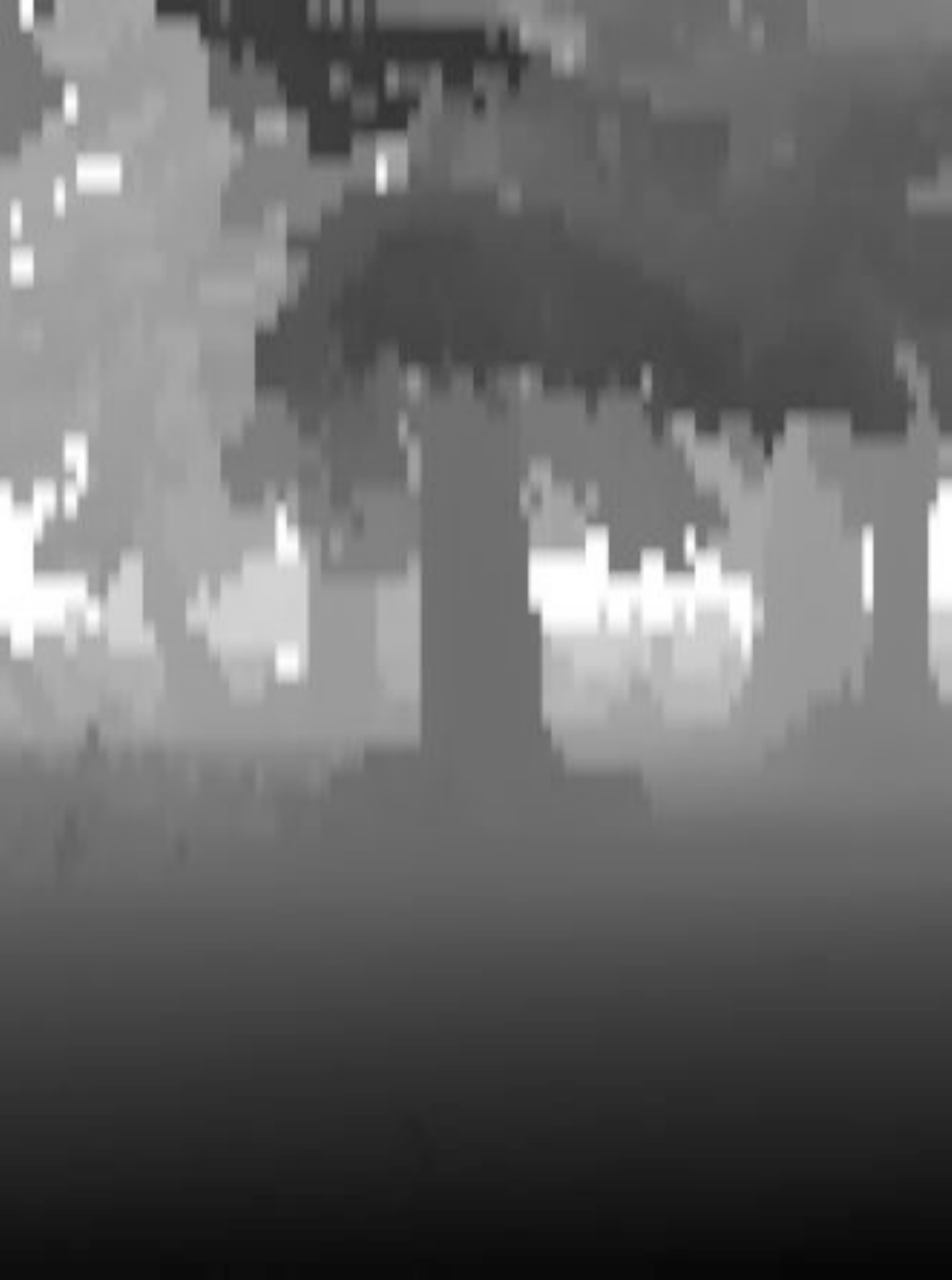}
\includegraphics[width=.235\linewidth]{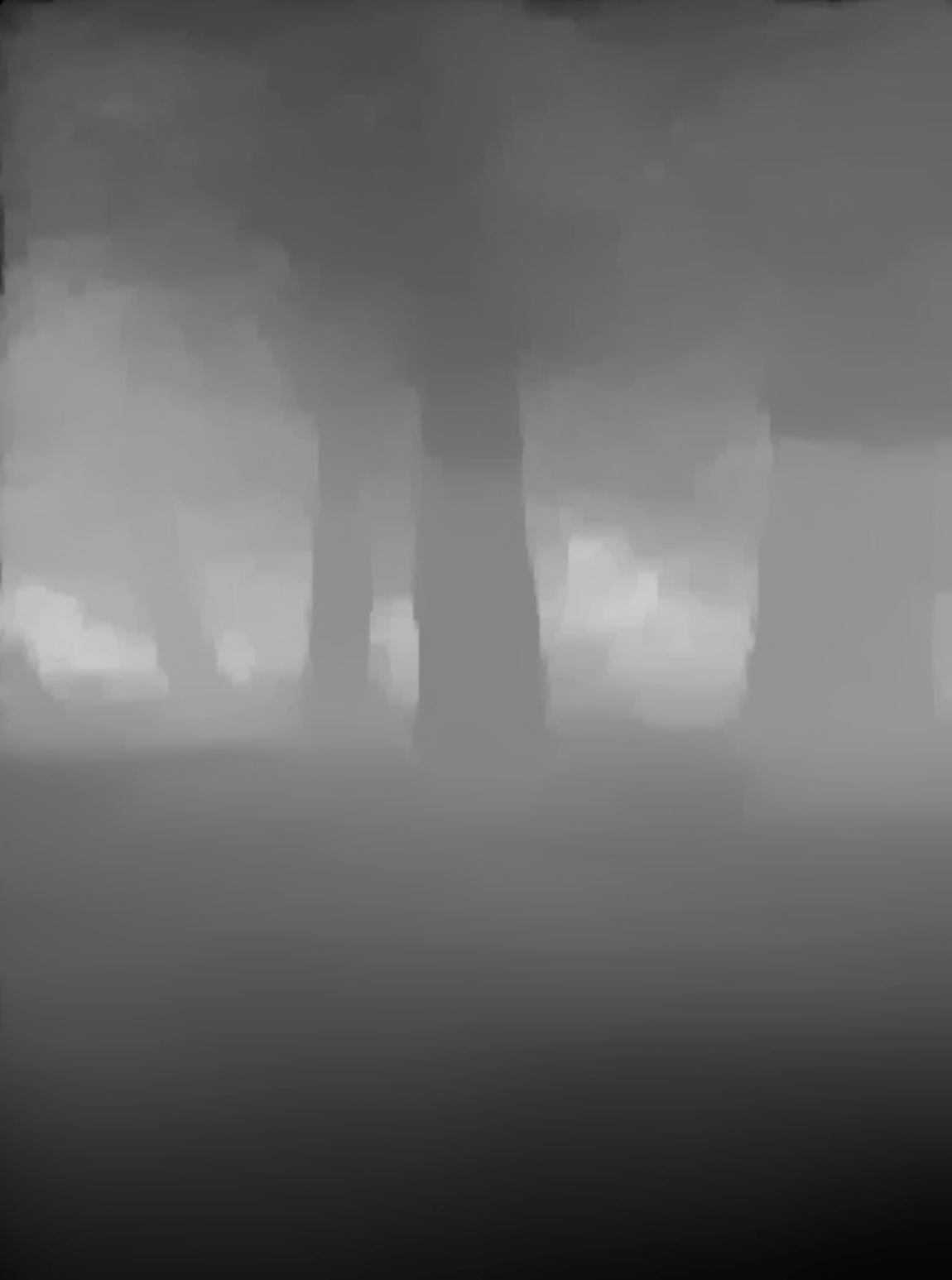}
\includegraphics[height=.315\linewidth]{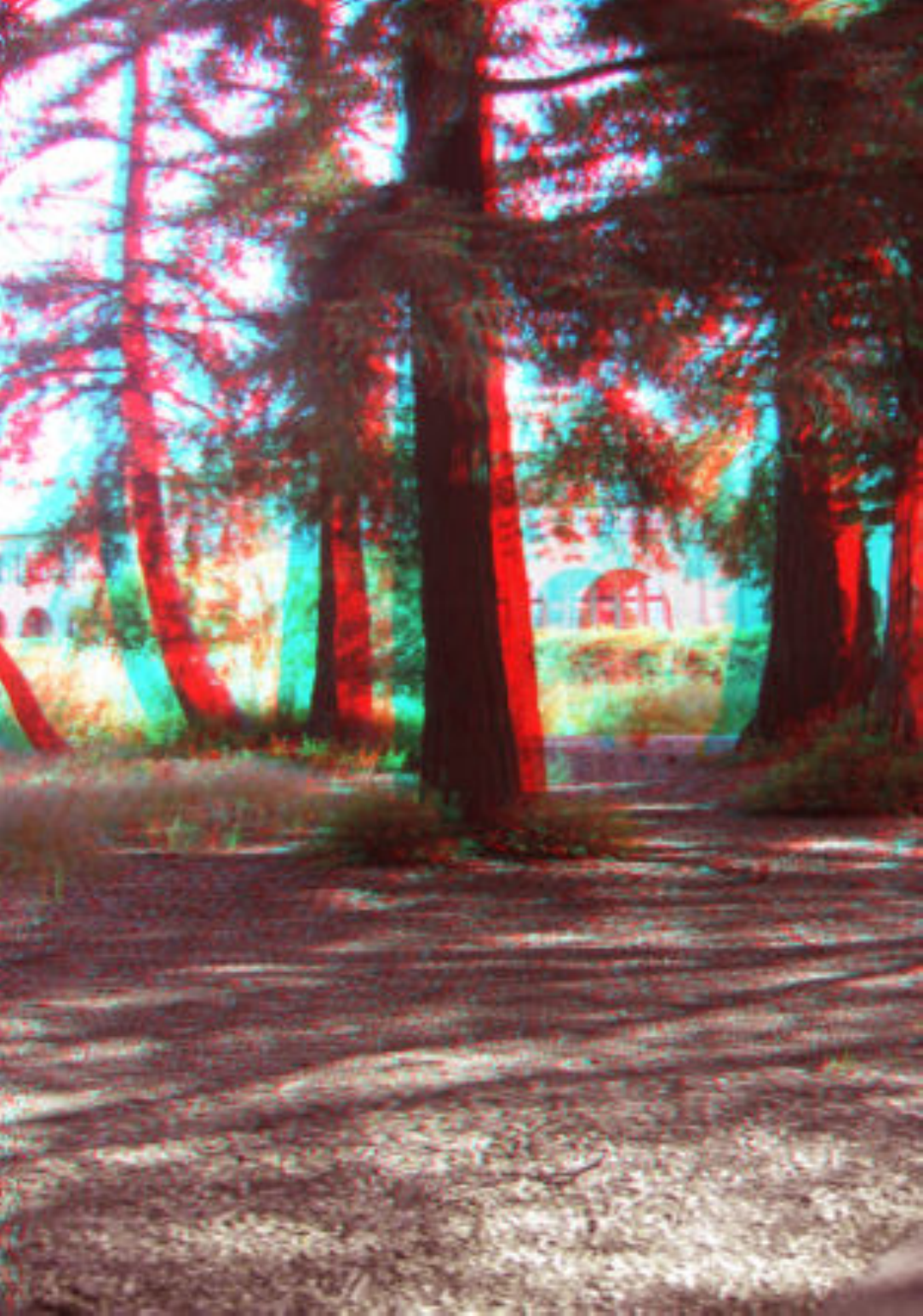}
\end{minipage}\\
\vspace{2mm}
%\rule{.995\linewidth}{.15mm}
%\\
\begin{minipage}[b]{0.49\linewidth}
\includegraphics[width=.24\linewidth]{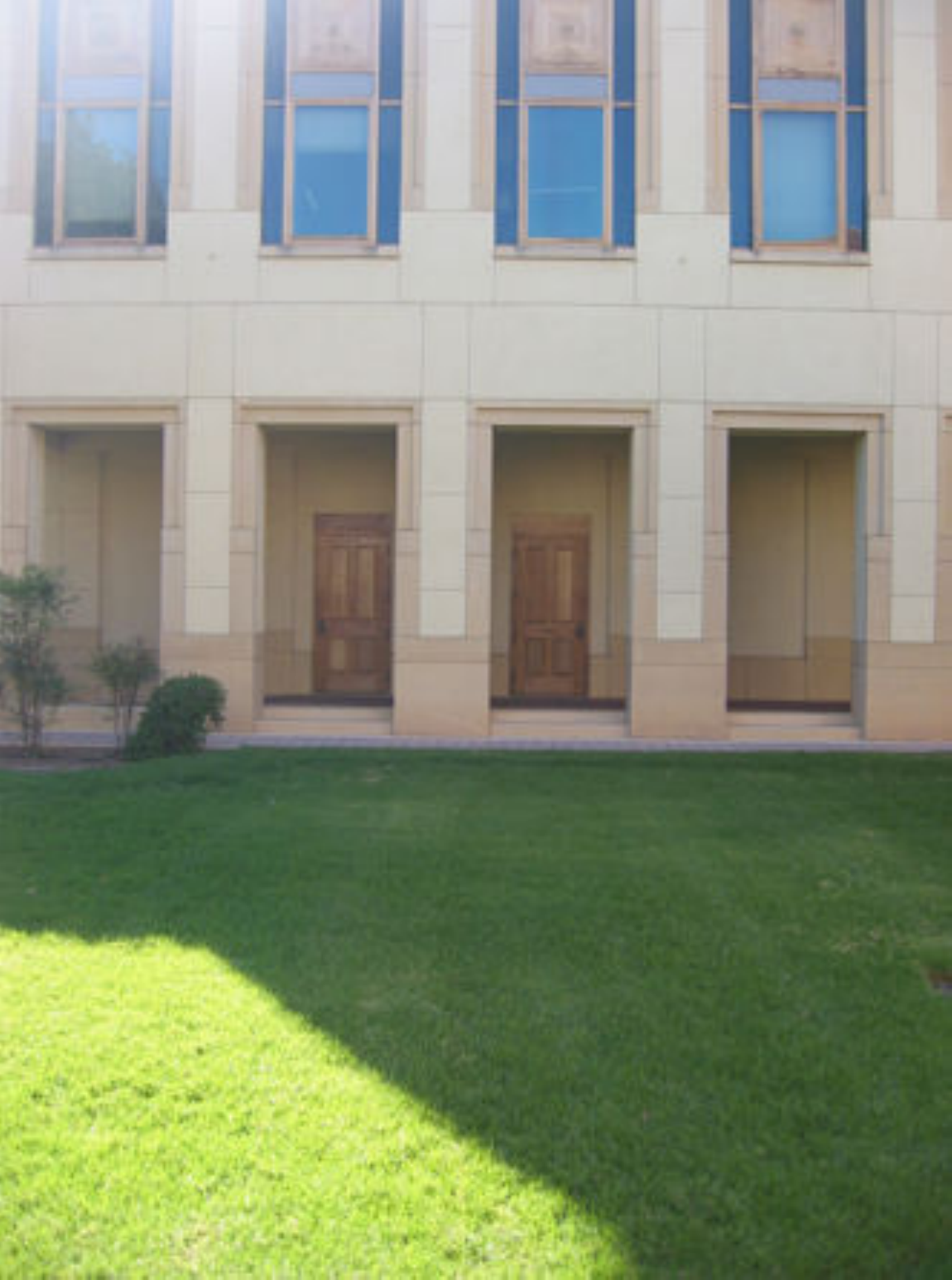}
\includegraphics[width=.24\linewidth]{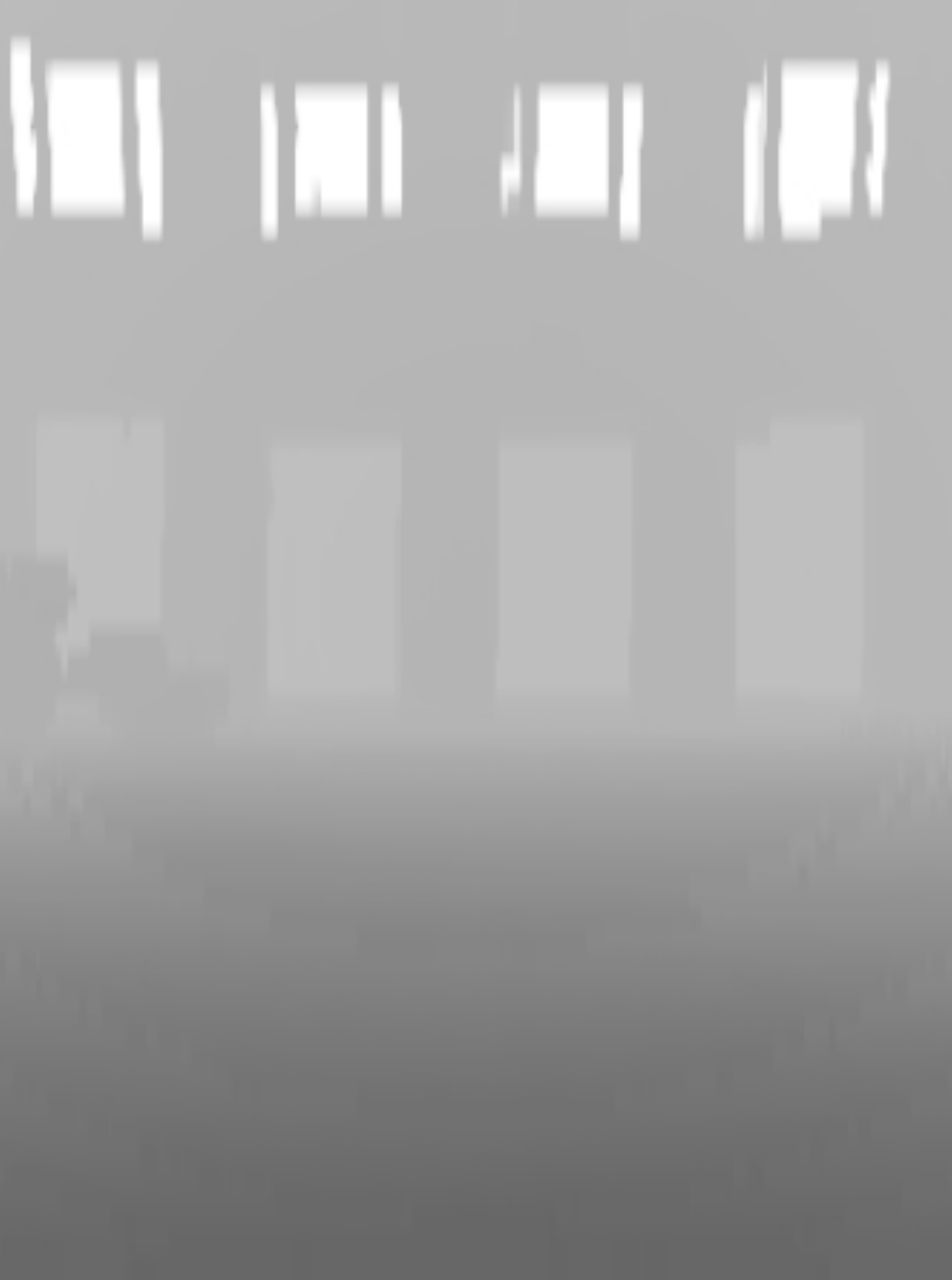}
\includegraphics[width=.24\linewidth]{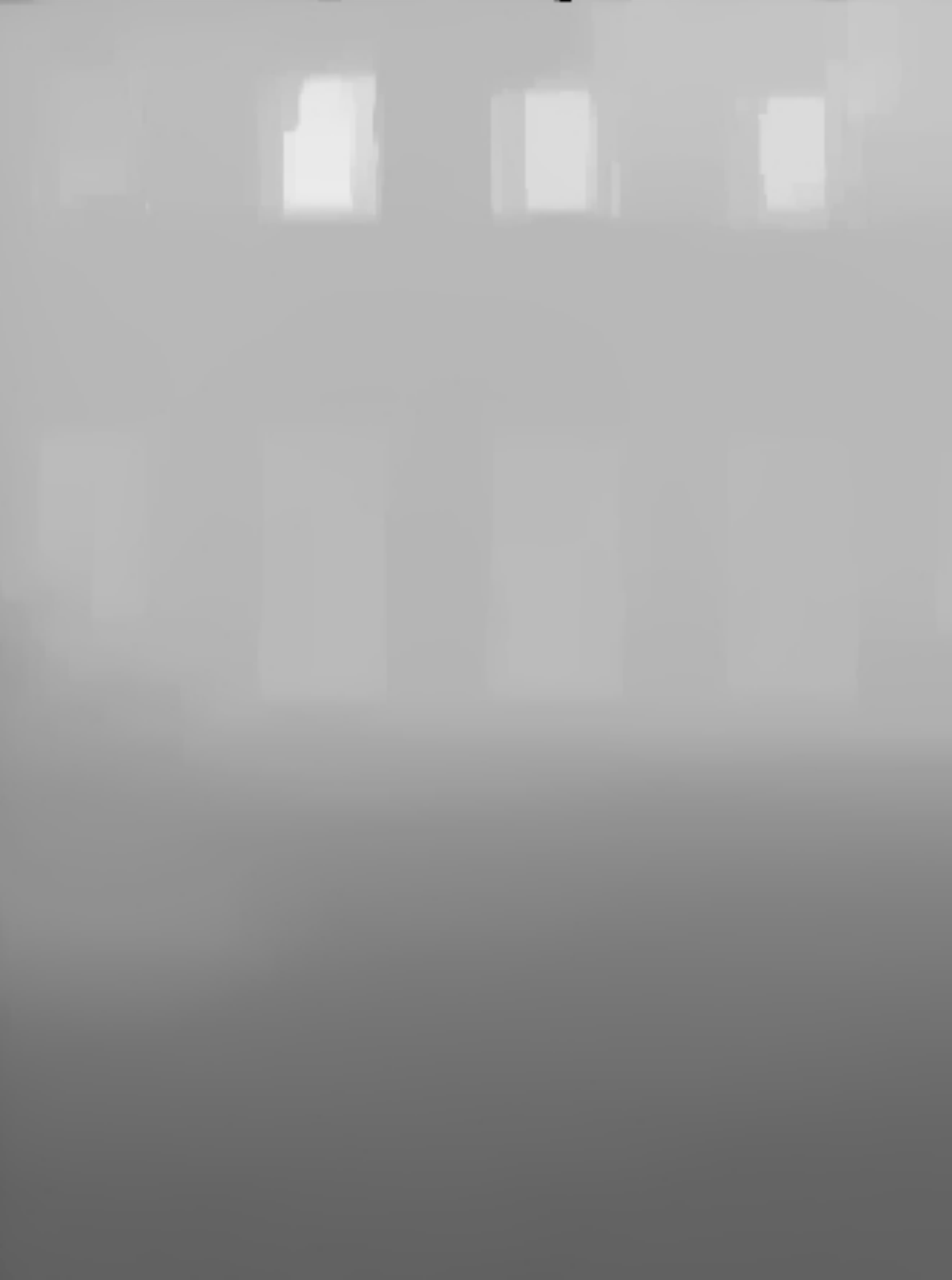}
\includegraphics[height=.3225\linewidth]{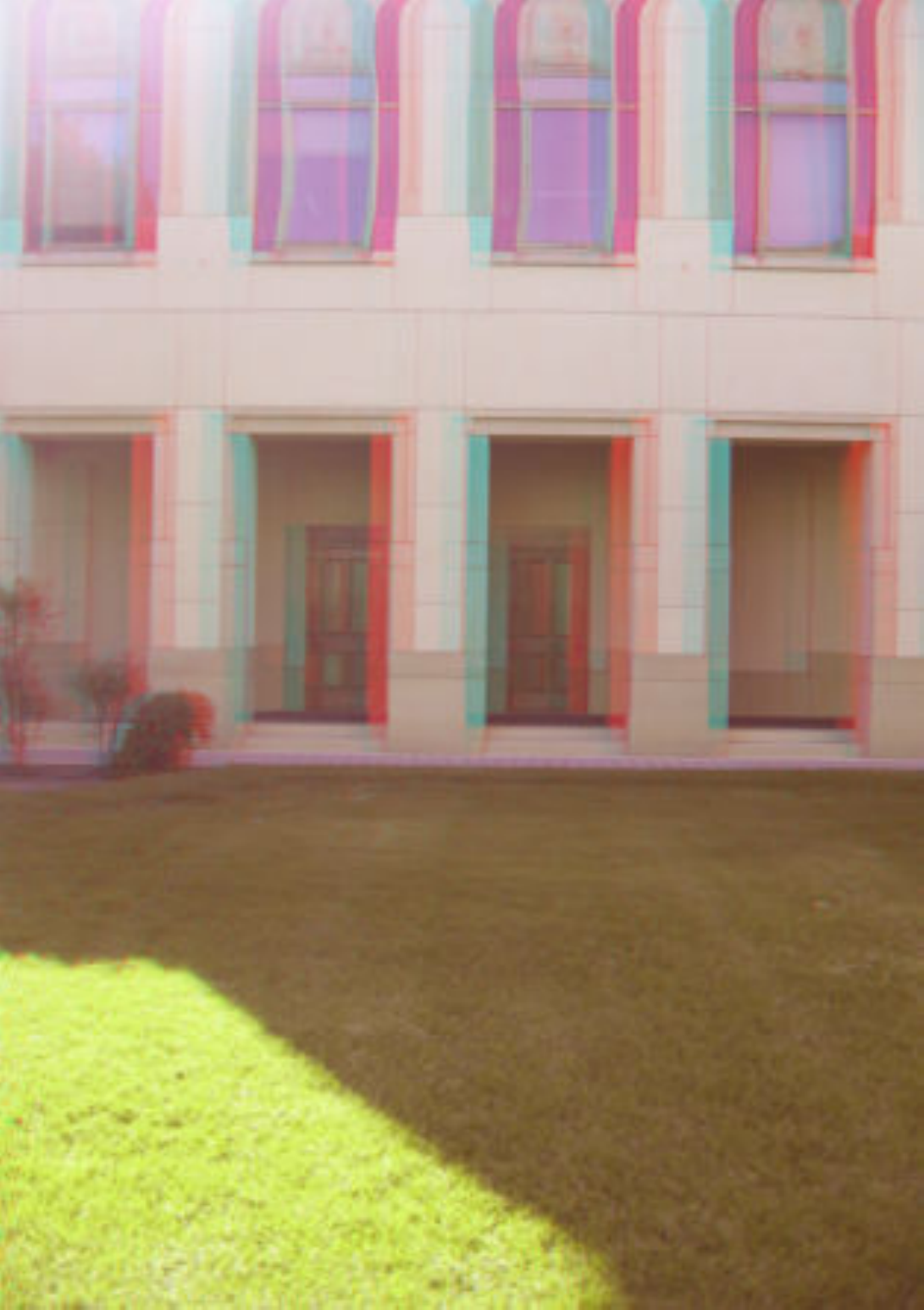}
\end{minipage}
\hfill
%\rule[-9mm]{.15mm}{.165\linewidth}
\hfill
\begin{minipage}[b]{0.49\linewidth}
\includegraphics[width=.235\linewidth]{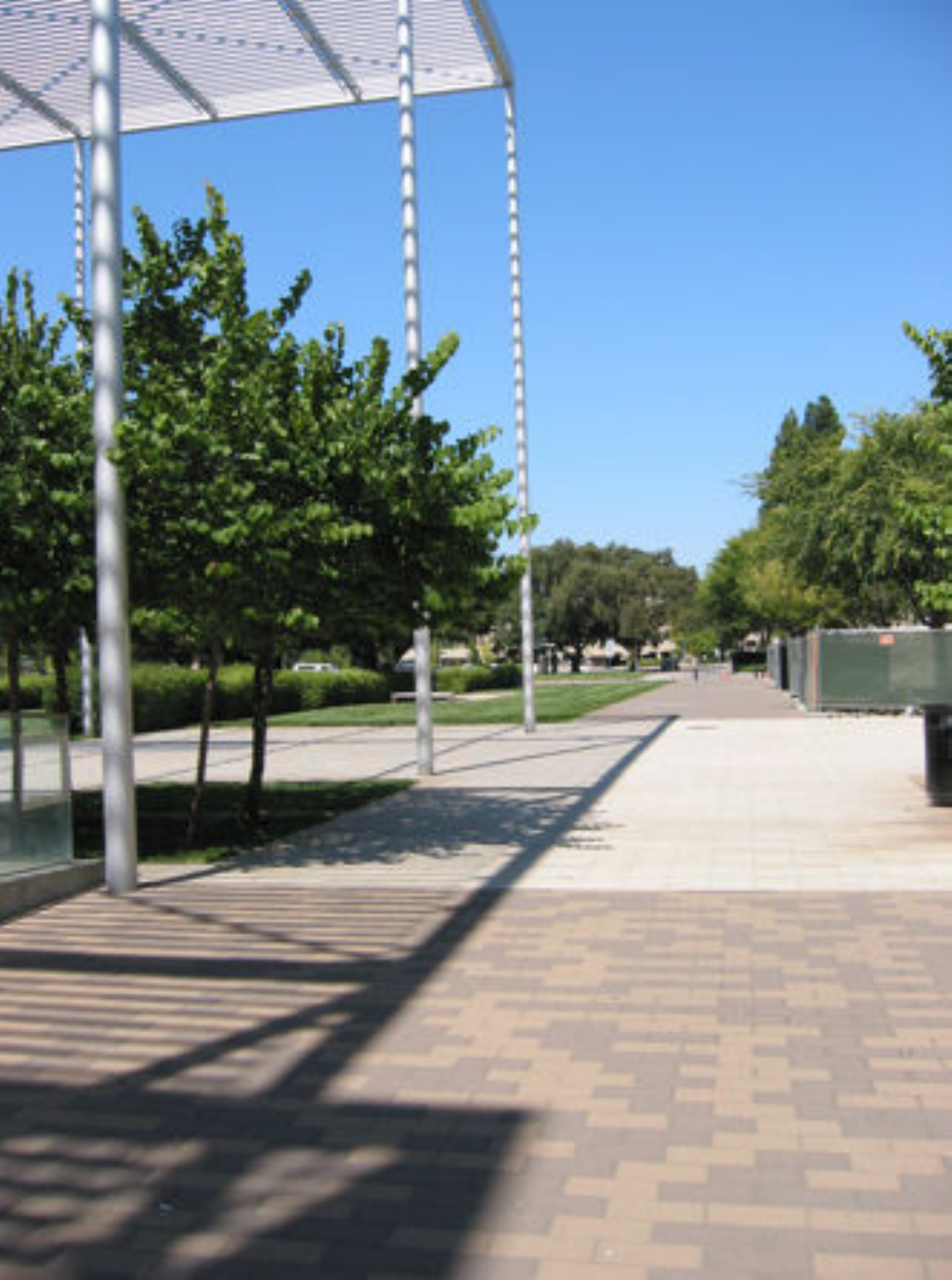}
\includegraphics[width=.235\linewidth]{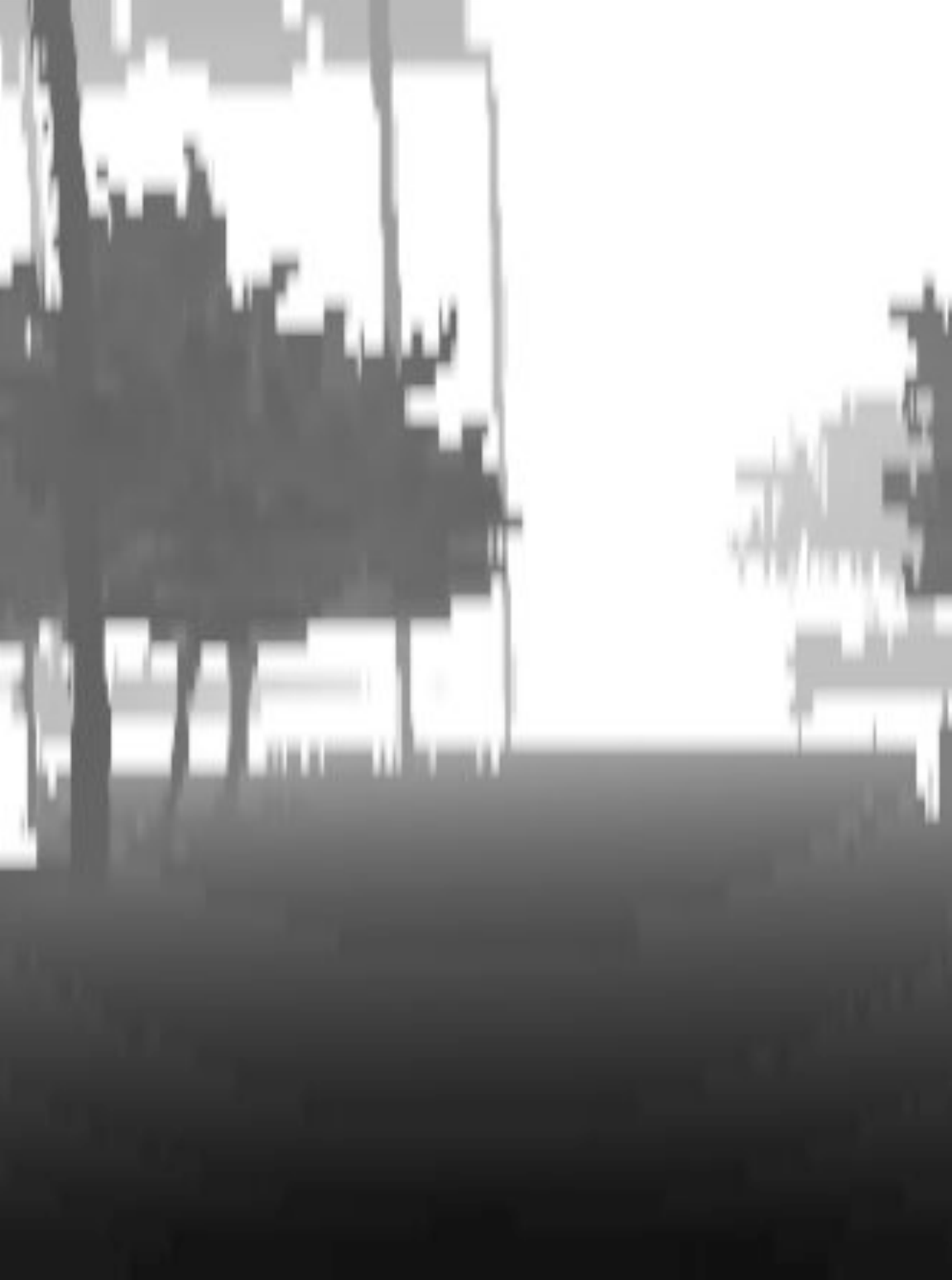}
\includegraphics[width=.235\linewidth]{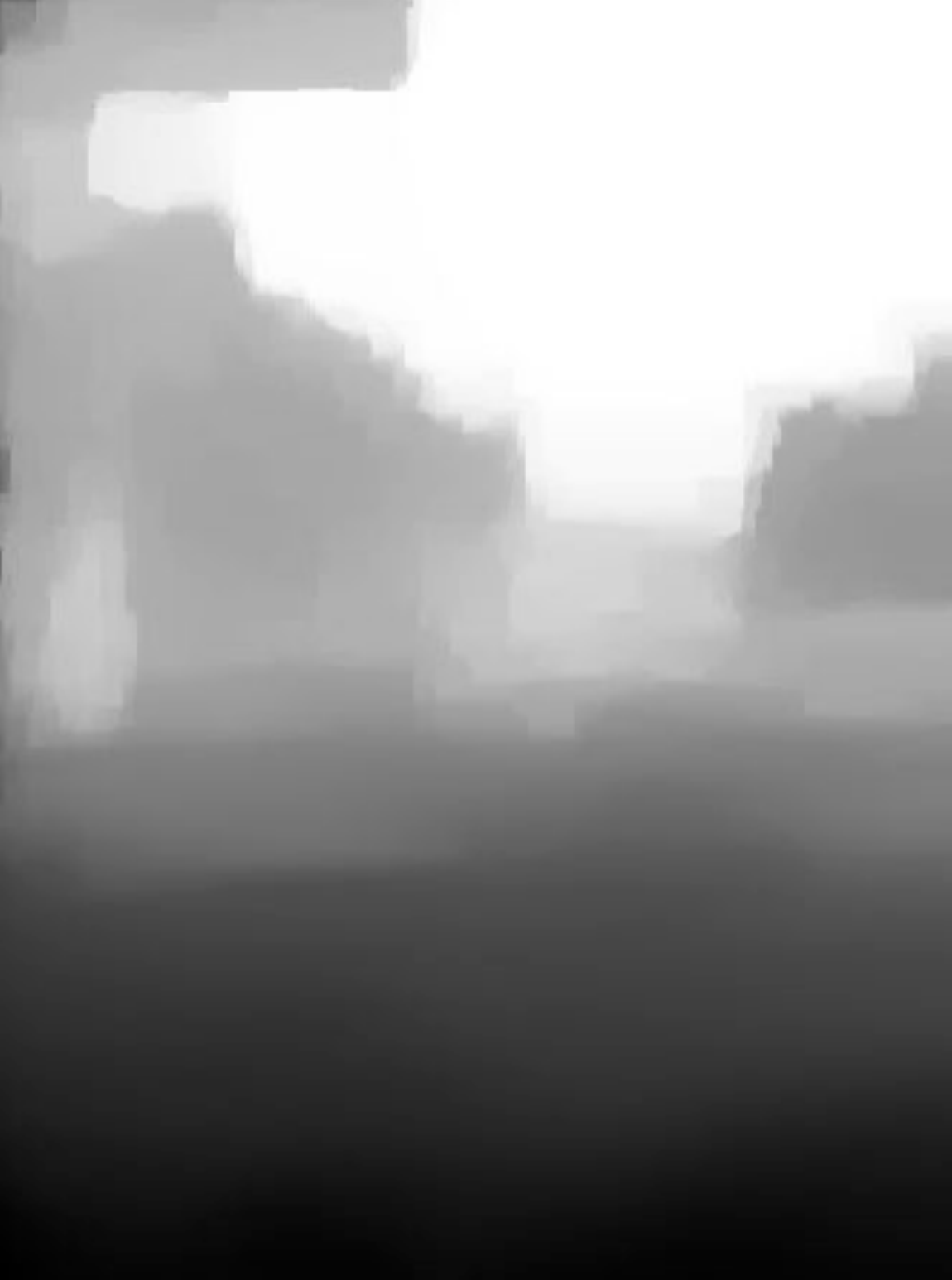}
\includegraphics[height=.32\linewidth]{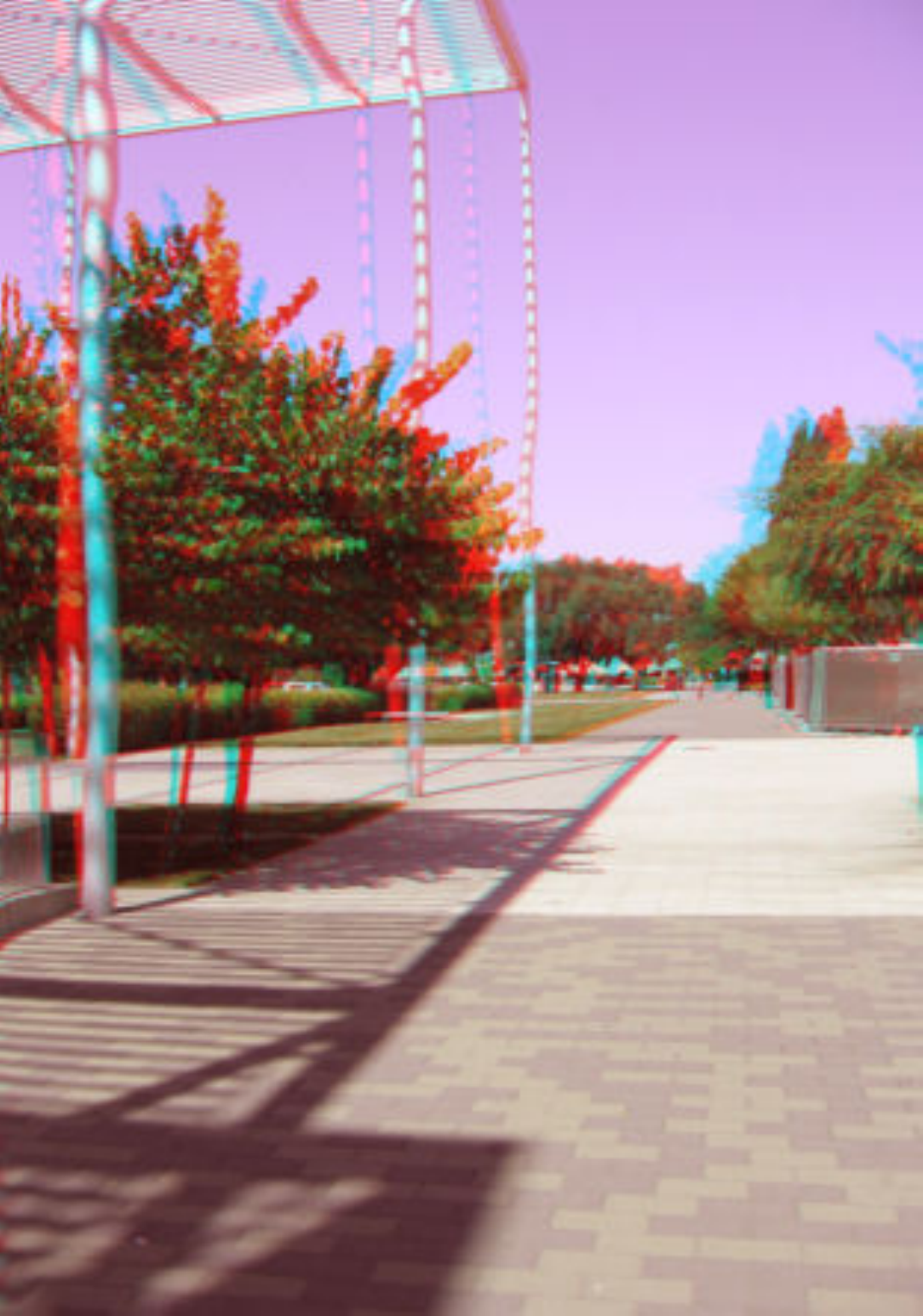}
\end{minipage}
\end{center}
\vspace{-3mm}
\caption{Single image results obtained on test images from the Make3D dataset. Each result contains the following four images (from left to right): original photograph,  ground truth depth from the dataset, our inferred depth, and our synthesized anaglyph \protect\includegraphics[height=5pt]{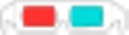} image. The depth images are shown in log scale. Darker pixels indicate nearby objects (black is roughly 1m away) and lighter pixels indicate objects farther away (white is roughly 80m away). Each pair of ground truth and inferred depths are displayed at the same scale.
}
\label{fig:outdoor_results}
\end{figure*}

Our method is as good as or better than the state-of-the-art for each metric. Note that previously no method achieved state-of-the-art results in more than one metric.  We show several examples in Fig~\ref{fig:outdoor_results}.
%In some cases, our method even produces better depth estimates than the ground truth (with low resolution and sensor errors). 
Thin structures (e.g., trees and pillars) are usually recovered well; however, fine structures are occasionally missed due to spatial regularization (such as the poles in the bottom-right image of Fig~\ref{fig:outdoor_results}).

%%%%%%%%%%%%%%%%%
\begin{table}[t]
\centerline{
\normalsize
\begin{tabular}{| l || c | c | c |} \hline
{\bf Method} & {\bf rel }& {\bf log$_{10}$} & {\bf RMS} \\ \hline \hline
 %Baseline & .698 & .334 & - \\ \hline
 Depth MRF~\cite{Saxena05learningdepth} & 0.530 & 0.198 & 16.7 \\ \hline
 Make3D~\cite{Saxena:09} & 0.370 & 0.187 & - \\ \hline
 Feedback Cascades~\cite{Li:2010} & - & - & 15.2 \\ \hline
 Semantic Labels~\cite{Liu:cvpr10} & 0.375 & {\bf 0.148} & - \\ \hline
 Depth Transfer (ours) & {\bf 0.361} & {\bf 0.148} & {\bf 15.1} \\ \hline
 \end{tabular}
 }
\vspace{2mm}
\caption{Comparison of depth estimation errors on the Make3D range image dataset. Using our single image technique, our method achieves state of the art results in each metric ({\bf rel} is relative error,  {\bf RMS} is root mean squared error; details in text).}
 \label{tab:make3d_results}
 \end{table}
%%%%%%%%%%%%%%%%%

%%%%%%%%%%%%%%%%%
%\begin{figure}
%\begin{center}
%\begin{minipage}{.9\linewidth}
%\includegraphics[width=.49\linewidth]{fig/kinect_plots/IndoorError_rel.pdf} \hfill
%\includegraphics[width=.49\linewidth]{fig/kinect_plots/IndoorError_log10.pdf}\\
%\includegraphics[width=.49\linewidth]{fig/kinect_plots/IndoorError_rmse.pdf} \hfill
%\includegraphics[width=.49\linewidth]{fig/kinect_plots/IndoorError_psnr.pdf}
%\end{minipage}
%\end{center}
%\caption{Box plots indicating the median, upper quartiles and lower quartiles for each metric. \todo{Need to add a legend and make Y axis numbers much bigger} \todo{Maybe this fig isn't even necessary?}
%}
% \label{fig:kinect_plots}
%\end{figure}
%%%%%%%%%%%%%%%%%

%%%%%%%%%%%%%%%%%
%%%%%%%%%%%%%%%%%

%%%%%%%%%%%%%%%%%
\begin{figure*}[t]
\begin{center}
\begin{minipage}[b]{0.49\linewidth}
\hspace{4mm} Input \hspace{6mm} True depth \hspace{5mm} Inferred${}^{\star}$ \hspace{4mm}  Inferred${}^{\star\star}$ \\
\includegraphics[width=.24\linewidth]{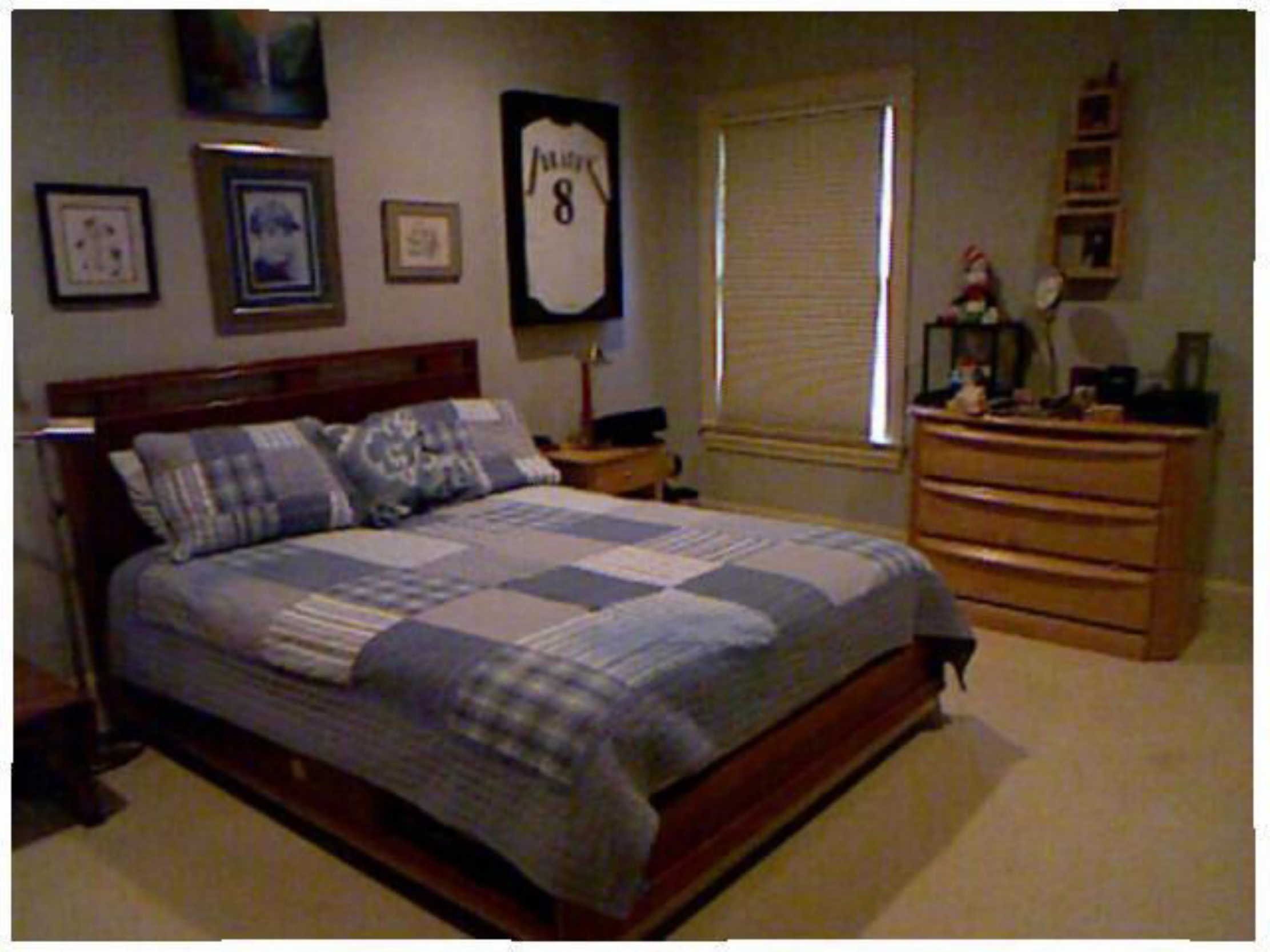}
\includegraphics[width=.24\linewidth]{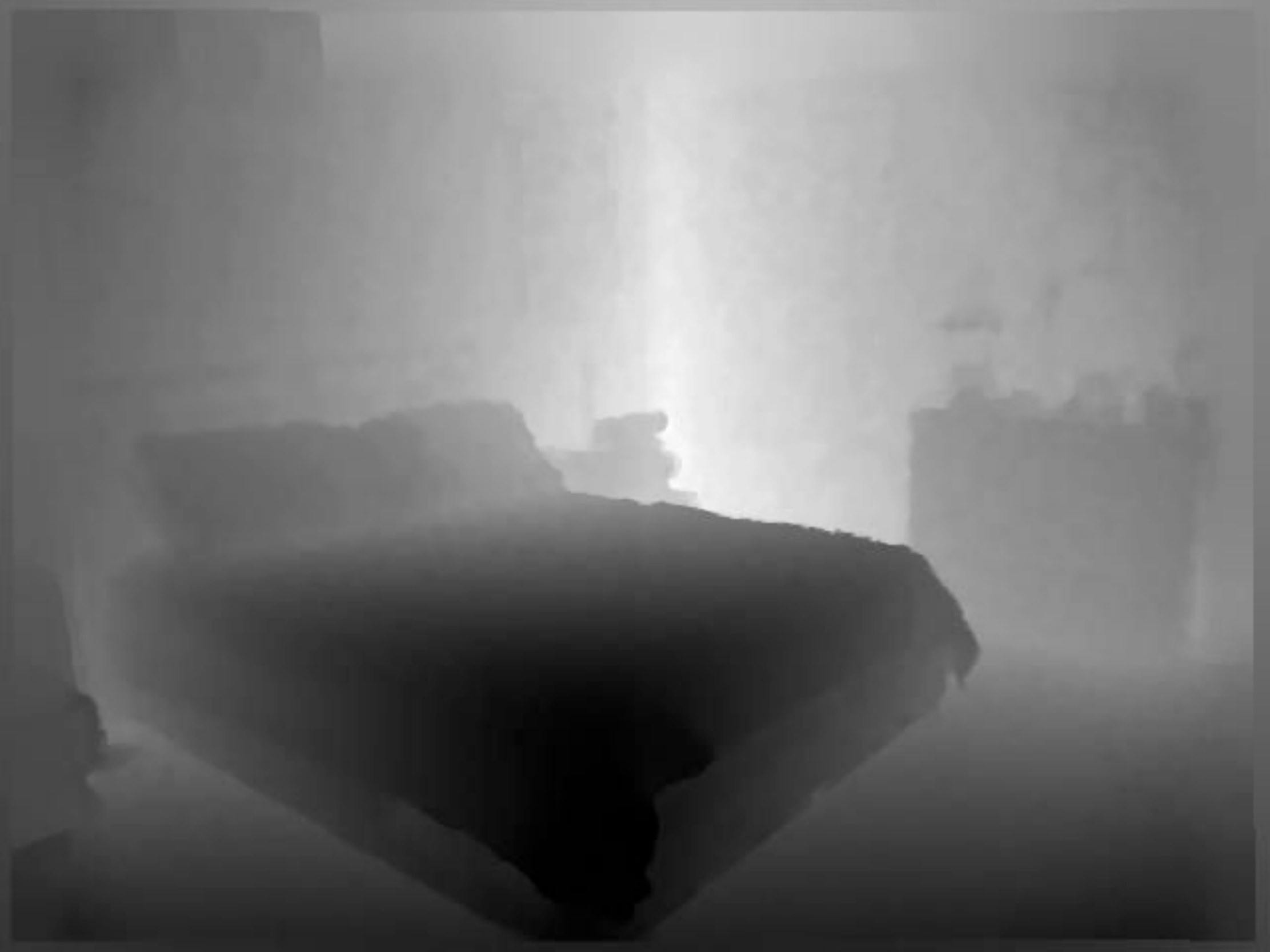}
\includegraphics[width=.24\linewidth]{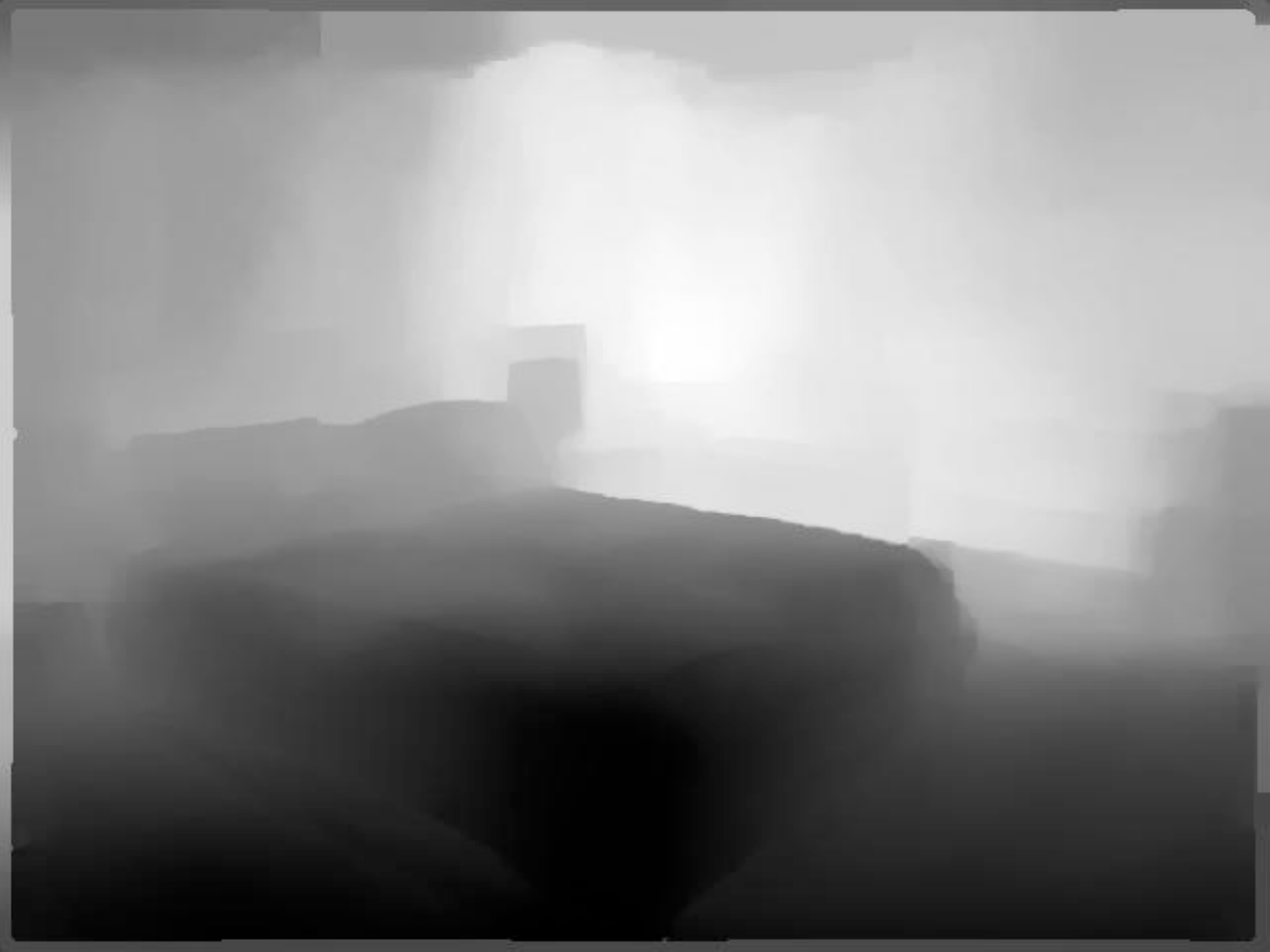}
\includegraphics[width=.24\linewidth]{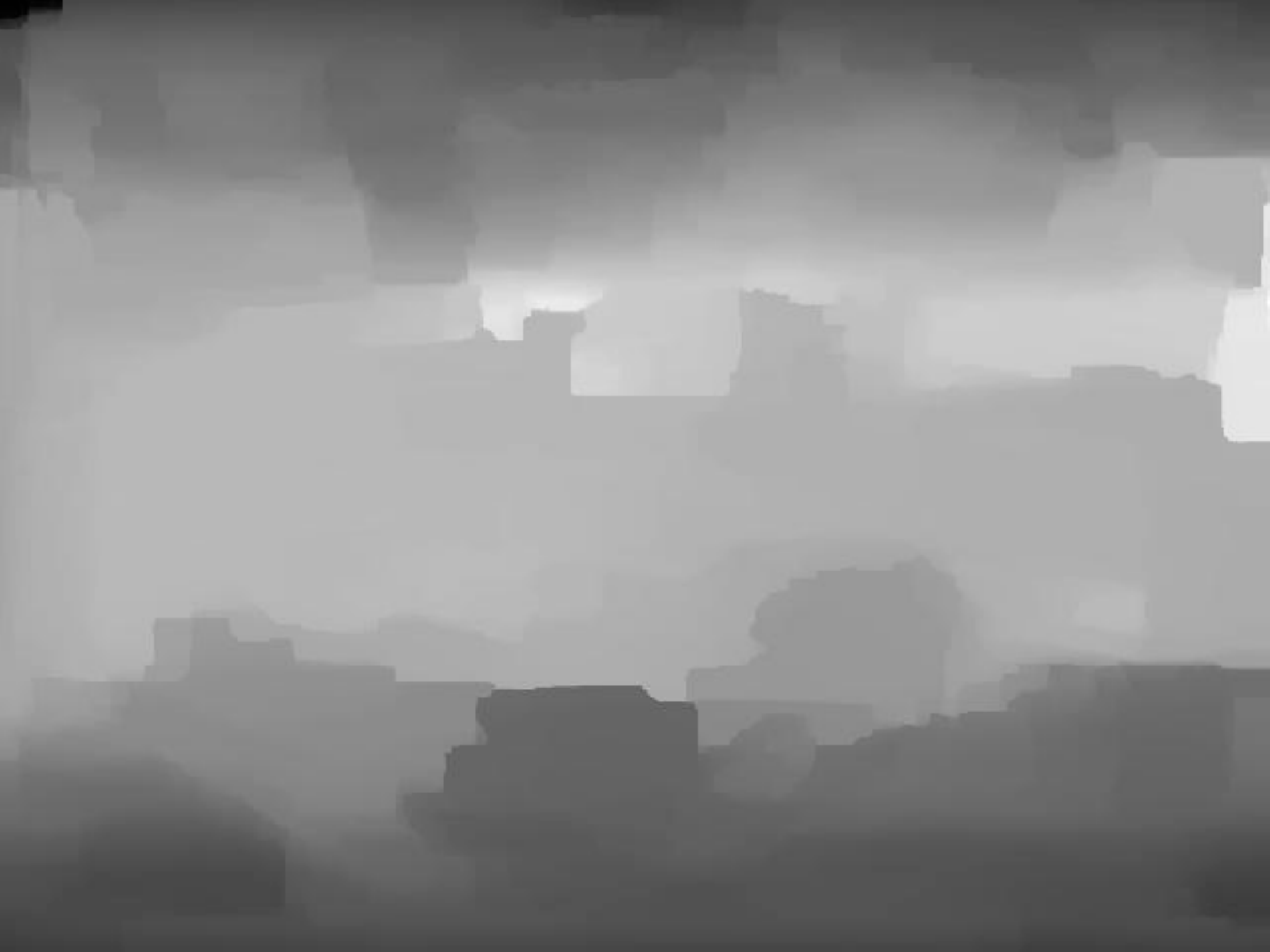}
\end{minipage}
\hfill
\begin{minipage}[b]{0.49\linewidth}
\hspace{4mm} Input \hspace{6mm} True depth \hspace{5mm} Inferred${}^{\star}$ \hspace{4mm}  Inferred${}^{\star\star}$ \\
\includegraphics[width=.24\linewidth]{fig/results/NYU/0141/im.pdf}
\includegraphics[width=.24\linewidth]{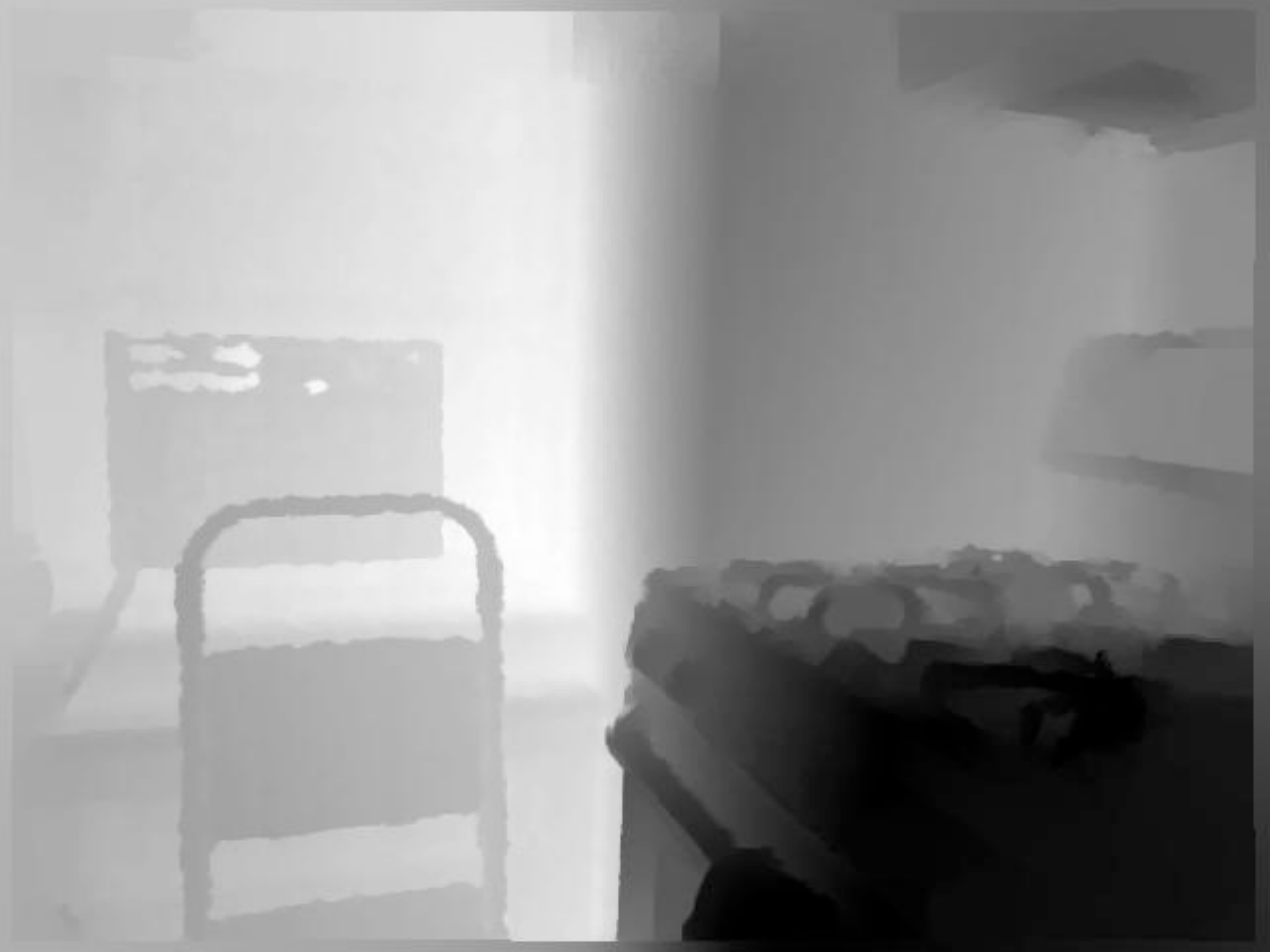}
\includegraphics[width=.24\linewidth]{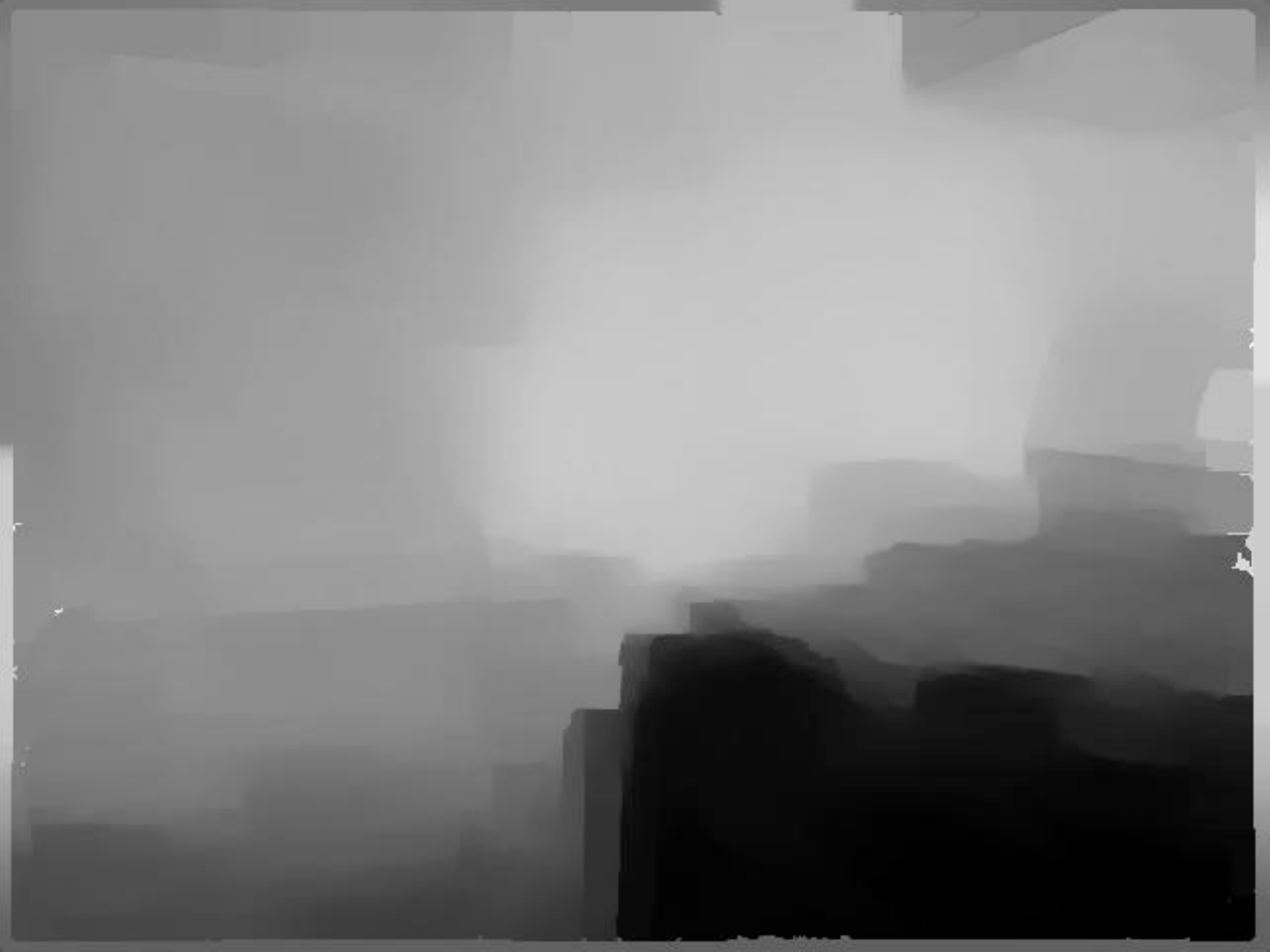}
\includegraphics[width=.24\linewidth]{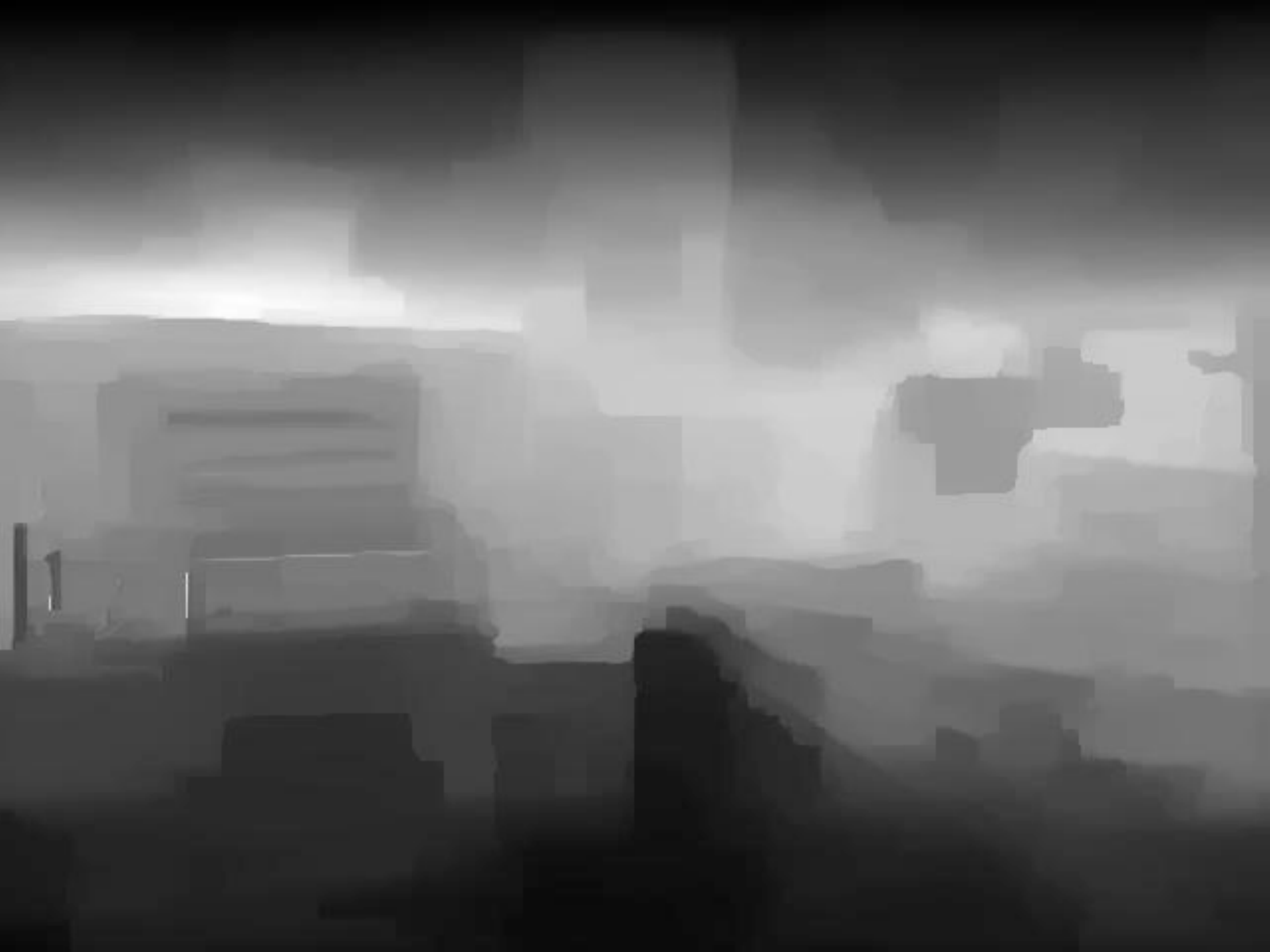}
\end{minipage}\\
\vspace{2mm}
\begin{minipage}[b]{0.49\linewidth}
\includegraphics[width=.24\linewidth]{fig/results/NYU/0891/im.pdf}
\includegraphics[width=.24\linewidth]{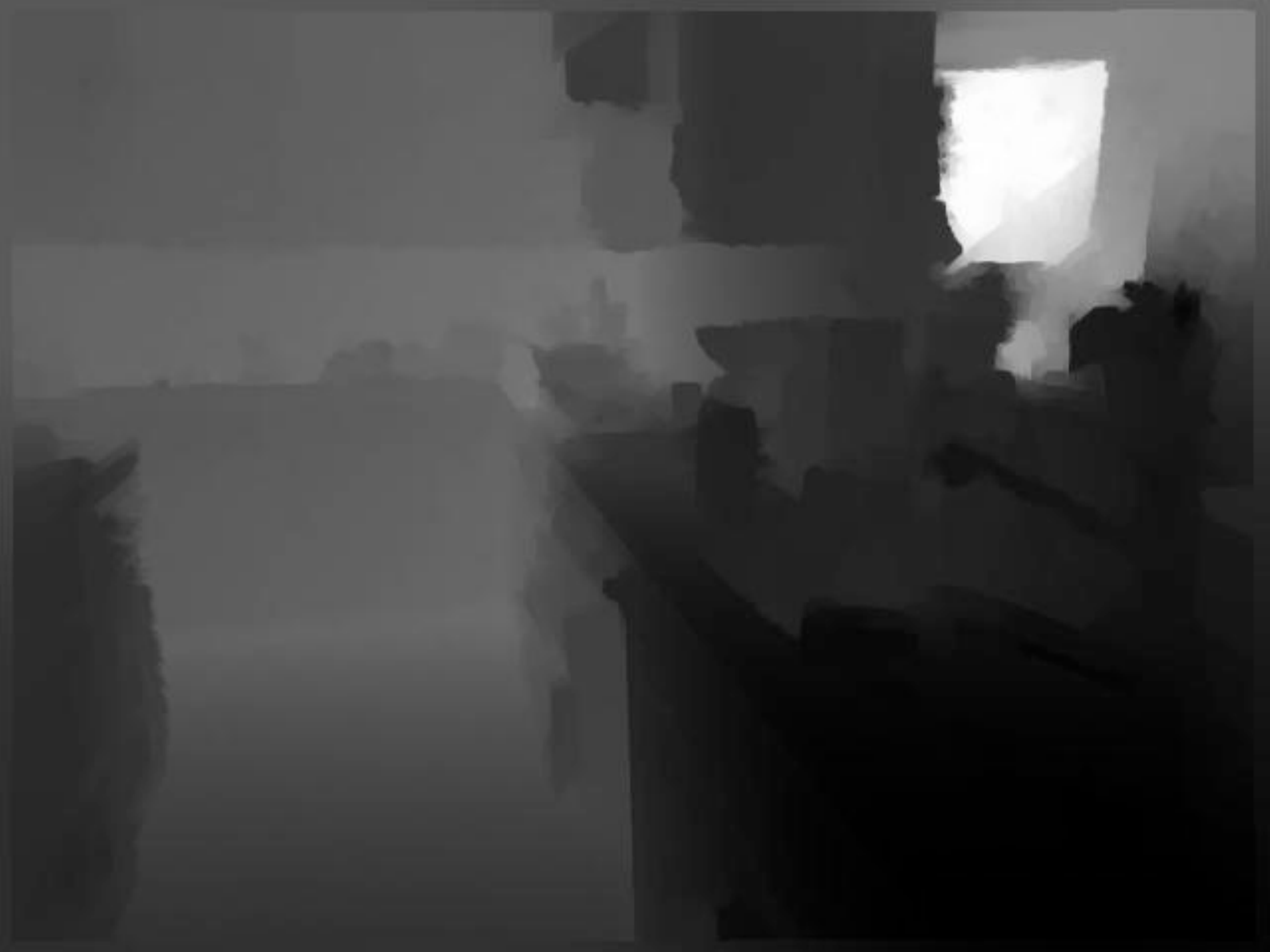}
\includegraphics[width=.24\linewidth]{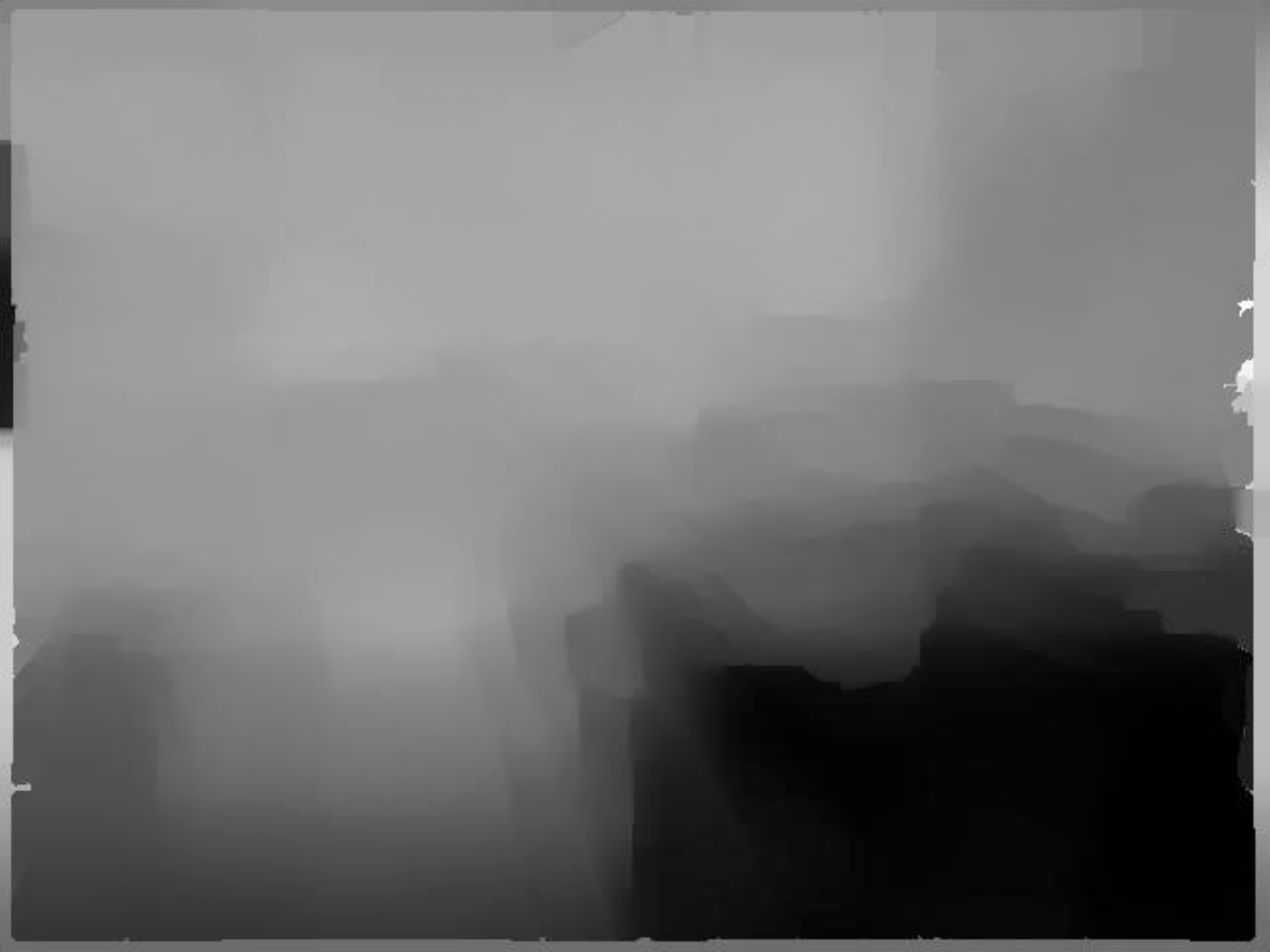}
\includegraphics[width=.24\linewidth]{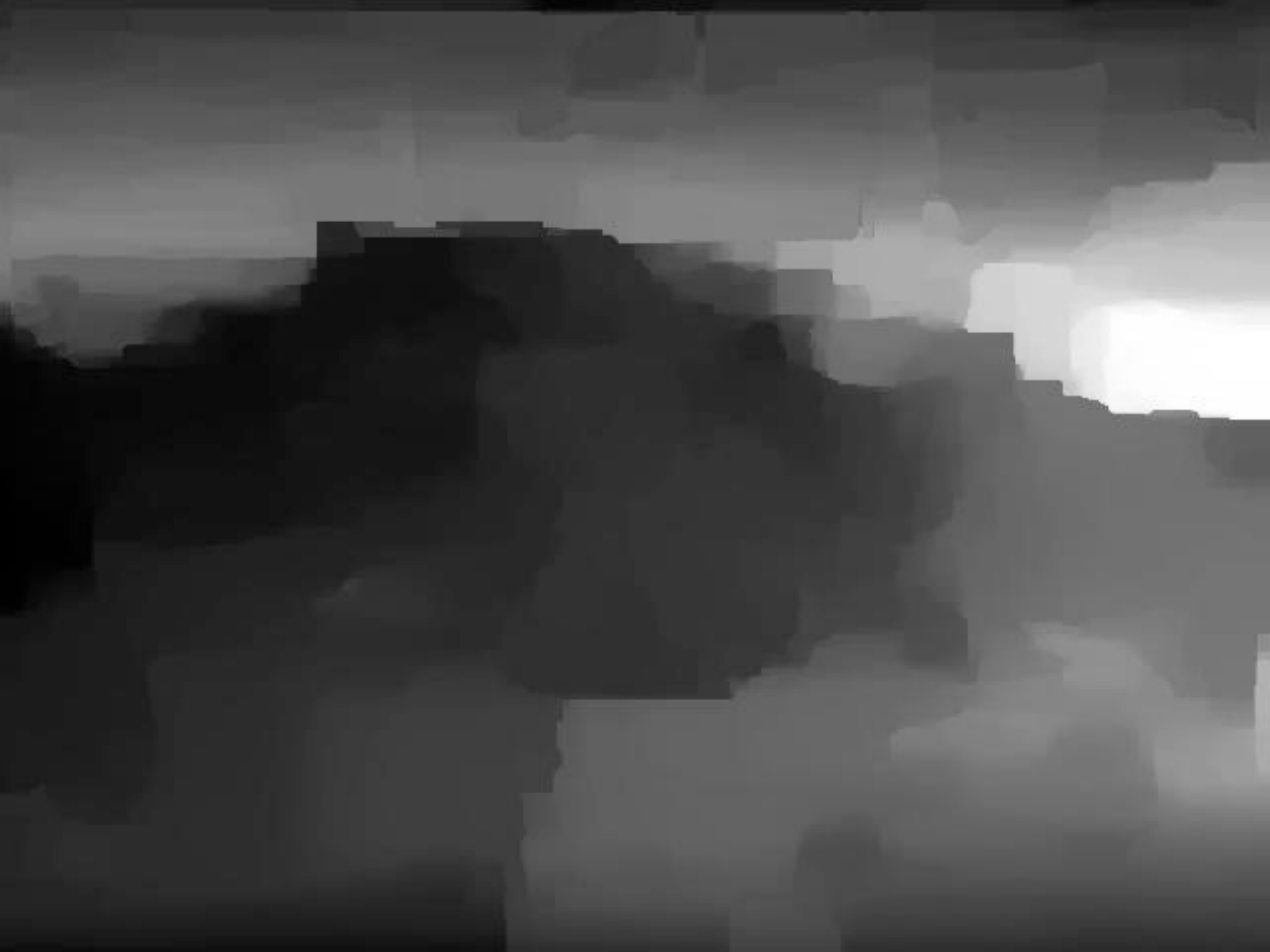}
\end{minipage}
\hfill
\begin{minipage}[b]{0.49\linewidth}
\includegraphics[width=.24\linewidth]{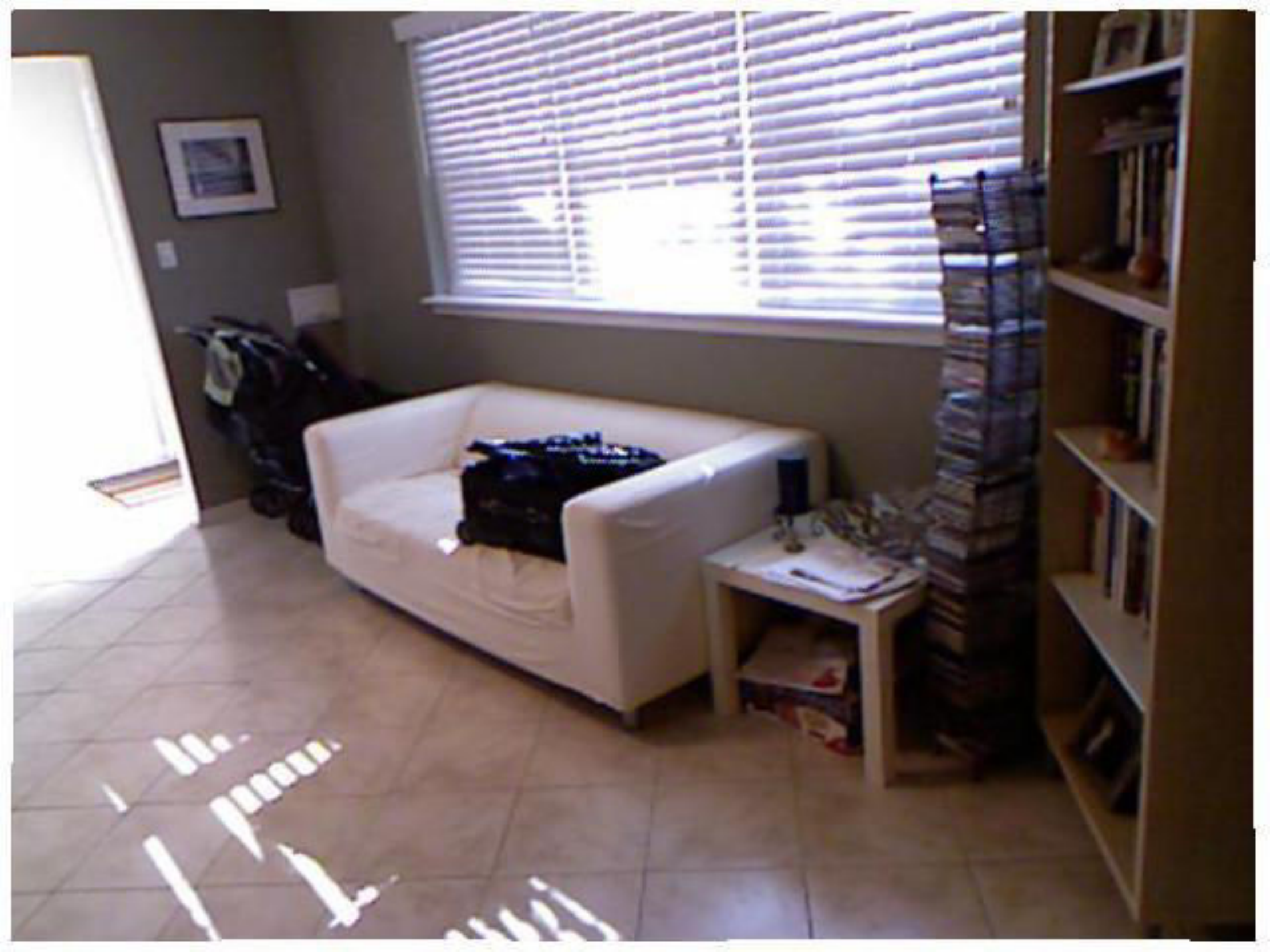}
\includegraphics[width=.24\linewidth]{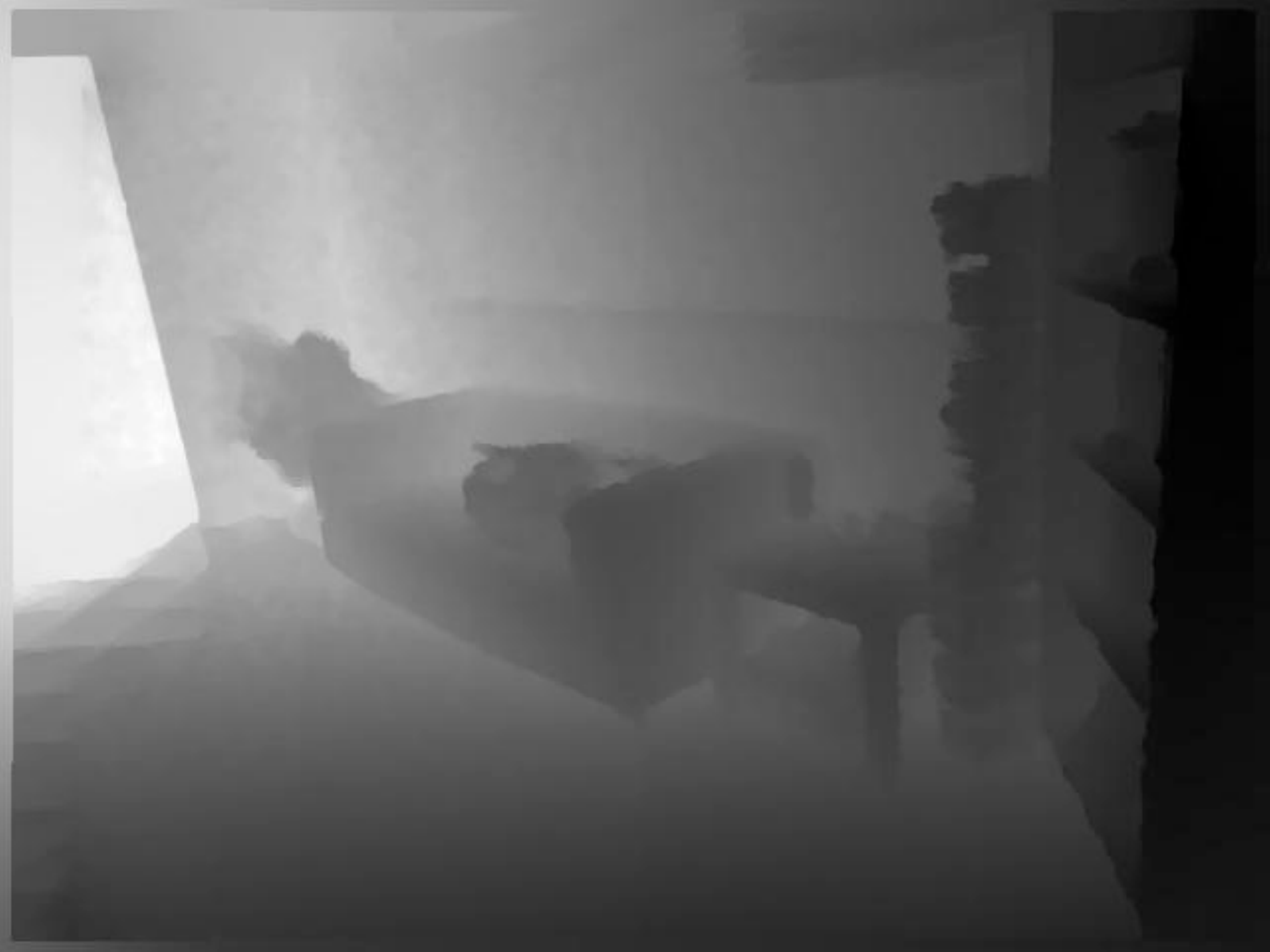}
\includegraphics[width=.24\linewidth]{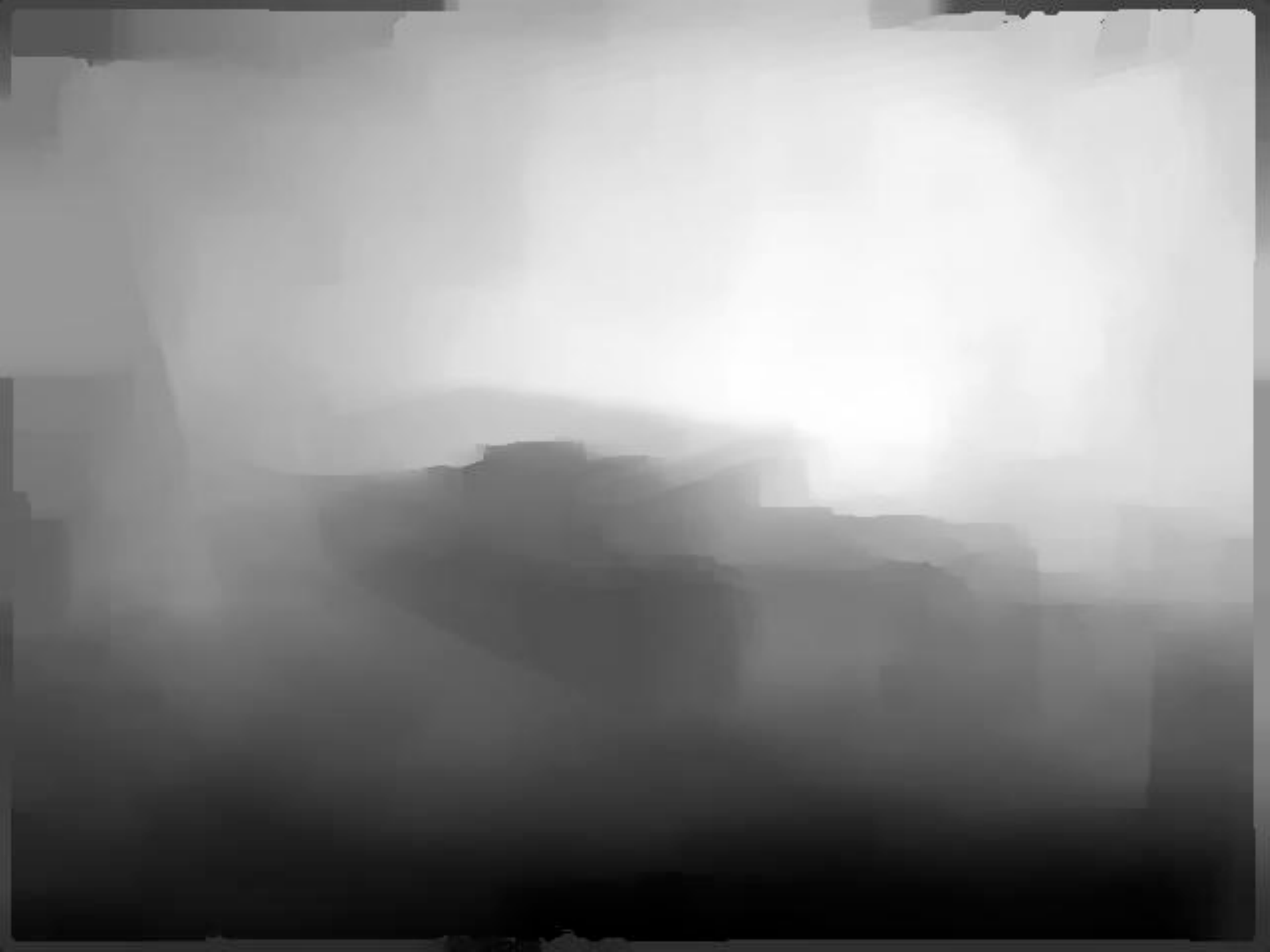}
\includegraphics[width=.24\linewidth]{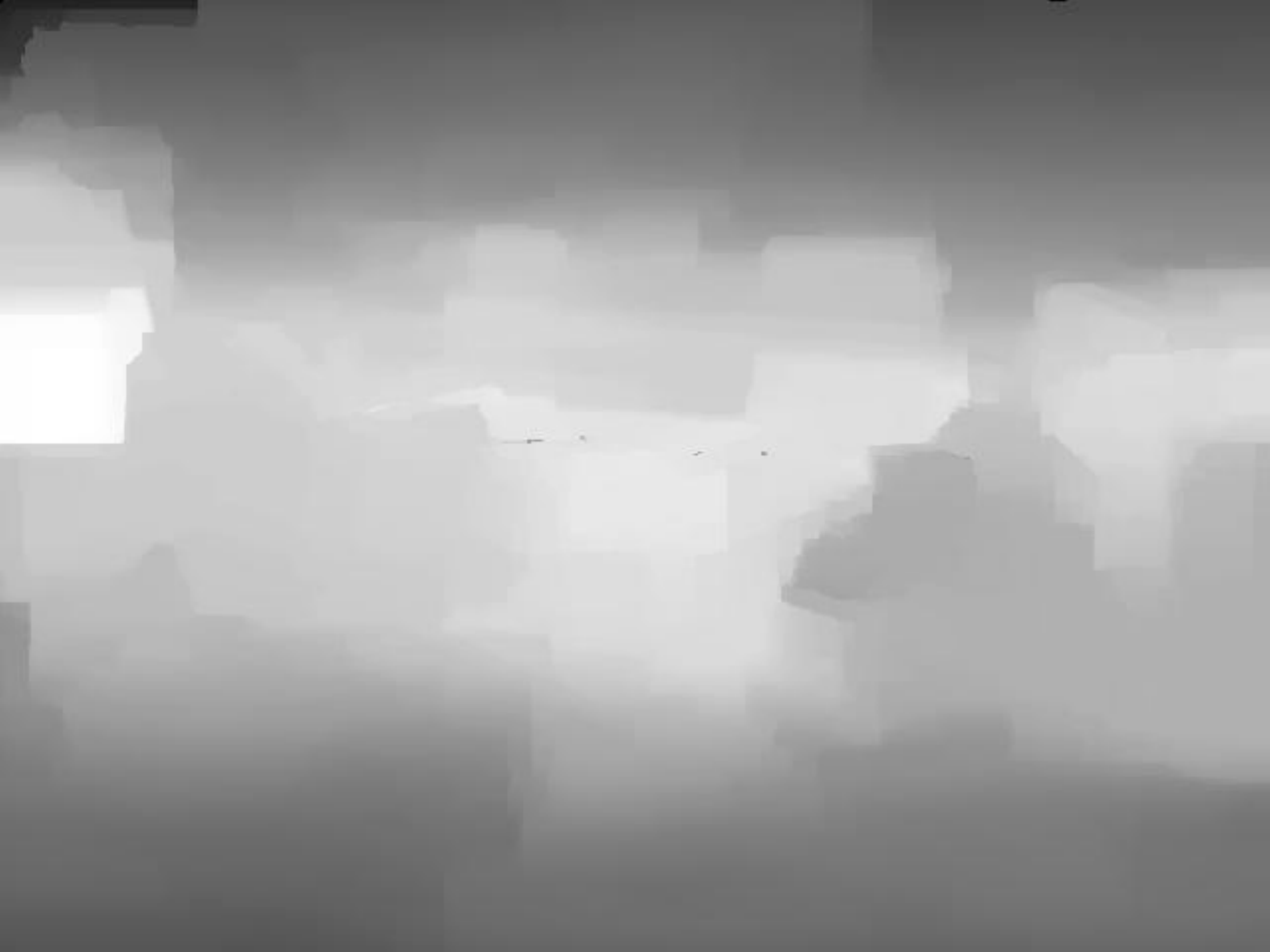}
\end{minipage}
\end{center}
\vspace{-3mm}
\caption{Single image results obtained on the NYU Depth Dataset. Each result contains the following four images (from left to right): original photograph,  ground truth depth from the dataset, our inferred depth trained on the NYU depth dataset (${}^{\star}$), holding out the particular image), and our inferred depth trained on our MSR-V3D dataset (${}^{\star\star}$). The top left result is in the fifth percentile (in terms of log${}_{10}$ error), the top right is in the bottom fifth percentile, and both bottom results are near the median. Notice how the NYU-trained results are much better; this is likely due to the high variation of indoor images (MSR-V3D  images appear very dissimilar to the NYU ones), and is also reflected in the quantitive results (Tab~\ref{tab:make3d_results}). Holes in the true depth maps have been filled using the algorithm in~\cite{Silberman:ECCV12}, and each pair of ground truth and inferred depths are displayed at the same scale.
}
\label{fig:nyu_results}
\vspace{-5mm}
\end{figure*}
%%%%%%%%%%%%%%%%%

\newtext{
\subsubsection{NYU Depth Dataset}
We report additional results for the NYU Depth Dataset~\cite{Silberman:ECCV12}, which consists of 1449 indoor RGBD images captured with a Kinect. Holes from the Kinect are disregarded during training (candidate searching and warping), and are not included in our error analysis.

%%%%%%%%%%%%%%%%%
\begin{table} 
\
\vspace{-0mm}
\centerline{
\normalsize
\begin{tabular}{| l || c | c | c |} \hline
{\bf Method} & {\bf rel }& {\bf log$_{10}$} & {\bf RMS} \\ \hline \hline
Depth Transfer ${}^\text{(NYU)}$ & {\bf 0.350} & {\bf 0.131} & {\bf 1.2} \\ \hline
Depth Transfer ${}^\text{(MSR-V3D)}$ & 0.806 & 0.231 & 3.7 \\ \hline
Depth Fusion~\cite{Konrad:SPIE:12} & 0.368 & 0.135 & 1.3 \\ \hline
Depth Fusion (no warp)~\cite{Konrad:3DCINE:12} & 0.371 & 0.137 & 1.3 \\ \hline
NYU average depth${}^{\dagger }$ & 0.491  & 0.327 &  4.3 \\ \hline
NYU per-pixel average${}^{\dagger \dagger}$ & 0.561 & 0.164 & 20.1 \\ \hline
 \end{tabular}
 }
\caption{Quantitative evaluation of our method and the methods of Konrad et al. on the NYU Depth Dataset.  All methods are trained on the NYU dataset using a hold-one-out scheme, except for Depth Transfer ${}^\text{(MSR-V3D)}$ which is trained using our own dataset (containing significantly different indoor scenes). Each method uses seven candidate images ($K=7$). 
It is interesting that the per-pixel average (${}^{\dagger \dagger}$) performs worse than a single depth value (${}^{\dagger}$), evidencing that these metrics are likely not very perceptual.
\vspace{-12mm} %Only way to make this stay on the page!
}
\label{tab:nyu_results}
\end{table}
%%%%%%%%%%%%%%%%%

%%%%%%%%%%%%%%%%%
\begin{figure}
\vspace{5mm}
\includegraphics[width=0.23\linewidth]{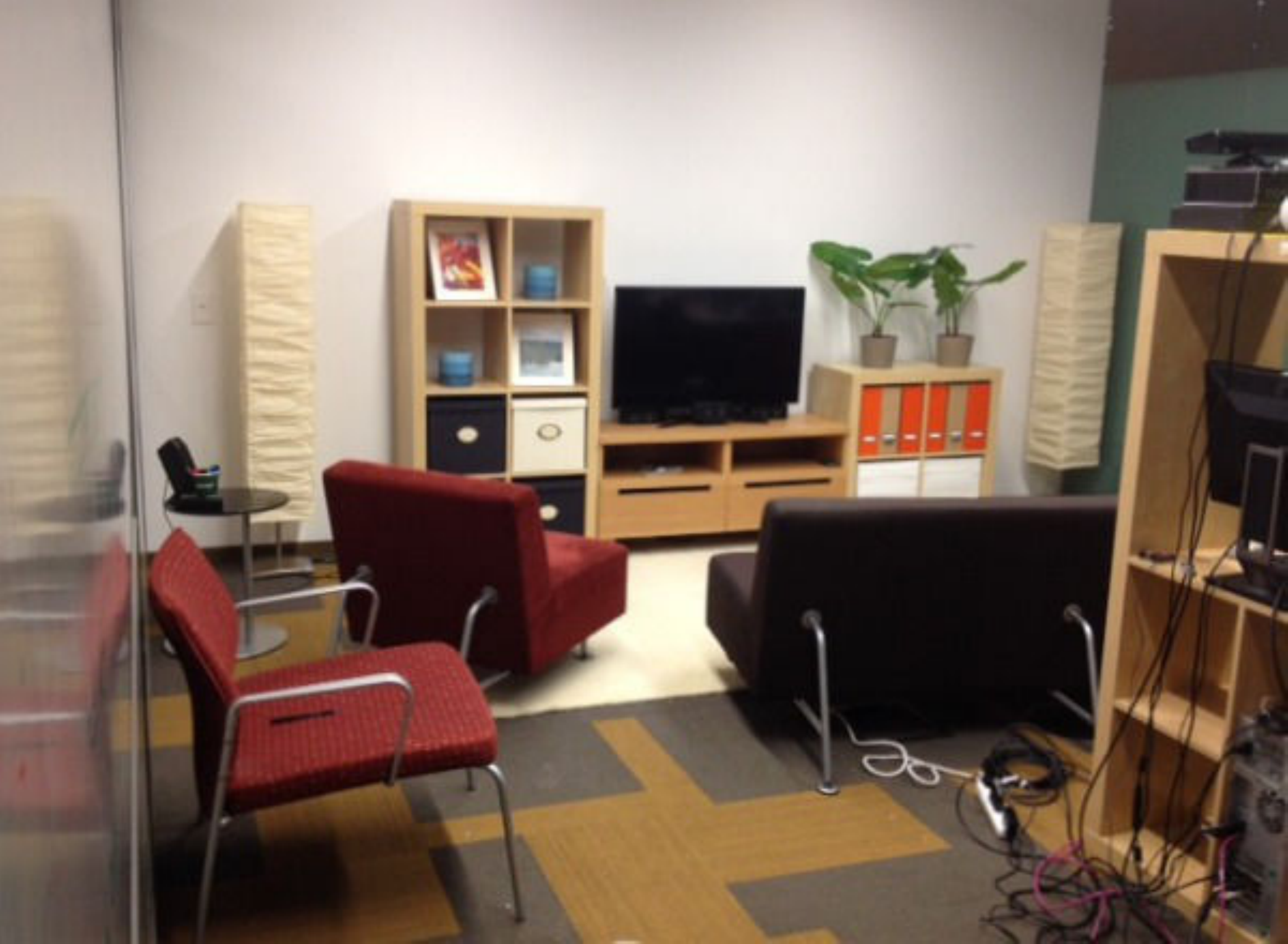}
\includegraphics[width=0.23\linewidth]{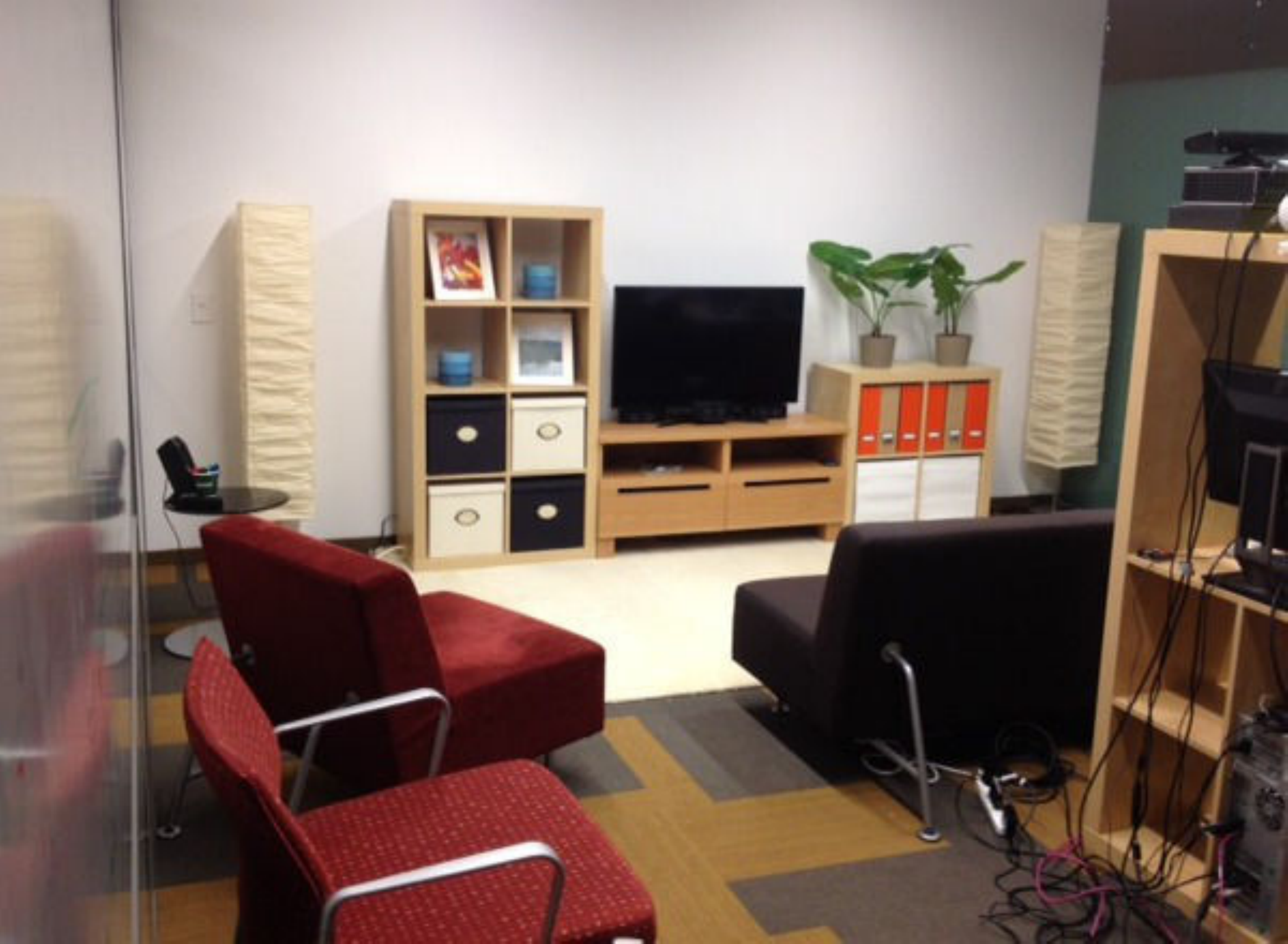}
\hfill
\includegraphics[width=0.23\linewidth,trim=0 0 0 10, clip=true]{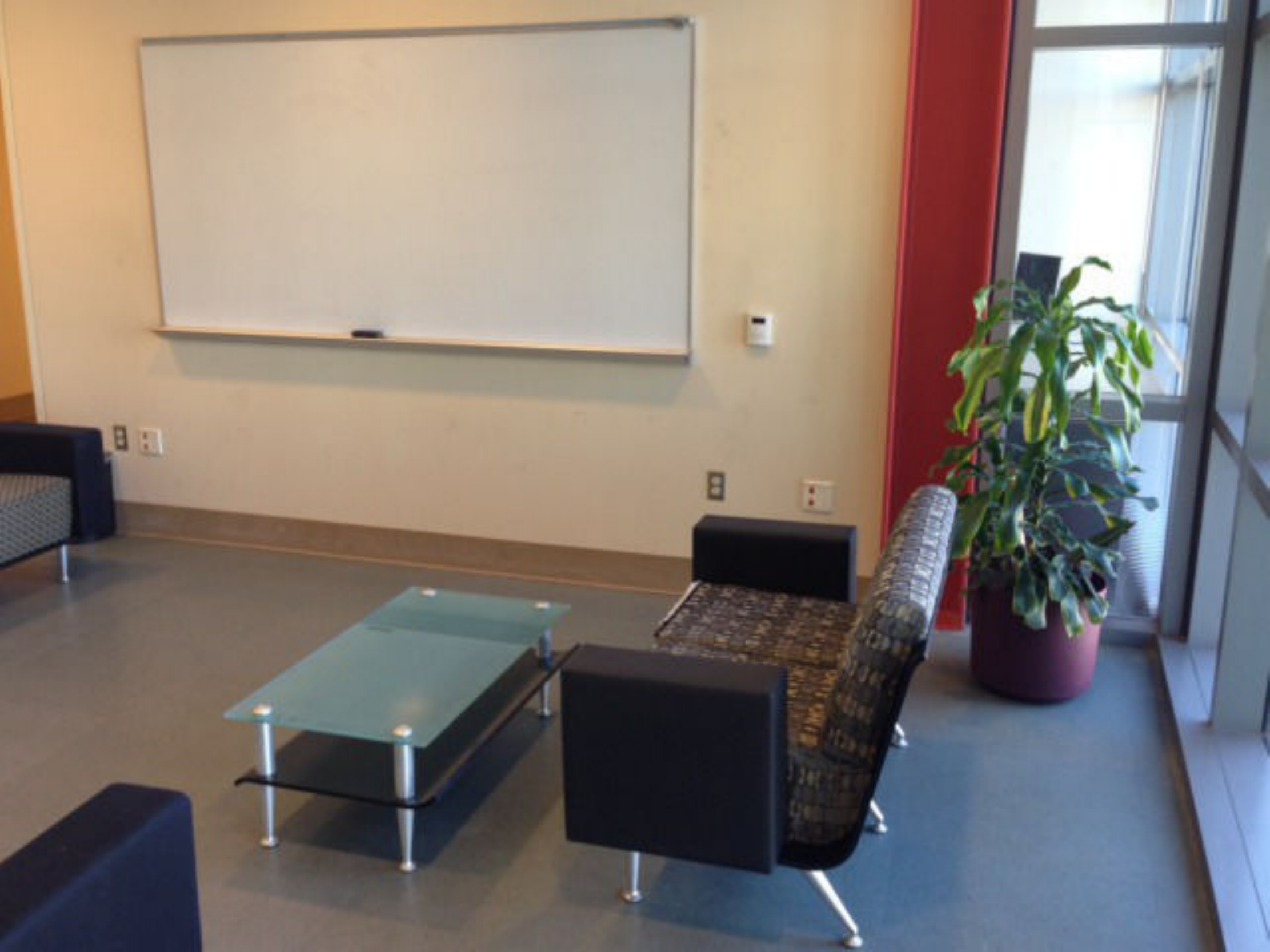}
\includegraphics[width=0.23\linewidth,trim=0 0 0 10, clip=true]{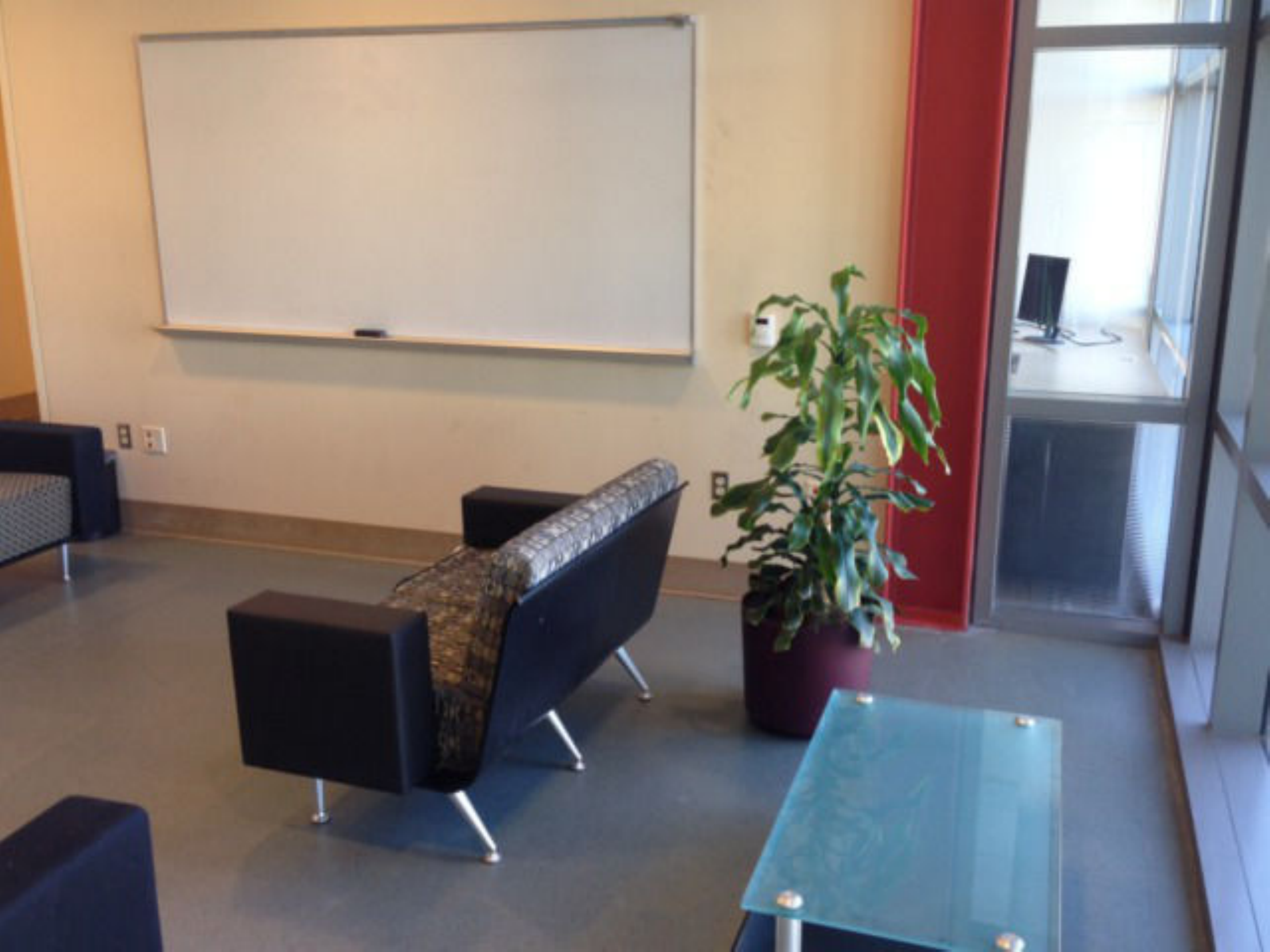}
\\
\includegraphics[width=0.23\linewidth]{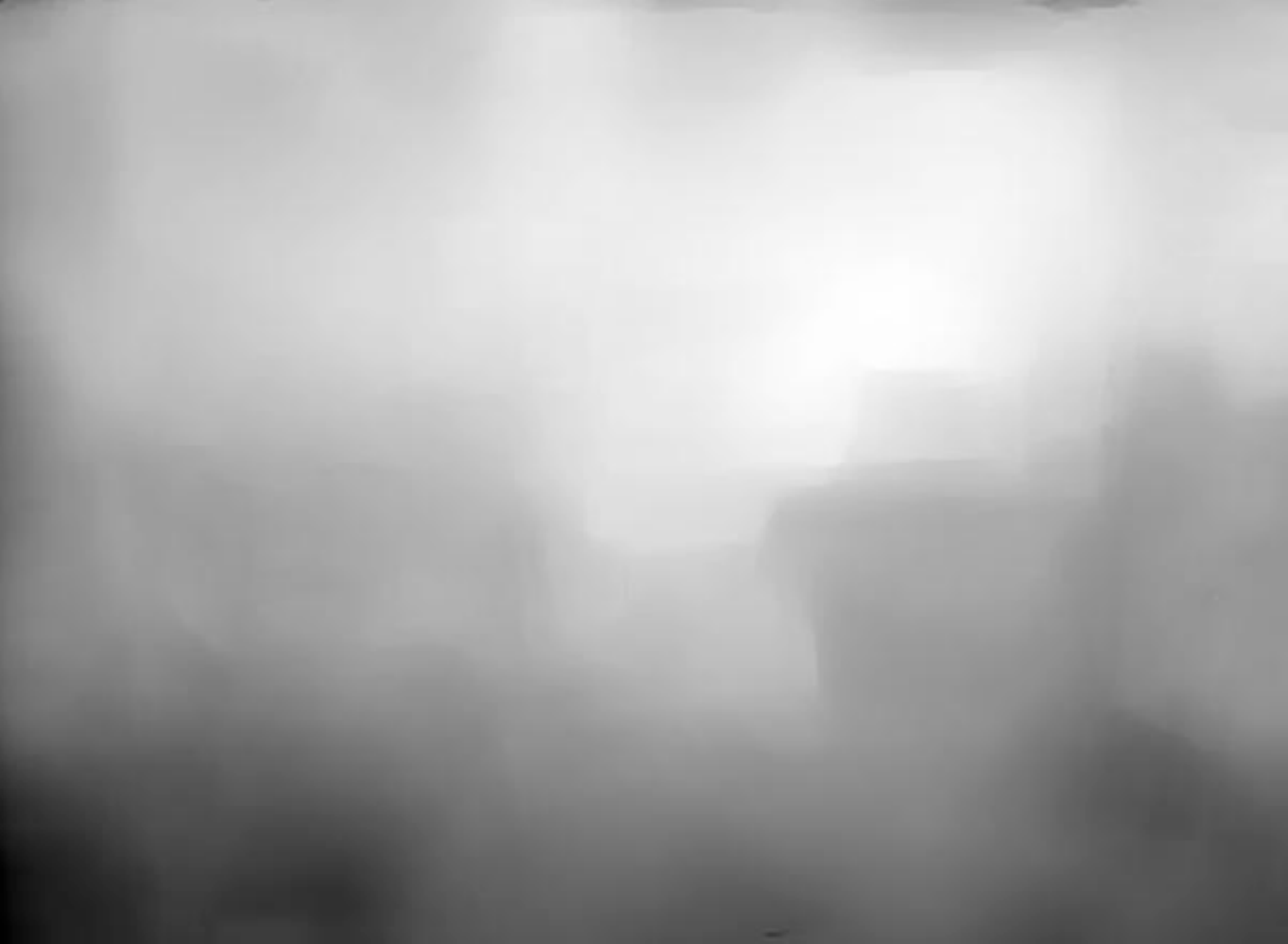}
\includegraphics[width=0.23\linewidth]{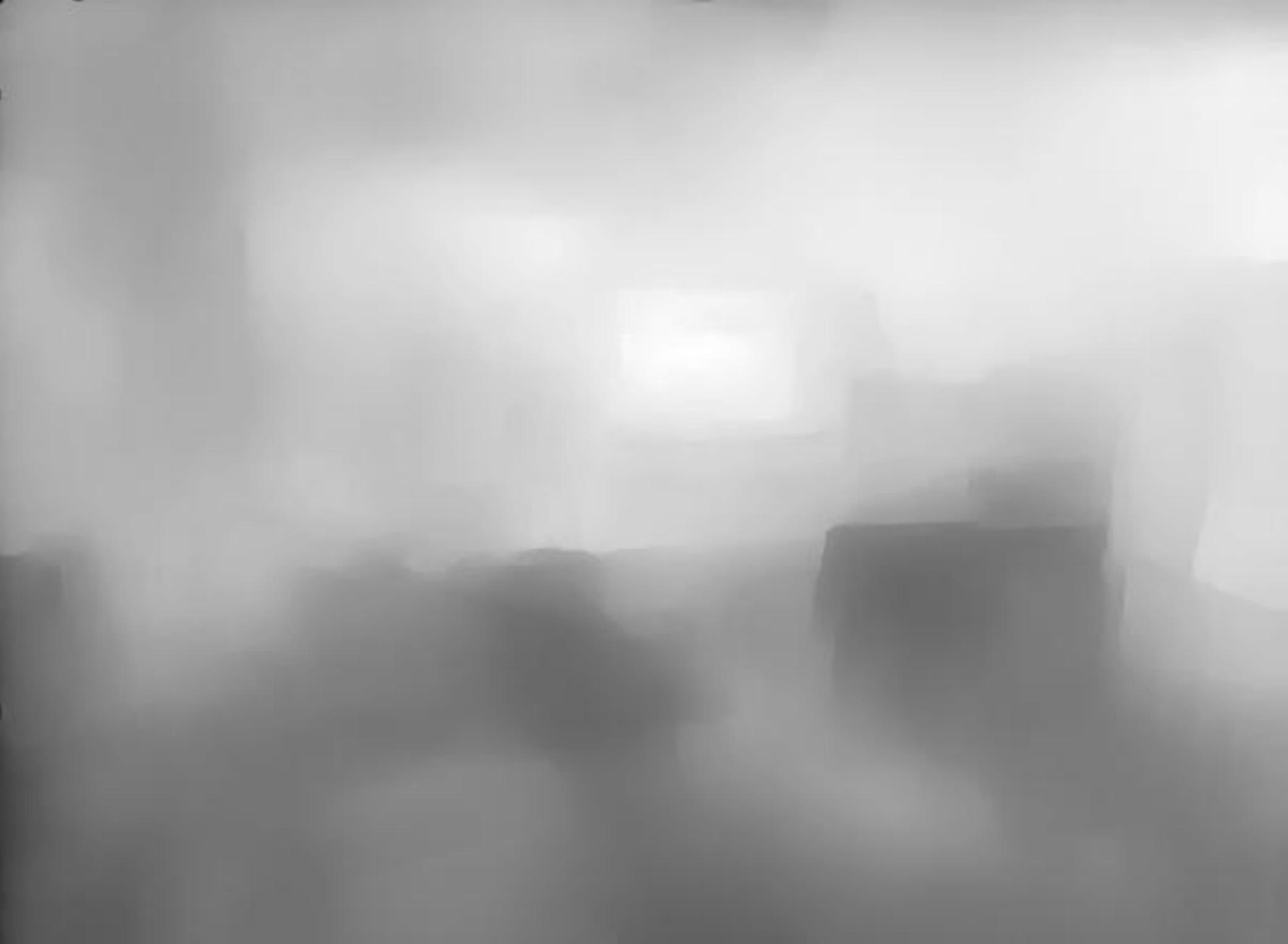}
\hfill
\includegraphics[width=0.23\linewidth,trim=0 0 0 10, clip=true]{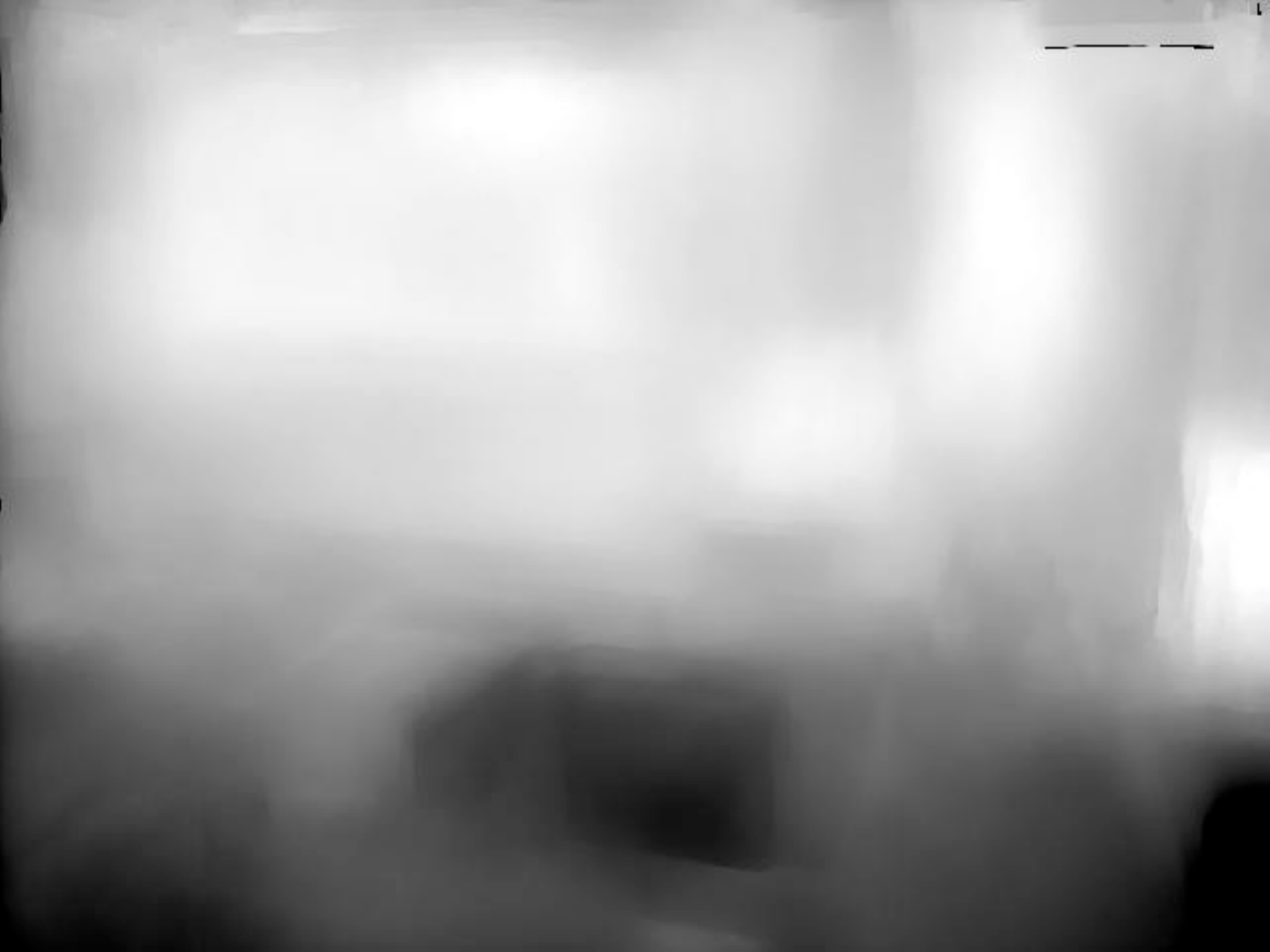}
\includegraphics[width=0.23\linewidth,trim=0 0 0 10, clip=true]{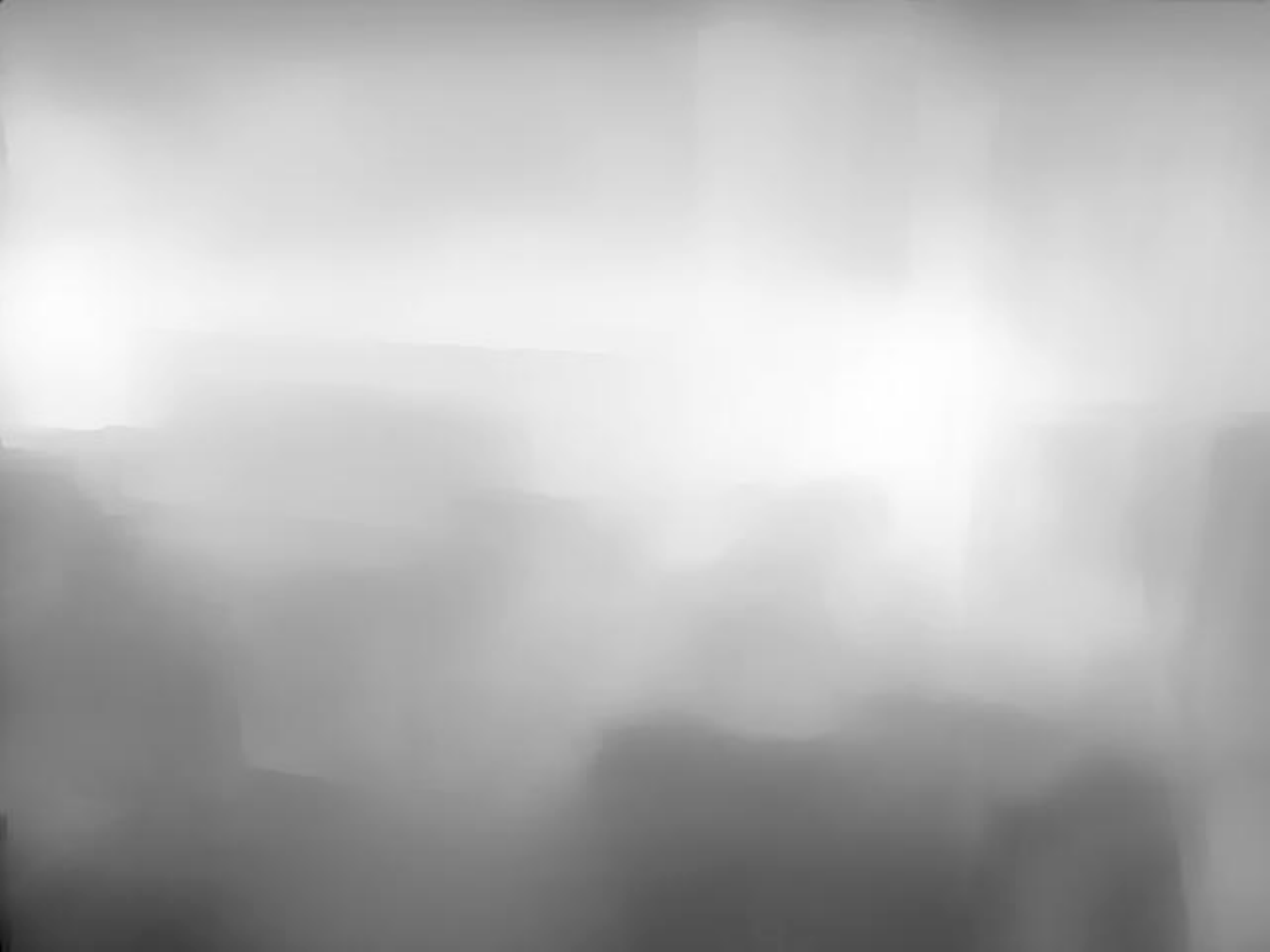}\\
\begin{minipage}{0.23\linewidth}\centerline{\small(a)}\end{minipage}
\begin{minipage}{0.23\linewidth}\centerline{\small(b)}\end{minipage}
\hfill
\begin{minipage}{0.23\linewidth}\centerline{\small(c)}\end{minipage}
\begin{minipage}{0.23\linewidth}\centerline{\small(d)}\end{minipage}
\caption{\newtextRev{Demonstration of our method on scenes where objects have been repositioned. The chairs and couch in (a) are moved closer in (b), and furniture in (c) are repositioned in (d). Although the depth maps contain noticeable errors, it is evident that our estimates are influenced by factors other than appearance. For example, although the chairs and couch in (a-b) have the same appearance, the estimated depth of (b) is noticeably closer in the chair/couch regions, as it should be.}}
\label{fig:furniture}
\end{figure}
%%%%%%%%%%%%%%%%%

Quantitative results are shown in Table~\ref{tab:nyu_results}, and Fig~\ref{fig:nyu_results} shows qualitative results. For comparison, we train our algorithm in two different ways and report results on each. One method trains (e.g., selects RGBD candidates) from the NYU depth dataset (holding out the particular example that is being tested), and the other method trains using all RGBD images found in MSR-V3D. We observe a significant degradation in results when training using our own dataset (MSR-V3D), likely because our dataset contains many fewer scenes than the NYU dataset, and a less diverse set of examples (NYU contains home, office, and many unique interiors, a total of 464; ours is primarily office-type scenes from four different buildings). 
This also suggests that generalization to interior scenes is much more difficult than outdoor, which coincides with the intuition that indoor scenes, on average, contain more variety than outdoor scenes.

Additionally, we compare our method to the single-image approaches of Konrad et al.~\cite{Konrad:SPIE:12,Konrad:3DCINE:12}. Both of these methods choose candidate RGBD images using global image features (as in our work), but dissimilar from our optimization, they compute depth as the median value of the candidate depth images (per-pixel), and post-process their depths with a cross-bilateral filter. \cite{Konrad:SPIE:12} applies SIFT flow to warp/align depth (similar to our method), while \cite{Konrad:3DCINE:12} opts not to for efficiency (their main focus is 2D-to-3D conversion rather than depth estimation). We show improvements over both of these methods, but more importantly, our depth and 2D-to-3D conversion  methods apply to videos rather than only single images.

Finally, we compare all of these results to two baselines: the average depth value over the entire NYU dataset (${}^\dagger$), and the per-pixel average of the NYU depth (${}^{\dagger \dagger}$). We observe that our method significantly outperforms these baselines in three metrics (relative, log${}_{10}$, and RMS error) when trained on the NYU dataset (hold-one-out). Our error rates increase by training on our dataset (MSR-V3D), but this training method still outperforms the baselines in several metrics. Note that these baselines do use some knowledge from the NYU dataset, unlike our method trained on MSR-V3D.
}

%%%%%%%%%%%%%%%%%
\begin{figure*}[t]
\begin{minipage}{0.24\linewidth}
\includegraphics[width=.49\linewidth]{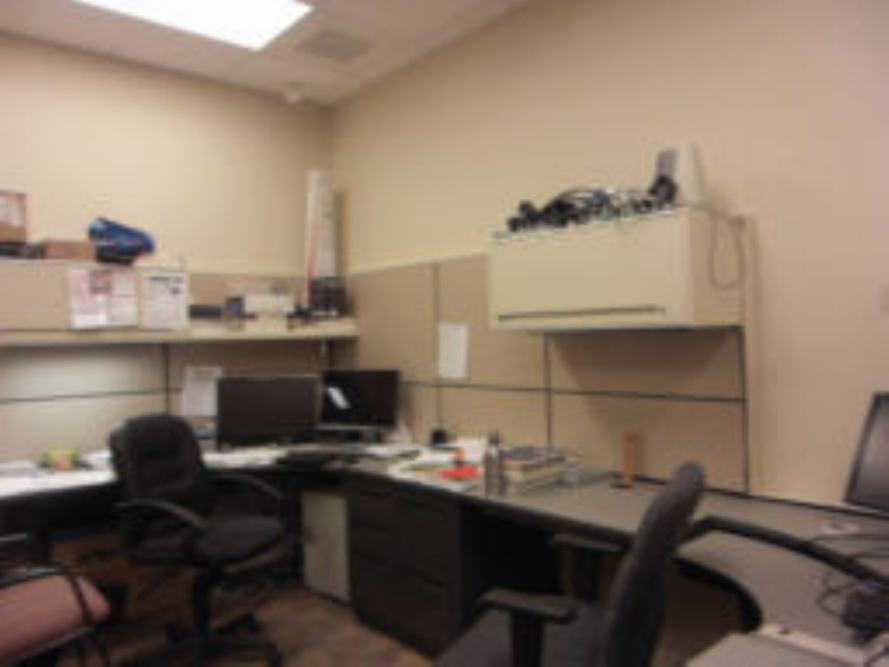}
\includegraphics[width=.49\linewidth]{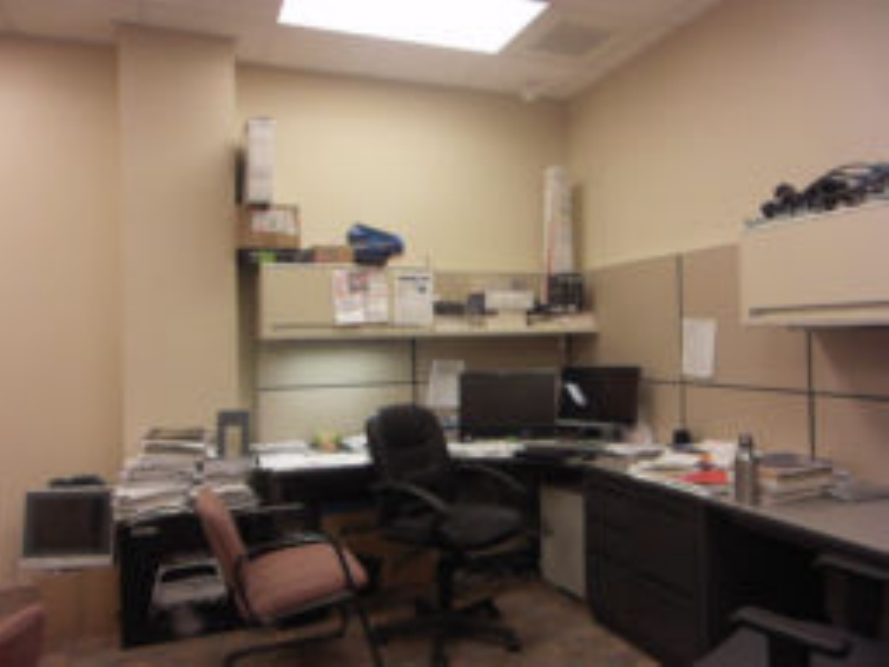}
\\
\includegraphics[width=.49\linewidth]{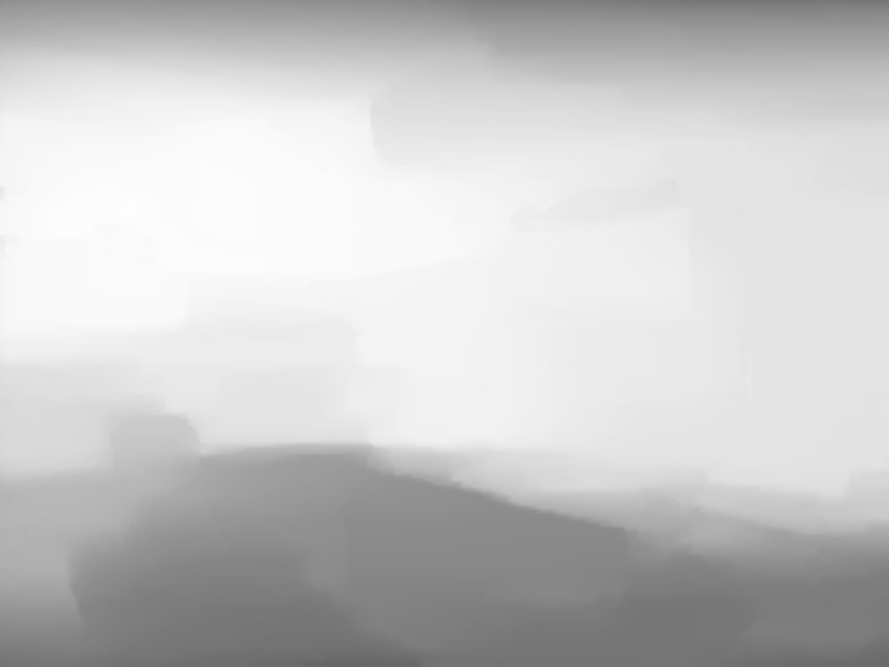}
\includegraphics[width=.49\linewidth]{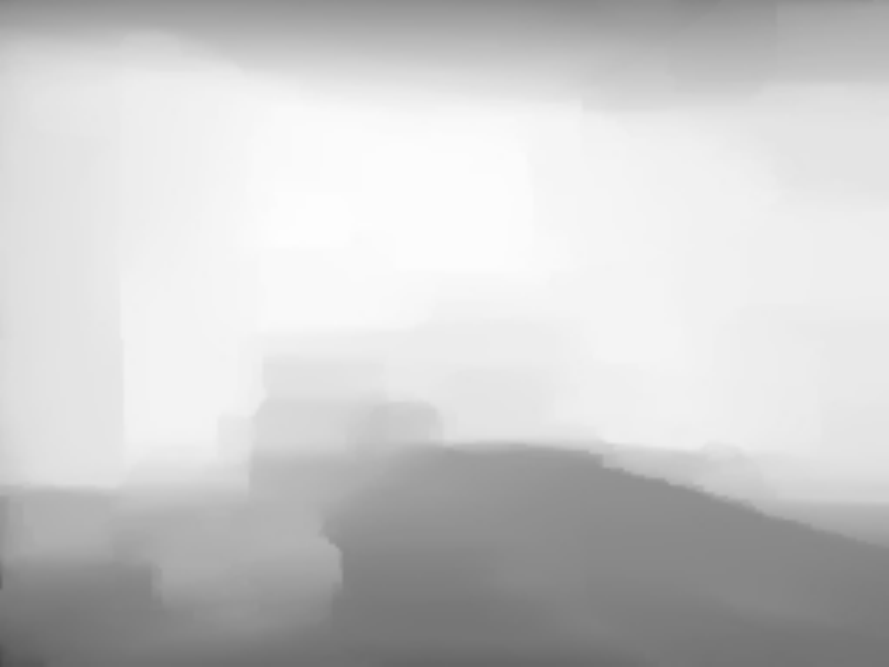}
\\
\includegraphics[width=.49\linewidth]{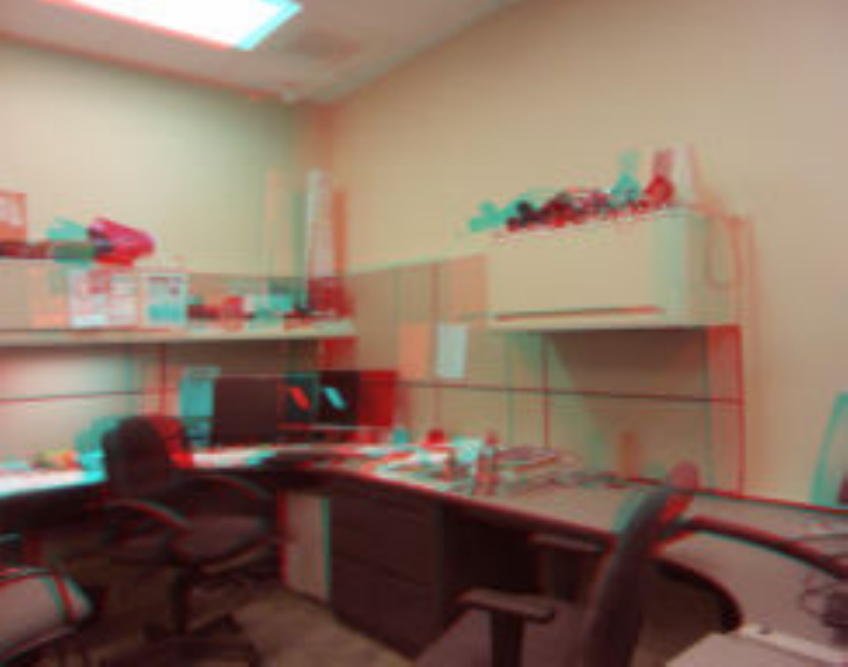}
\includegraphics[width=.49\linewidth]{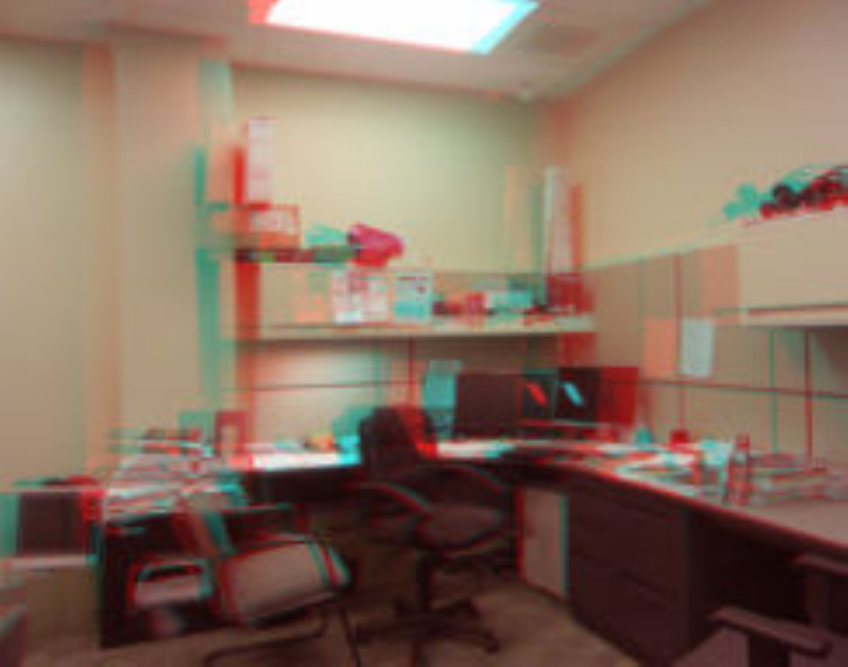}
\end{minipage}
\hfill
\begin{minipage}{0.24\linewidth}
\includegraphics[width=.49\linewidth]{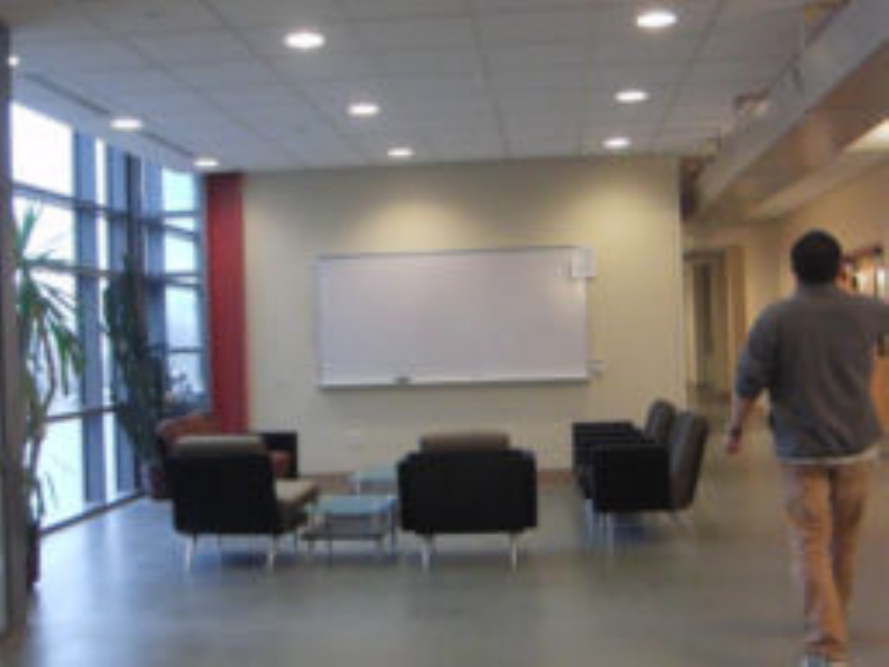}
\includegraphics[width=.49\linewidth]{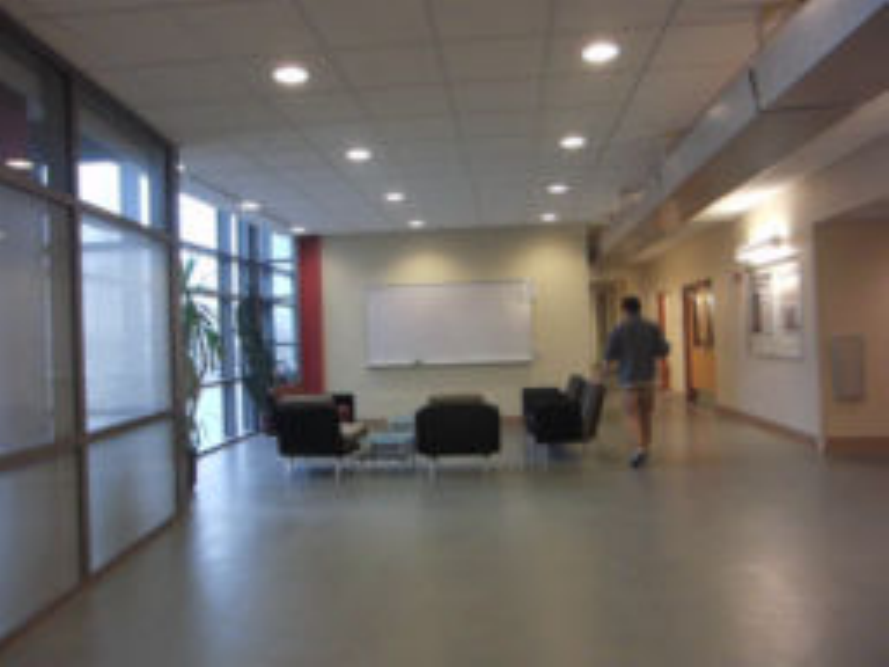}
\\
\includegraphics[width=.49\linewidth]{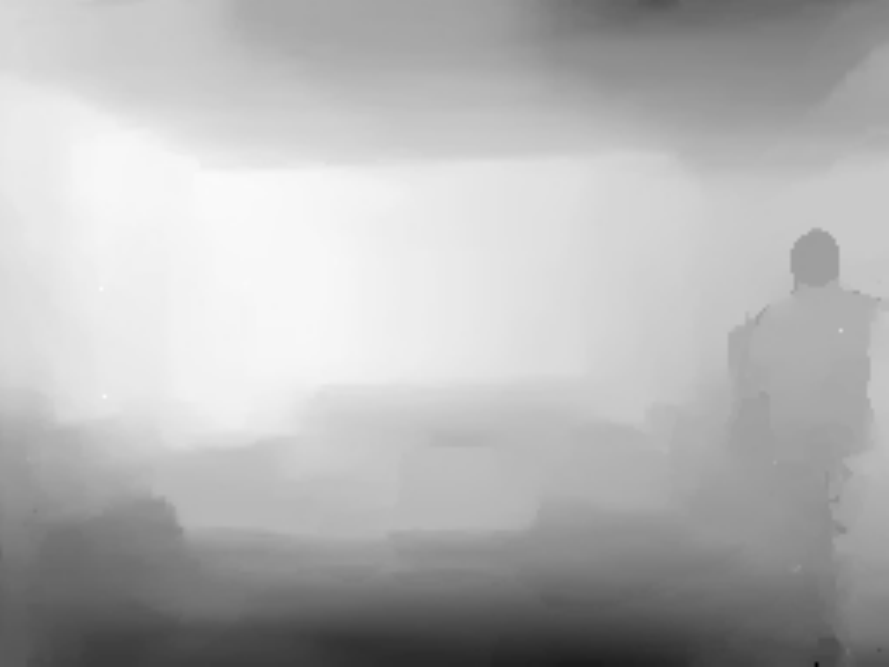}
\includegraphics[width=.49\linewidth]{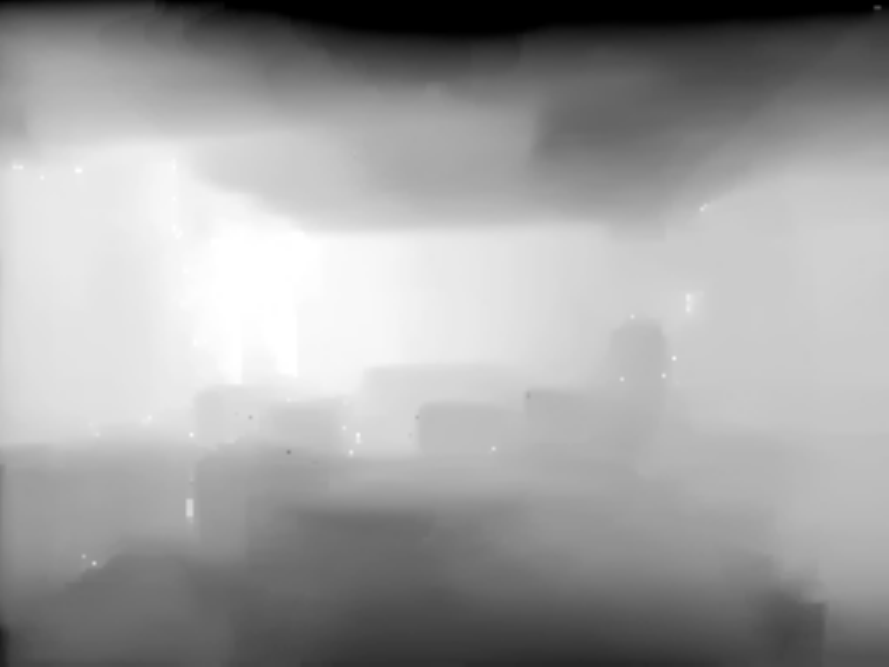}
\\
\includegraphics[width=.49\linewidth]{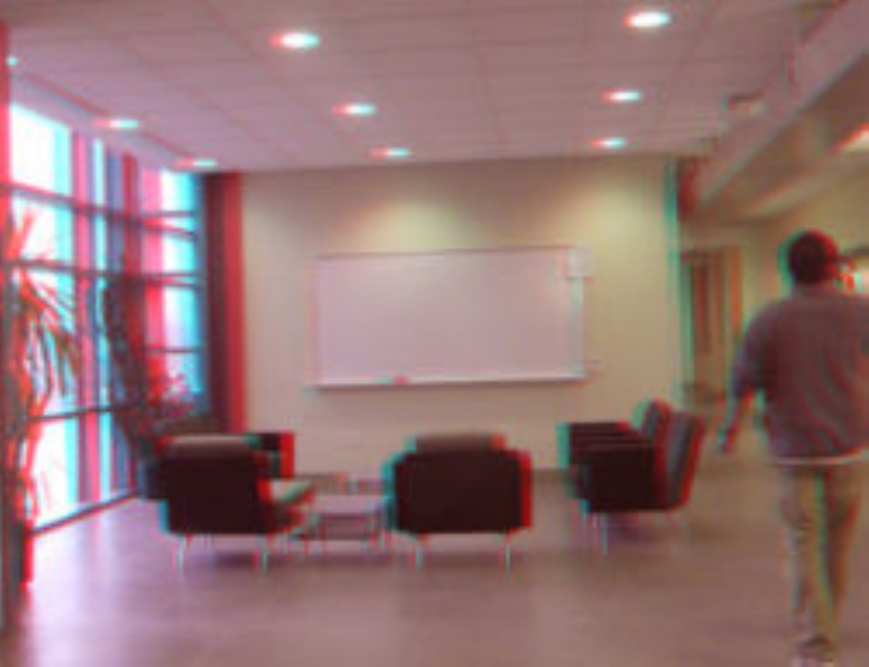}
\includegraphics[width=.49\linewidth]{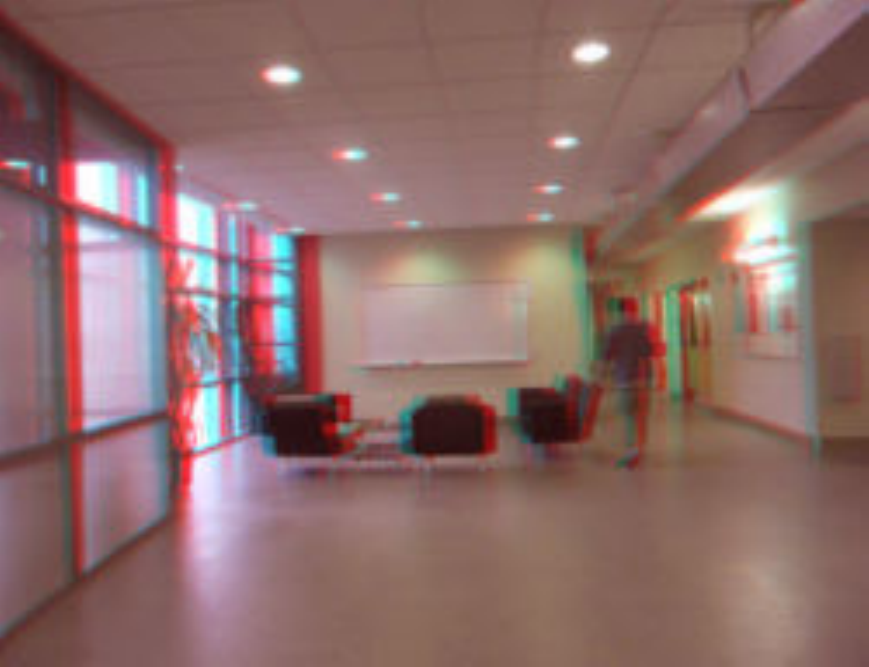}
\end{minipage}
\hfill
\begin{minipage}{0.24\linewidth}
\includegraphics[width=.49\linewidth]{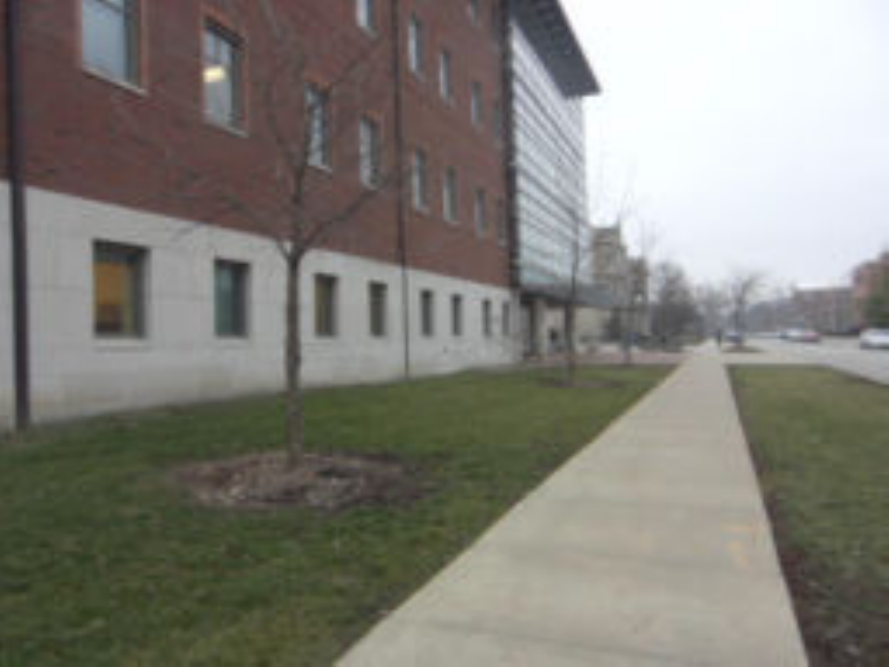}
\includegraphics[width=.49\linewidth]{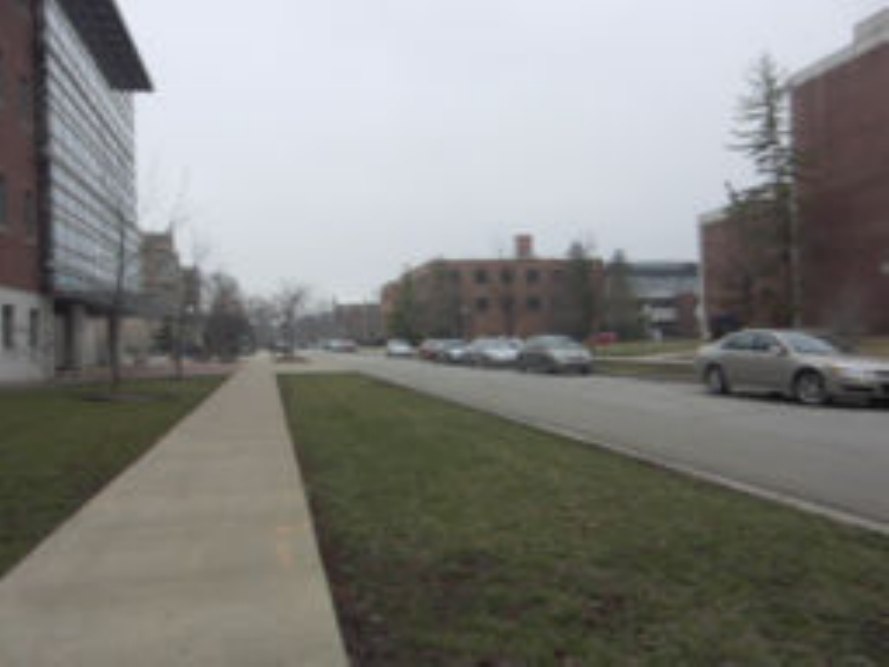}
\\
\includegraphics[width=.49\linewidth]{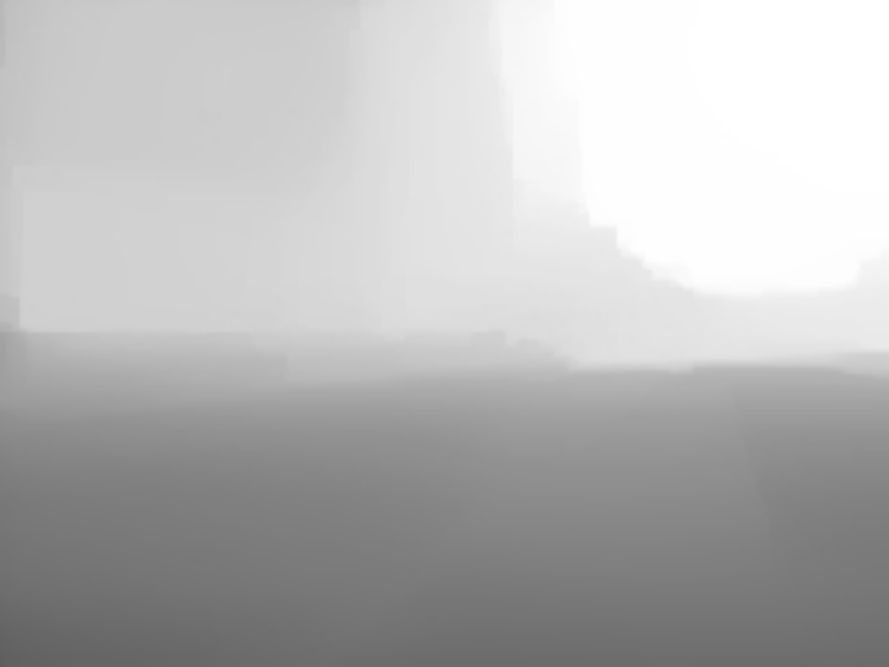}
\includegraphics[width=.49\linewidth]{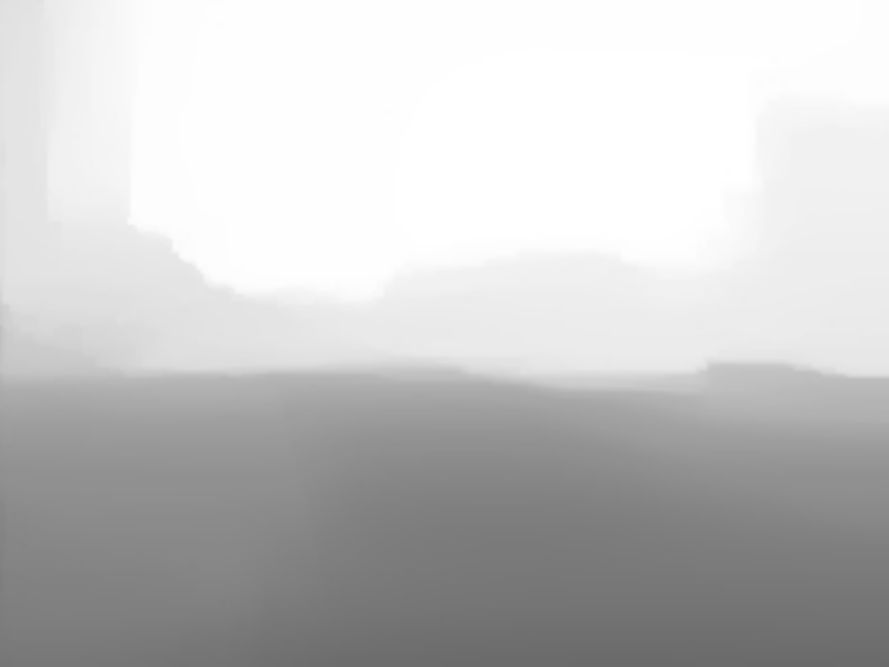}
\\
\includegraphics[width=.49\linewidth]{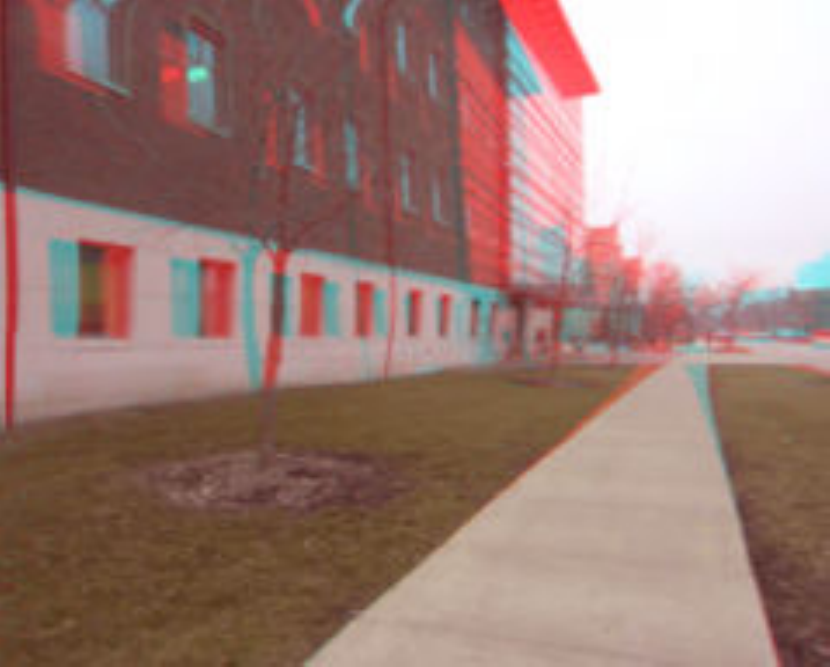}
\includegraphics[width=.49\linewidth]{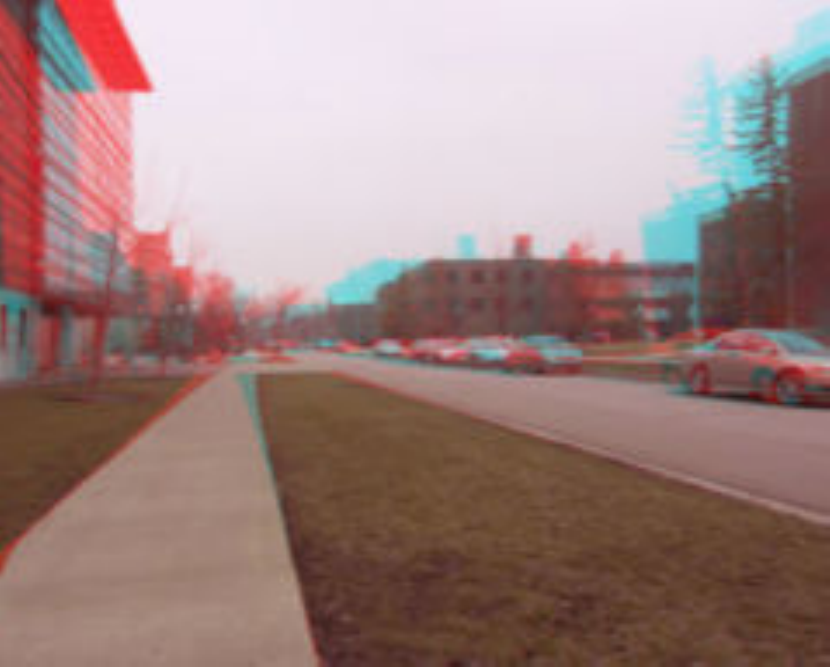}
\end{minipage}
\hfill
\begin{minipage}{0.24\linewidth}
\includegraphics[width=.49\linewidth]{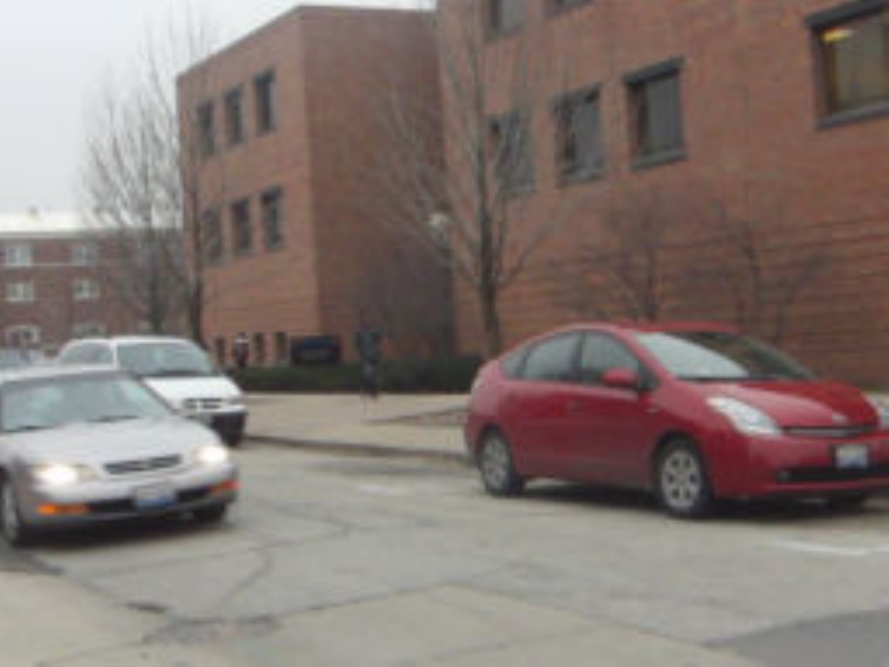}
\includegraphics[width=.49\linewidth]{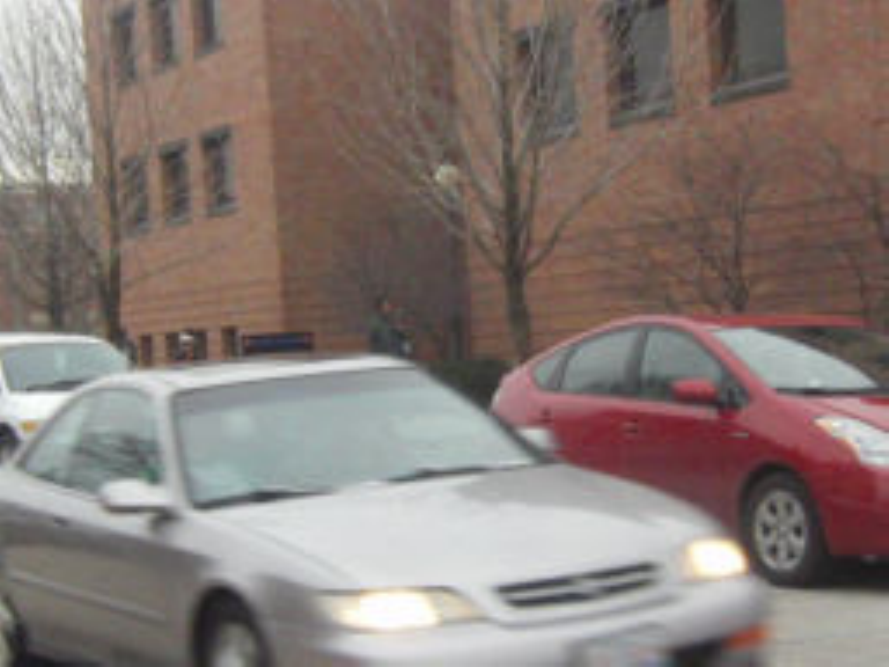}
\\
\includegraphics[width=.49\linewidth]{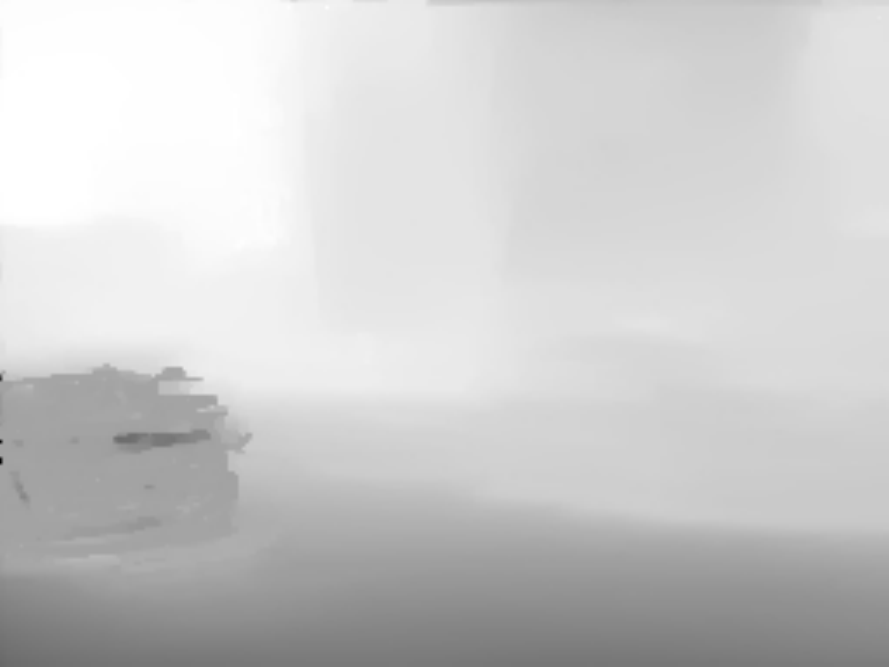}
\includegraphics[width=.49\linewidth]{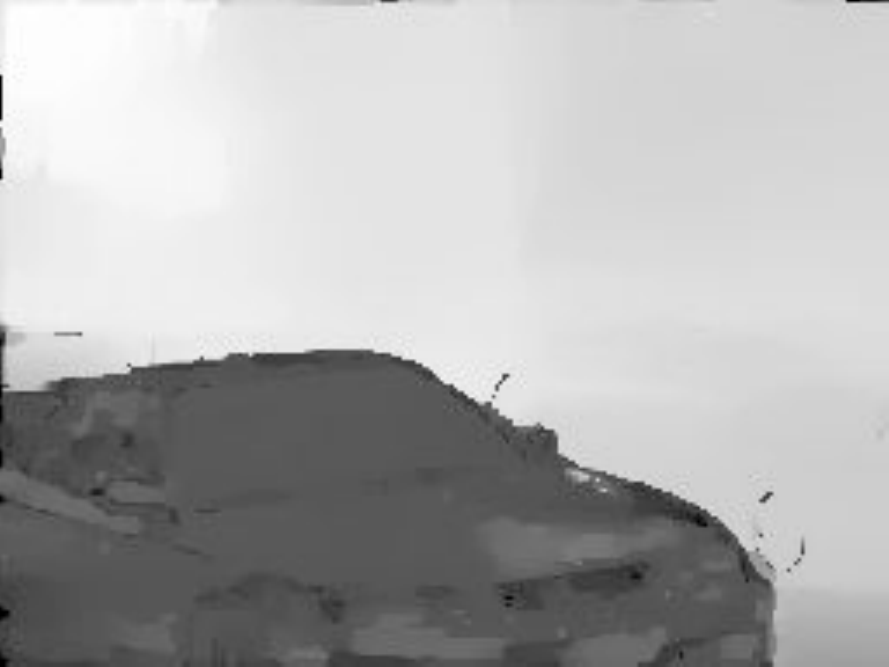}
\\
\includegraphics[width=.49\linewidth]{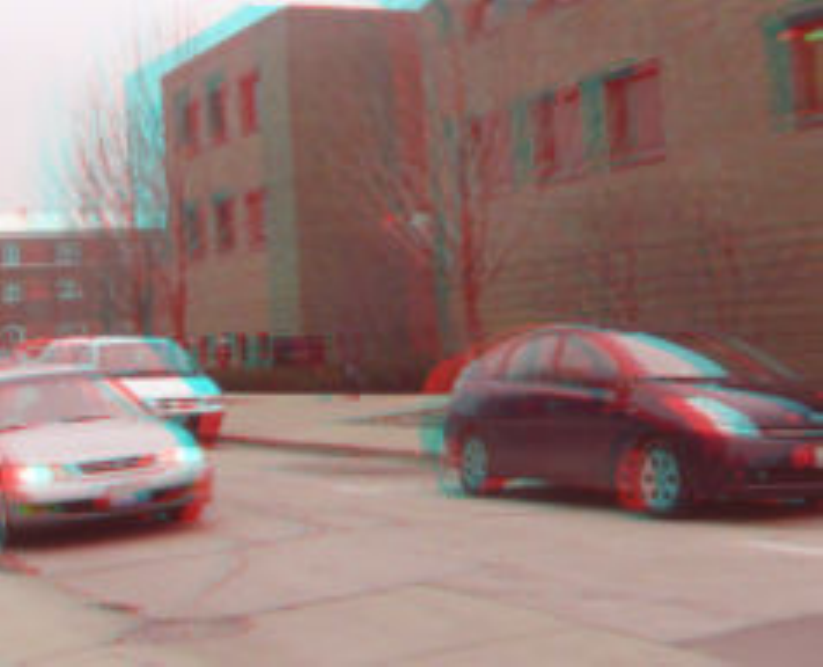}
\includegraphics[width=.49\linewidth]{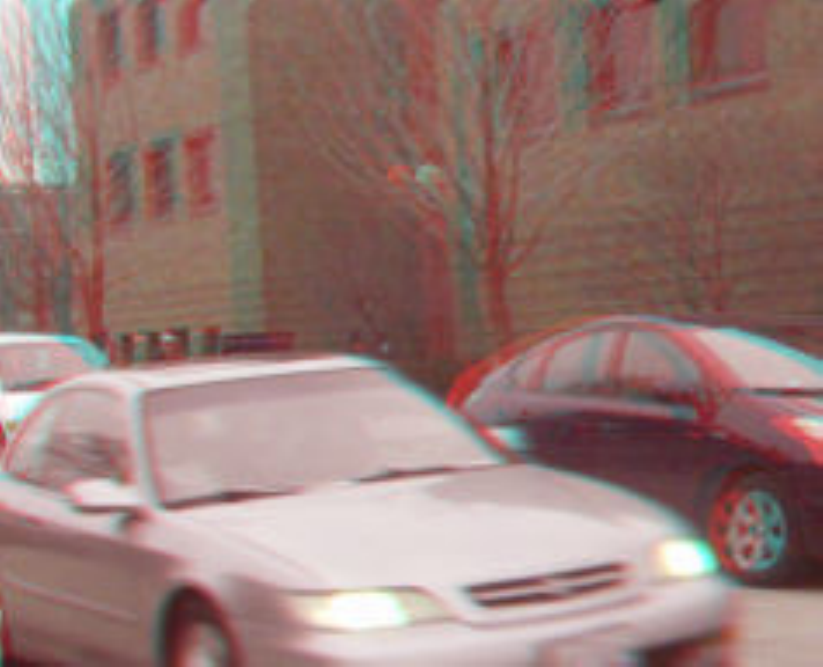}
\end{minipage}
\caption{Results obtained on four different sequences captured with a rotating camera and/or variable focal length. We show the input frames (\emph{top}), inferred depth (\emph{middle}) and inferred 3D anaglyph (\emph{bottom}). Notice that the sequences are time-coherent and that moving objects are not ignored.
}
\label{fig:rot_zoom_results}
\end{figure*}
%%%%%%%%%%%%%%%%%

%%%%%%%%%%%%%%%%%
\begin{figure*}
\begin{center}
\begin{minipage}{1.0\linewidth}
%\centerline{\hspace{.05\linewidth} Input \hspace{.075\linewidth}  True depth \hspace{.03\linewidth}  Our depth \hspace{.04\linewidth}  True 3D \hspace{.05\linewidth}  Our 3D \hspace{.01\linewidth} }
\begin{minipage}{0.49\linewidth}
\includegraphics[width=.195\linewidth]{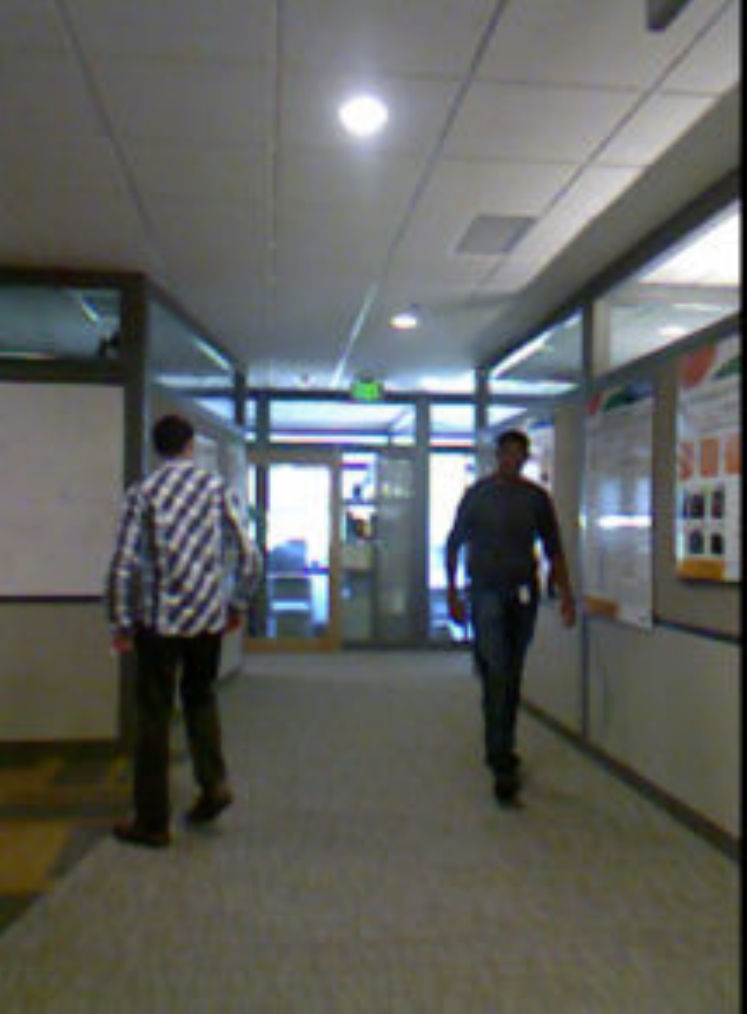}
\includegraphics[width=.195\linewidth]{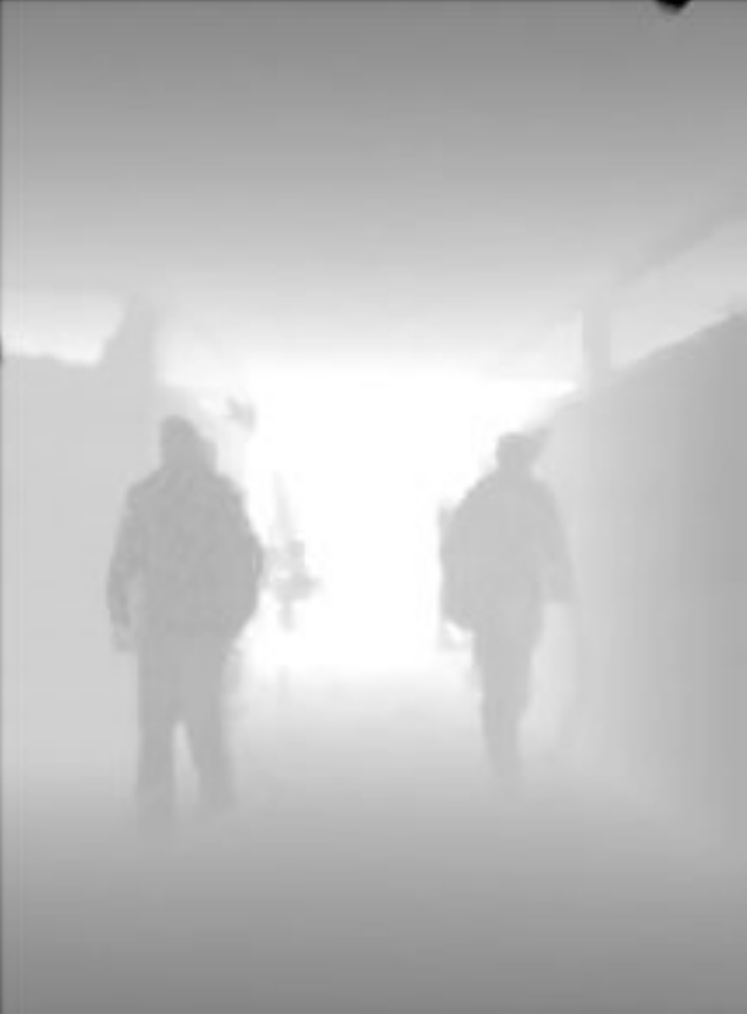}
\includegraphics[width=.195\linewidth]{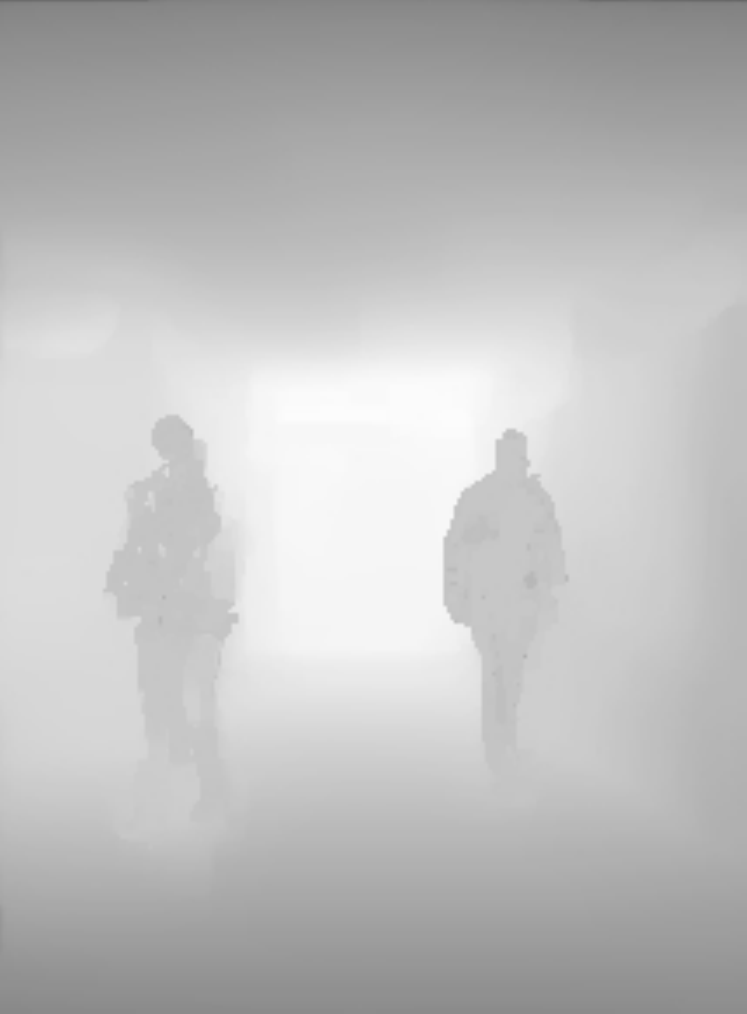}
\includegraphics[height=.265\linewidth]{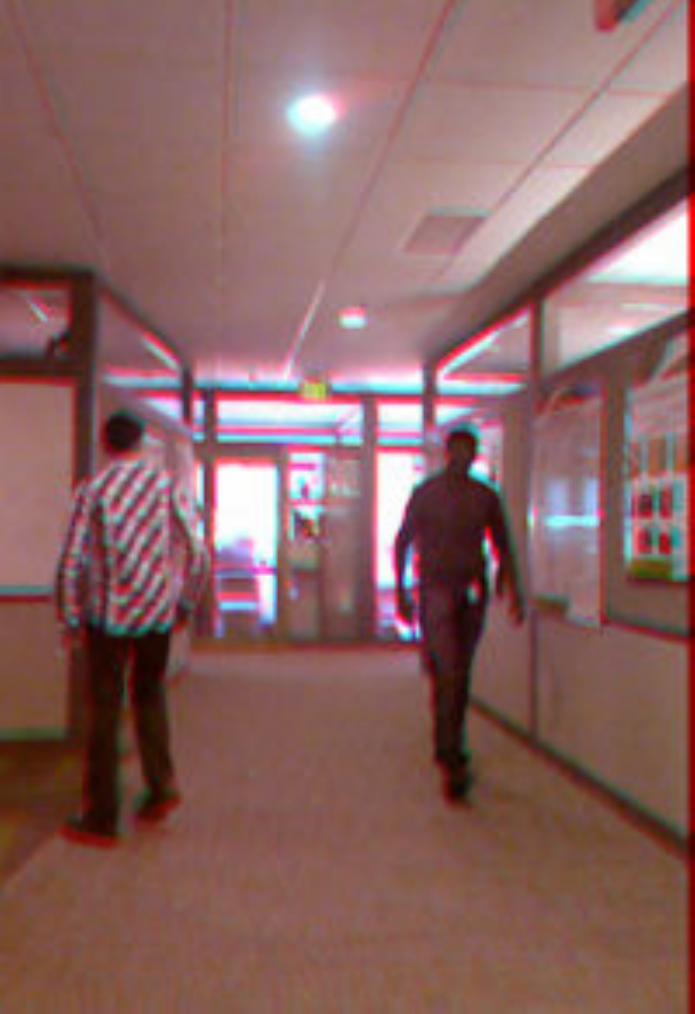}
\includegraphics[height=.265\linewidth]{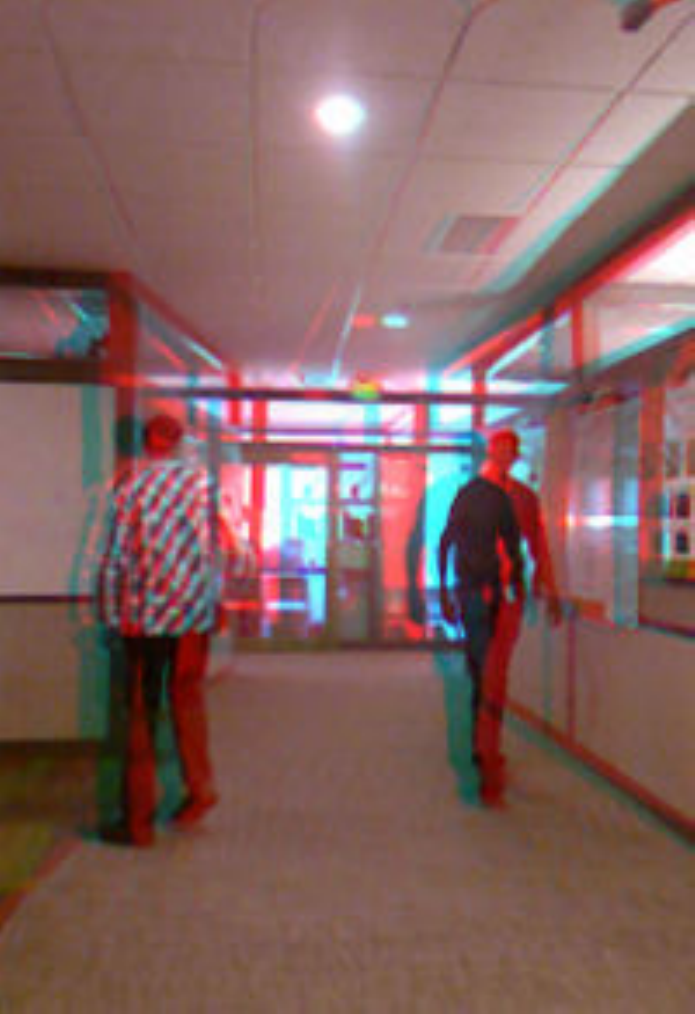}
\end{minipage}
\hfill
%\rule[-7.5mm]{.15mm}{.13\linewidth}
\hfill
\begin{minipage}{.49\linewidth}
\includegraphics[width=.195\linewidth]{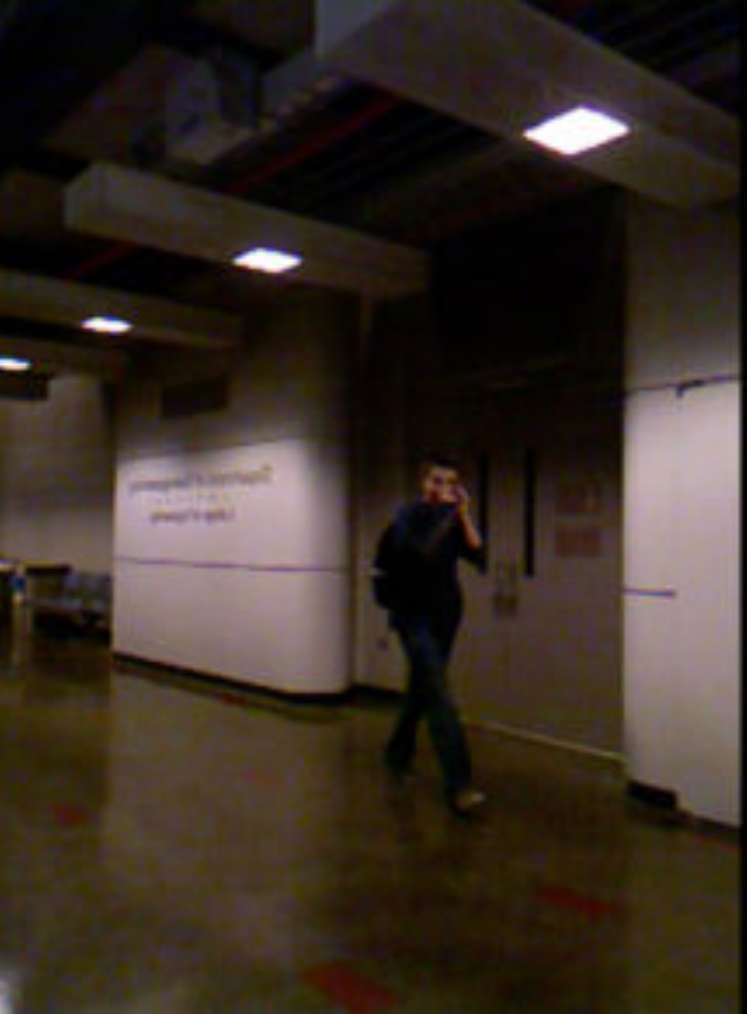}
\includegraphics[width=.195\linewidth]{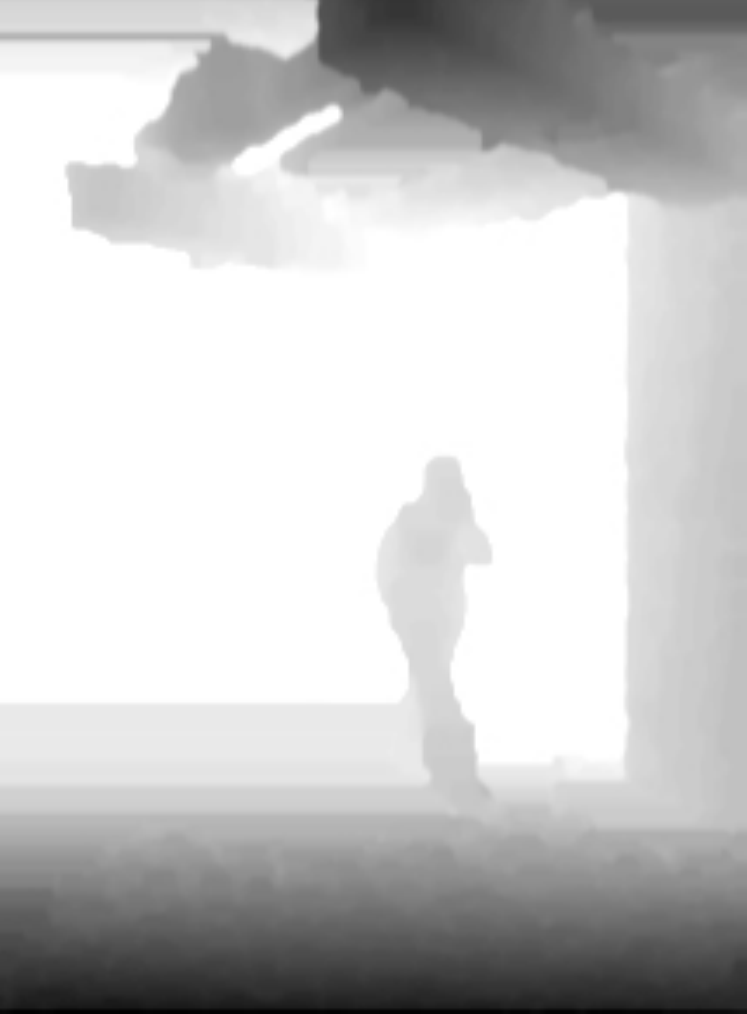}
\includegraphics[width=.195\linewidth]{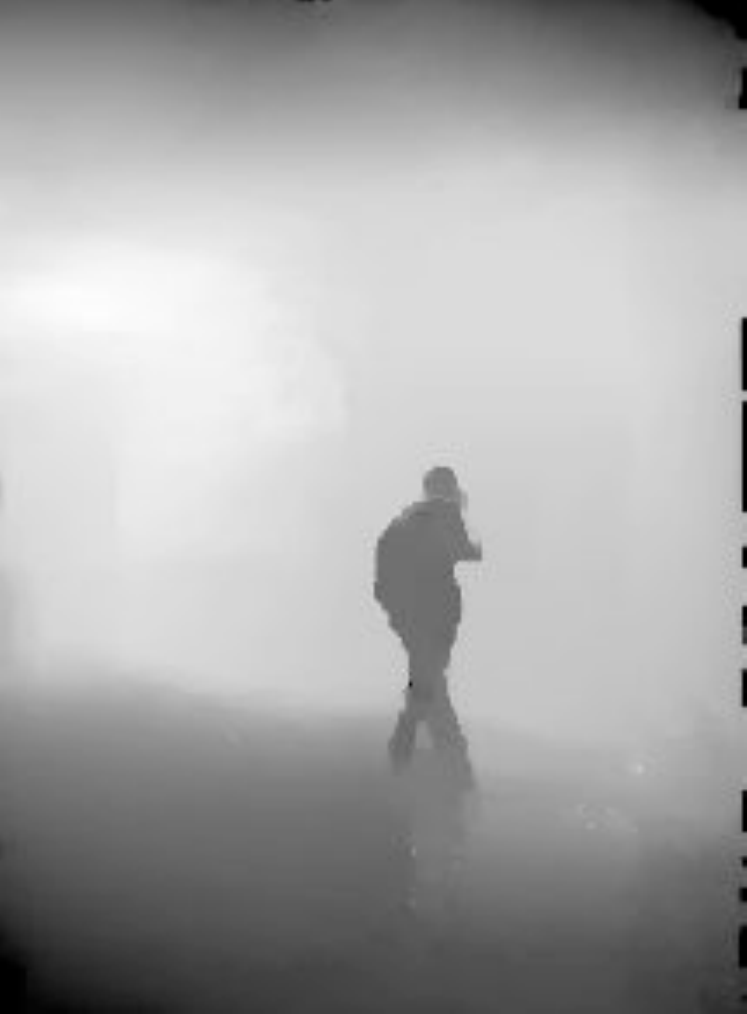}
\includegraphics[height=.265\linewidth, trim=0pt 0pt 0pt 0pt, clip=true]{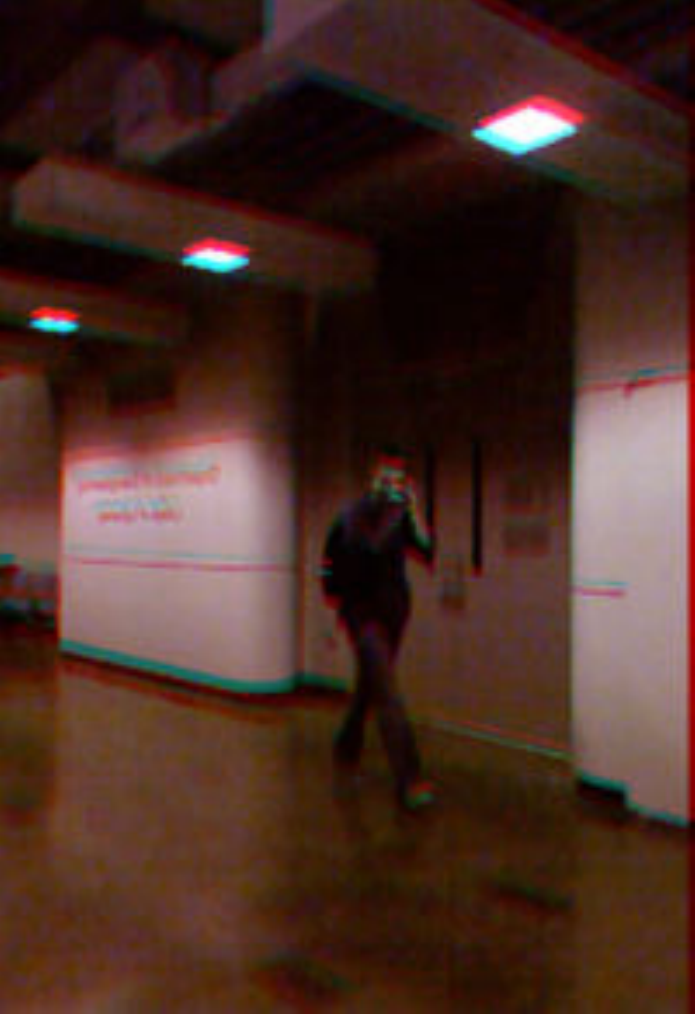}
\includegraphics[height=.265\linewidth, trim=0pt 0pt 0pt 0pt, clip=true]{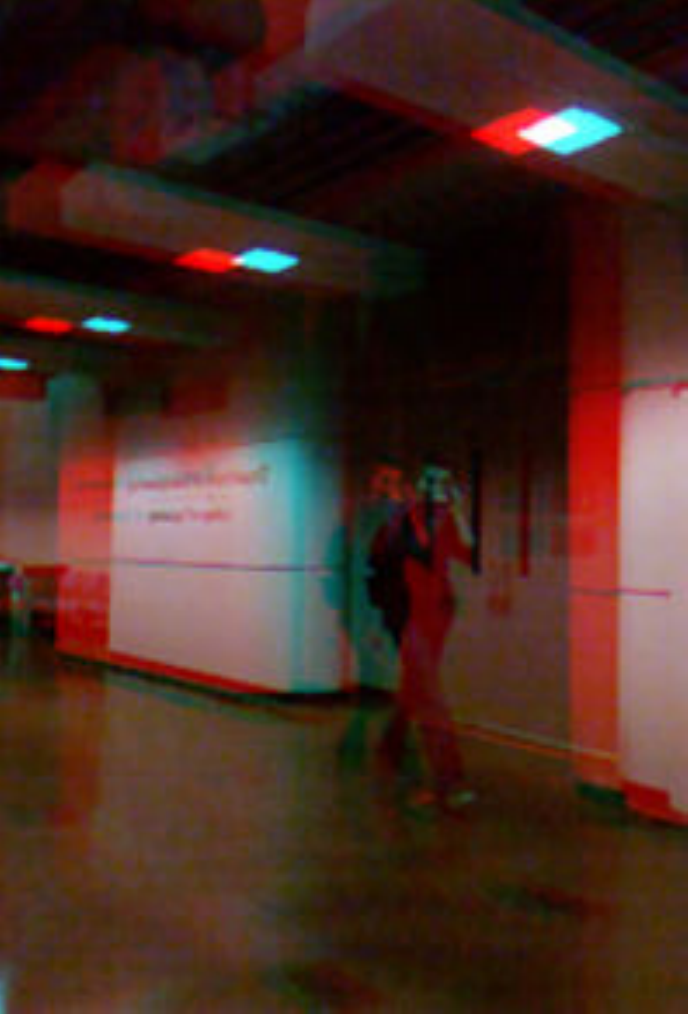}
\end{minipage} \\ \vspace{2mm}

%\rule{\linewidth}{.15mm}
%\\
\begin{minipage}{.49\linewidth}
\includegraphics[width=.195\linewidth]{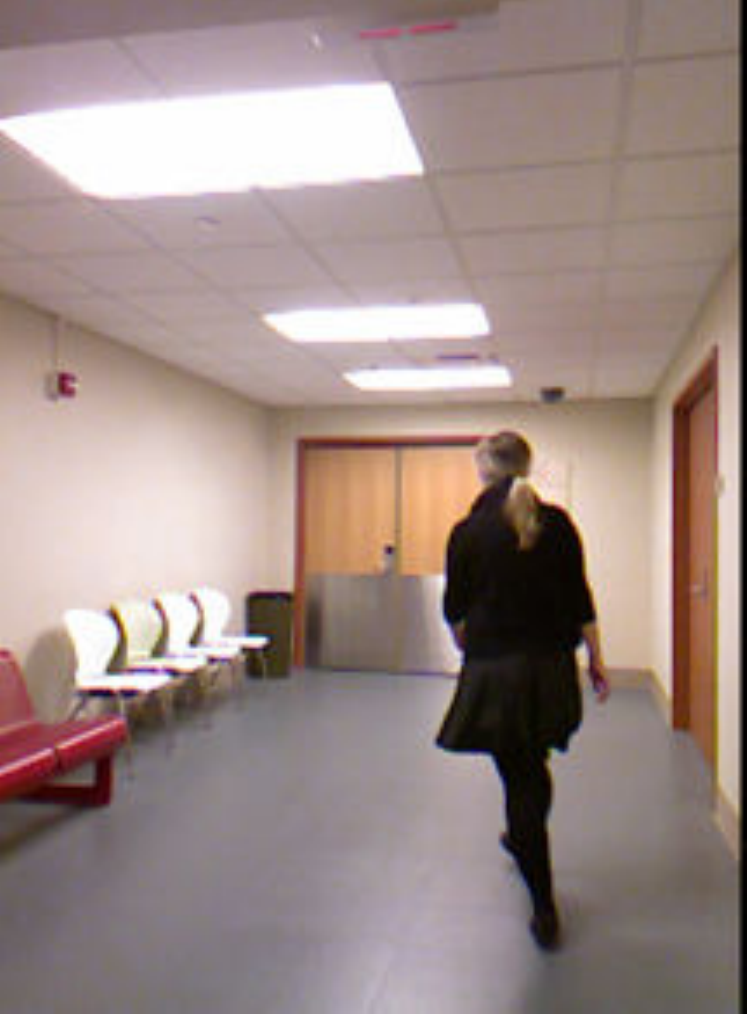}
\includegraphics[width=.195\linewidth]{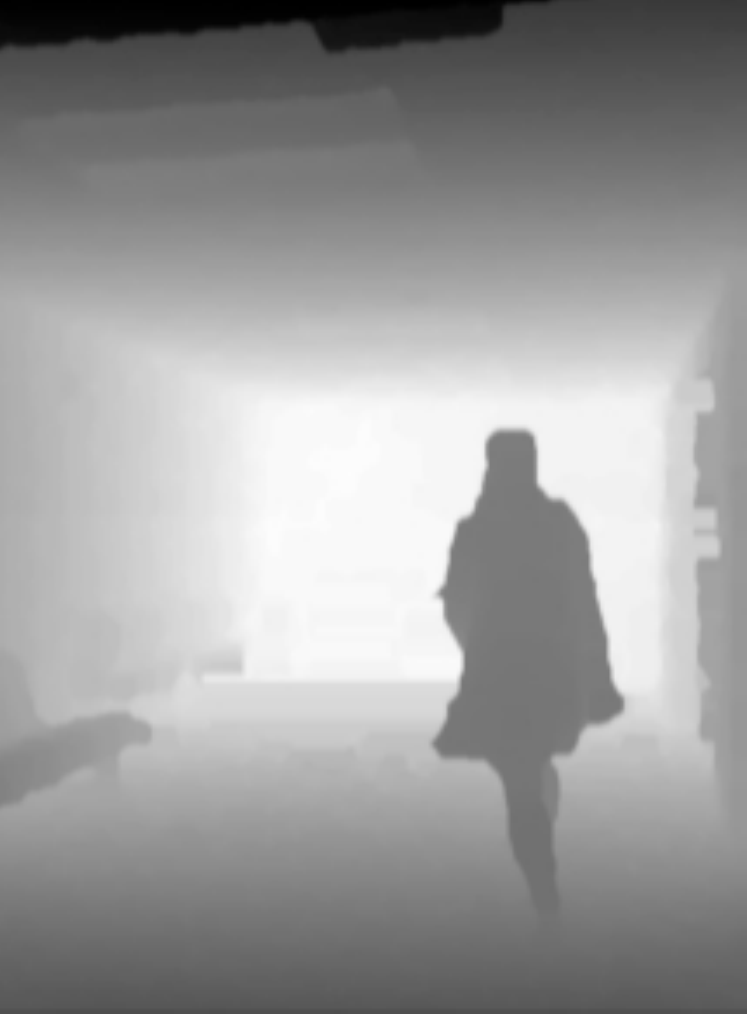}
\includegraphics[width=.195\linewidth]{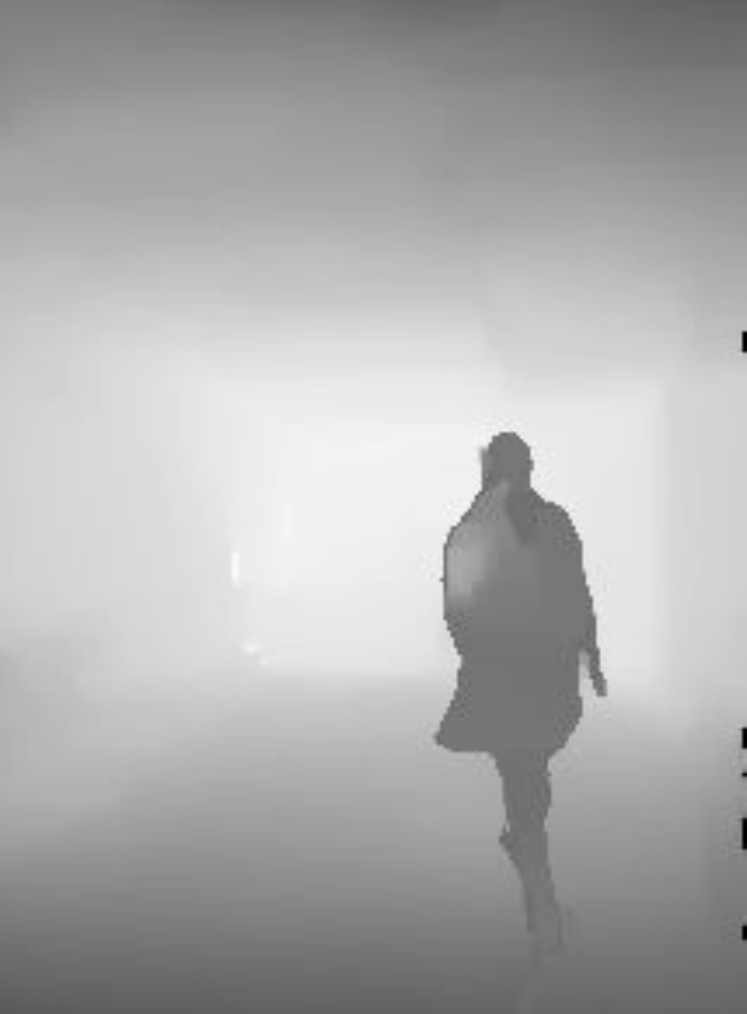}
\includegraphics[height=.265\linewidth]{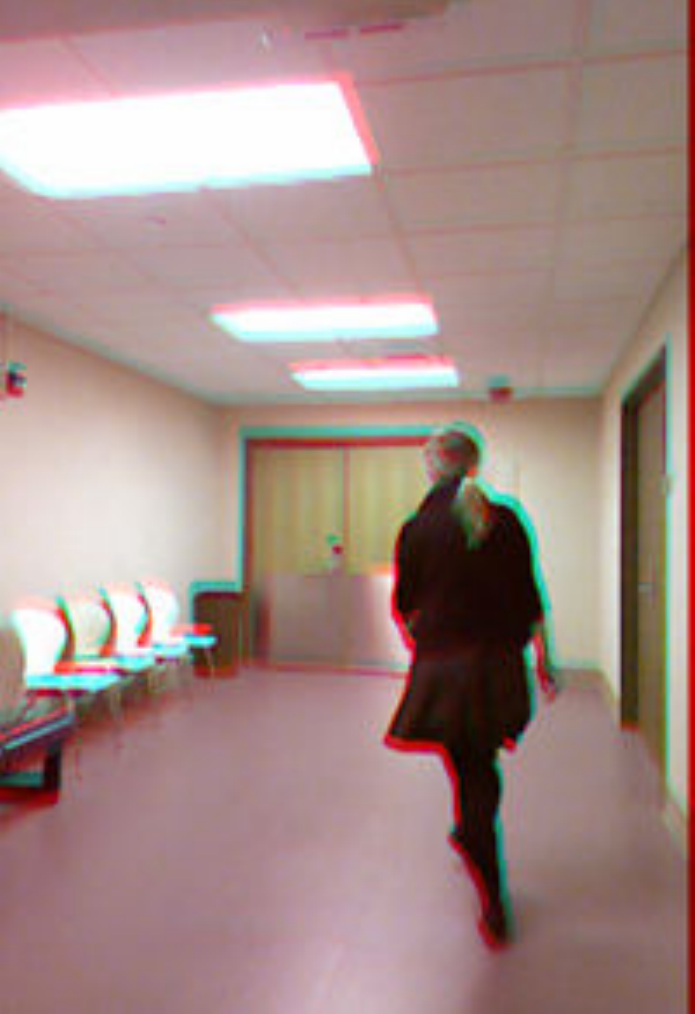}
\includegraphics[height=.265\linewidth]{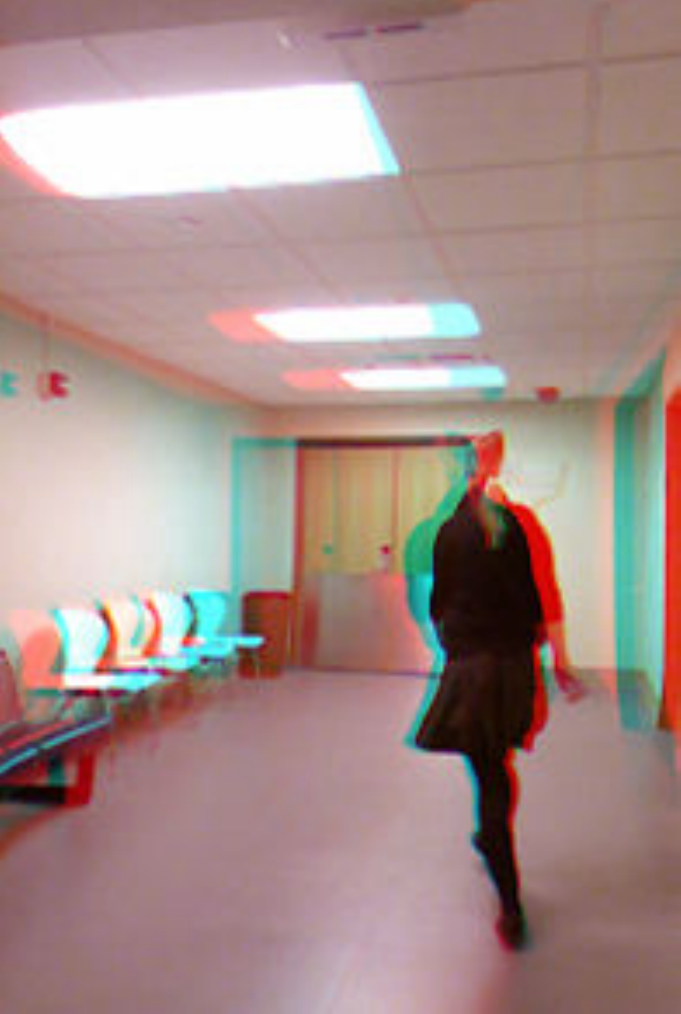}
\end{minipage}
\hfill
%\rule[-7mm]{.15mm}{.13\linewidth}
\hfill
\begin{minipage}{.49\linewidth}
\includegraphics[width=.195\linewidth]{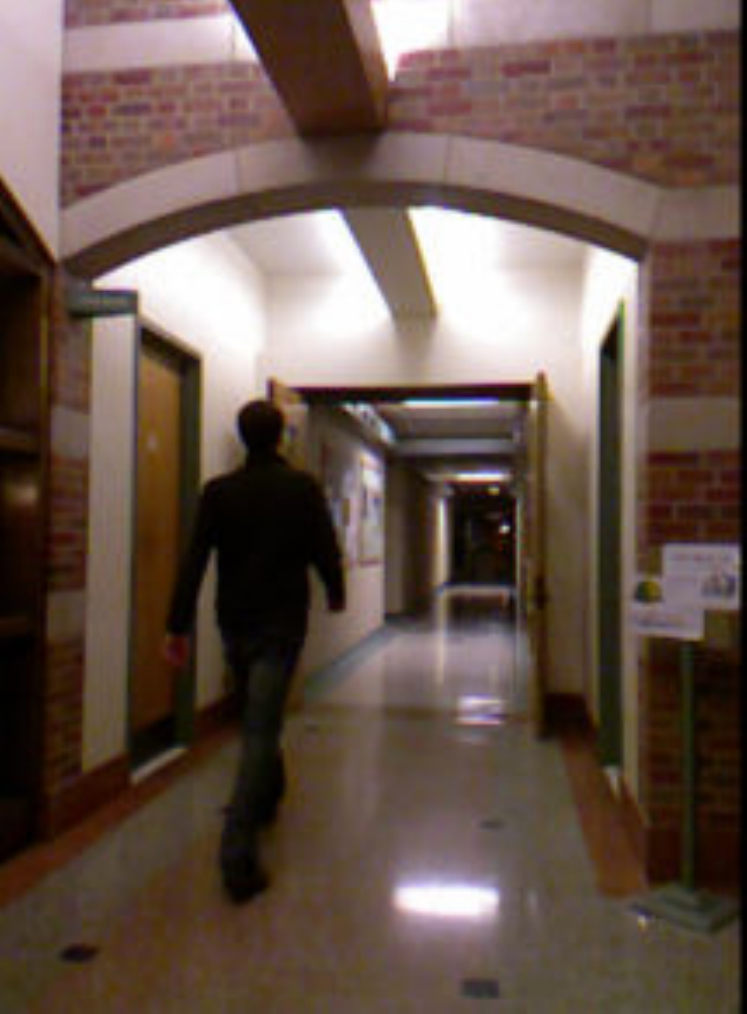}
\includegraphics[width=.195\linewidth]{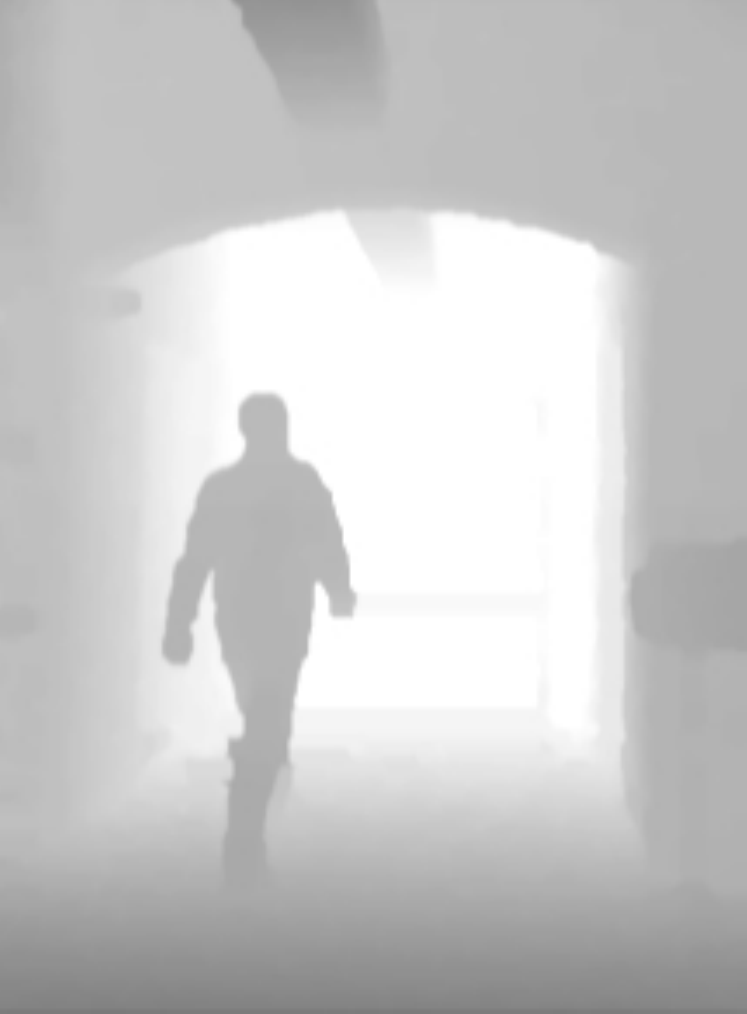}
\includegraphics[width=.195\linewidth]{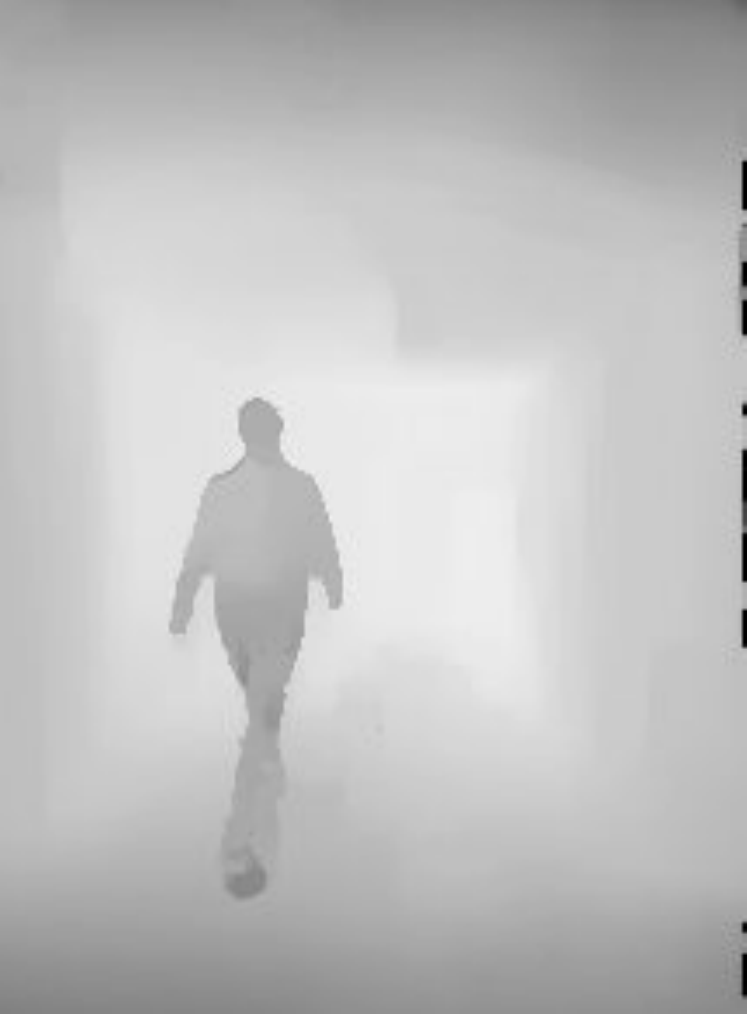}
\includegraphics[height=.265\linewidth, trim=0pt 0pt 0pt 0pt, clip=true]{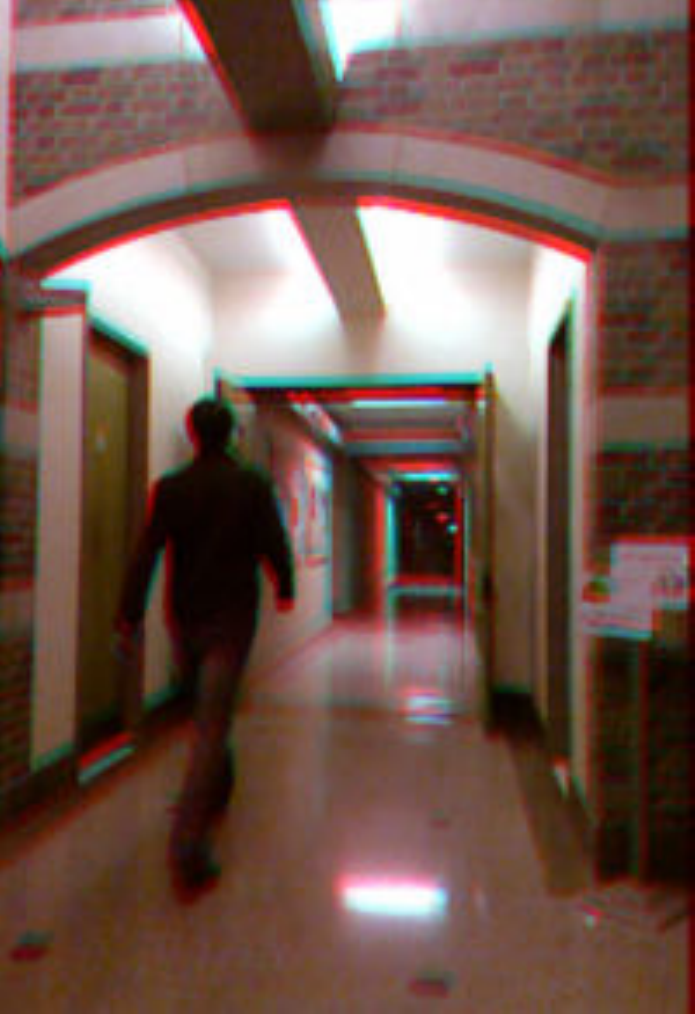}
\includegraphics[height=.265\linewidth, trim=0pt 0pt 4pt 0pt, clip=true]{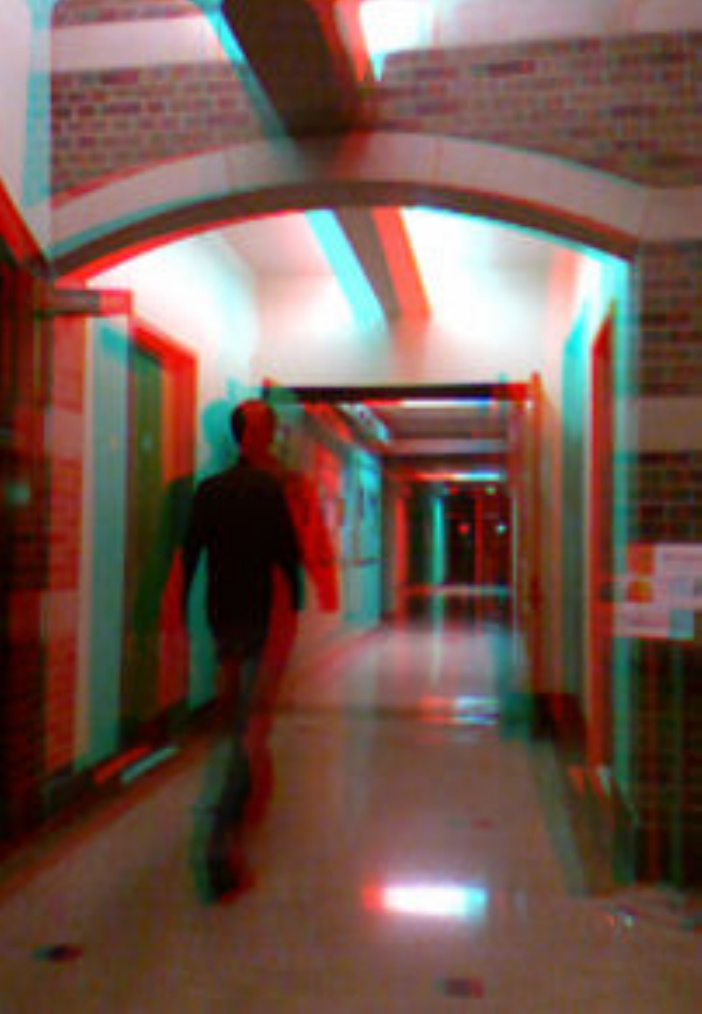}
\end{minipage}
\end{minipage}
\end{center}
\vspace{-2mm}
\caption{Video results obtained on test images for each building in our stereo-RGBD dataset (buildings 1-4, from left to right and top to bottom). For each result (from left to right): original photograph, ground truth depth, our inferred depth, ground truth anaglyph \protect\includegraphics[height=5pt]{fig/anaglyph.pdf} image, and our synthesized anaglyph image. Because the ground truth 3D images were recorded with a fixed interocular distance (roughly 5cm), we cannot control the amount of ``pop-out,'' and the 3D effect is subtle. However, this is a parameter we can set using our automatic approach to achieve a desired effect, which allows for an enhanced 3D experience. Note also that our algorithm can handle multiple moving objects (\emph{top}).
%We captured data in four different buildings, with data from only one building used for training. Results in the left column are from the building used for training (obtained by holding each of the left sequences out of the training database), and the results on the right are from other buildings not used in training.
% Each pair of inferred and ground truth depths are displayed in log space at the same scale.
\vspace{2mm}
}
\label{fig:indoor_results}
\end{figure*}
%%%%%%%%%%%%%%%%%

%%%%%%%%%%%%%%%%%
\begin{figure*}[!htb]
\begin{minipage}{0.085\linewidth}
\centerline{Input} \vspace{10mm} 
\centerline{Ours} \vspace{8mm} 
\centerline{Ours} \centerline{\small(SfM prior)} \vspace{6mm} 
\centerline{Zhang} \centerline{et al.~\cite{Zhang:09}}
%\begin{sideways} Zhang et al.~\cite{Zhang:09} | Ours (sfm) | Ours | Input \end{sideways}
\end{minipage}
\begin{minipage}{0.445\linewidth}
\includegraphics[width=0.32\linewidth]{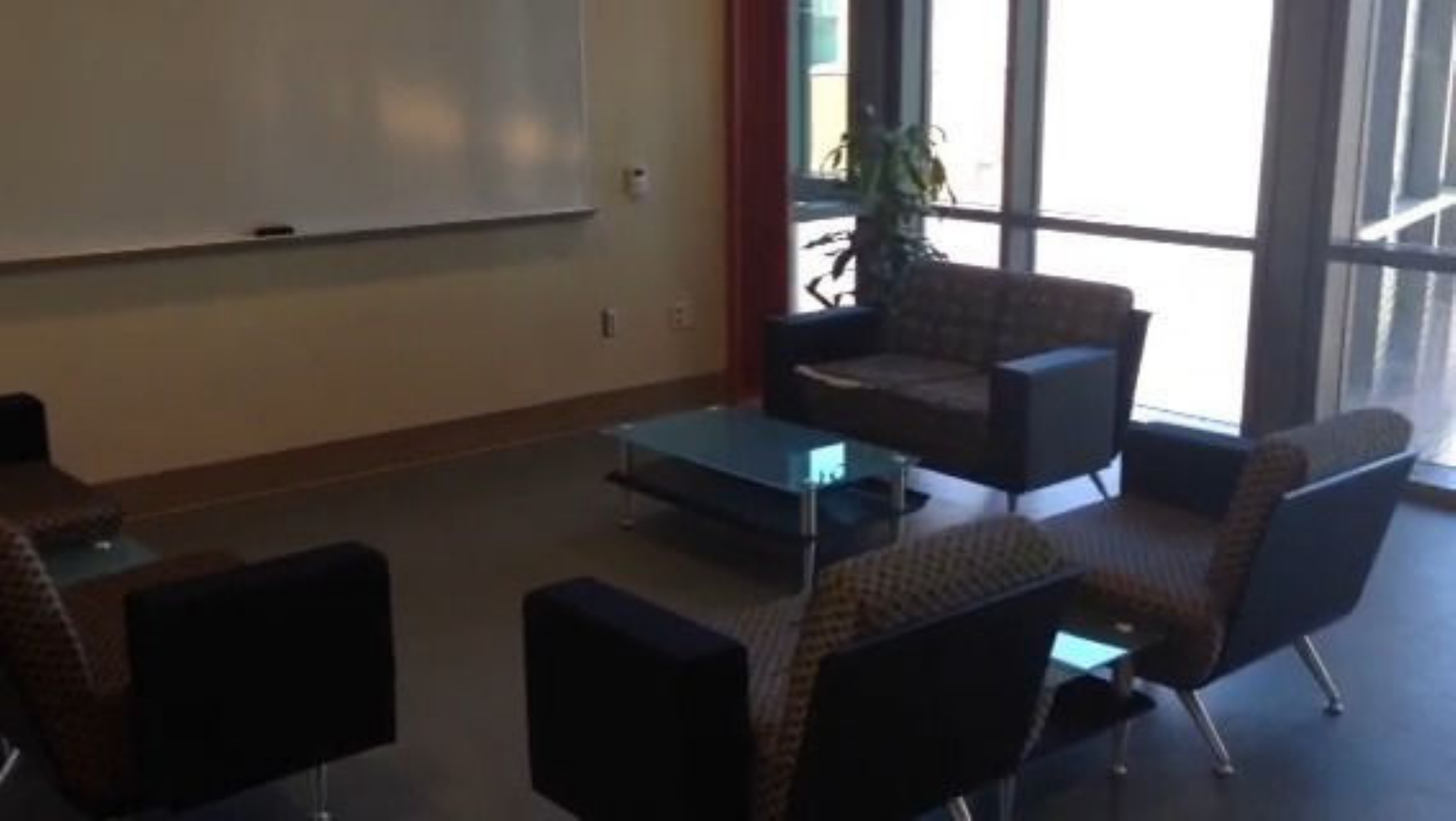}
\includegraphics[width=0.32\linewidth]{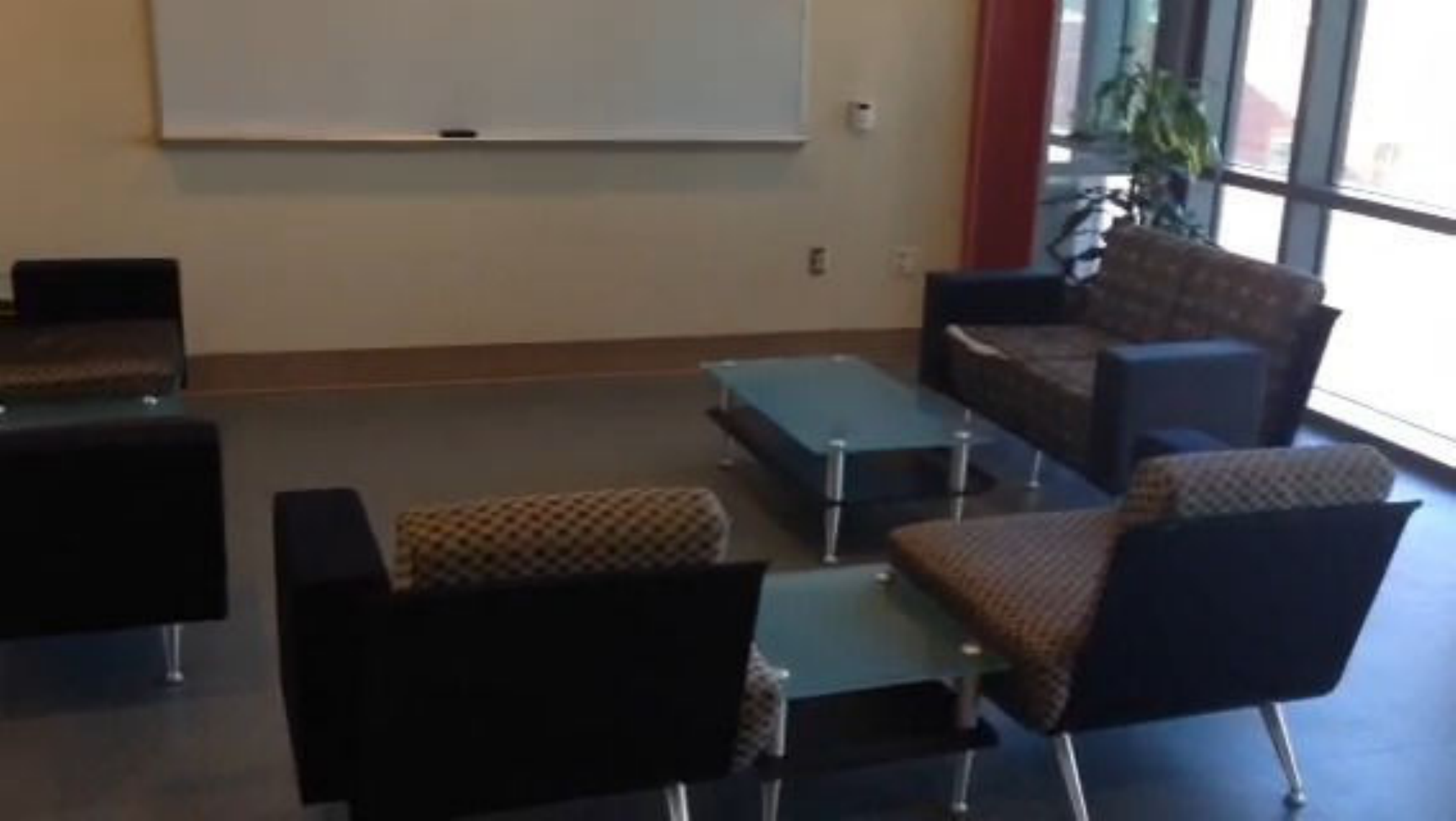}
\includegraphics[width=0.32\linewidth]{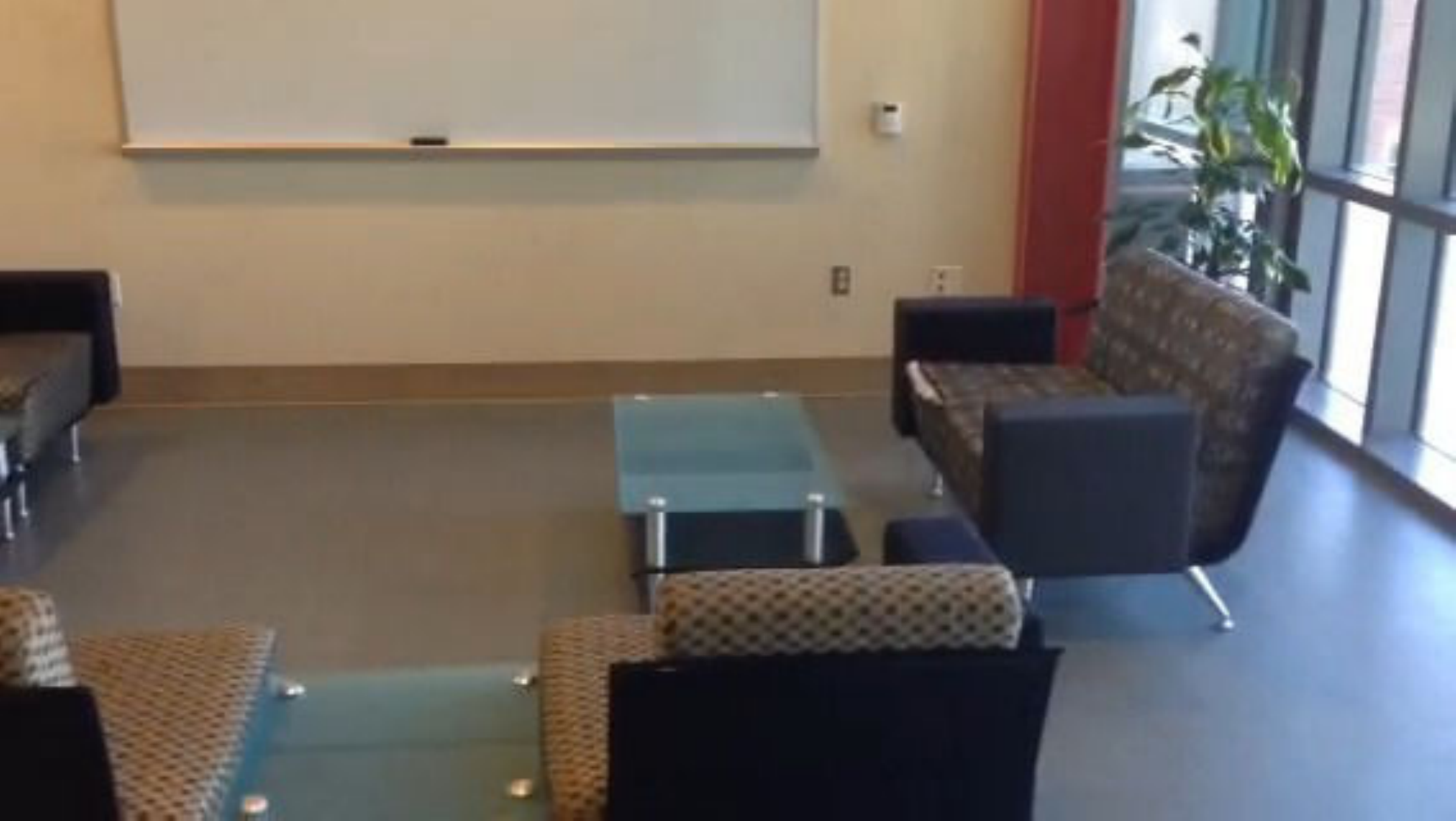}
\\
\includegraphics[width=0.32\linewidth]{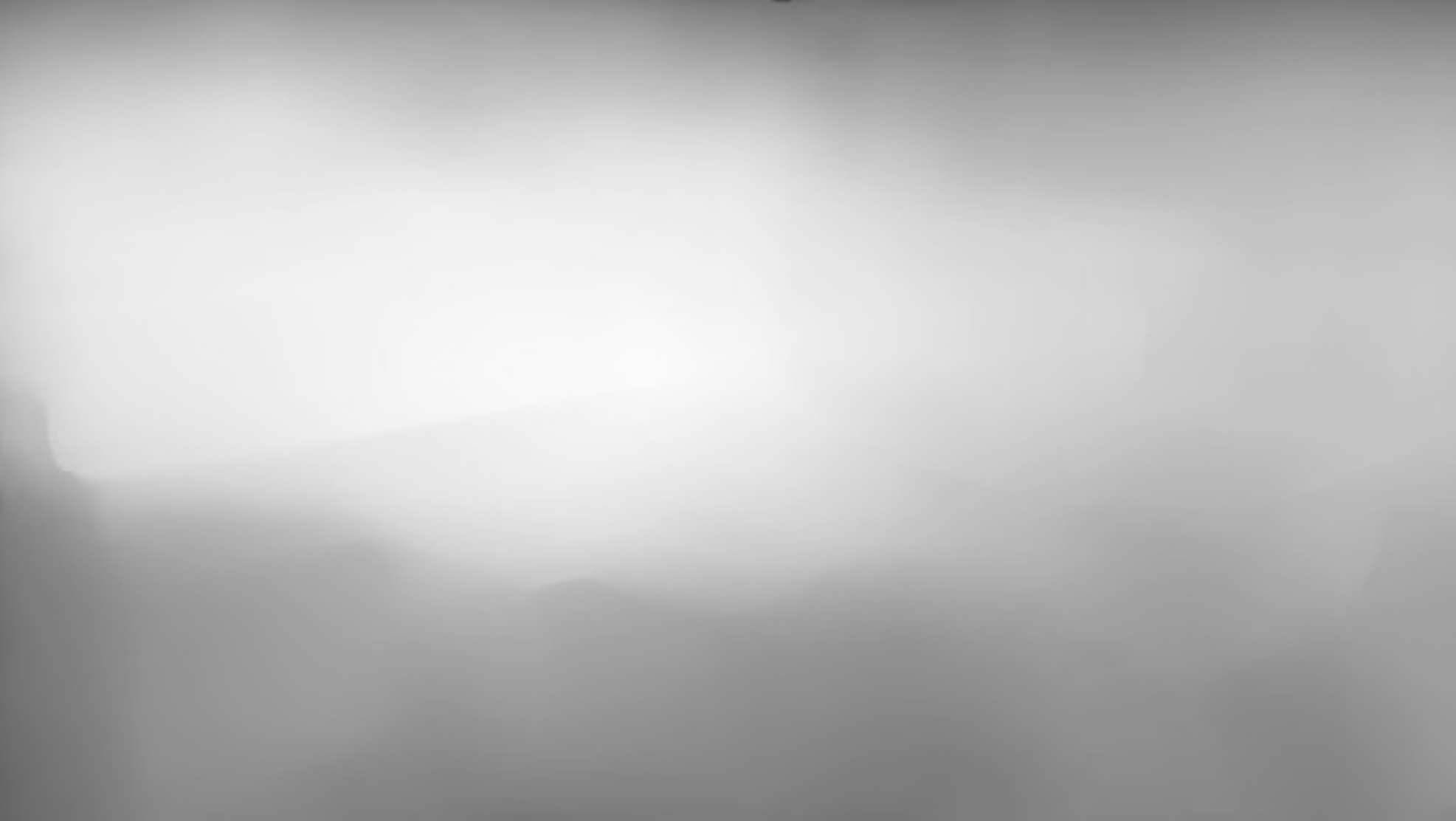}
\includegraphics[width=0.32\linewidth]{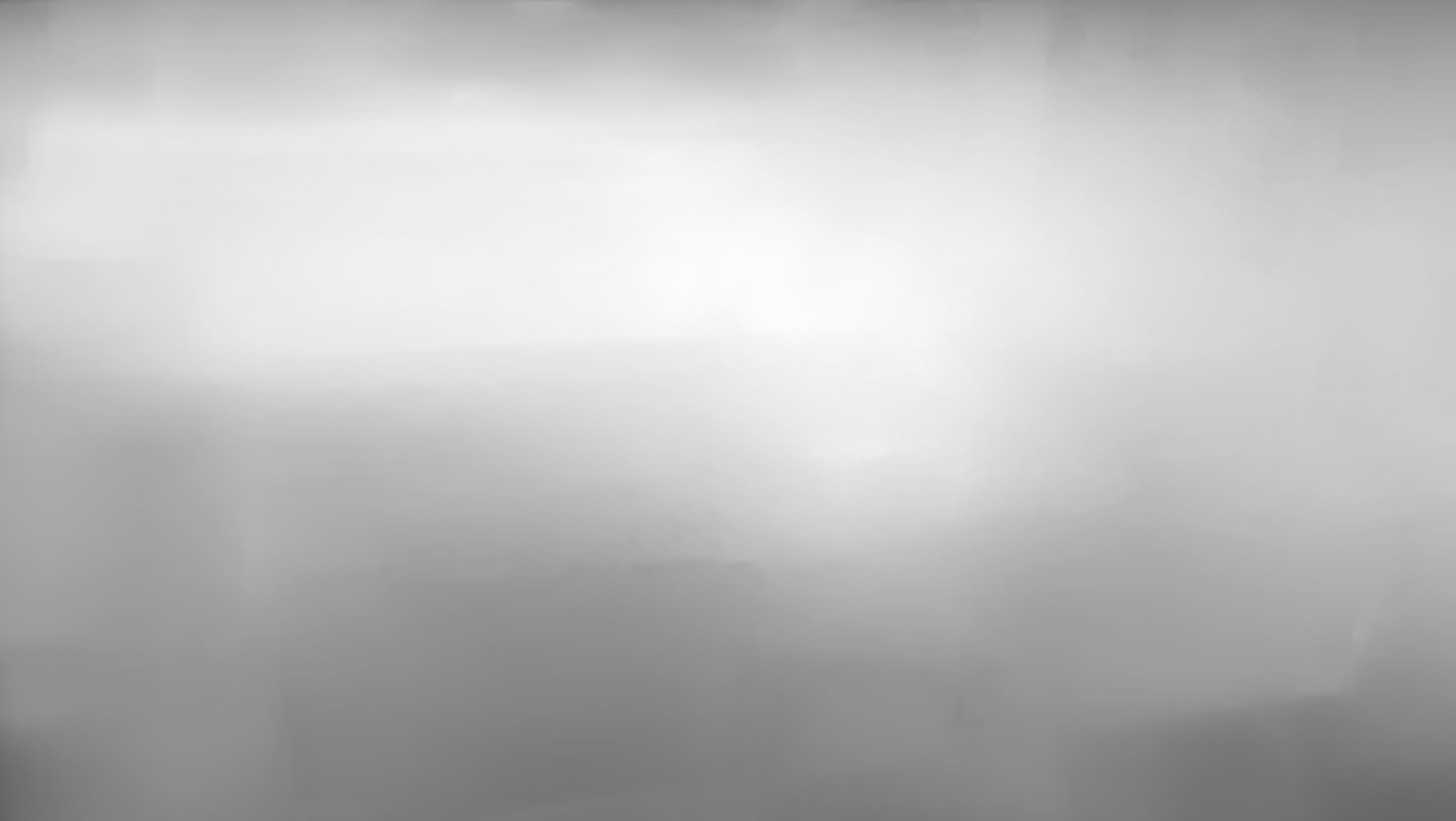}
\includegraphics[width=0.32\linewidth]{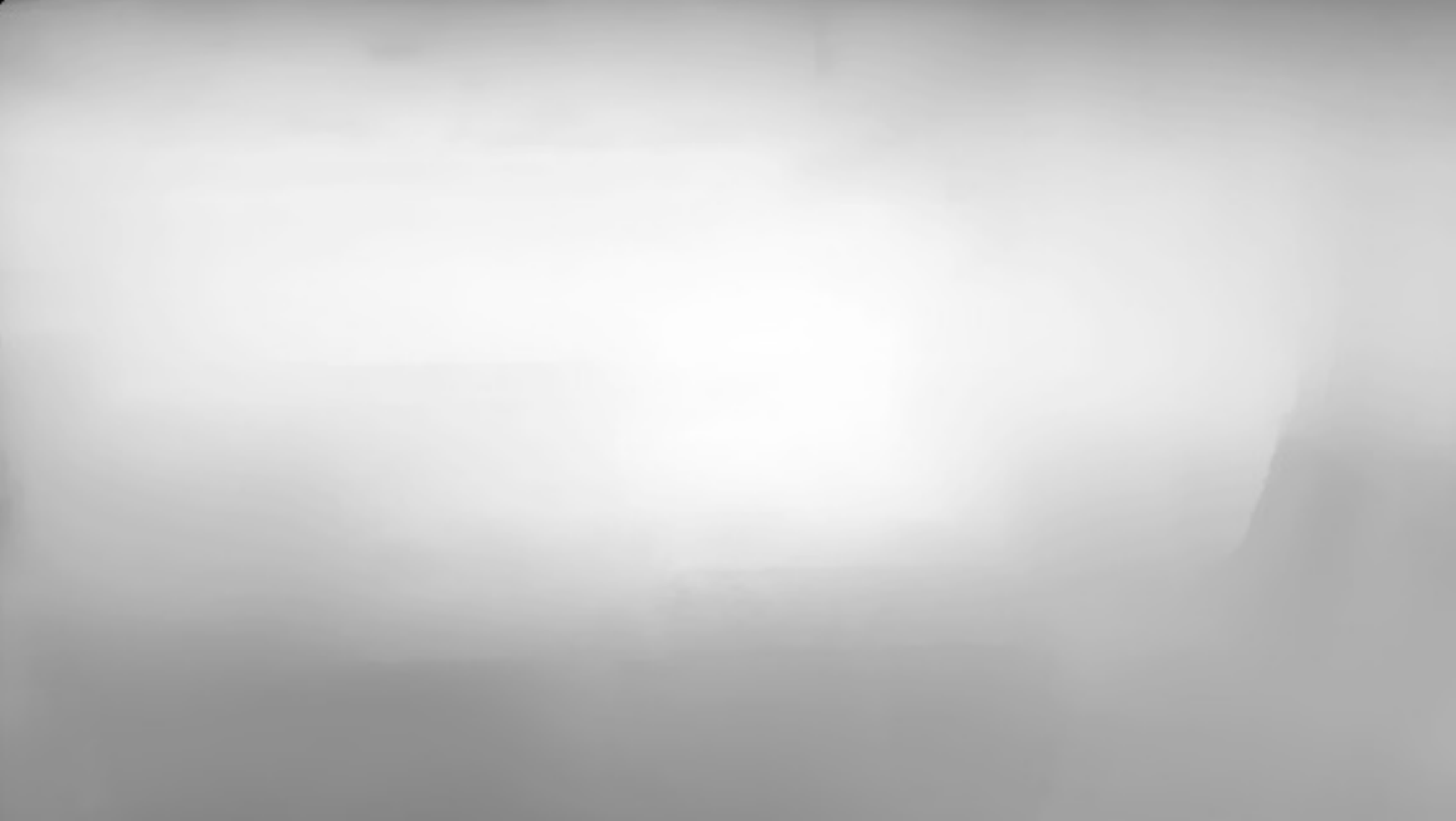}
\\
\includegraphics[width=0.32\linewidth]{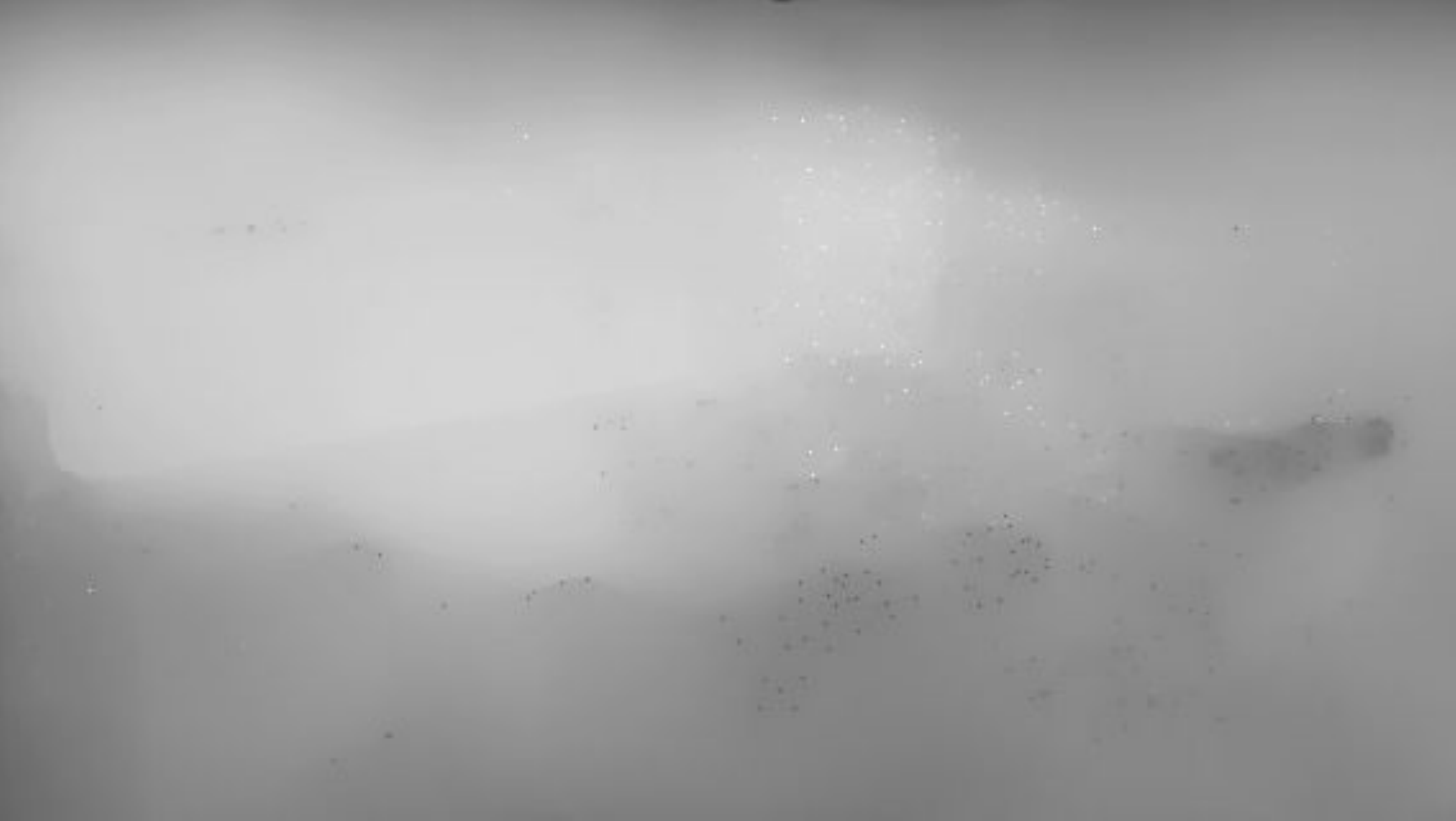}
\includegraphics[width=0.32\linewidth]{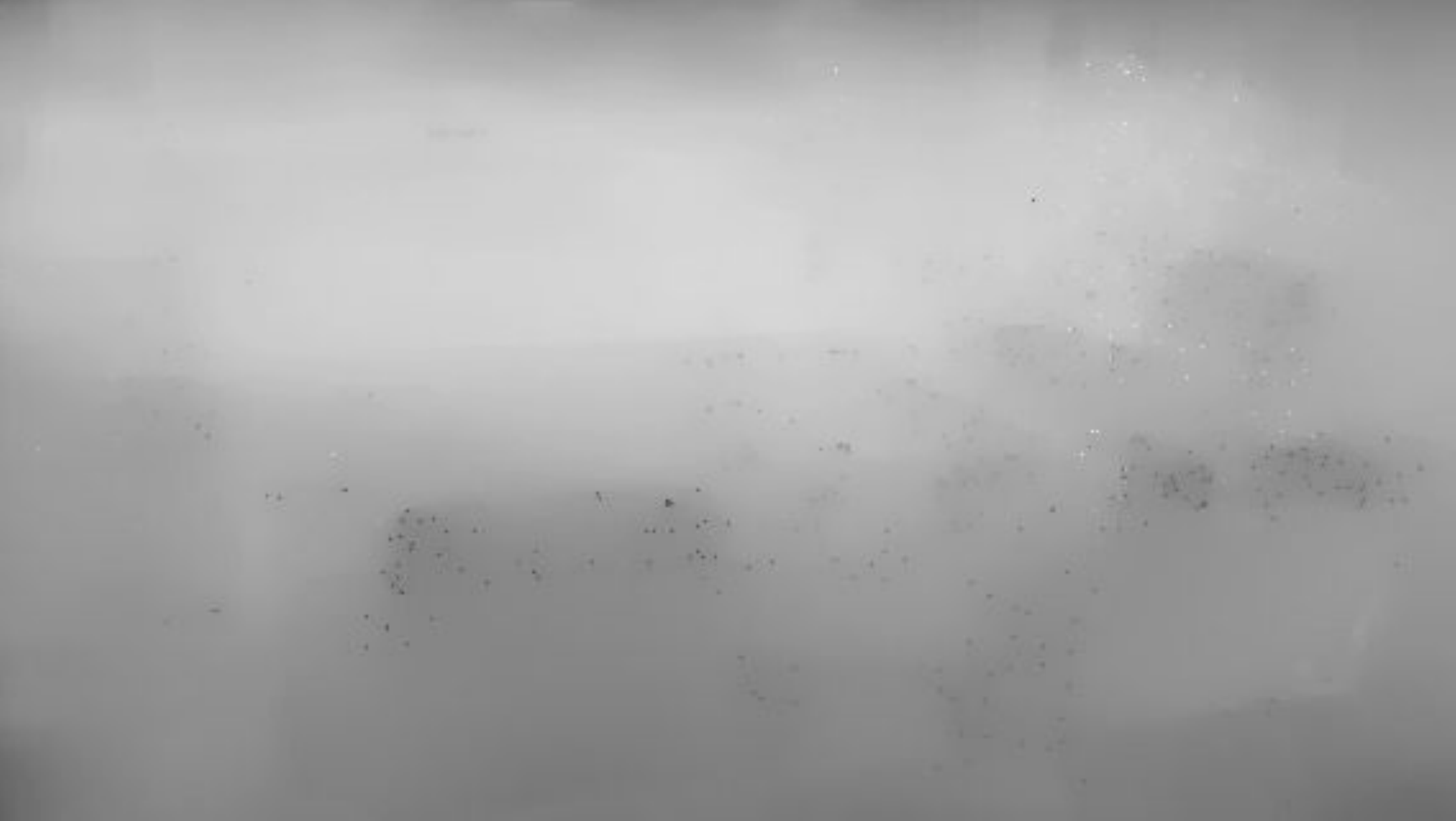}
\includegraphics[width=0.32\linewidth]{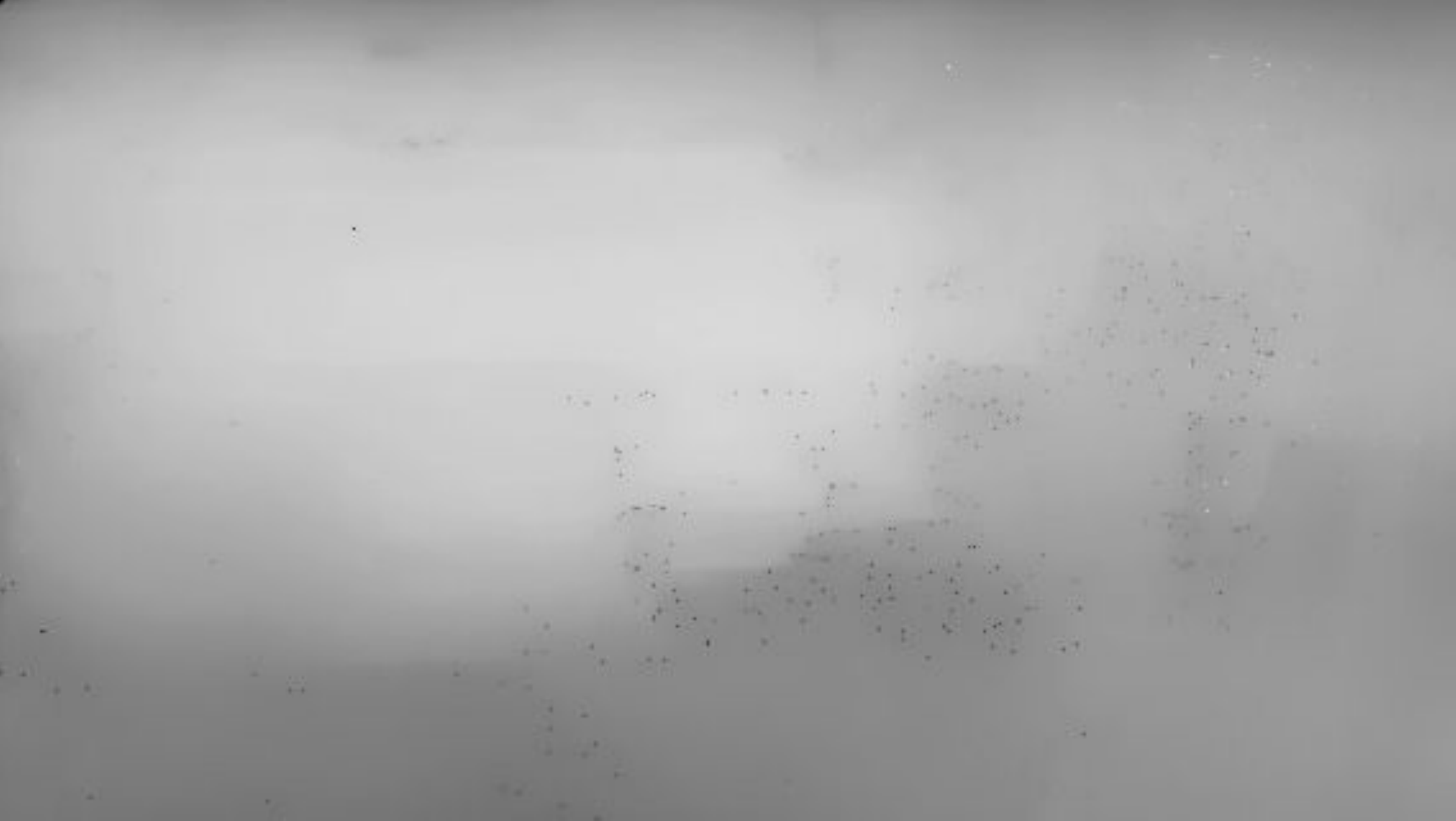}
\\
\includegraphics[width=0.32\linewidth]{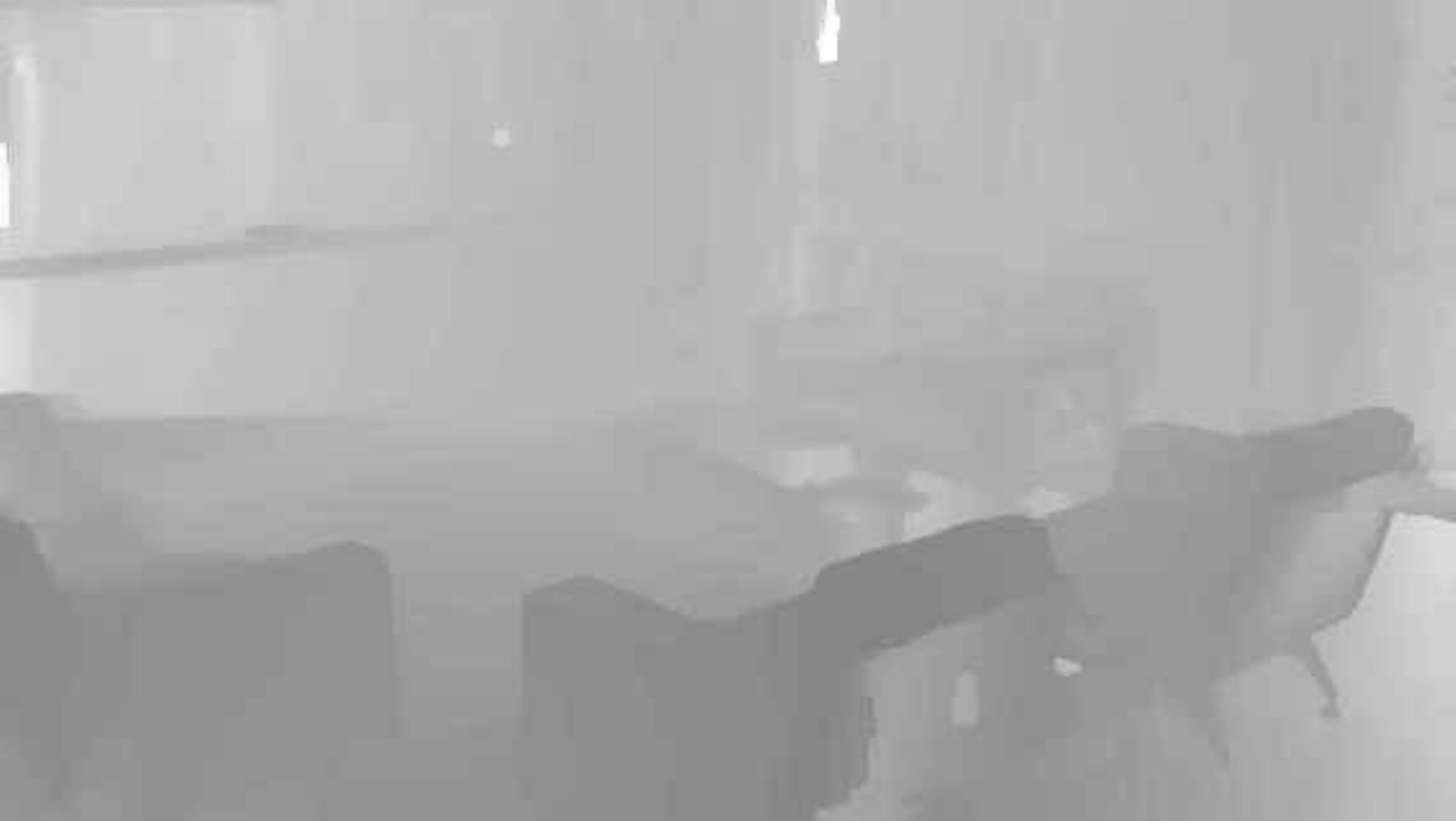}
\includegraphics[width=0.32\linewidth]{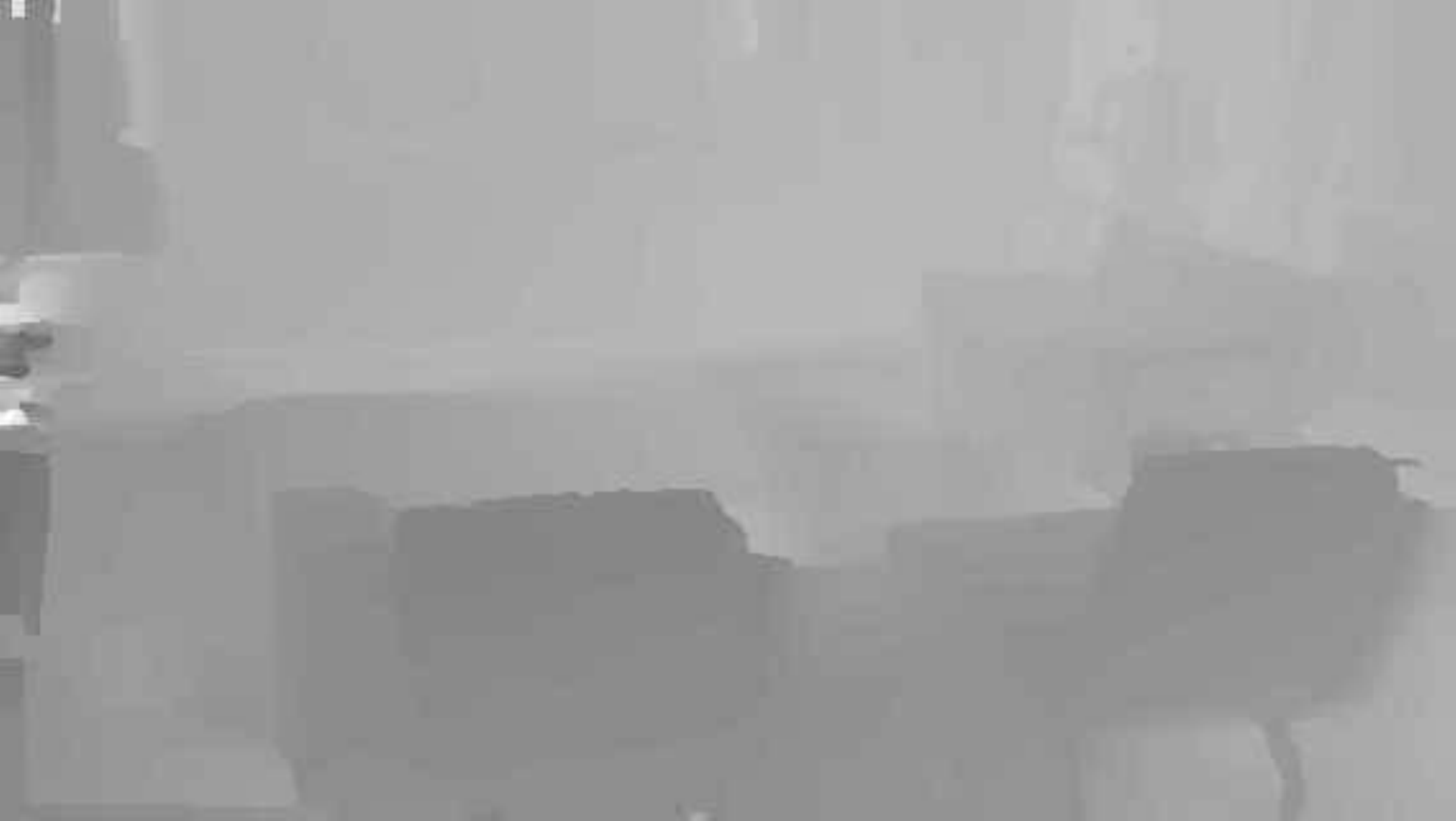}
\includegraphics[width=0.32\linewidth]{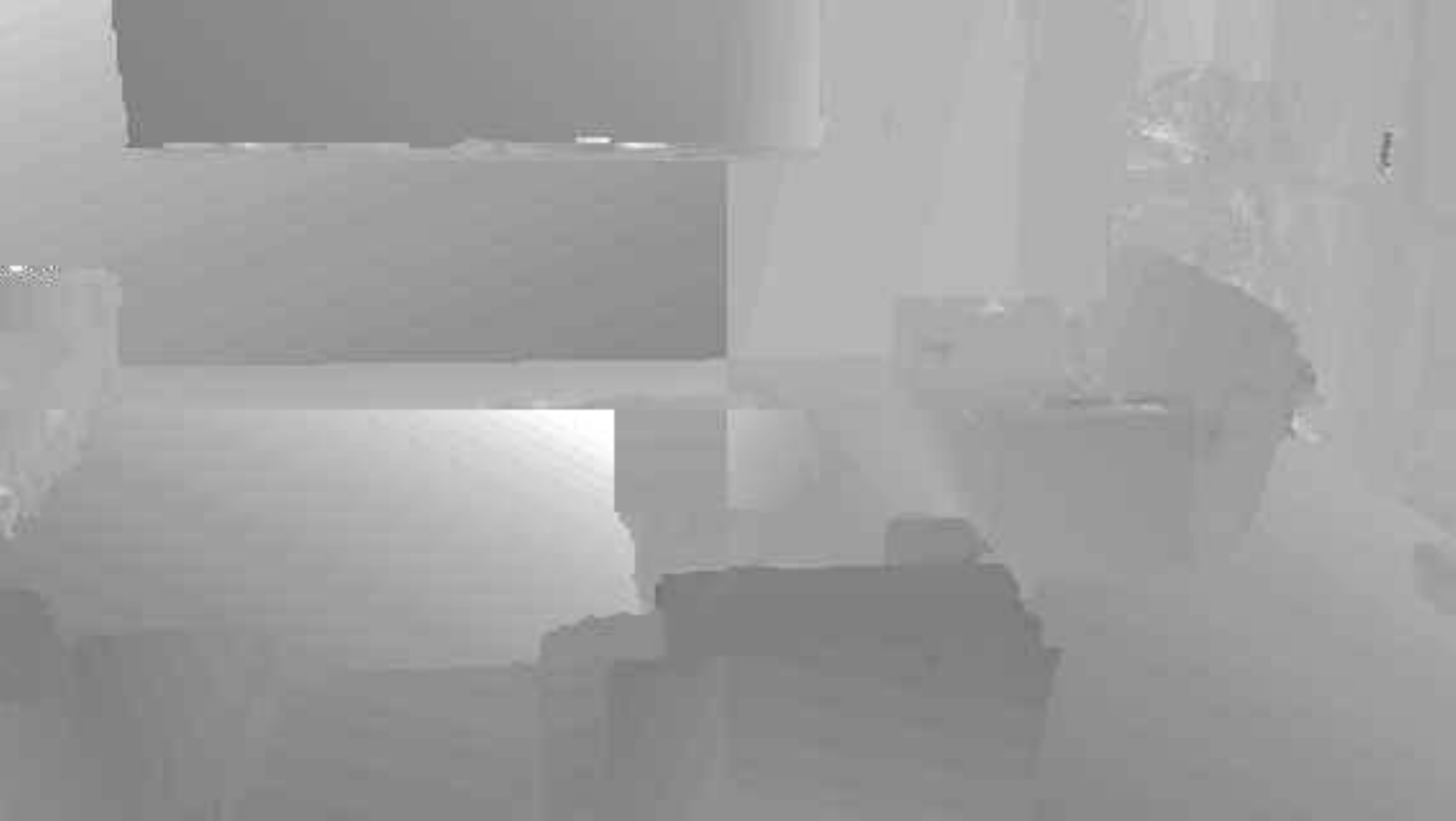}
\end{minipage}
\hfill
\begin{minipage}{0.445\linewidth}
\includegraphics[width=0.24\linewidth]{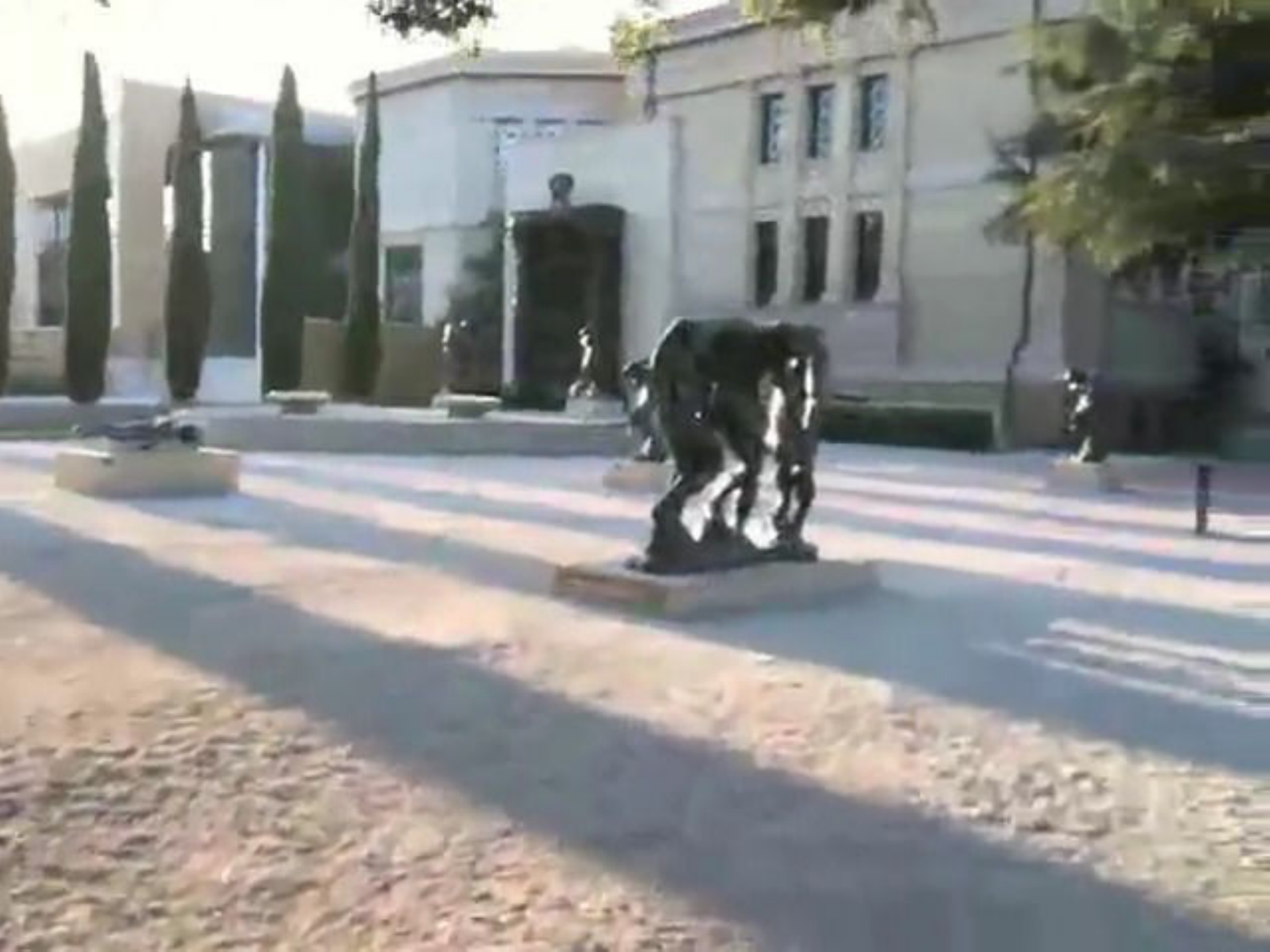}
\includegraphics[width=0.24\linewidth]{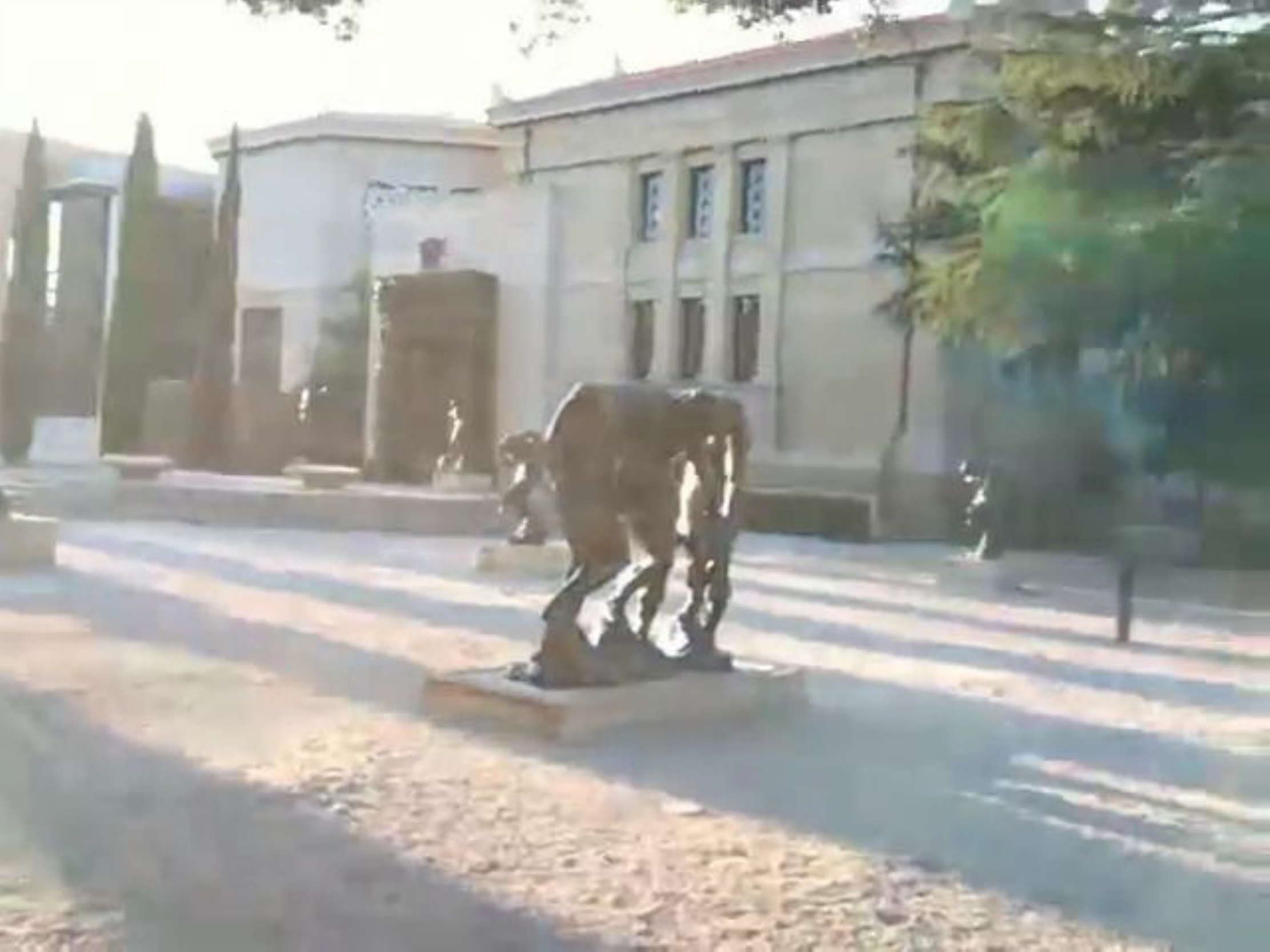}
\includegraphics[width=0.24\linewidth]{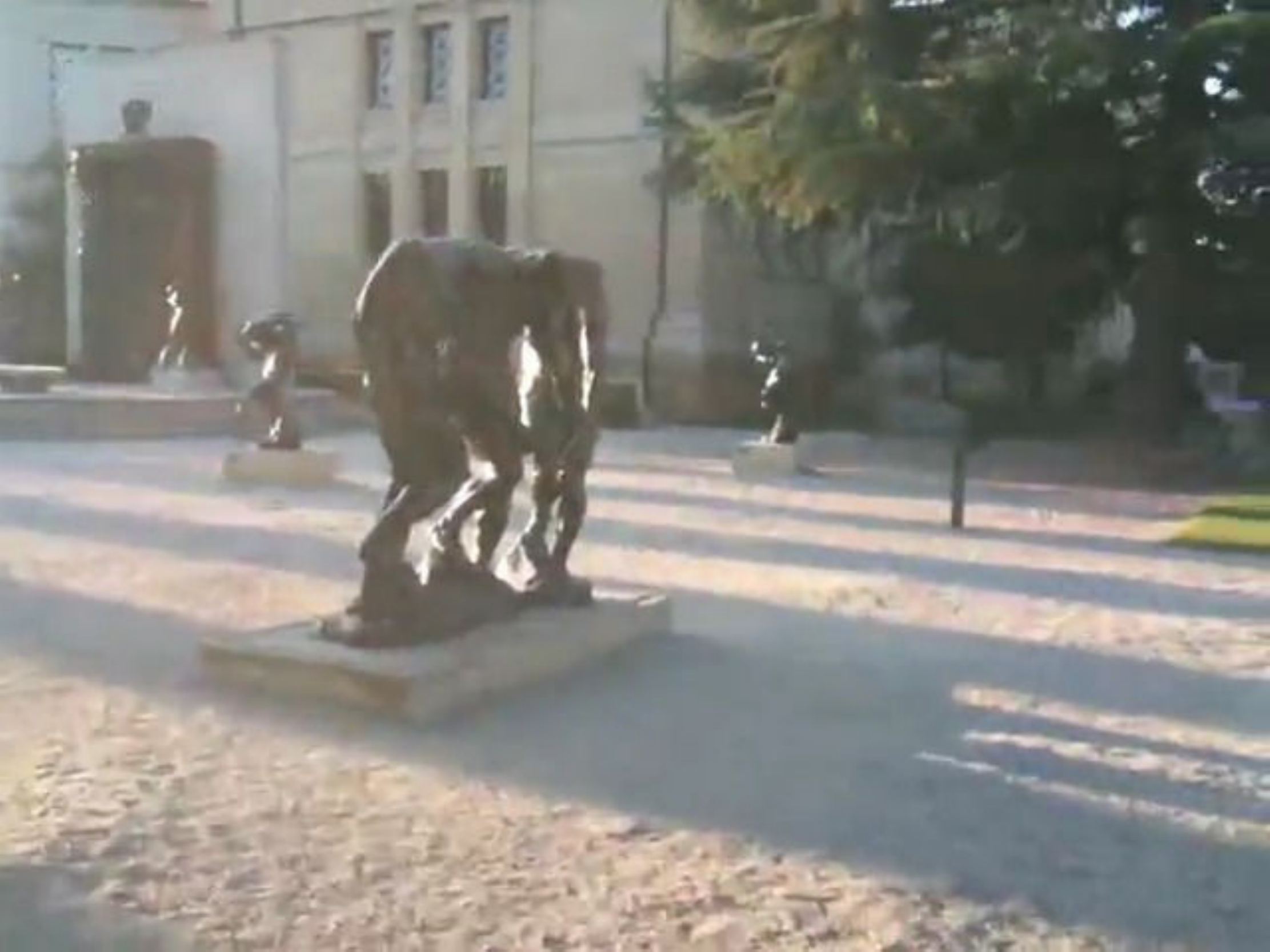}
\includegraphics[width=0.24\linewidth]{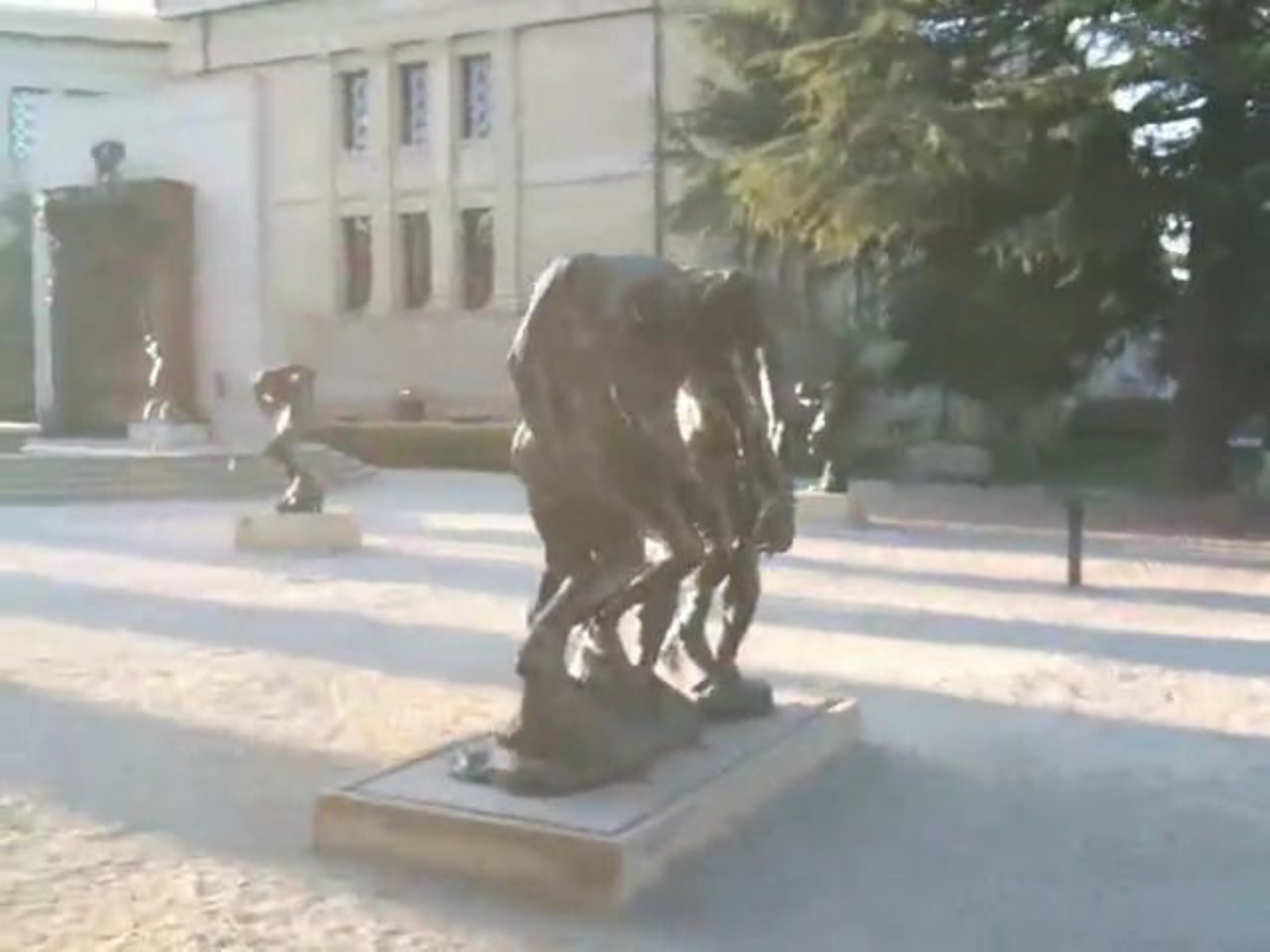}
\\
\includegraphics[width=0.24\linewidth]{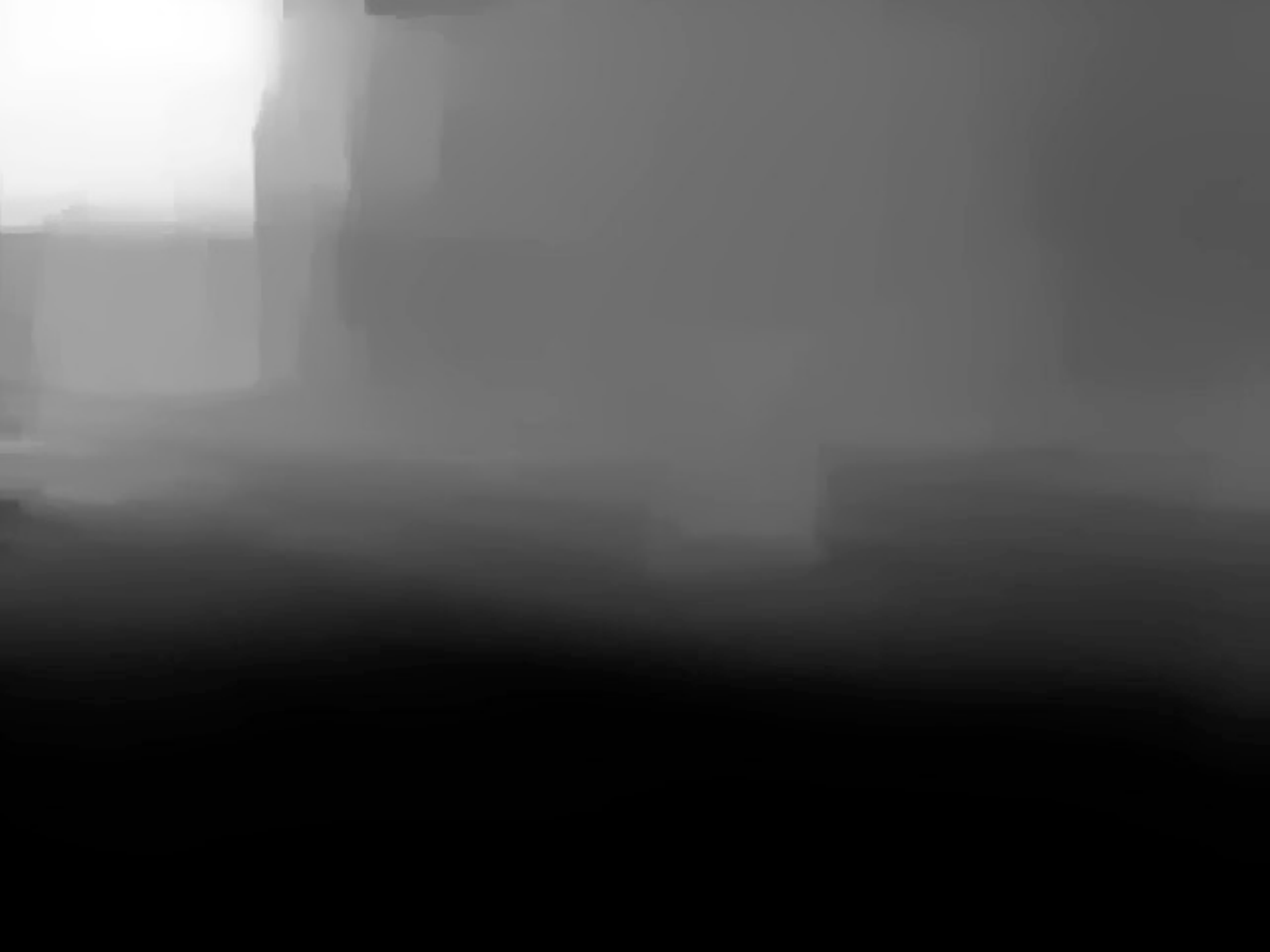}
\includegraphics[width=0.24\linewidth]{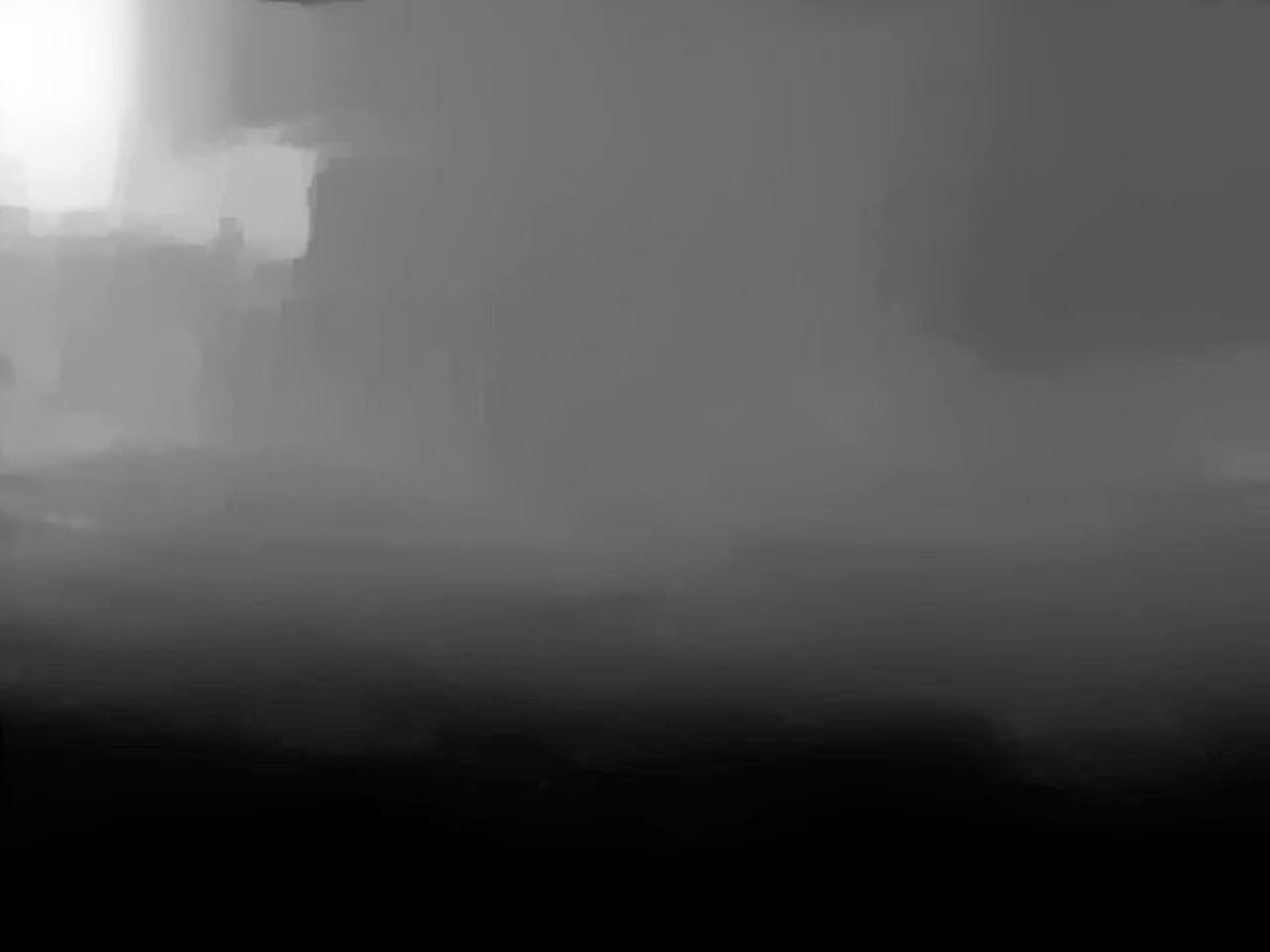}
\includegraphics[width=0.24\linewidth]{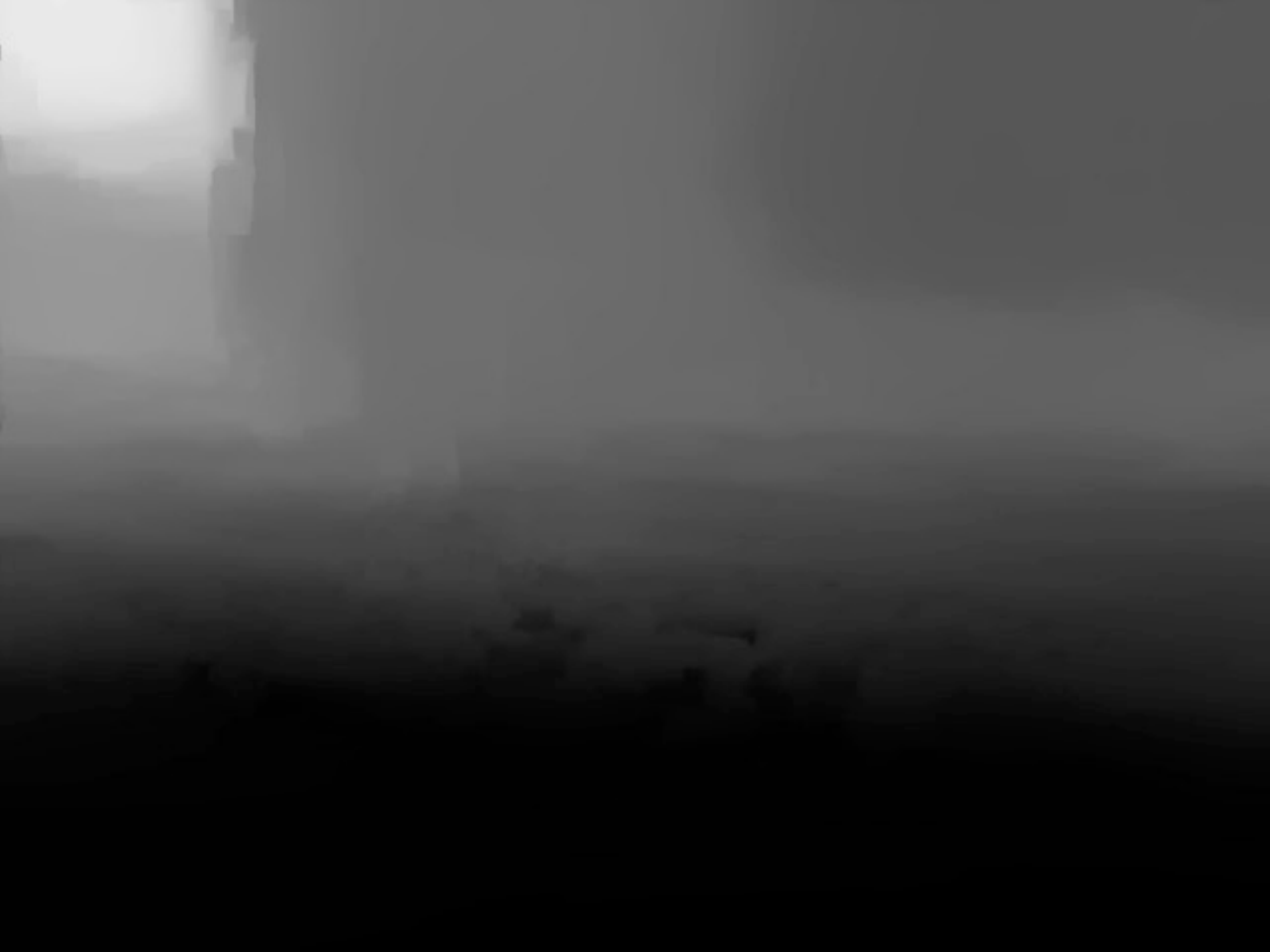}
\includegraphics[width=0.24\linewidth]{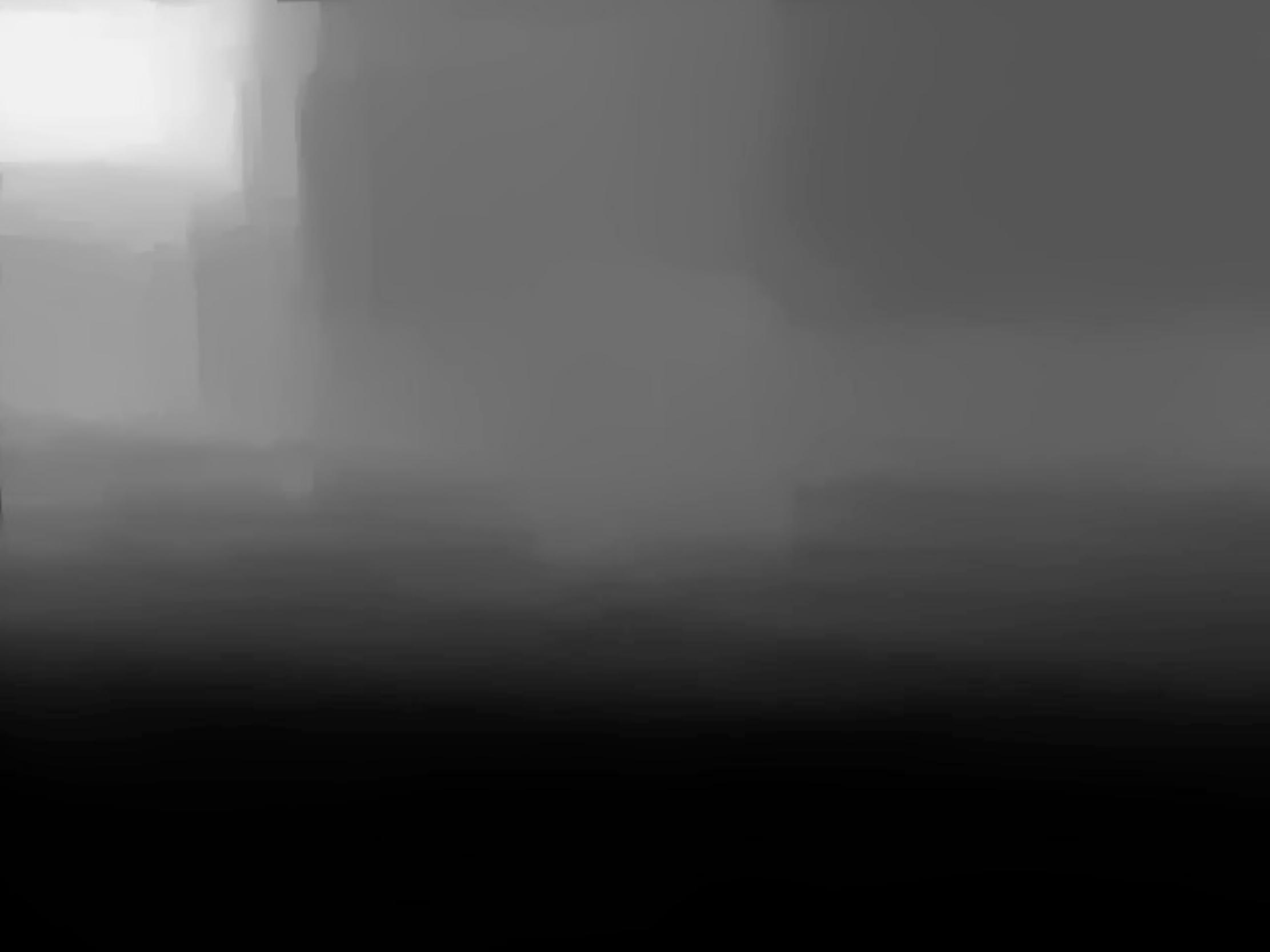}
\\
\includegraphics[width=0.24\linewidth]{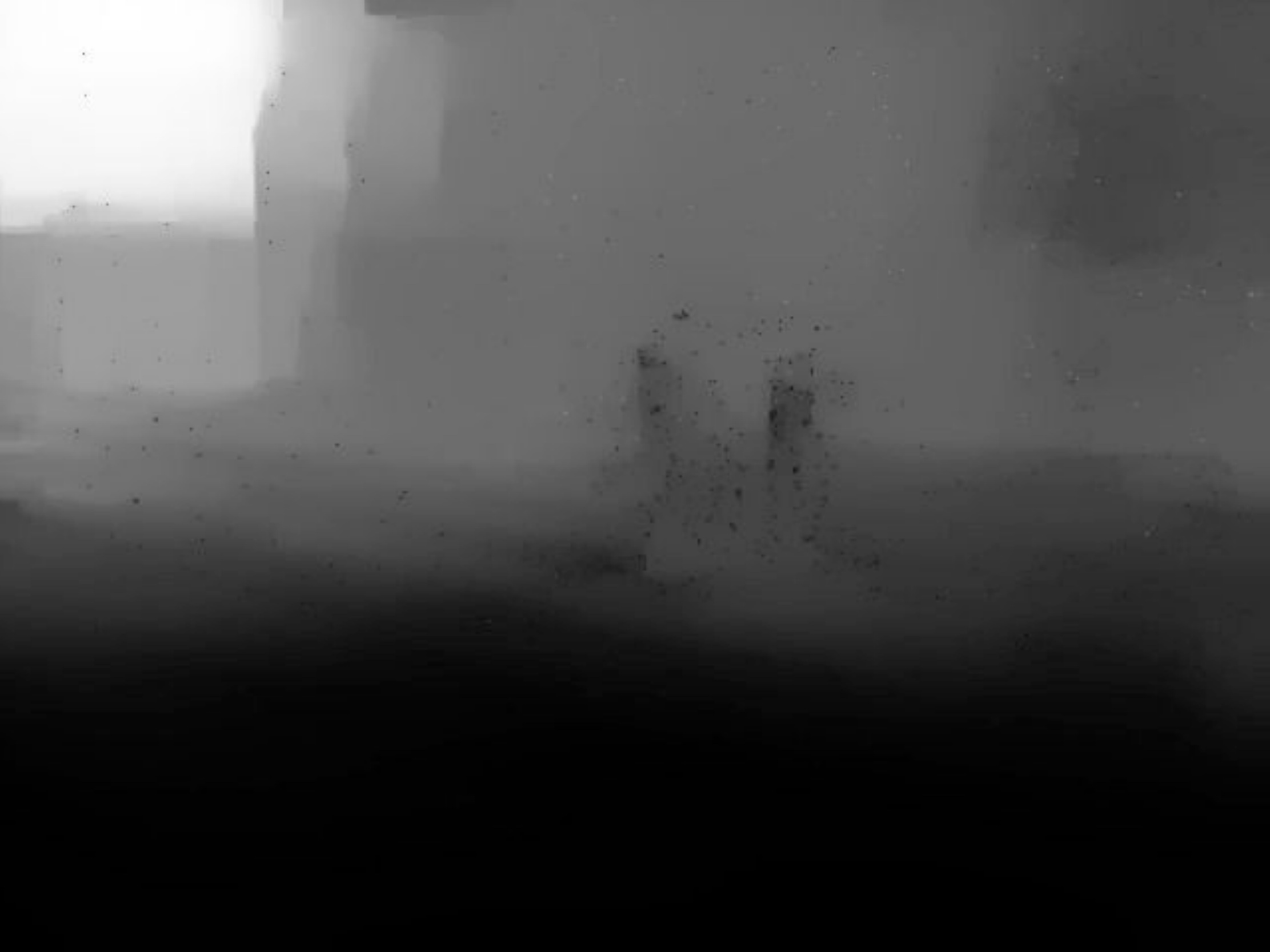}
\includegraphics[width=0.24\linewidth]{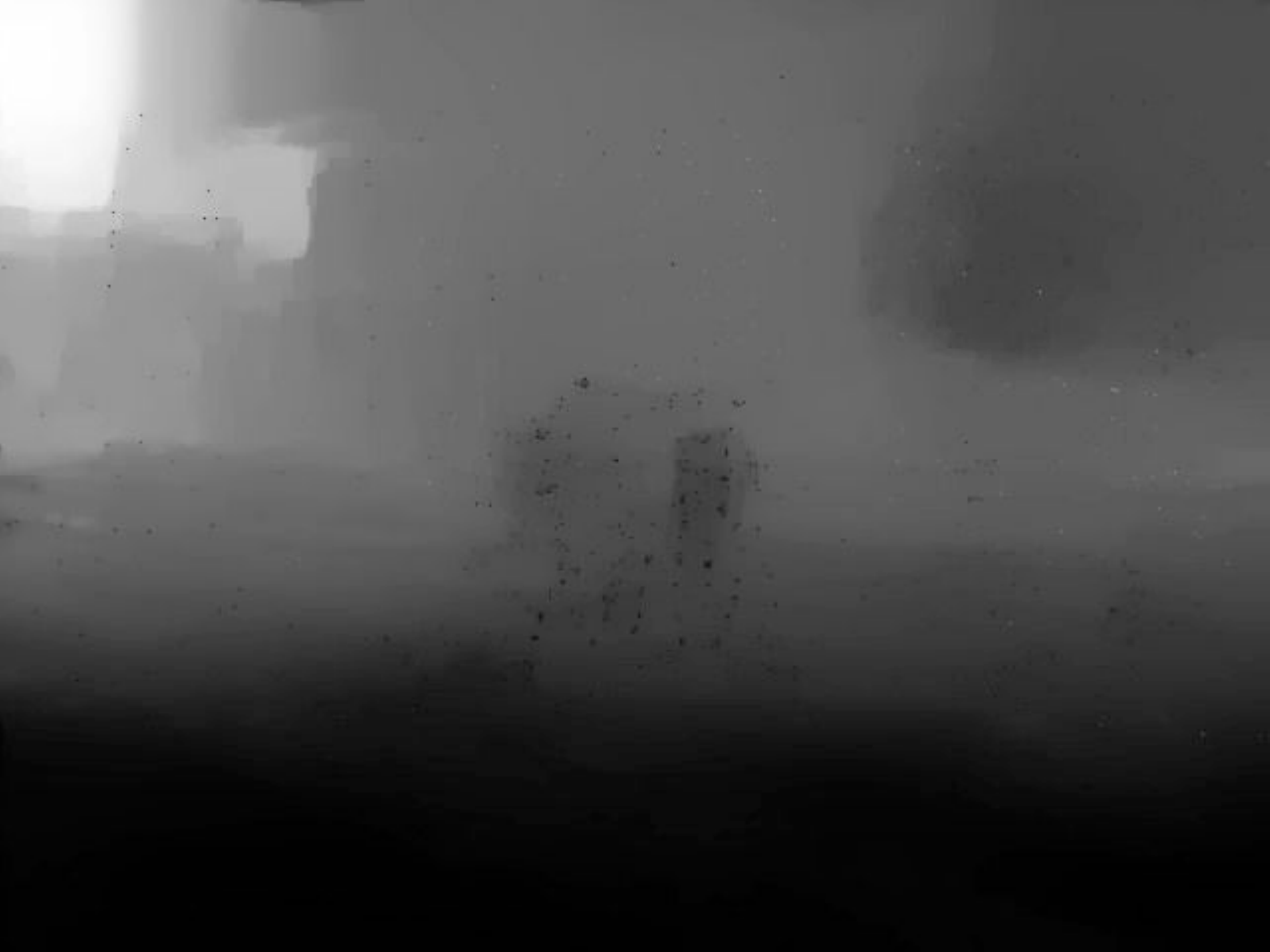}
\includegraphics[width=0.24\linewidth]{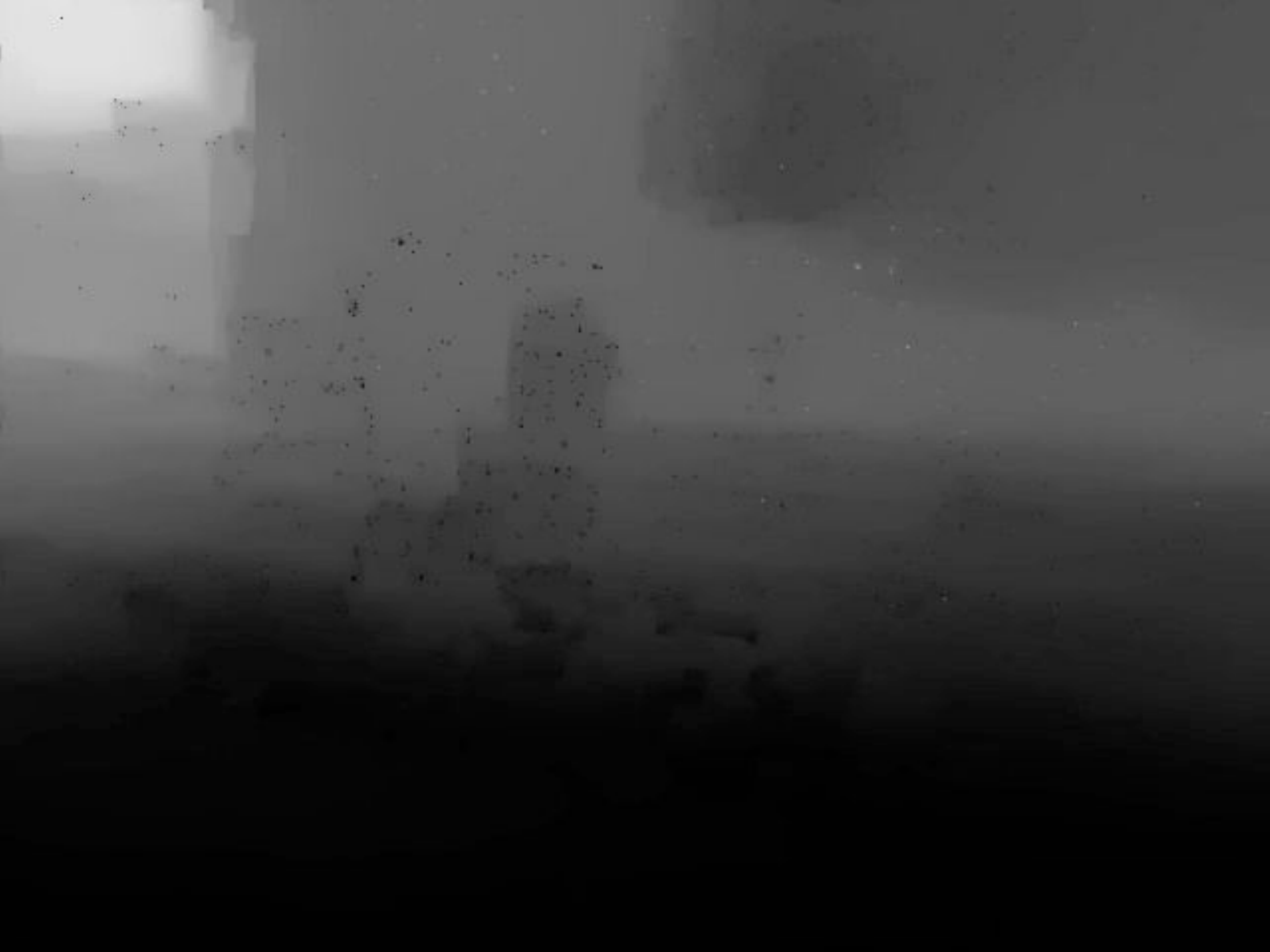}
\includegraphics[width=0.24\linewidth]{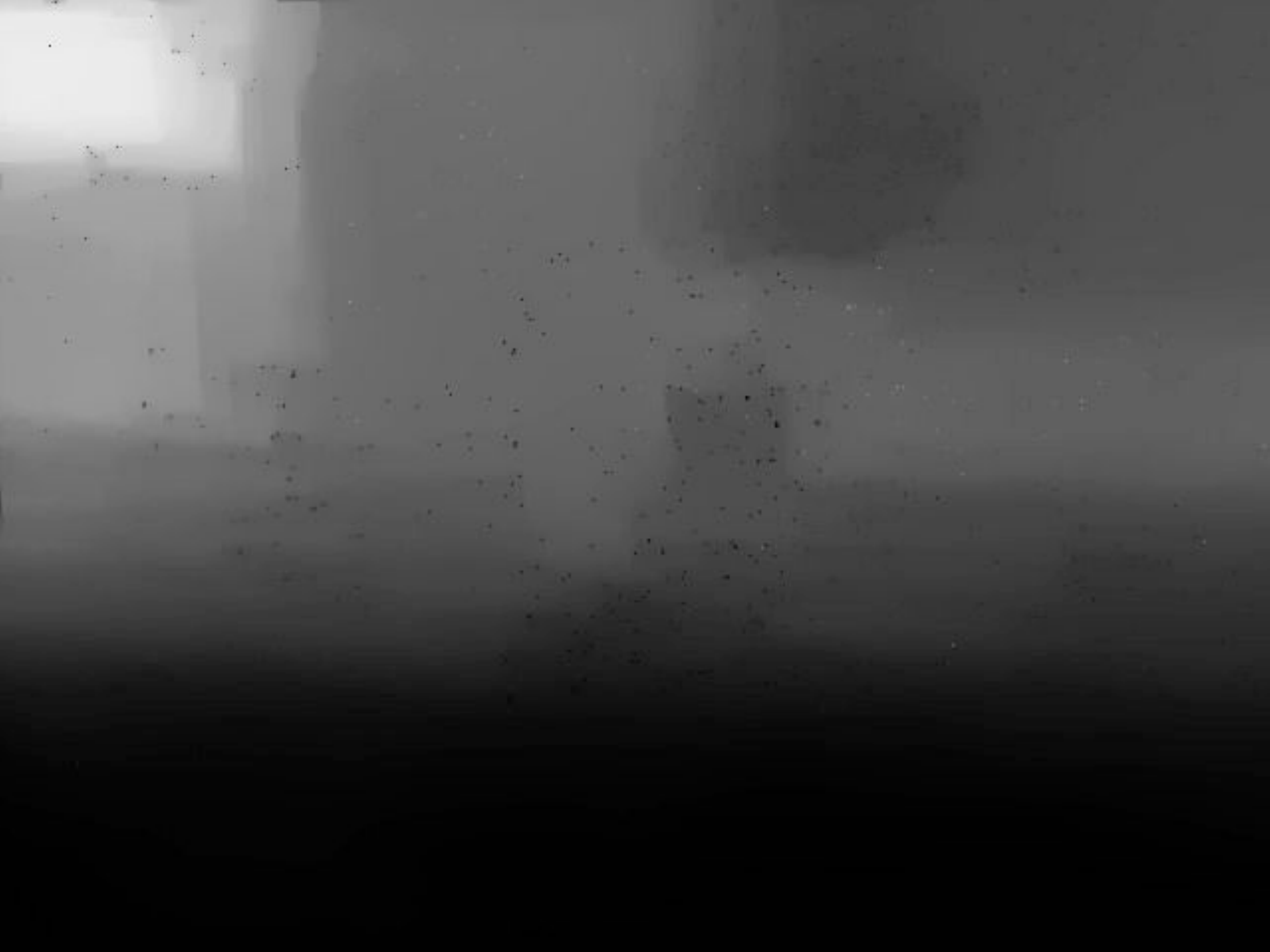}
\\
\includegraphics[width=0.24\linewidth]{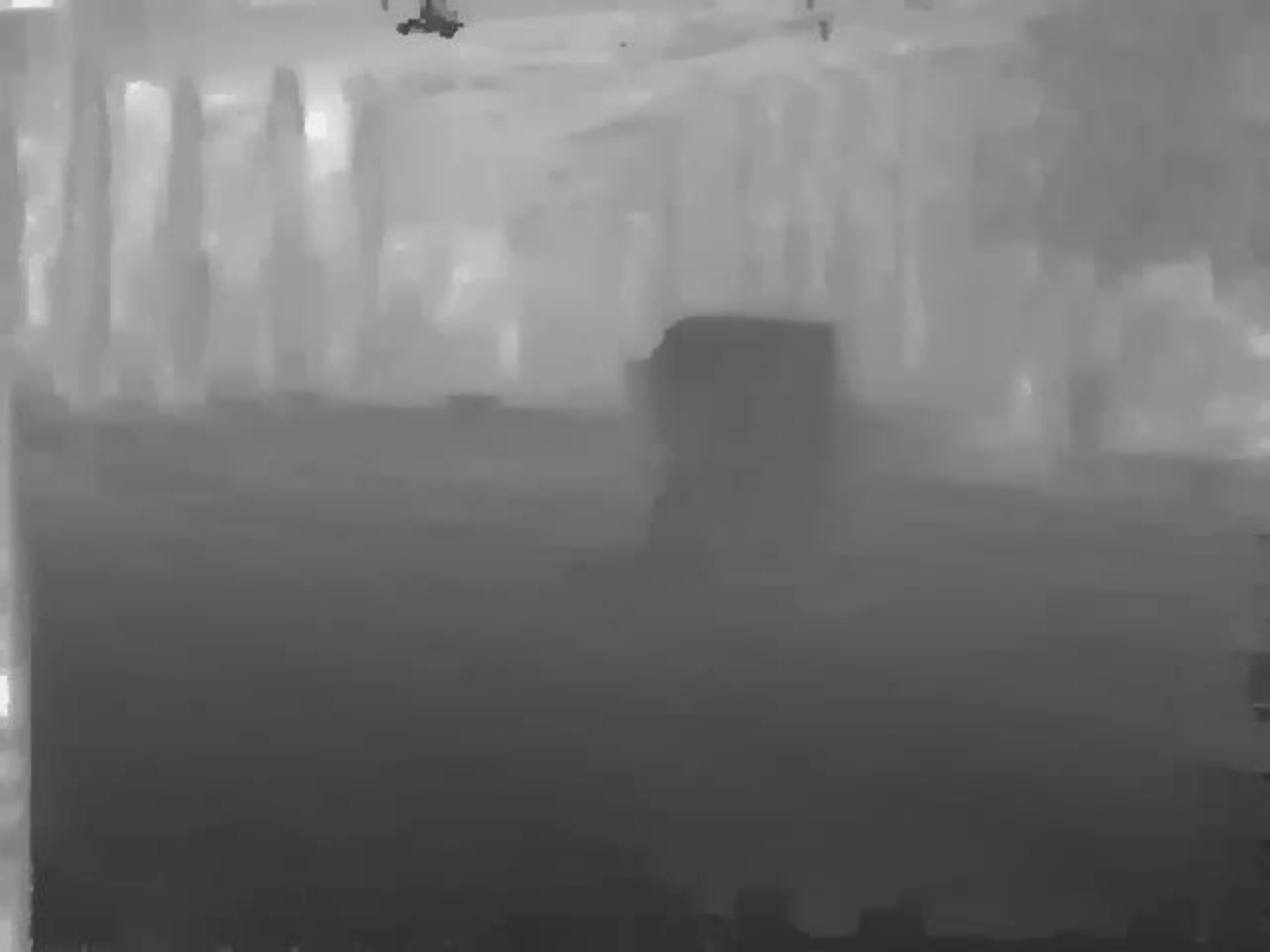}
\includegraphics[width=0.24\linewidth]{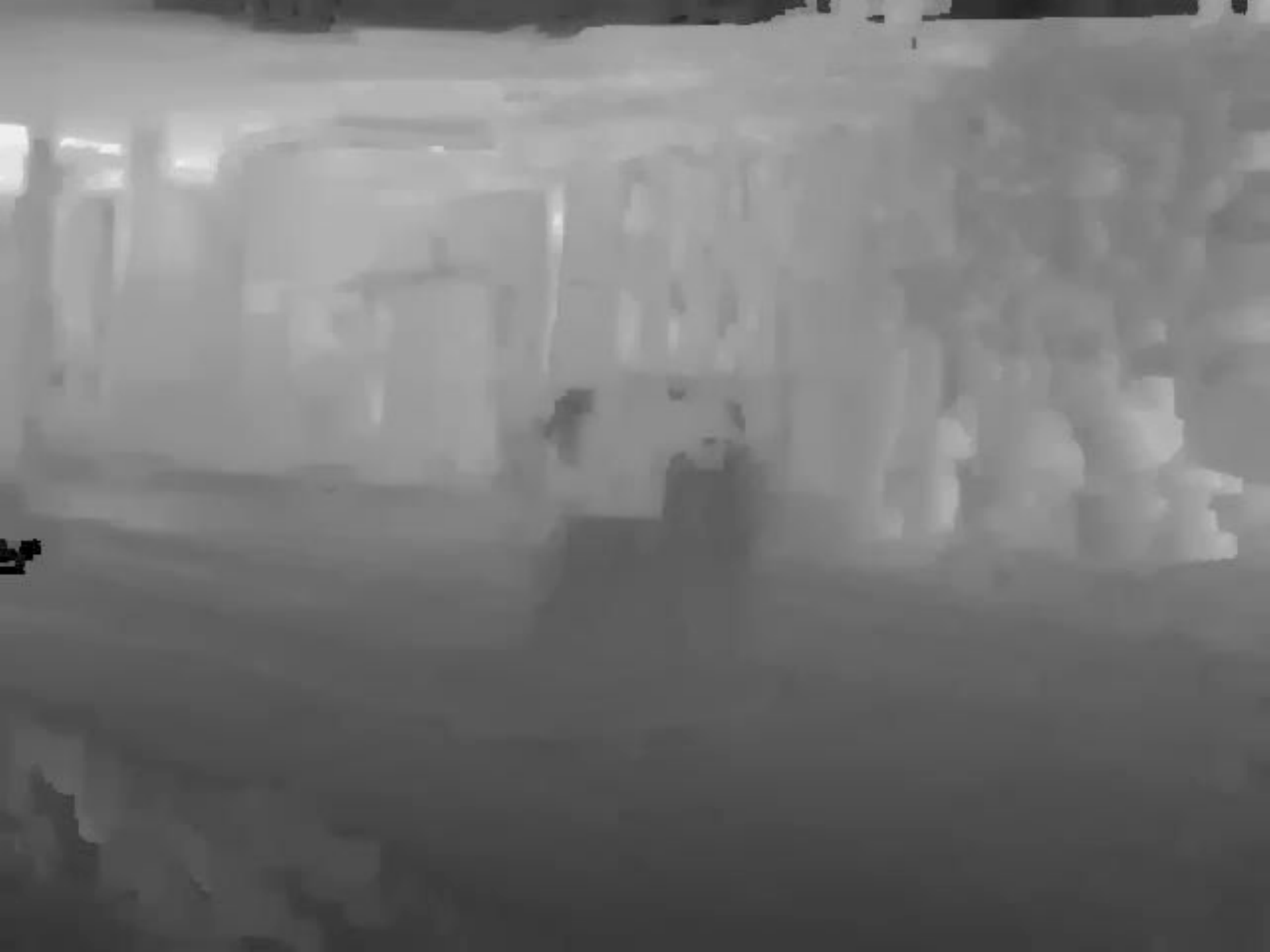}
\includegraphics[width=0.24\linewidth]{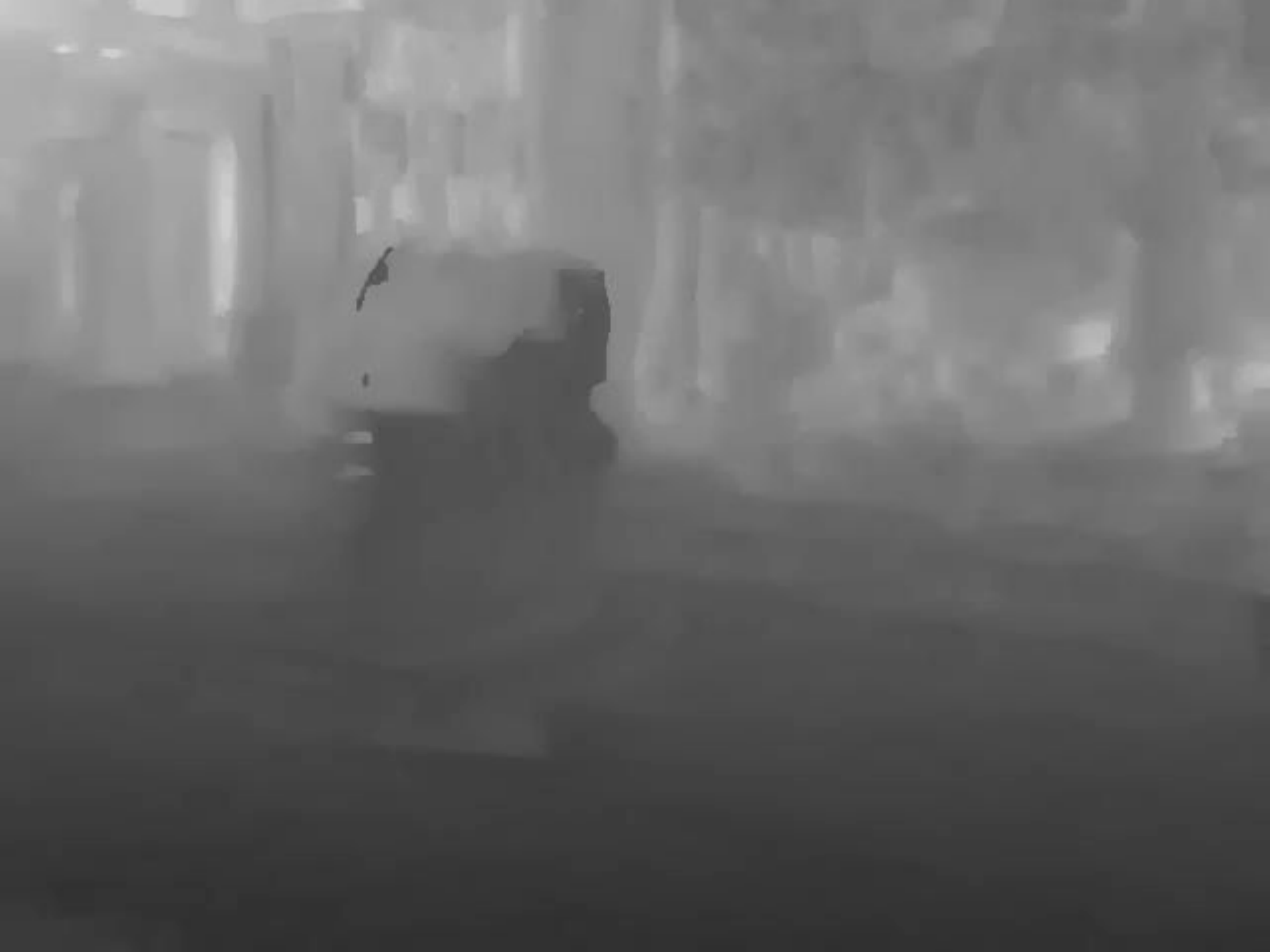}
\includegraphics[width=0.24\linewidth]{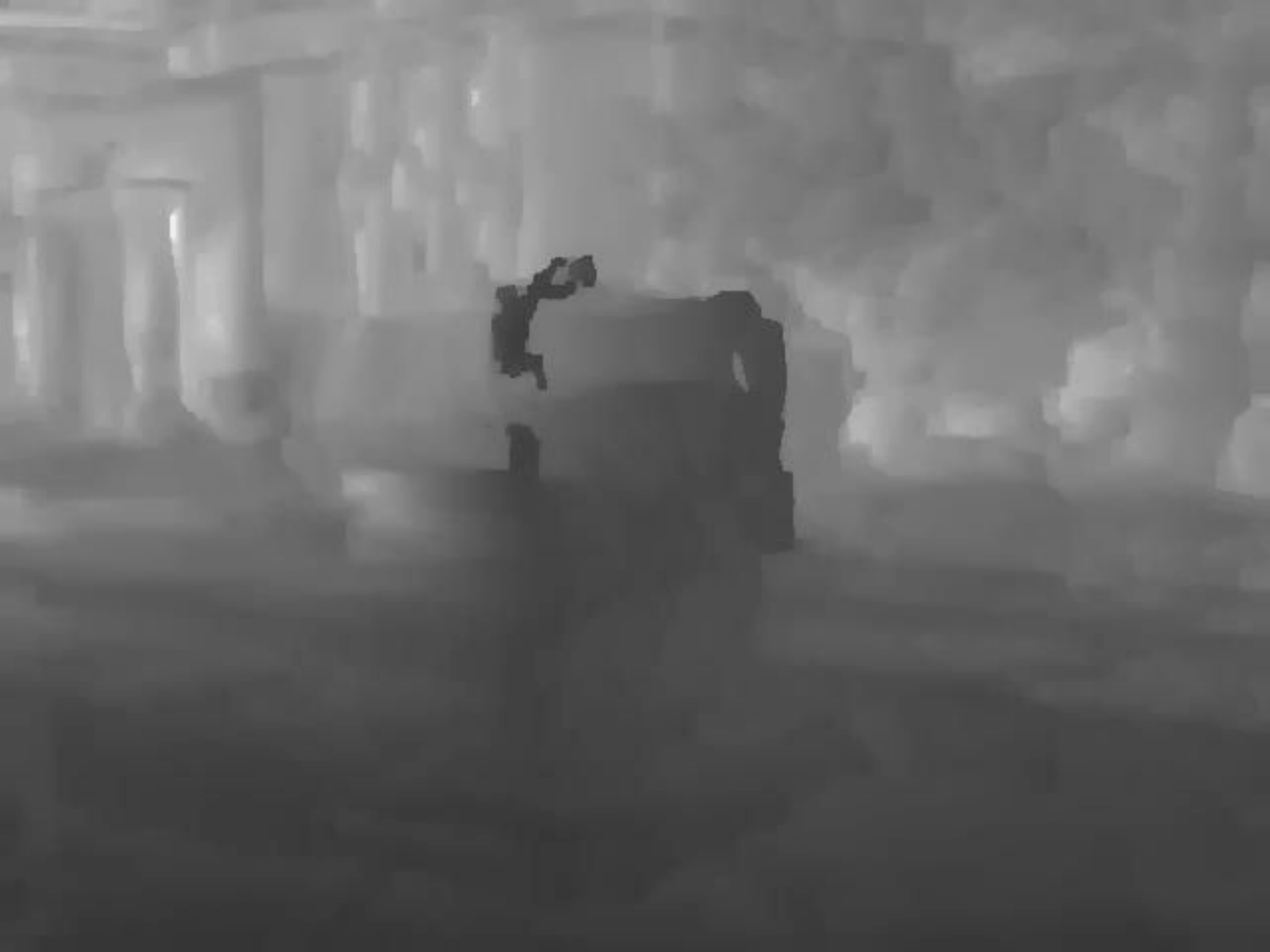}
\end{minipage}
\caption{\newtextRev{
Comparison of our method on videos containing parallax. For each input, we show the depth estimated using our method with no modification, and the depth estimated when our method is bootstrapped with SfM (e.g. a sparse depth map, calculated using SfM, is used as the prior). We also compare these results to the dense depth estimates of Zhang et al.~\cite{Zhang:09}, whose method works only for videos with parallax. When SfM estimates are poor (left sequence), multi-view stereo methods may perform worse than our method, but can be quite good given decent SfM estimates and mostly Lambertian scenes (right sequence). Overall, SfM seems to help estimate relative depth near boundaries, whereas our method seems to better estimate global structure.}
}
\label{fig:parallax}
\end{figure*}
%%%%%%%%%%%%%%%%%

\newtextRev{
One concern of our method may be its reliance on appearance. In Fig~\ref{fig:furniture}, we demonstrate that although appearance partially drives our depth estimates, other factors such as spatial relationships, scale, and orientation necessarily contribute. For example, photographing the same scene with objects in different configurations will lead to different depth estimates, as would be expected.
}

\subsection{Video results}
Our technique works well for videos of many types scenes and video types (Figs~\ref{fig:teaser},~\ref{fig:temporal_benefit},~\ref{fig:motionseg_homography},~\ref{fig:rot_zoom_results},~\ref{fig:indoor_results},~\ref{fig:parallax}). We use the dataset we collected in Sec~\ref{sec:dataset} to validate our method for videos (we know of no other existing methods/datasets to compare to). This dataset contains ground truth depth and stereo image sequences for four different buildings (referred to as Buildings 1, 2, 3, and 4), and to gauge our algorithm's ability to generalize, {\em we only use data from Building 1 for training}. We still generate results for Building 1 by holding each particular example out of the training set during inference.

%%%%%%%%%%%%%%%%%
\begin{table}[t]
\centerline{
\normalsize
\begin{tabular}{| l || c | c | c || c |} \hline
{\bf Dataset} & {\bf rel }& {\bf log$_{10}$} & {\bf RMS} & {\bf PSNR}\\ \hline \hline
% Baseline & X & X & X & X \\ \hline
 Building 1$^{\Diamond}$ & 0.196 & 0.082 & 8.271 & 15.6 \\ \hline
 Building 2 & 0.394 & 0.135 & 11.7 & 15.6 \\ \hline
 Building 3 & 0.325 & 0.159 & 15.0 & 15.0\\ \hline
 Building 4 & 0.251 & 0.136 & 15.3 & 16.4 \\ \hline
 Outdoors$^{\Diamond\Diamond}$ & - & - & - & 15.2 \\ \hline \hline
 All & 0.291 & 0.128 & 12.6 & 15.6 \\ \hline
 \end{tabular}
 }
 \vspace{2mm}
\caption{Error averaged over our stereo-RGBD dataset.
%Ground truth and inferred depths are scaled to match the global range of the Make3D depth data (1-81m) to make these results comparable to our single image results.
%We use PSNR to measure the similarity of our synthesized right view with ground truth.
${}^{\Diamond}$Building used for training (results for Building 1 trained using a hold-one-out scheme). $^{\Diamond\Diamond}$No ground truth depth available.
}
 \label{tab:results_kinect}
\end{table}
%%%%%%%%%%%%%%%%%

We show quantitative results in Table~\ref{tab:results_kinect} and qualitative results in Fig~\ref{fig:indoor_results}. We calculate error using the same metrics as in our single image experiments, and to make these results comparable with Table~\ref{tab:make3d_results}, we globally rescale the ground truth and inferred depths to match the range of the Make3D database (roughly 1-81m).
%Our video extension provides a significant boost in performance, outperforming in each of the three metrics we evaluated our single image algorithm on by a large margin.
As expected, the results from Building 1 are the best, but our method still achieves reasonable errors for the other buildings as well.

Fig~\ref{fig:indoor_results} shows a result from each building in our dataset (top left is Building 1). As the quantitative results suggest, our algorithm performs very well for this building. In the remaining examples, we show results of videos captured in the other three buildings, all which contain vastly different colors, surfaces and structures from the Building 1. Notice that even for these images our algorithm works well, as evidenced by the estimated depth and 3D images.

\newtextRev{
We demonstrate our method on videos containing parallax in Fig~\ref{fig:parallax}. Such videos can be processed with structure from motion (SfM) and multi-view stereo algorithms; e.g. the dense depth estimation method of Zhang et al.~\cite{Zhang:09}. We visually compare our method to the method of Zhang et al., as well as a version of our method where the depth prior comes from a sparse SfM point cloud. Specifically, we solve for camera pose, track and triangulate features using publicly available code from Zhang et al. Then, we project the triangulated features into each view and compute their depth; this sparse depth map is used as the prior term in Eq~\ref{eq:prior}, and for videos with parallax, we increase the prior weight ($\beta=10^3$) and turn off motion segmentation ($\eta=0$). We note that in most cases of parallax, a multi-view stereo algorithm is preferable to our solution, yet in some cases our method seems to perform better qualitatively.
}

As further evaluation, a qualitative comparison between our technique and the publicly available version of Make3D is shown in Fig~\ref{fig:make3d_comparison}. Unlike Make3D, our technique is able to extract the depth of the runner throughout the sequence, in part because Make3D does not incorporate temporal information.

Our algorithm also does not require video training data to produce video results. We can make use of static RGBD images (e.g., Make3D dataset) to train our algorithm for video input, and we show several outdoor video results in Figs~\ref{fig:teaser},~\ref{fig:temporal_benefit},~\ref{fig:motionseg_homography}, and~\ref{fig:parallax}. Even with static data from another location, our algorithm is usually able to infer accurate depth and stereo views.

Since we collected ground truth stereo images in our dataset, we also compare our synthesized right view (from our 2D-to-3D algorithm, see Sec.~\ref{sec:stereoest}) with the actual right view. We use peak signal-to-noise ratio (PSNR) to measure the quality of the reconstructed views, as shown in Table~\ref{tab:results_kinect}. We could not acquire depth outdoors, and we use this metric to compare our outdoor and indoor results.

\newtext{

%%%%%%%%%%%%%%%%%
%\begin{table}[t]
%\centerline{
%\begin{tabular}{| l || c | c | c |} \hline
%$k$ & {\bf rel }& {\bf log$_{10}$} & {\bf RMS} \\ \hline \hline
% 1 & 0.522 & 0.182 & 17.4 \\ \hline
% 3 & 0.413 & 0.157 & 15.9 \\ \hline
% 7 & {\bf 0.361} & {\bf 0.148} & {\bf 15.1} \\ \hline
% 11 & 0.362 & {\bf 0.148} & 15.2 \\ \hline
% 15 & 0.369 & 0.150 & 15.3 \\ \hline
% \end{tabular}
% }
%\vspace{2mm}
%\caption{Effect of the number of chosen candidates ($K$). Errors are reported on the Make3D dataset with varying values of $K$. For this dataset, $K=7$ is optimal, but increasing $K$ beyond 7 does not  significantly degrade results.}
% \label{tab:changek}
% \end{table}

%%%%%%%%%%%%%%%%%
\begin{figure}[t]
\begin{center}
\begin{minipage}{\columnwidth}
\hfill
\begin{minipage}{.17\linewidth}
\centerline{ {\bf Input}}
\centerline{215$\times$292}
\vspace{1mm}
\includegraphics[width=\linewidth]{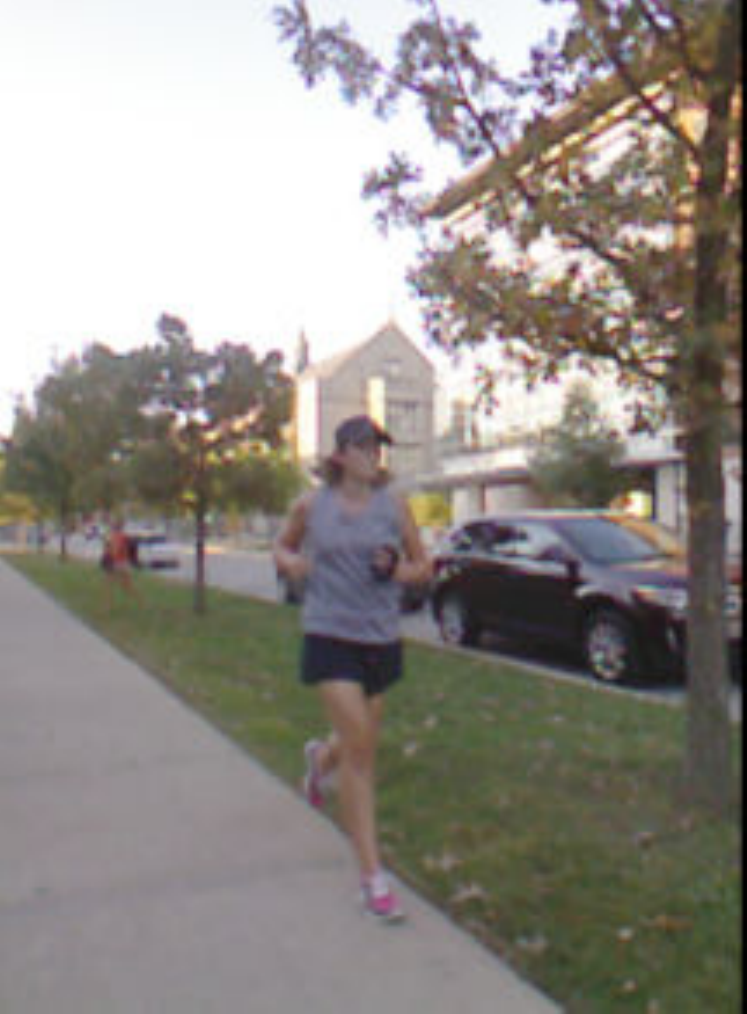}
\includegraphics[width=\linewidth]{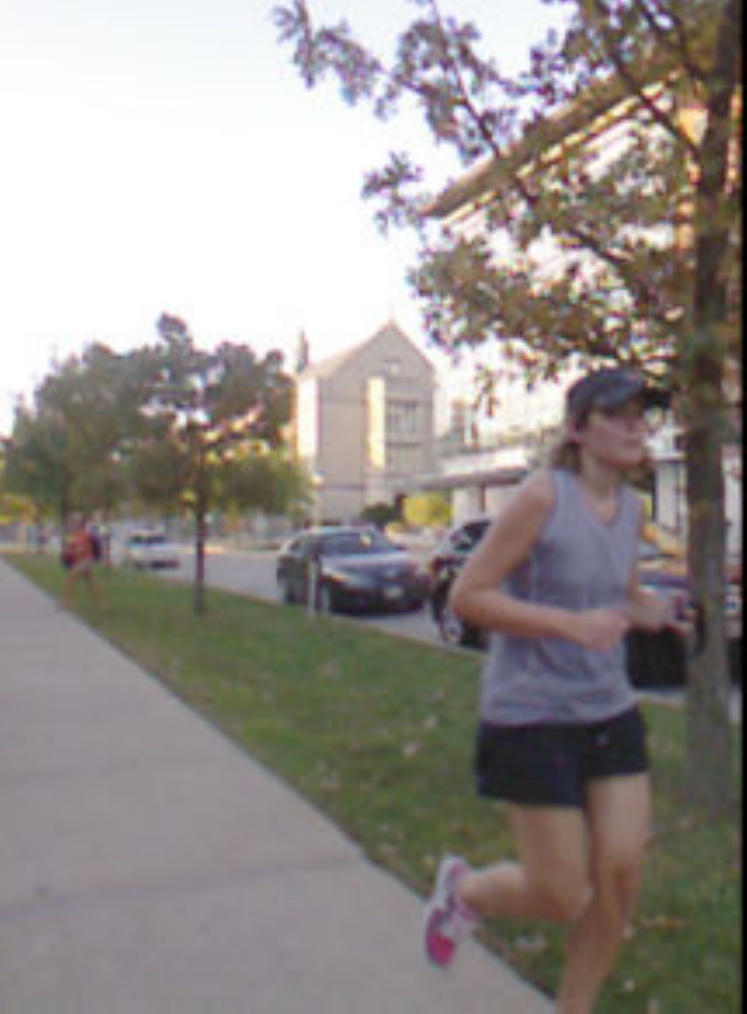}
\end{minipage}
\hfill
\begin{minipage}{.34\linewidth}
\centerline{ {\bf Make3D}}
\centerline{305$\times$55}
\vspace{1mm}
\begin{minipage}{.5\linewidth}
\includegraphics[width=\linewidth]{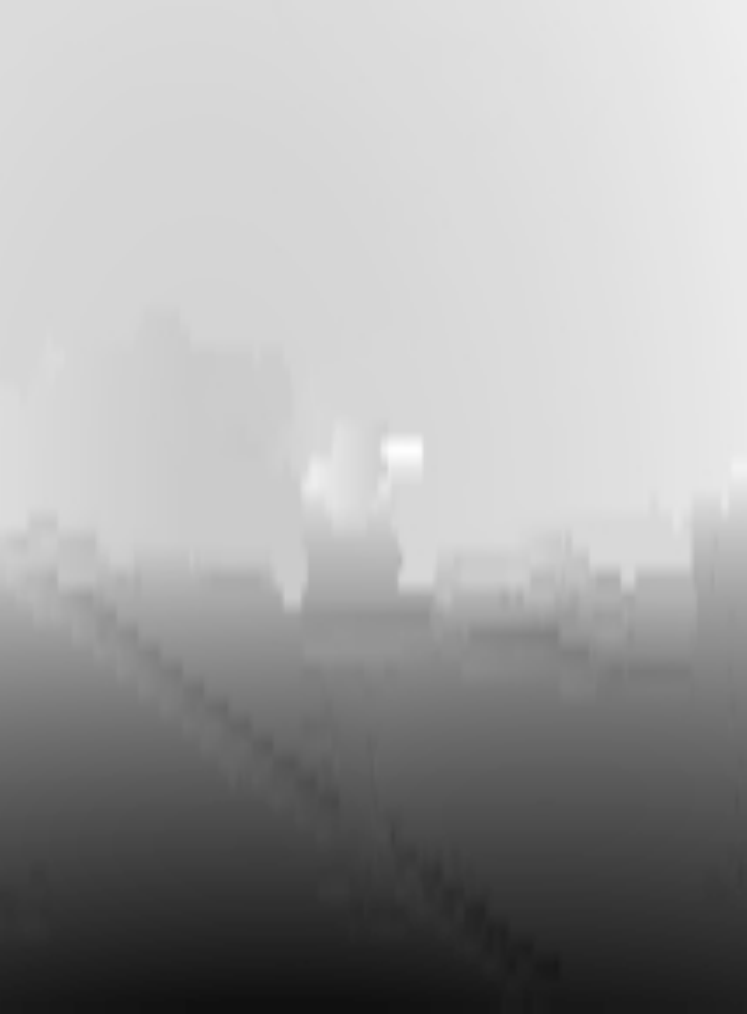}
\includegraphics[width=\linewidth]{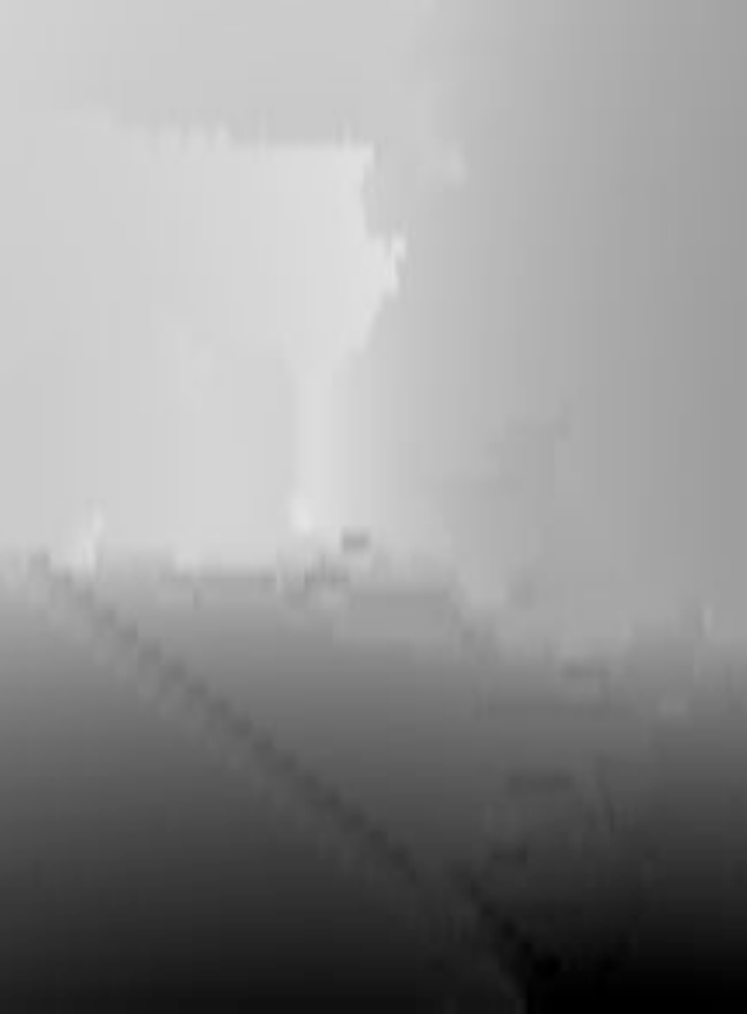}
\end{minipage}
\begin{minipage}{.465\linewidth}
\includegraphics[width=\linewidth]{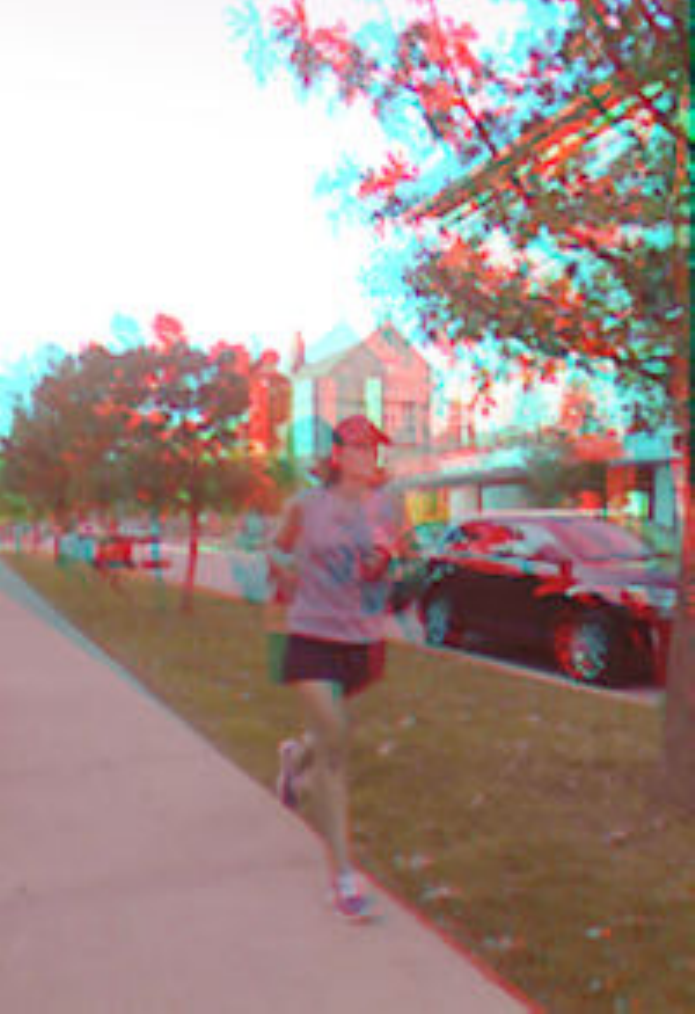}
\includegraphics[width=\linewidth]{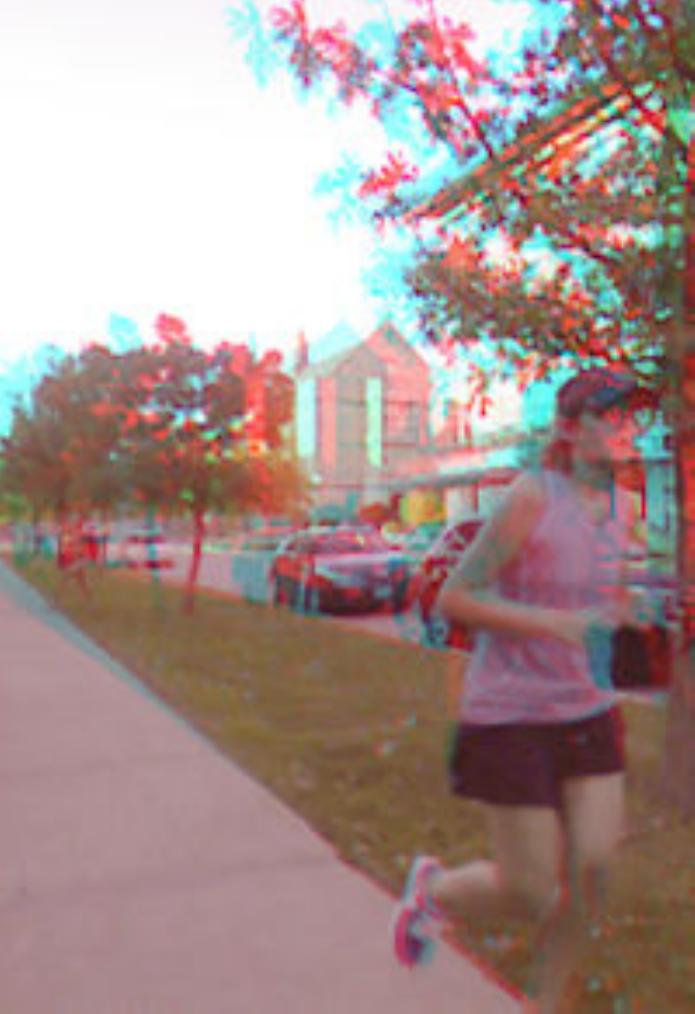}
\end{minipage}
\end{minipage}
\hfill
\begin{minipage}{.34\linewidth}
\centerline{{\bf Depth transfer (ours)}}
\centerline{215$\times$292}
\vspace{1mm}
\begin{minipage}{.5\linewidth}
\includegraphics[width=\linewidth]{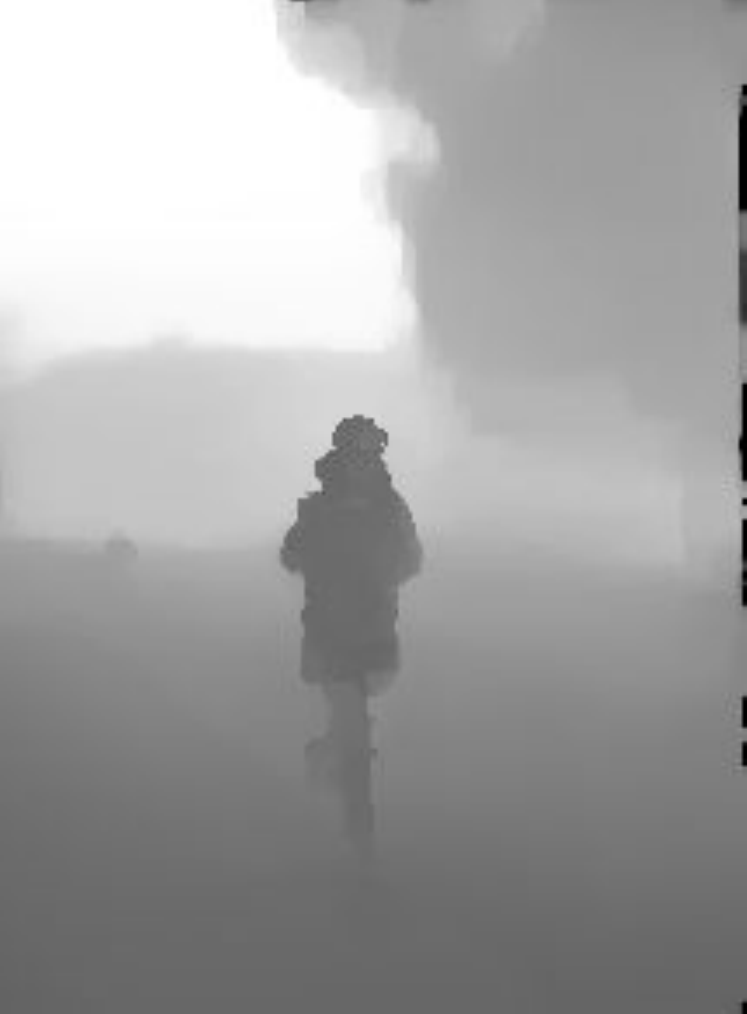}
\includegraphics[width=\linewidth]{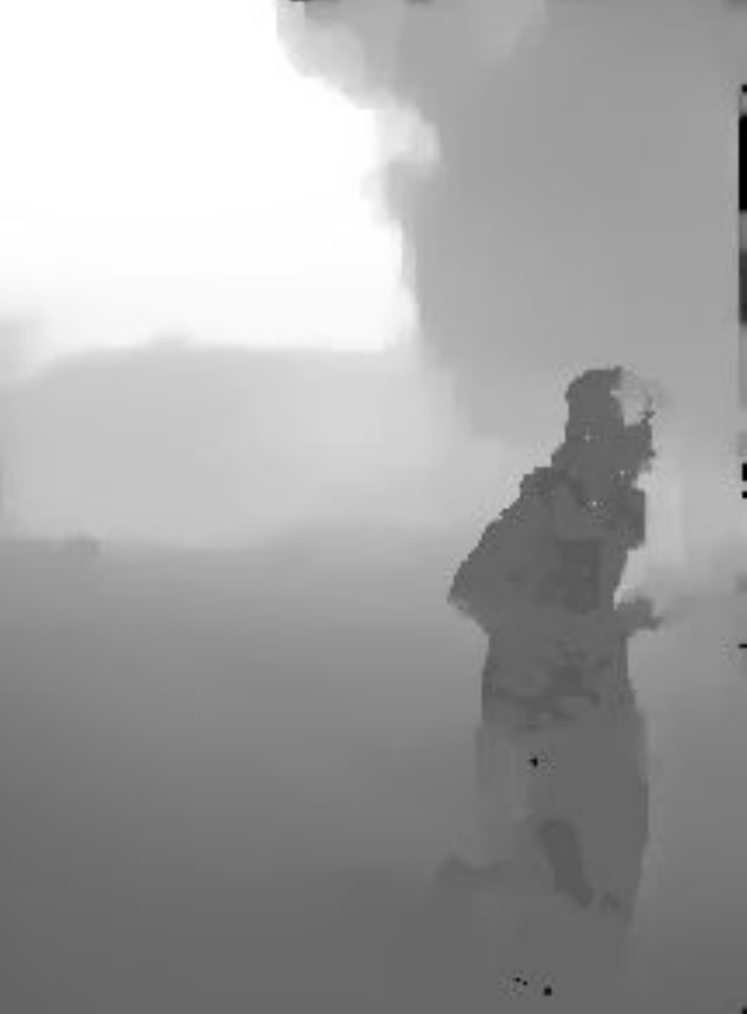}
\end{minipage}
\begin{minipage}{.445\linewidth}
\includegraphics[width=\linewidth]{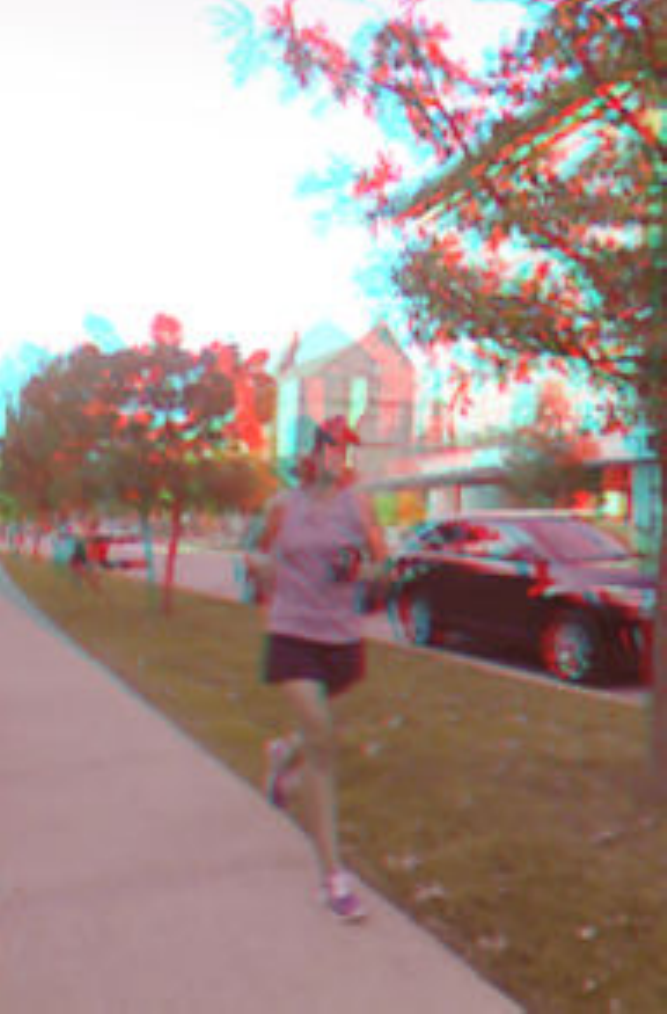}
\includegraphics[width=\linewidth]{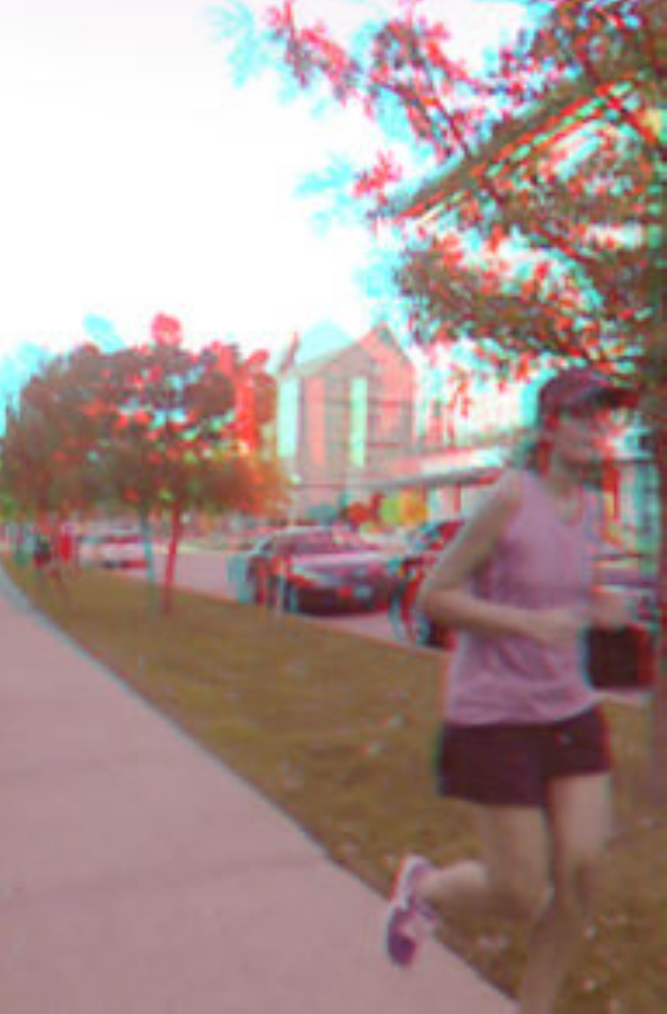}
\end{minipage}
\end{minipage}
\hfill
%\vspace{-6mm}
\end{minipage}
\end{center}
\caption{Comparison between our technique and the publicly available version of Make3D (\protect\url{http://make3d.cs.cornell.edu/code.html}). Make3D depth inference is trained to produce depths of resolution $55\times 305$ (bilinearly resampled for visualization), and we show results of our algorithm at the input native resolution. The anaglyph images are produced using the technique in Sec~\ref{sec:stereoest}. Depths displayed at same scale.
\vspace{2mm}
}
\label{fig:make3d_comparison}
\end{figure}
%%%%%%%%%%%%%%%%%

%%%%%%%%%%%%%%%%%
\begin{table}[t]
\centerline{
\normalsize
\begin{tabular}{| l l || c | c | c |} \hline
Training set & ($N=64$) & {\bf rel }& {\bf log$_{10}$} & {\bf RMS} \\ \hline \hline
GIST candidates & ($K=7$) & 0.431 & 0.164 & 15.9 \\ \hline
Full dataset & ($K=64$) & 0.475 & 0.207 & 20.3 \\ \hline
 \end{tabular}
 }
\vspace{2mm}
\caption{Importance of training set size. We select a subset of 64 images from the Make3D dataset, and compare our method using only 7 candidate images (selected using GIST features) versus using the entire dataset (64 candidates). The results suggest that selectively re-training based on the target image's content can significantly improve results.}
 \label{tab:train_size}
 \end{table}
%%%%%%%%%%%%%%%%%

%%%%%%%%%%%%%%%%%
\begin{figure}[t]
\includegraphics[width=1\columnwidth]{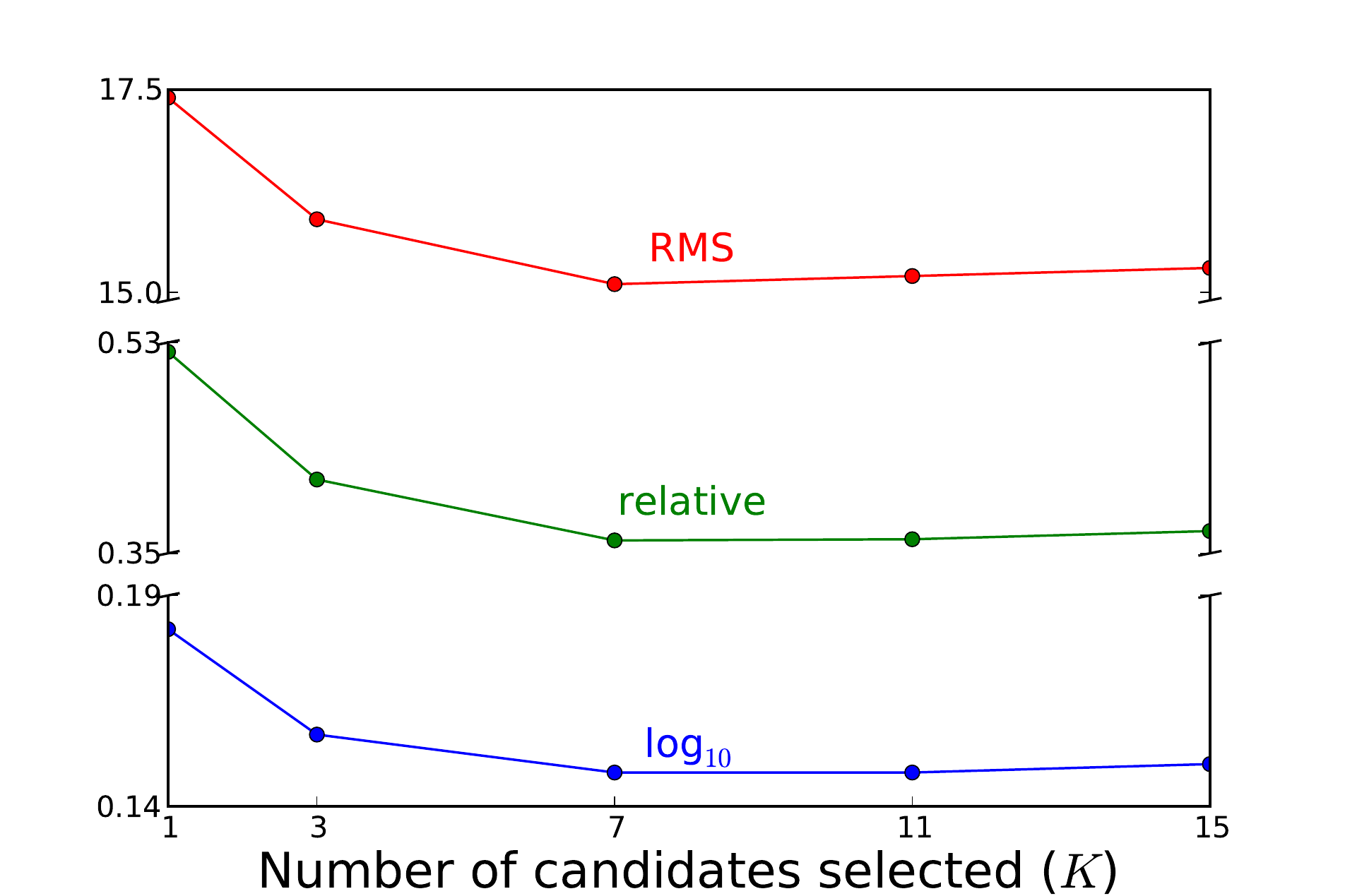}
\caption{Effect of the number of chosen candidates ($K$). Errors are reported on the Make3D dataset with varying values of $K$. For this dataset, $K=7$ is optimal, but increasing $K$ beyond 7 does not  significantly degrade results.
}
 \label{fig:changek}
 \end{figure}
%%%%%%%%%%%%%%%%%

\subsection{Importance of training scale}
One implicit hypothesis our algorithm makes is that {\it training with only a few ``similar'' images is better than training with a large set of arbitrary images}. This is encoded by our nearest neighbor candidate search: we choose $k$ similar images (in our work, based on GIST features), and use {\it only} these images to sample/transfer depth from. Conversely, Saxena et al.~\cite{Saxena:09} train a parametric model for predicting depth using their entire dataset (Make3D); Liu et al.~\cite{Liu:cvpr10} found that training a different models for each semantic classes (tree, sky, etc.) improves results. The results in Table~\ref{tab:train_size} are good evidence that training on only similar {\it scenes} improves results (rather than only similar semantic classes as in~\cite{Liu:cvpr10}, and on the entire dataset~\cite{Saxena:09}).

We verify this further with another experiment: we created a sub-dataset by randomly choosing 64 images from the Make3D dataset, then we compared the results of our method using only similar images for training (7 nearest neighbors) and the results of our method trained using the entire dataset (64 nearest neighbors) during inference. As the results in Table~\ref{tab:train_size} suggest, training using only similar scenes greatly improves quantitative results.
In addition, since our ``training'' is simply a nearest neighbor query (which can be done quickly in seconds), it can be orders of magnitude faster than retraining parametric models.

\subsection{Effect of $K$ (number of candidates)}
We also evaluate how our technique behaves given different values of $K$, i.e., how many candidate images are selected prior to inference. On the Make3D dataset, we evaluate three error metrics (same as above: relative, log${}_{10}$, and RMS) averaged over the entire dataset using different values of $K$. Fig~\ref{fig:changek} shows the results, and we see that $K=7$ is optimal for this dataset, but we still achieve comparable results with $K\geq 7$.

Empirically, we find that $K$ acts as a smoothing parameter. This fits with the intuition that more candidate images will likely lead to diversity in the candidate set, and since the inferred depth is in some sense sampled from all candidates, the result will be smoother as $K$ increases.
}

%%%%%%%%%%%%%%%%%%%%%%%%%%%%%%%%%%%%%%
\section{Application: 2D-to-3D}
\label{sec:stereoest}

In recent years, 3D\footnote{The presentation of stereoscopic (left+right) video to convey the sense of depth.} videos have become increasingly popular. Many feature films are now available in 3D, and increasingly more personal photography devices are now equipped with stereo capabilities (from point-and-shoots to attachments for video cameras and SLRs). Distributing user-generated content is also becoming easier. Youtube has recently incorporated 3D viewing and uploading features, and many software packages have utilities for handling and viewing 3D file formats, e.g., Fujifilm's FinePixViewer.

As 3D movies and 3D viewing technology become more widespread,
it is desirable to have techniques that can convert legacy 2D movies to 3D in an efficient and inexpensive way. Currently, the movie industry uses expensive solutions that tend to be manual-intensive. For example, it was reported that the cost of converting (at most) 20 minutes of footage for the movie ``Superman Returns'' was \$10 million\footnote{See \url{http://en.wikipedia.org/wiki/Superman\_Returns}.}.

Our technique can be used to automatically generate the depth maps necessary to produce the stereoscopic video (by warping each input frame using its corresponding depth map). To avoid generating holes at disocclusions in the view synthesis step, we adapt and extend Wang~\ea's technique~\cite{Wang:sbim11}. They developed a method that takes as input a single image and per-pixel disparity values, and intelligently warps the input image based on the disparity such that highly salient regions remain unmodified. Their method was applied only to single images; we extend this method to handle video sequences as well. 

%Details of our view synthesis technique can be found in the appendix.

%%%%%%%%%%%%%%%%%%%%%%%%%%%%%%%%%%%%%%%%%%%%%
\newtext{
%\section*{Appendix: automatic stereoscopic view synthesis}
\subsection{Automatic stereoscopic view synthesis}
After estimating depth for a video sequence (or a single image), we perform depth image-based rendering (DIBR) to synthesize a new view for stereoscopic display. A typical strategy for DIBR is simply reprojecting pixels based on depth values to a new, synthetic camera, but such methods are susceptible to large ``holes'' at disocclusions. Much work has been done to fill these holes (e.g.,~\cite{Colombari_continuousparallax,Gunnewiek:09,Luo:09,Zhang:11}), but visual artifacts still remain in the case of general scenes.

We propose a novel extension to a recent DIBR technique which uses image warping to overcome problems such as disocclusions and hole filling. Wang~\ea~\cite{Wang:sbim11} developed a method that takes as input a single image and per-pixel disparity values, and intelligently warps the input image based on the disparity such that highly salient regions remain unmodified. This method is illustrated in Fig~\ref{fig:viewsynthoverview}. The idea is that people are less perceptive of errors in low saliency regions, and thus disocclusions are covered by ``stretching'' the input image where people tend to not notice artifacts. This method was only applied to single images, and we show how to extend this method to handle video sequences in the following text.

%%%%%%%%%%%%%%%%%
\begin{figure}[t]
\centerline{\includegraphics[width=.99\columnwidth]{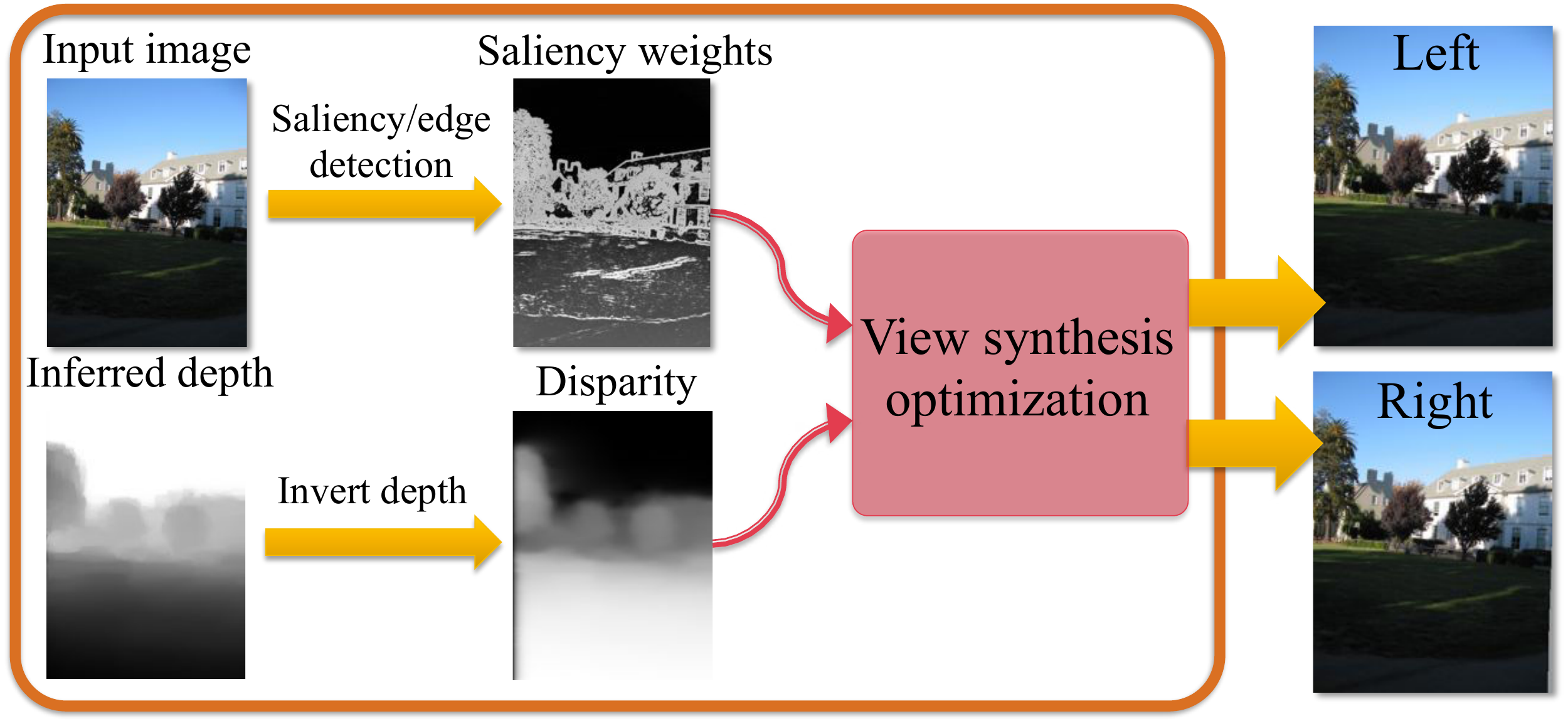}}
\vspace{-2mm}
\caption{Summary of the view synthesis procedure for a single image, as described by Wang~\ea~\cite{Wang:sbim11}. Given an image and corresponding depth, we compute salient regions and disparity, and compute stereoscopic images by warping the input image. We extend this method to handle videos, as in equation~\ref{eq:viewsynth}.
%\vspace{-1mm}
}
\label{fig:viewsynthoverview}
\end{figure}
%%%%%%%%%%%%%%%%%

Given an input image and depth values, we first invert the depth to convert it to disparity, and scale the disparity by the maximum disparity value:
\begin{equation}
\label{eq:disparity}
\ve{W}_0 = \frac{\ve{W}_{\text{max}}}{\ve{D}+\epsilon},
\end{equation}
where $\ve{W}_0 = \{ W_1, \ldots, W_n \}$, $\ve{D} = \{ D_1, \ldots, D_n \}$  is the initial disparity and depth (resp) for each of the $n$ frames of the input, and $\ve{W}_{\text{max}}$ is a parameter which modulates how much objects ``pop-out'' from the screen when viewed with a stereoscopic device. Increasing this value enhances the ``3D'' effect, but can also cause eye strain or problems with fusing the stereo images if set too high. We set $\epsilon=0.01$.

%%%%%%%%%%%%%%%%%
\begin{figure*}
\newcommand{\linespace}{\vspace{-1.em}}
\begin{minipage}{.325\linewidth}
\includegraphics[width=.49\linewidth]{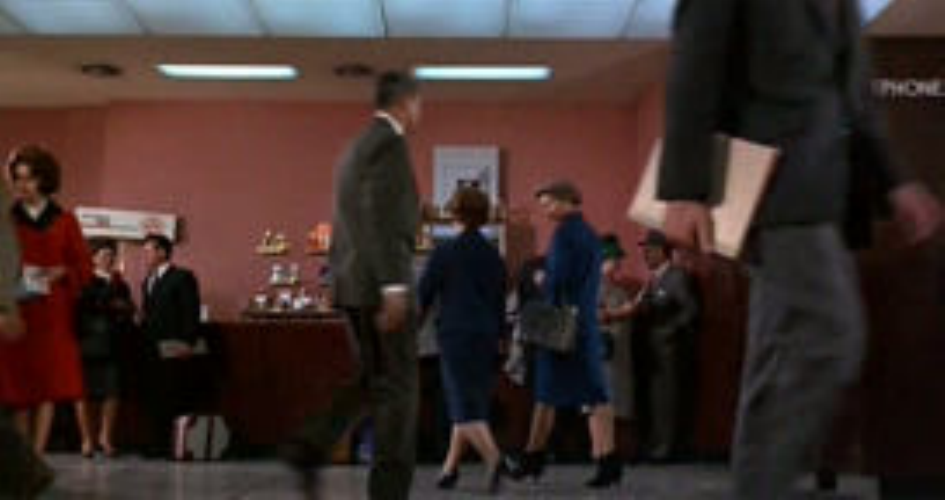}
\includegraphics[width=.49\linewidth]{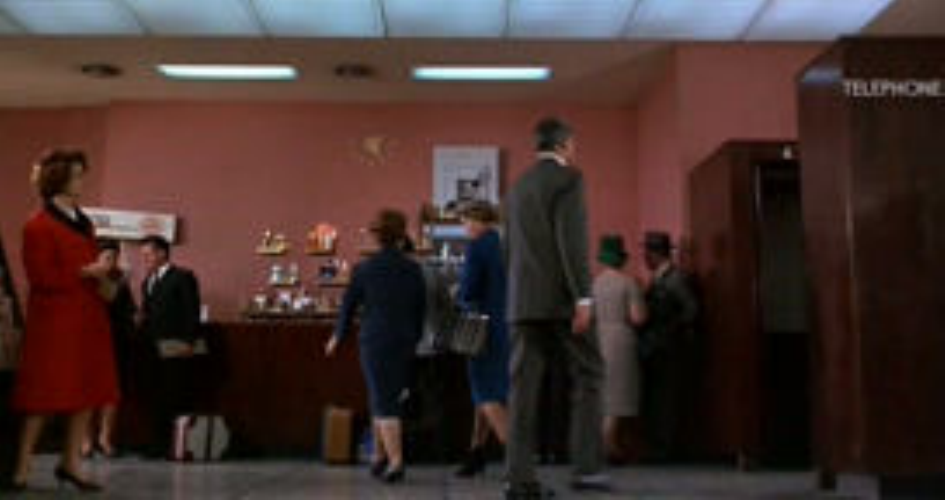}
\linespace\\
\includegraphics[width=.49\linewidth]{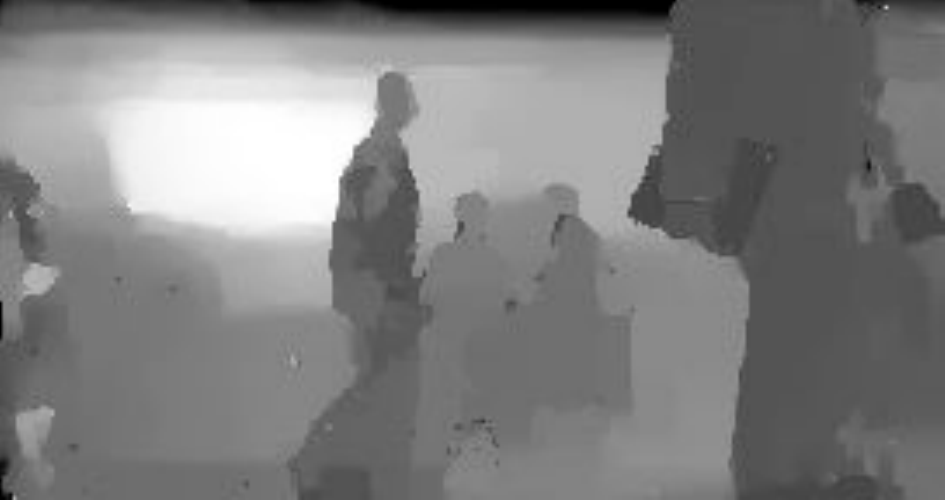}
\includegraphics[width=.49\linewidth]{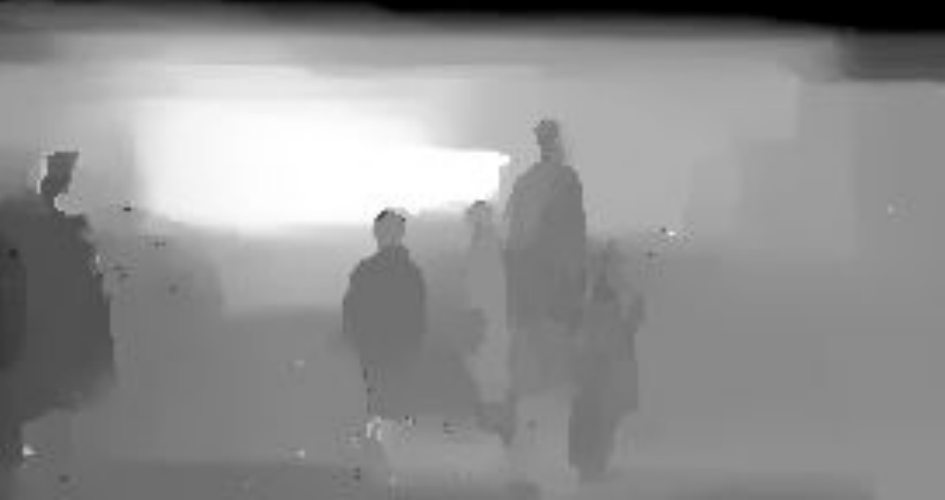}
\linespace\\
\includegraphics[width=.49\linewidth]{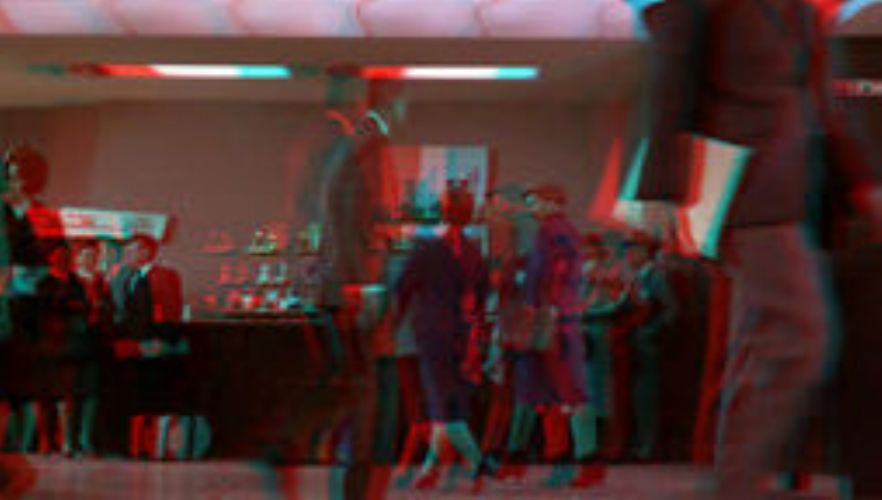}
\includegraphics[width=.49\linewidth]{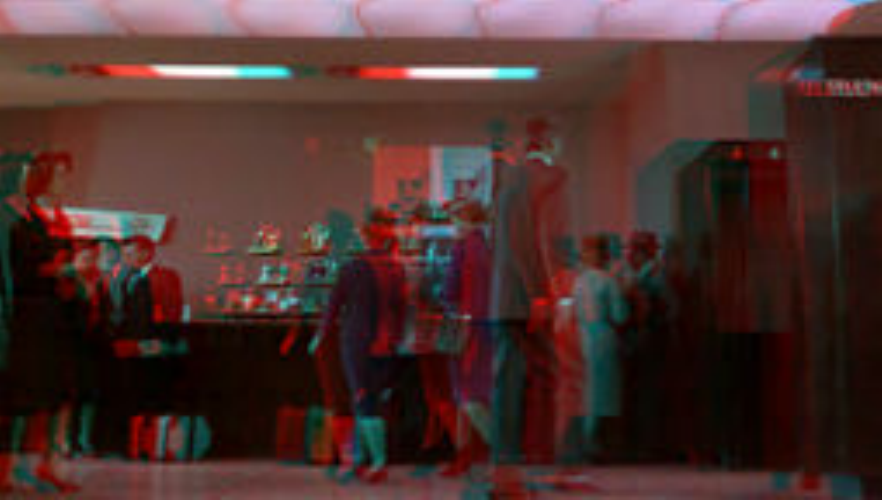}
\end{minipage}
\hfill
\begin{minipage}{.325\linewidth}
\includegraphics[width=.49\linewidth]{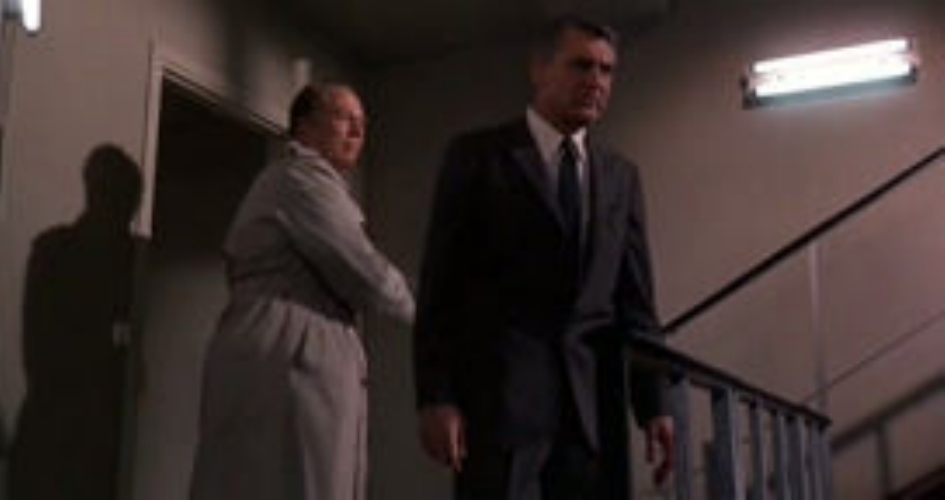}
\includegraphics[width=.49\linewidth]{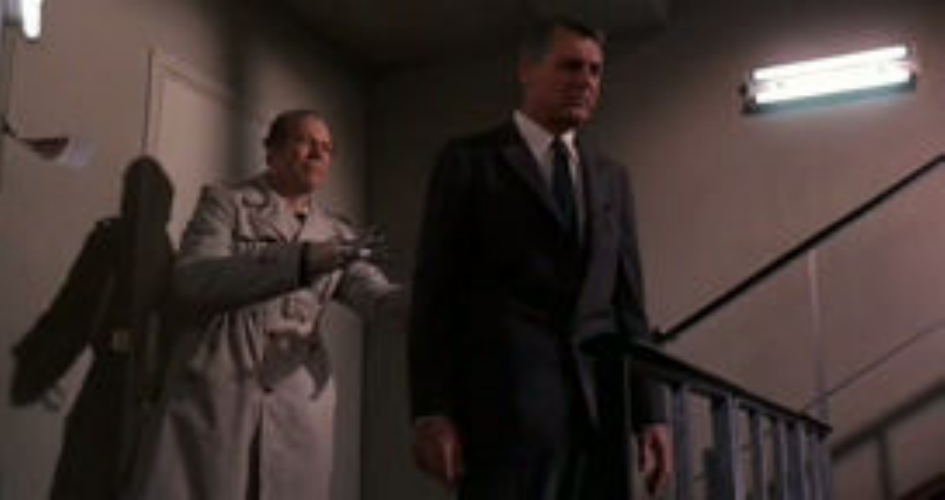}
\linespace\\
\includegraphics[width=.49\linewidth]{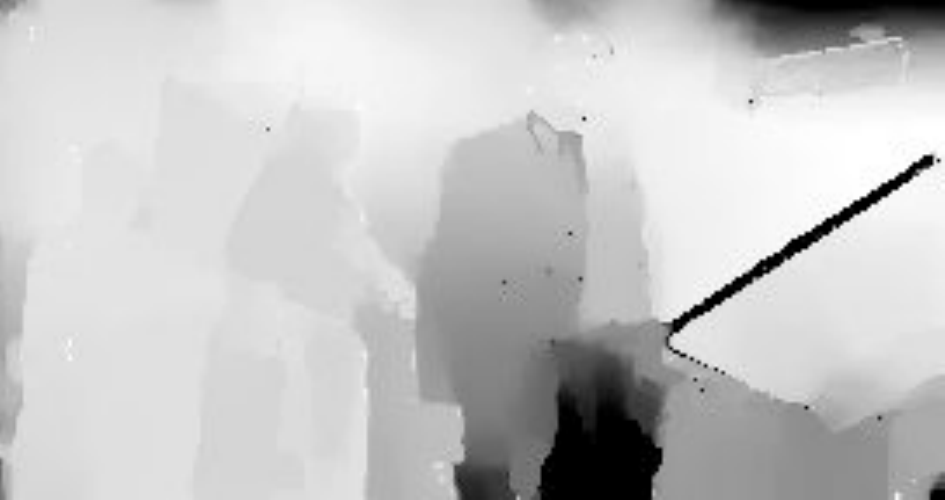}
\includegraphics[width=.49\linewidth]{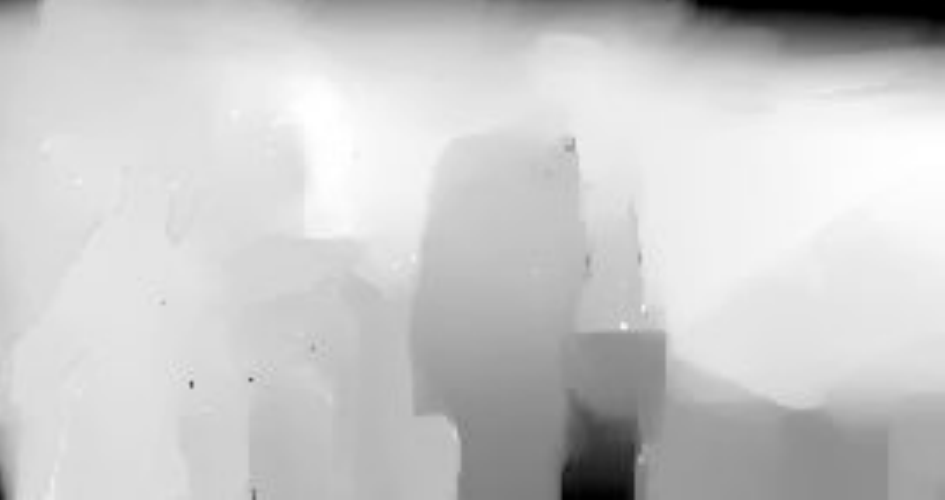}
\linespace\\
\includegraphics[width=.49\linewidth]{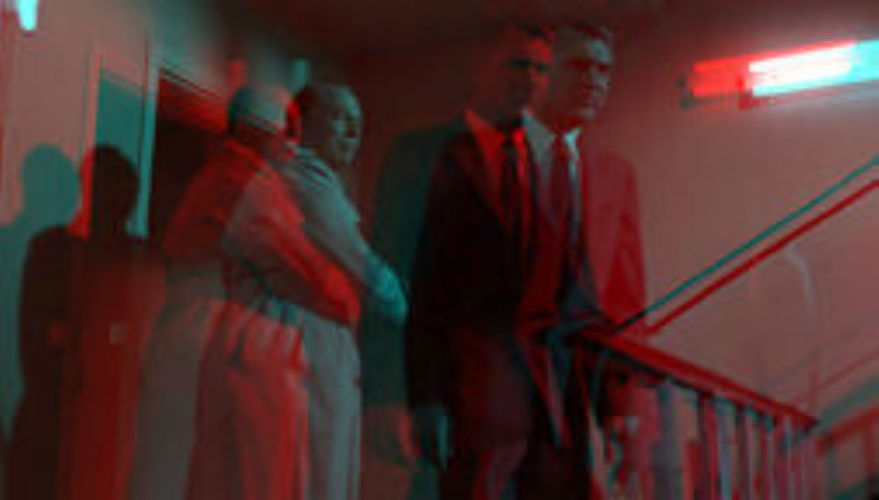}
\includegraphics[width=.49\linewidth]{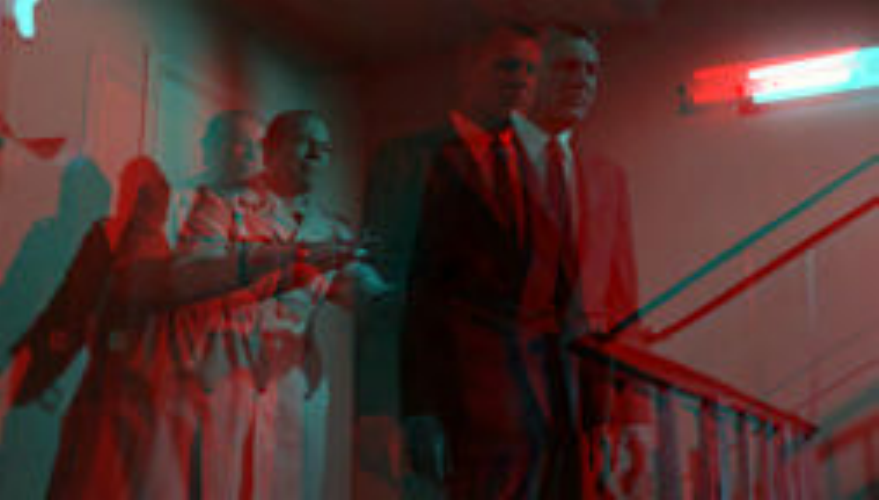}
\end{minipage}
\hfill
\begin{minipage}{.325\linewidth}
\includegraphics[width=.49\linewidth]{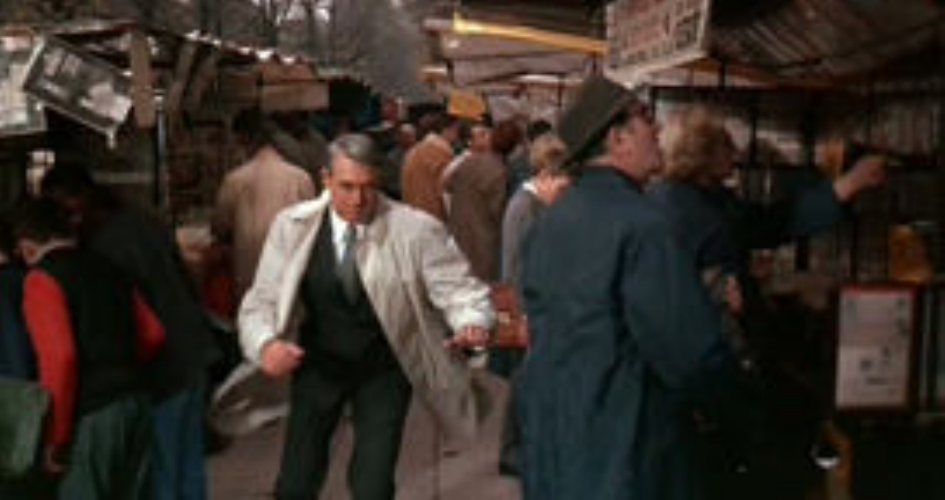}
\includegraphics[width=.49\linewidth]{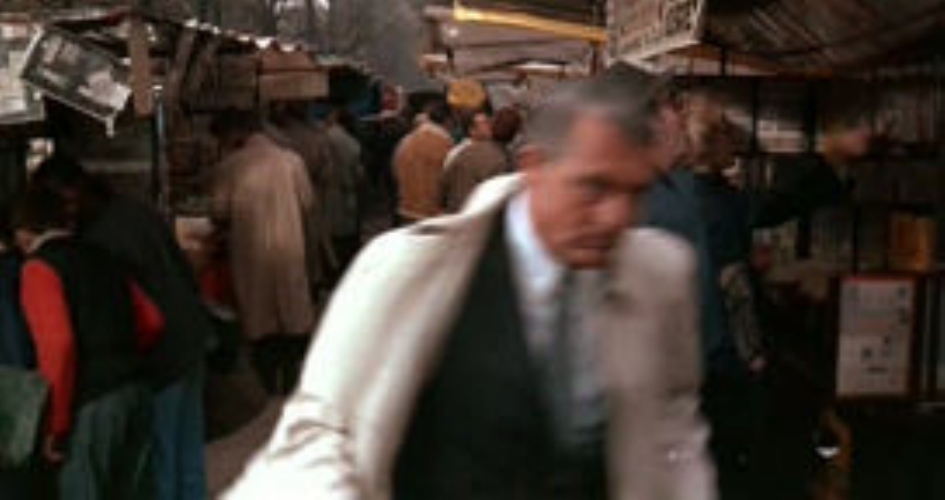}
\linespace\\
\includegraphics[width=.49\linewidth]{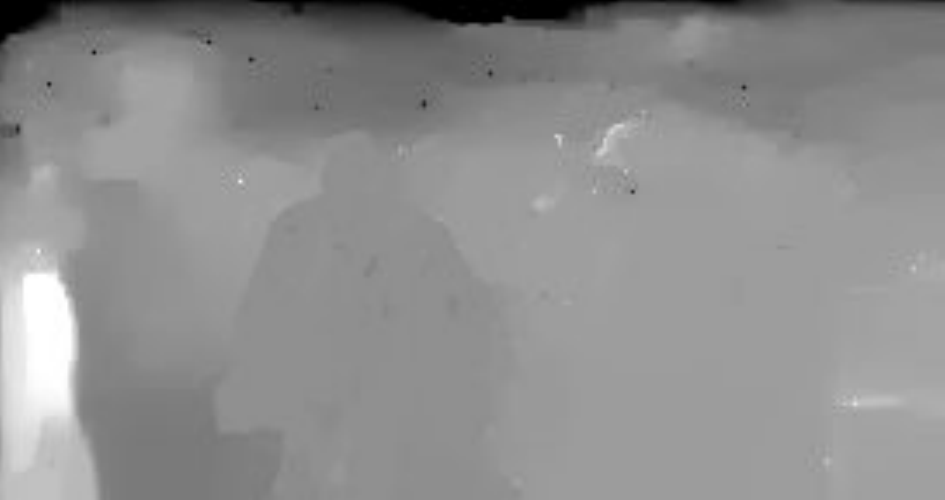}
\includegraphics[width=.49\linewidth]{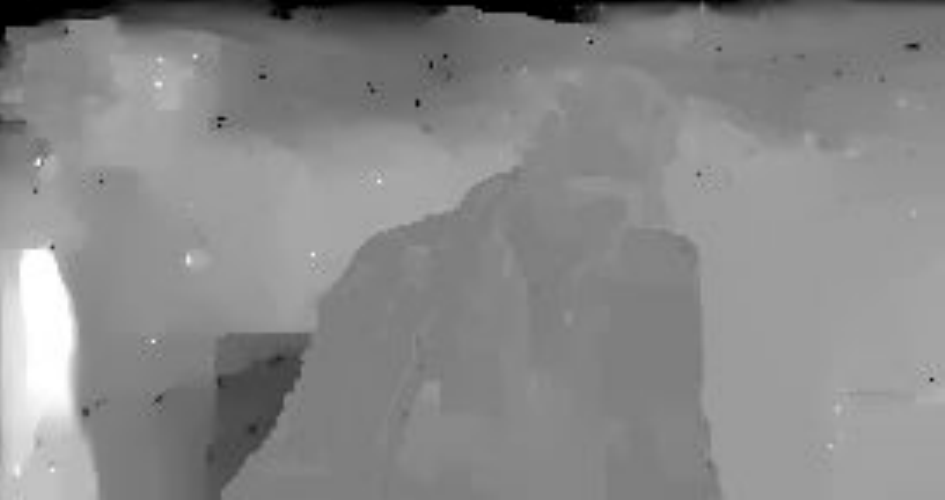}
\linespace\\
\includegraphics[width=.49\linewidth]{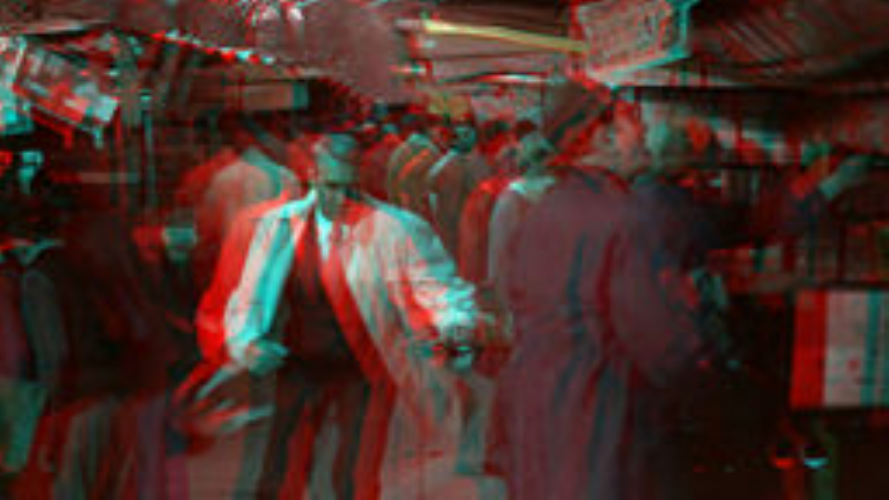}
\includegraphics[width=.49\linewidth]{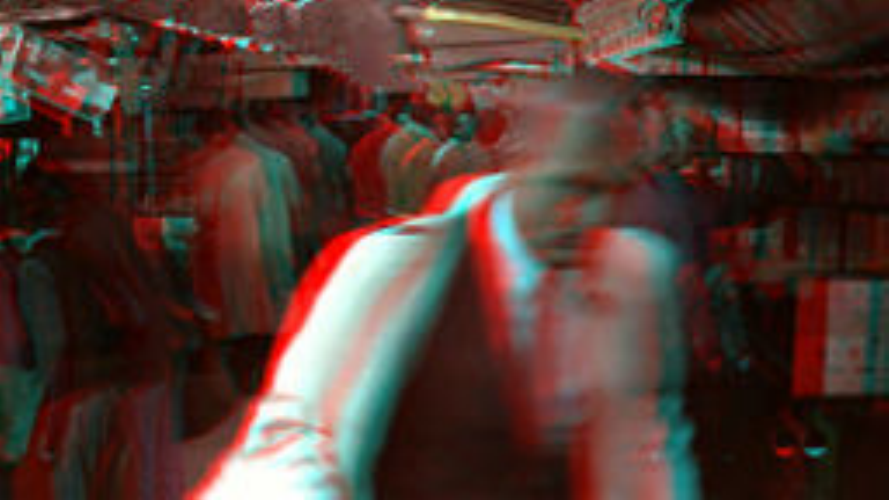}
\end{minipage}
%\vspace{-2mm}
\caption{Several clips from the feature film {\it Charade}. Each result contains (from top to bottom): the original frames, estimated depth, and estimated anaglyph \protect\includegraphics[height=5pt]{fig/anaglyph.pdf} automatically generated by our algorithm. Some imperfections in depth are conveniently masked in the 3D image due to textureless or less salient regions.
%(trained again using only Building 1 data).
%While not perfect, our depth maps are suitable for convincing 3D results. Depth errors in less salient regions do not affect the 3D result much.
%Here, we show two clips from the movie {\it Charade}. Each result shows several frames from each clip and contains three rows (top to bottom): input frame, estimated depth, estimated anaglyph \protect\includegraphics[height=5pt]{fig/anaglyph.pdf} automatically generated by our algorithm (trained again using only Building 1 data).
}
%\vspace{-8mm}
\label{fig:charade}
\end{figure*}
%%%%%%%%%%%%%%%%%%%%%%%%%%%%%%%%%%%%%%

Then, to implement the saliency-preserving warp (which in turn defines two newly synthesized views), minimize the following unconstrained, quadratic objective:
\begin{equation}
\label{eq:viewsynth}
\begin{split}
Q(\ve{W}_i) &=\ \sum_{i\in\text{pixels}} Q_{\text{data}}(\ve{W}_i) + Q_{\text{smooth}}(\ve{W}_i),\\
Q_{\text{data}}(\ve{W}_i)  &= l_i(\ve{W}_i-\ve{W}_{0,i})^2,\\
Q_{\text{smooth}}(\ve{W}_i)  &= \lambda (s_{x,i} ||\nabla_x \ve{W}_i||^2+ s_{y,i}||\nabla_y \ve{W}_i||^2) + \\ & \hspace{5mm} \mu s_{t,i} ||\nabla_{flow}\ve{W}_i||^2,
\end{split}
\end{equation}
where $l_i$ is a weight based on image saliency and initial disparity values that constrains disparity values corresponding to highly salient regions and very close objects to remain unchanged, and is set to $l_i = \frac{W_{0,i}}{W_{\text{max}}} + (1+e^{-(||\nabla \ve{L}_i||-0.01)/0.002})^{-1}$. The $Q_{\text{smooth}}$ term contains the same terms as in our spatial and temporal smoothness functions in our depth optimization's objective function, and $\lambda$ and $\mu$ control the weighting of these smoothness terms in the optimization; we set $\lambda = \mu = 10$. As in Eq~\ref{eq:smoothweights}, we set $s_{x,i} = (1+e^{(||\nabla_x \ve{L}_i||-0.05)/.01})^{-1}$, $s_{y,i} = (1+e^{(||\nabla_y \ve{L}_i|| - \mu_L)/\sigma_L})^{-1}$, and $s_{t,i} = (1+e^{-(||\nabla_{flow} \ve{L}_i|| - \mu_L)/\sigma_L})^{-1}$, where $\nabla_x \ve{L}_i$ and $\nabla_y \ve{L}_i$ are image gradients in the respective dimensions (at pixel $i$), and $\nabla_{flow} \ve{L}_i$ is the flow difference across neighboring frames (gradient in the flow direction). We set $\mu_L = 0.05$ and $\sigma_L = 0.01$.
With this formulation, we ensure spatial and temporal coherence and most importantly that highly salient regions remain intact during view warping.

After optimization, we divide the disparities by two ($\ve{W} \gets \frac{\ve{W}}{2}$), and use these halved values to render the input frame(s) into two new views (corresponding to the stereo left and right views). We choose this method, as opposed to only rendering one new frame with larger disparities, because people are less perceptive of a many small artifacts when compared with few large artifacts~\cite{Wang:sbim11}. For rendering, we use the anisotropic pixel splatting method described by Wang~\ea~\cite{Wang:sbim11}, which ``splats'' input pixels into the new view (based on $\ve{W}$) as weighted, anisotropic Gaussian blobs.

With the two synthesized views, we can convert to any 3D viewing format, such as anaglyph or interlaced stereo. For the results in this paper, we use the anaglyph format as cyan/red anaglyph glasses are more widespread than polarized/autostereoscopic displays (used with interlaced 3D images). To reduce eye strain, we shift the left and the right images such that the nearest object has zero disparity, making the nearest object appear at the display surface, and all other objects appear {\it behind} the display. This is commonly known as the ``window'' metaphor~\cite{Koppal:cga11}.
}

\subsection{2D-to-3D Results}

Our estimated depth is good enough to generate compelling 3D images, and representative results are shown in (Figs~\ref{fig:rot_zoom_results},~\ref{fig:indoor_results}). We also demonstrate that our algorithm may be suitable for feature films in Fig~\ref{fig:charade}. More diverse quantities of training are required to achieve commercial-quality conversion; however, even with a small amount of data, we can generate plausible depth maps and create convincing 3D sequences automatically.

Recently, Youtube has released an automatic 2D-to-3D conversion tool, and we compared our method to theirs on several test sequences. Empirically, we noticed that the Youtube results have a much more subtle 3D effect. Both results are available online at \url{http://kevinkarsch.com/depthtransfer}.

Our algorithm takes roughly one minute per 640$\times$480 frame (on average) using a parallel implementation on a quad-core 3.2GHz processor.

%%%%%%%%%%%%%%%%%%%%%%%%%%%%%%%%%%%%%%
\section{Discussion}
\begin{figure}[t]
\begin{center}
\begin{minipage}{.218\linewidth}
\centerline{\bf\small Input}\vspace{0.5mm}
\includegraphics[width=\linewidth]{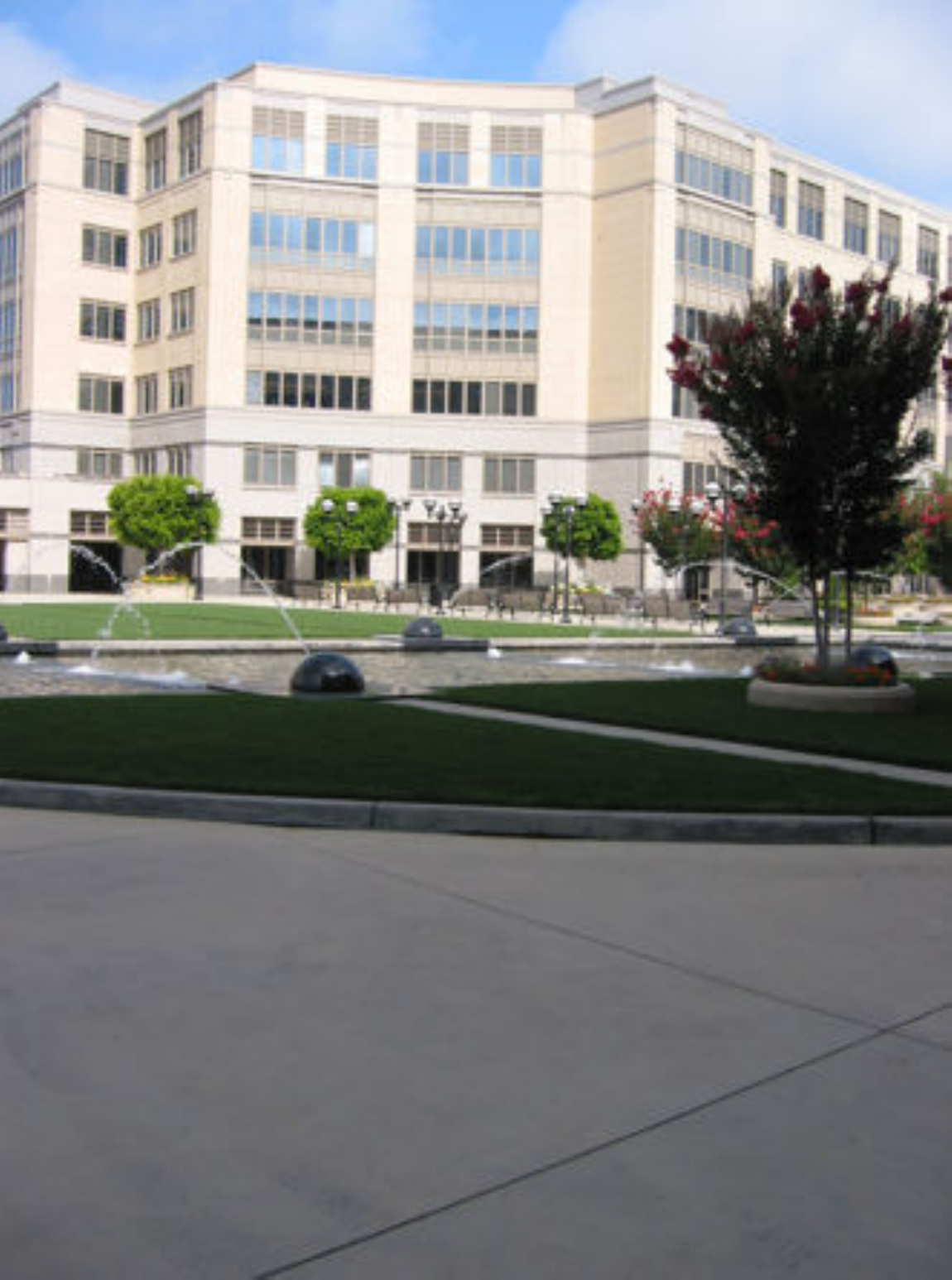}\vspace{0.5mm}\\
\includegraphics[width=\linewidth]{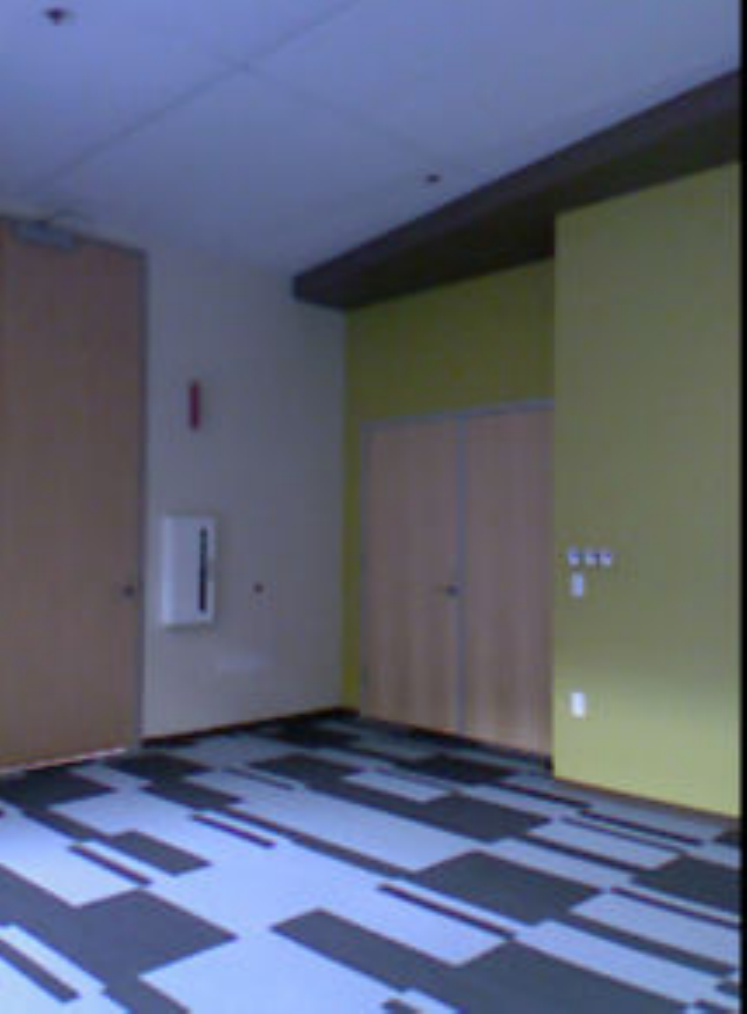}
\end{minipage}
\begin{minipage}{.22\linewidth}
\centerline{\bf\small Outdoor}\vspace{0.7mm}
\includegraphics[width=\linewidth]{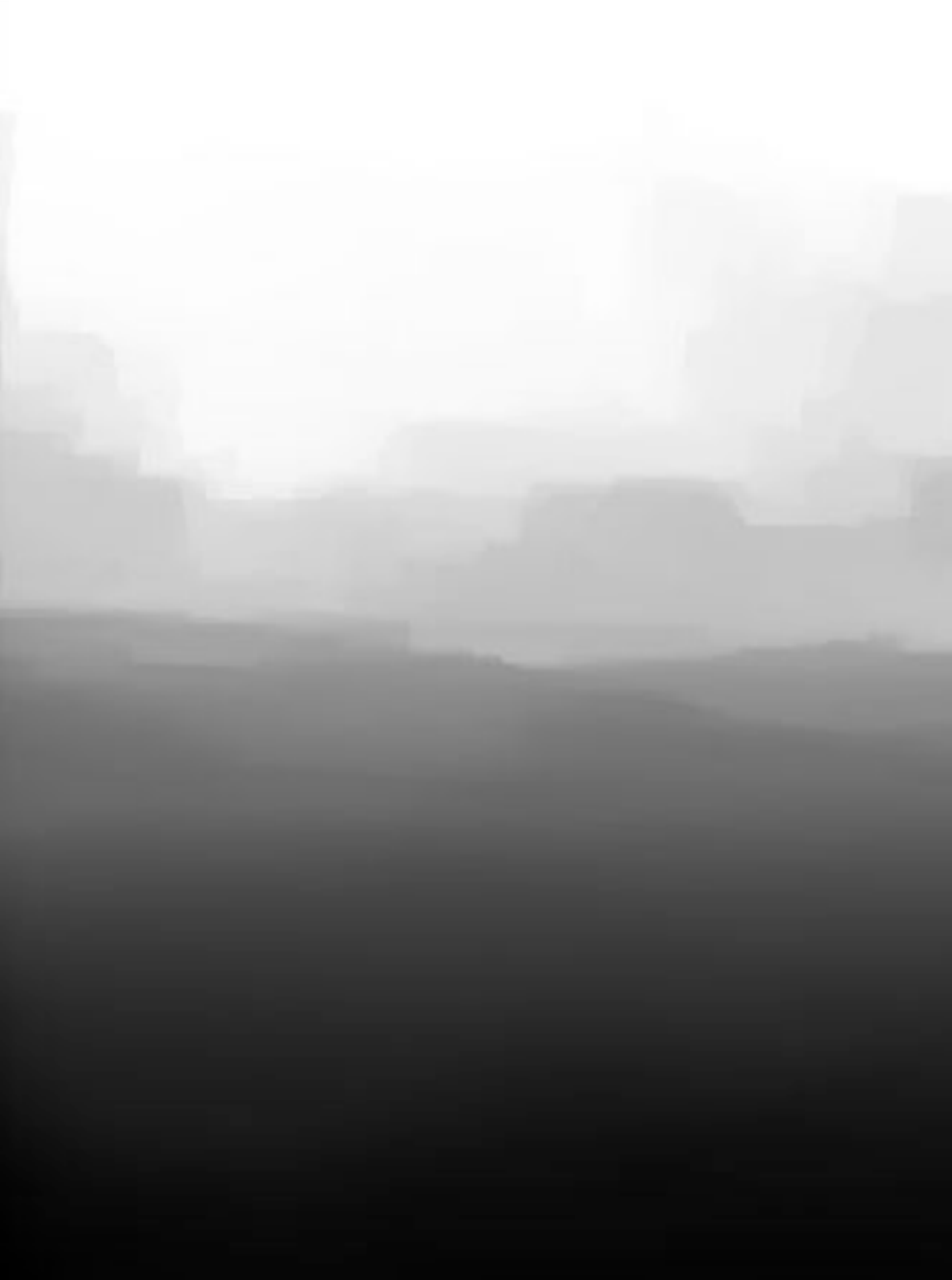}\vspace{0.5mm}\\
\includegraphics[width=\linewidth]{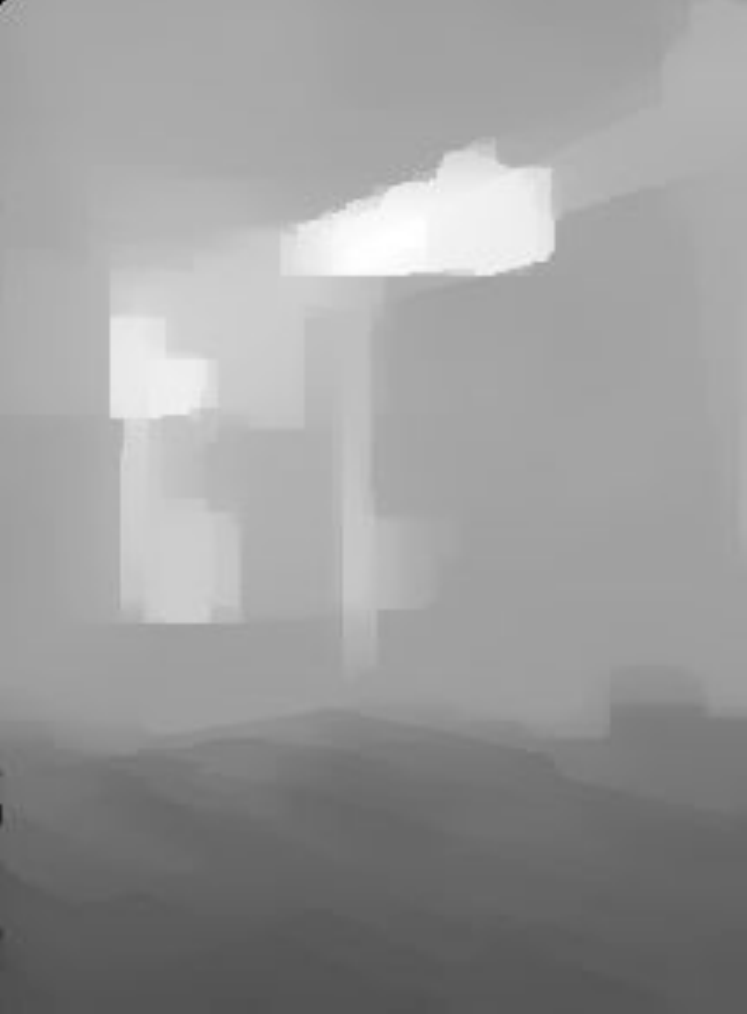}
\end{minipage}
\begin{minipage}{.22\linewidth}
\centerline{\bf\small Indoor}\vspace{0.5mm}
\includegraphics[width=\linewidth]{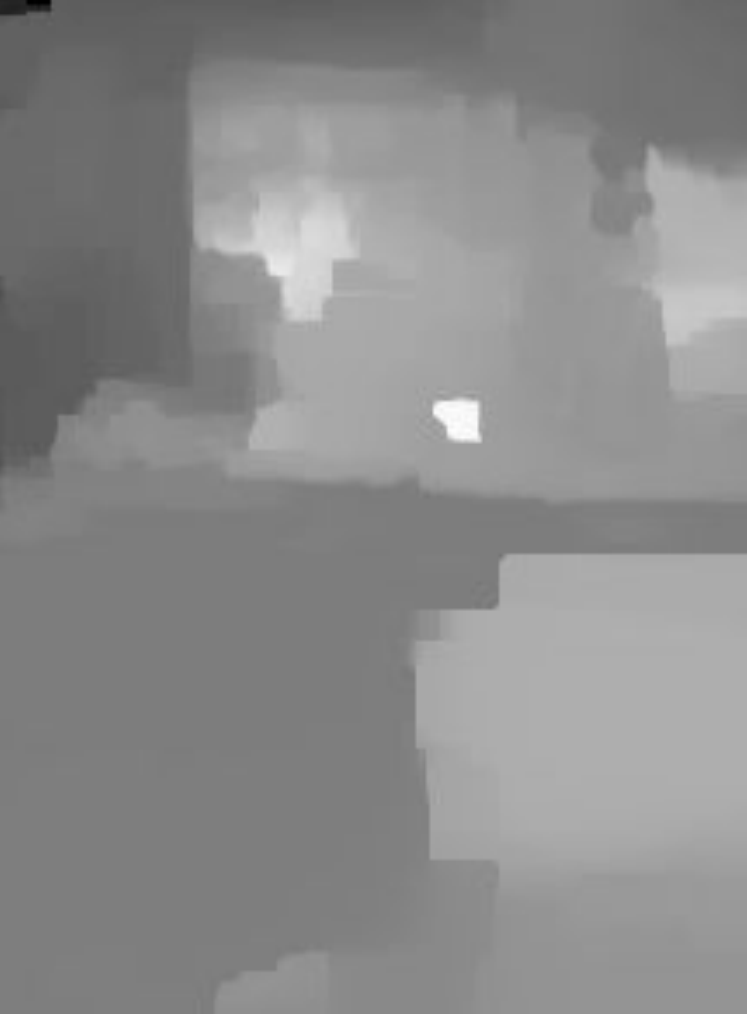}\vspace{0.5mm}\\
\includegraphics[width=\linewidth]{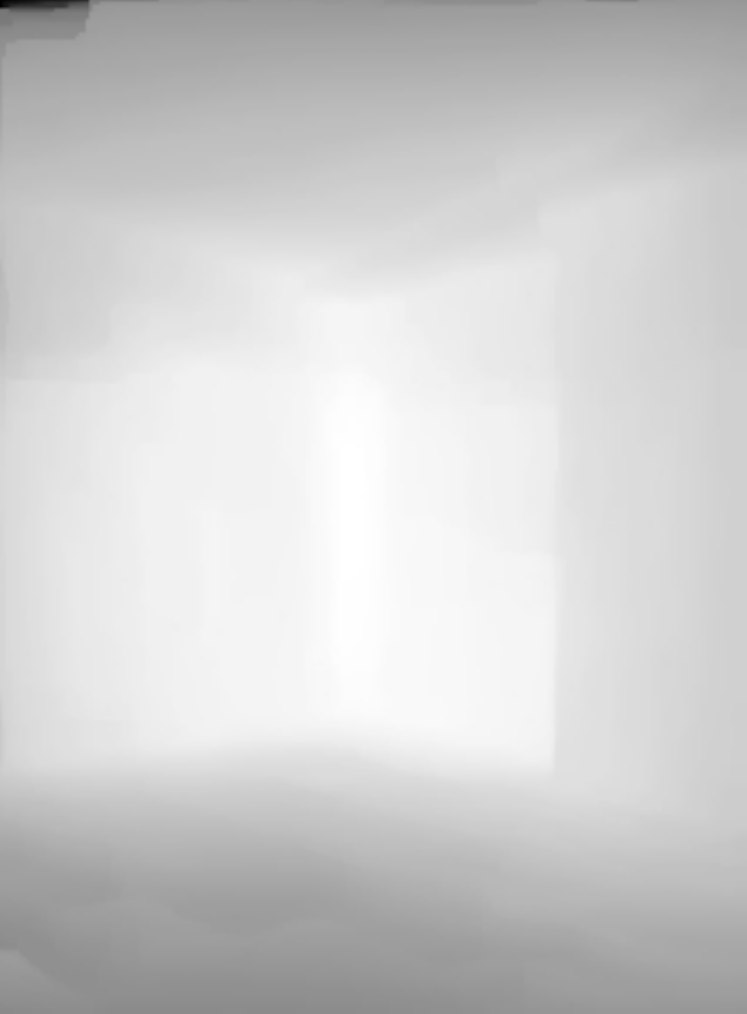}
\end{minipage}
\begin{minipage}{.22\linewidth}
\centerline{\bf\small All data}\vspace{0.7mm}
\includegraphics[width=\linewidth]{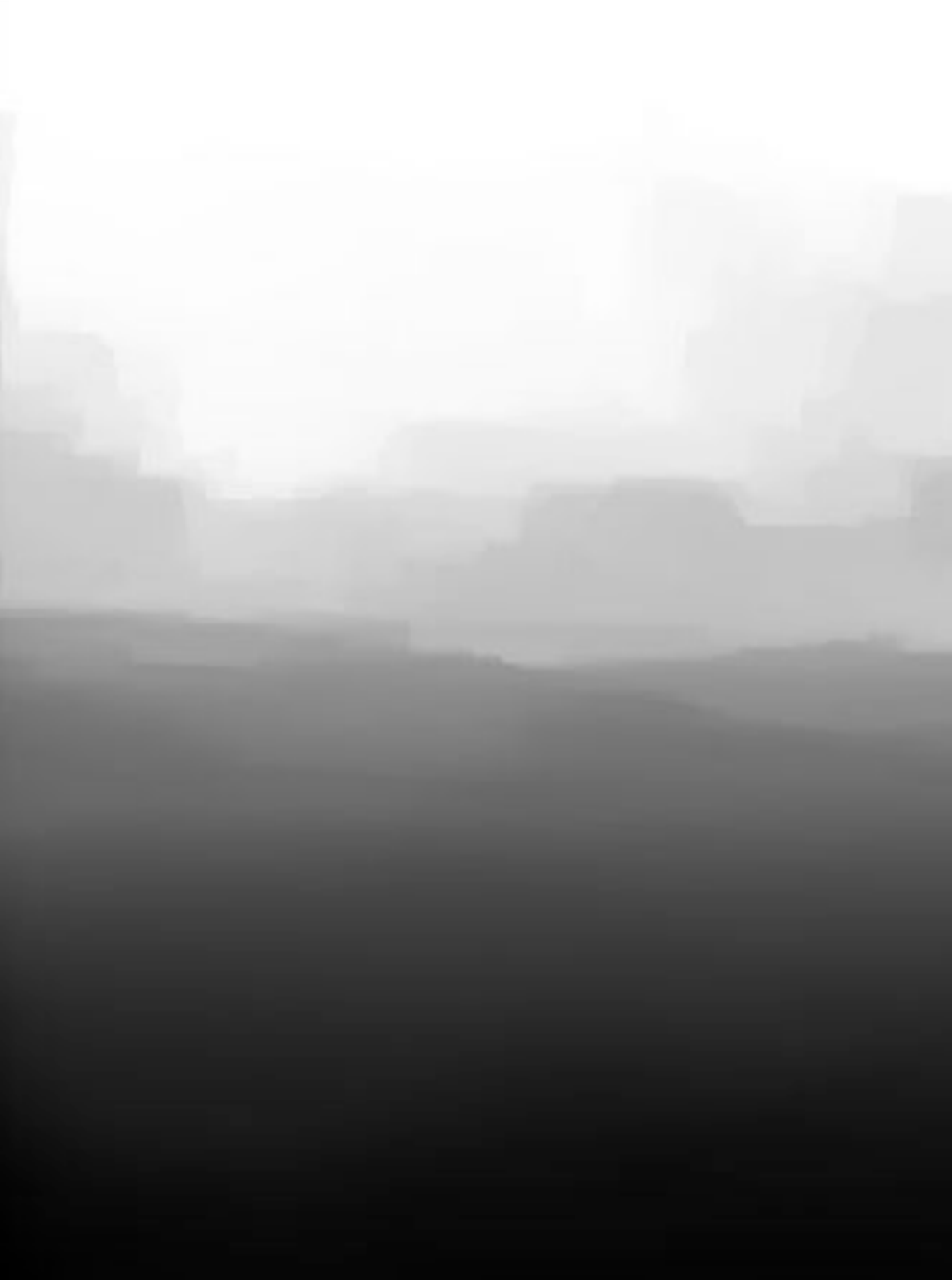}\vspace{0.5mm}\\
\includegraphics[width=\linewidth]{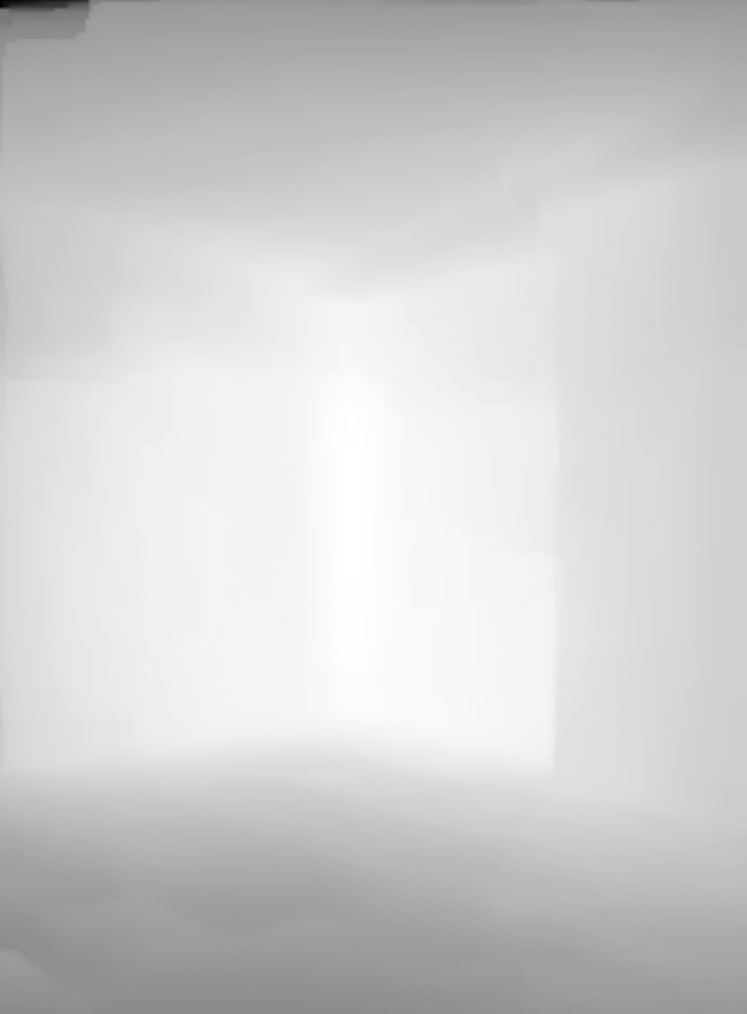}
\end{minipage}
\end{center}
\caption{Effect of using different training data for indoor and outdoor images. While the results are best if the proper dataset is used, we also get good results even if we combine all of the datasets.
}
\label{fig:train_example}
\end{figure}

Our results show that our depth transfer algorithm works for a large variety of indoor and outdoor sequences using a practical amount of training data. Note that our algorithm works for arbitrary videos, not just those with no parallax. However, videos with arbitrary camera motion and static scenes are best handled with techniques such as~\cite{Zhang:2011pami}. In Fig~\ref{fig:train_example}, we show that our algorithm requires some similar data in order to produce decent results (i.e., training with outdoor images for an indoor query is likely to fail). However, our algorithm can robustly handle large amounts of depth data with little degradation of output quality. The only issue is that more data requires more comparisons in the candidate search.

\begin{figure*}[t]

\begin{minipage}{.146\linewidth}
\centerline{\scriptsize Input}
\includegraphics[width=\linewidth]{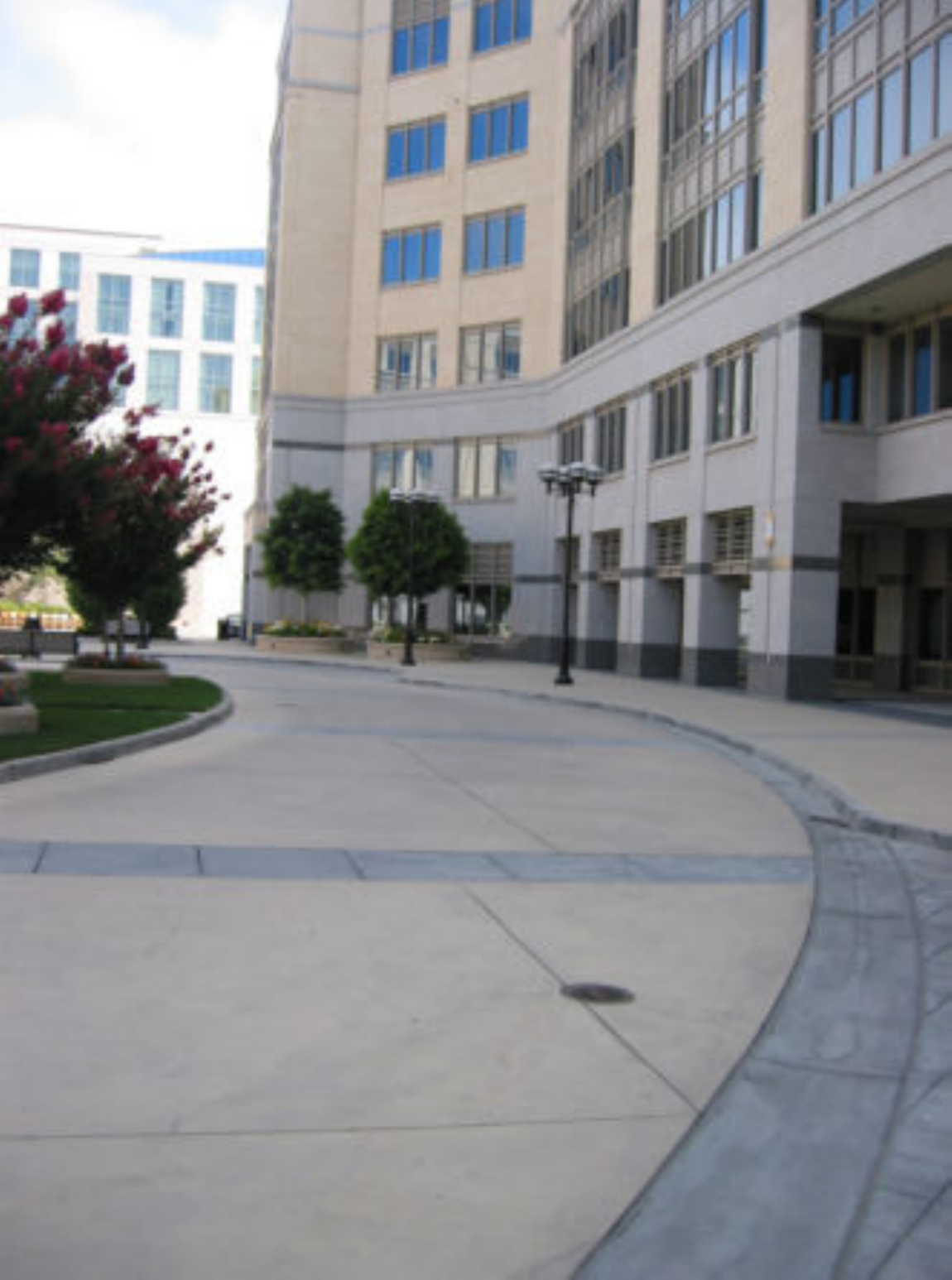}
\end{minipage}
\begin{minipage}{.535\linewidth}
\centerline{\scriptsize Candidate images and depths}
\includegraphics[width=.133\linewidth]{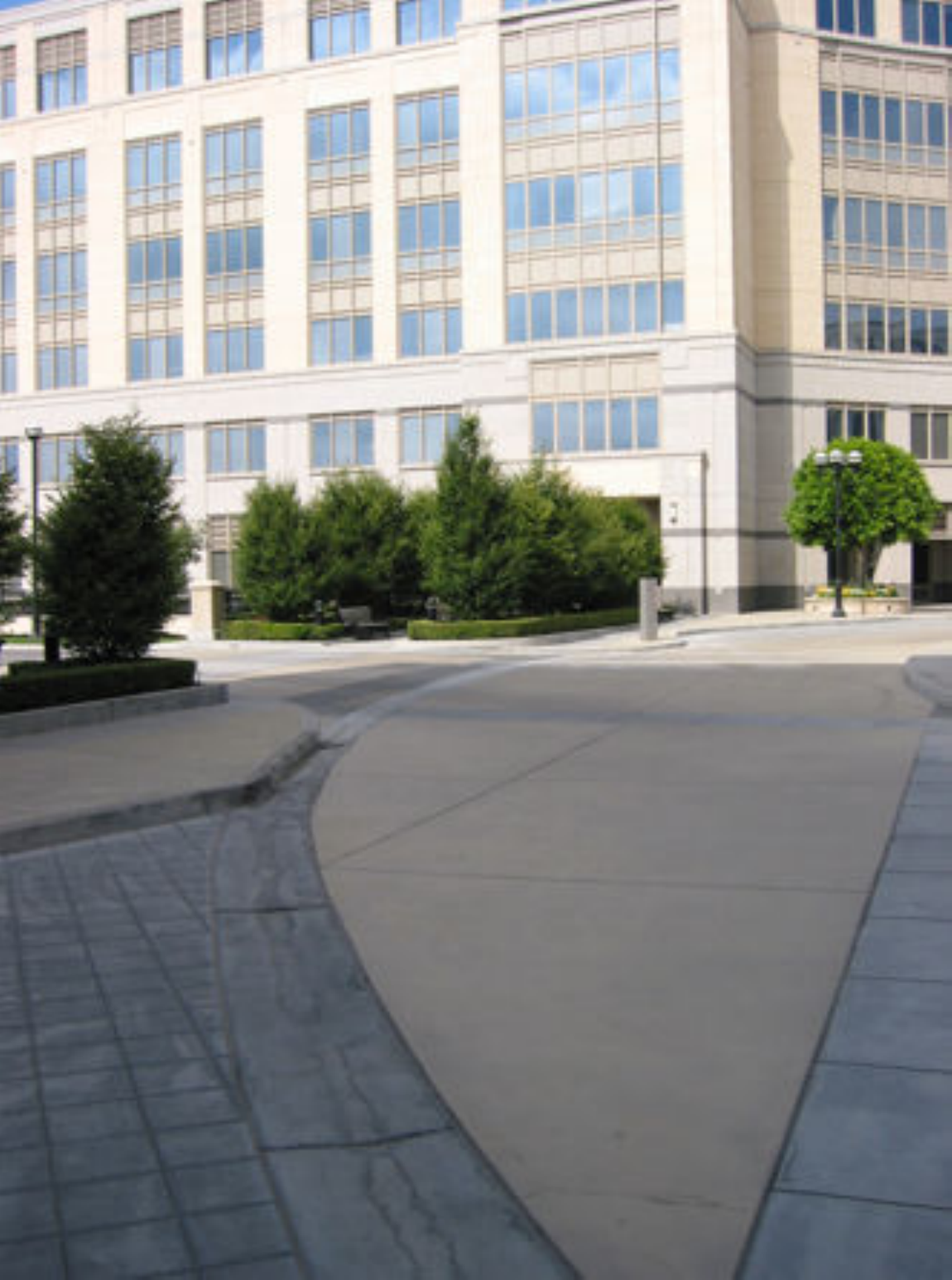}
\includegraphics[width=.133\linewidth]{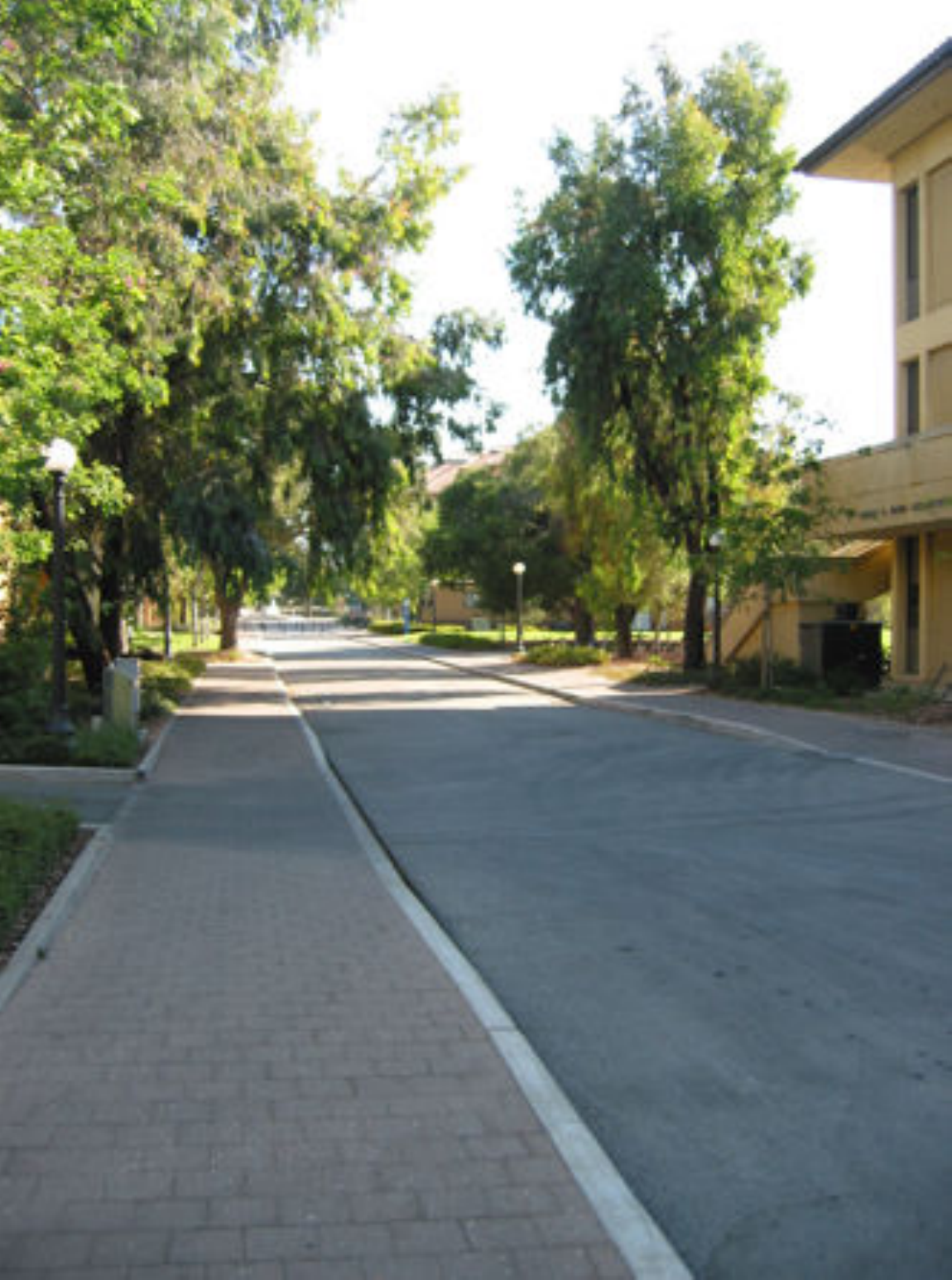}
\includegraphics[width=.133\linewidth]{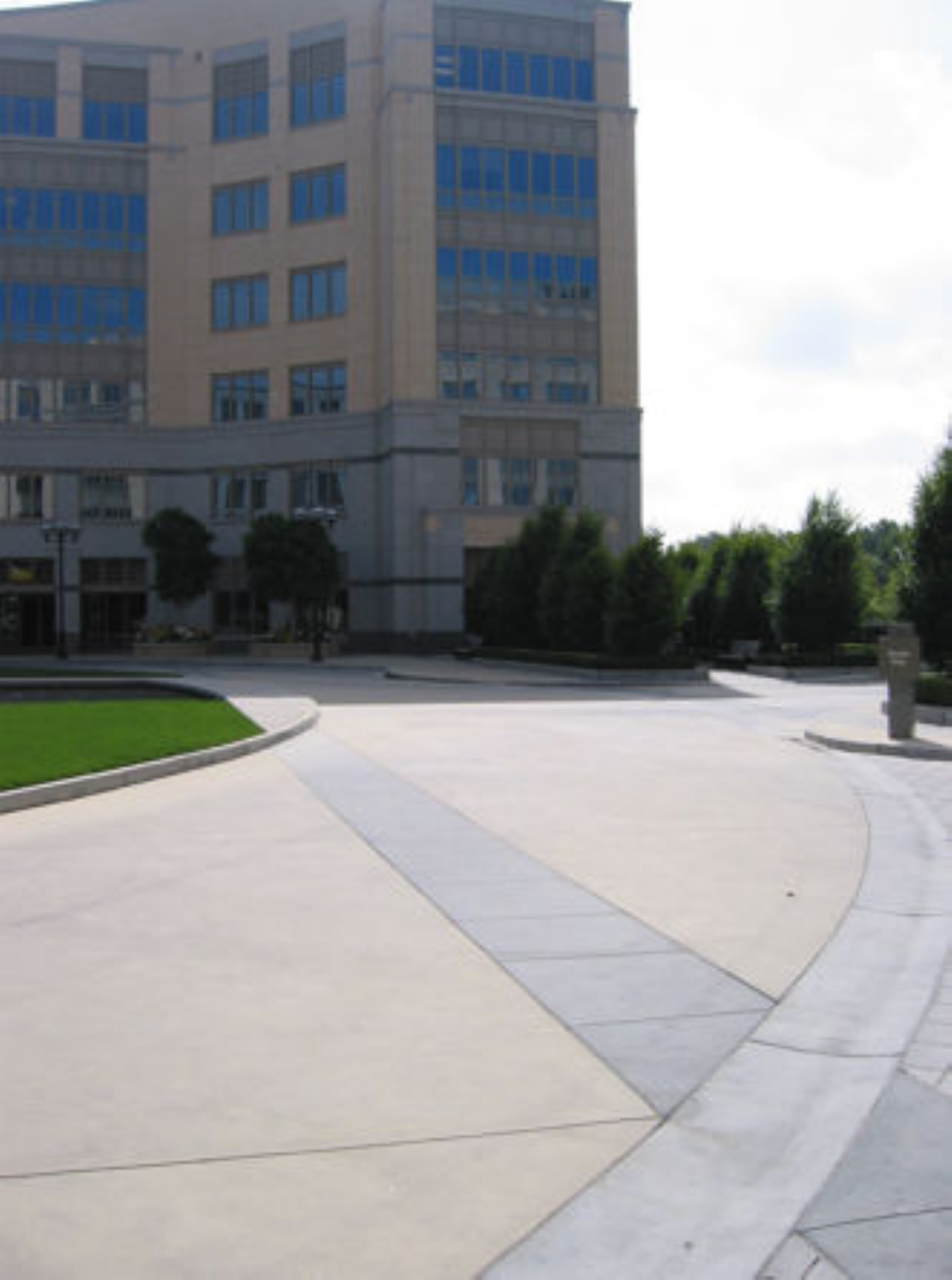}
\includegraphics[width=.133\linewidth]{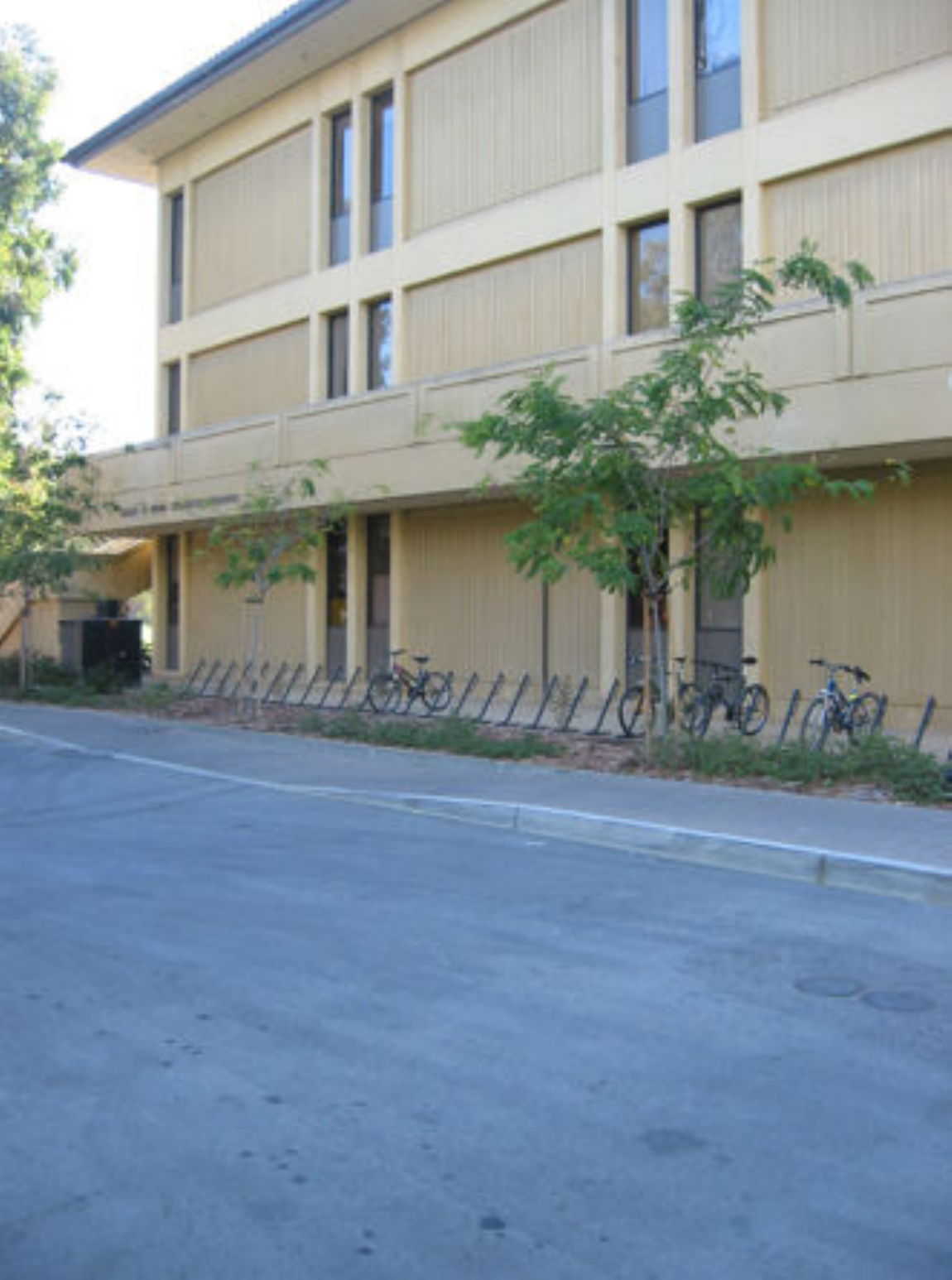}
\includegraphics[width=.133\linewidth]{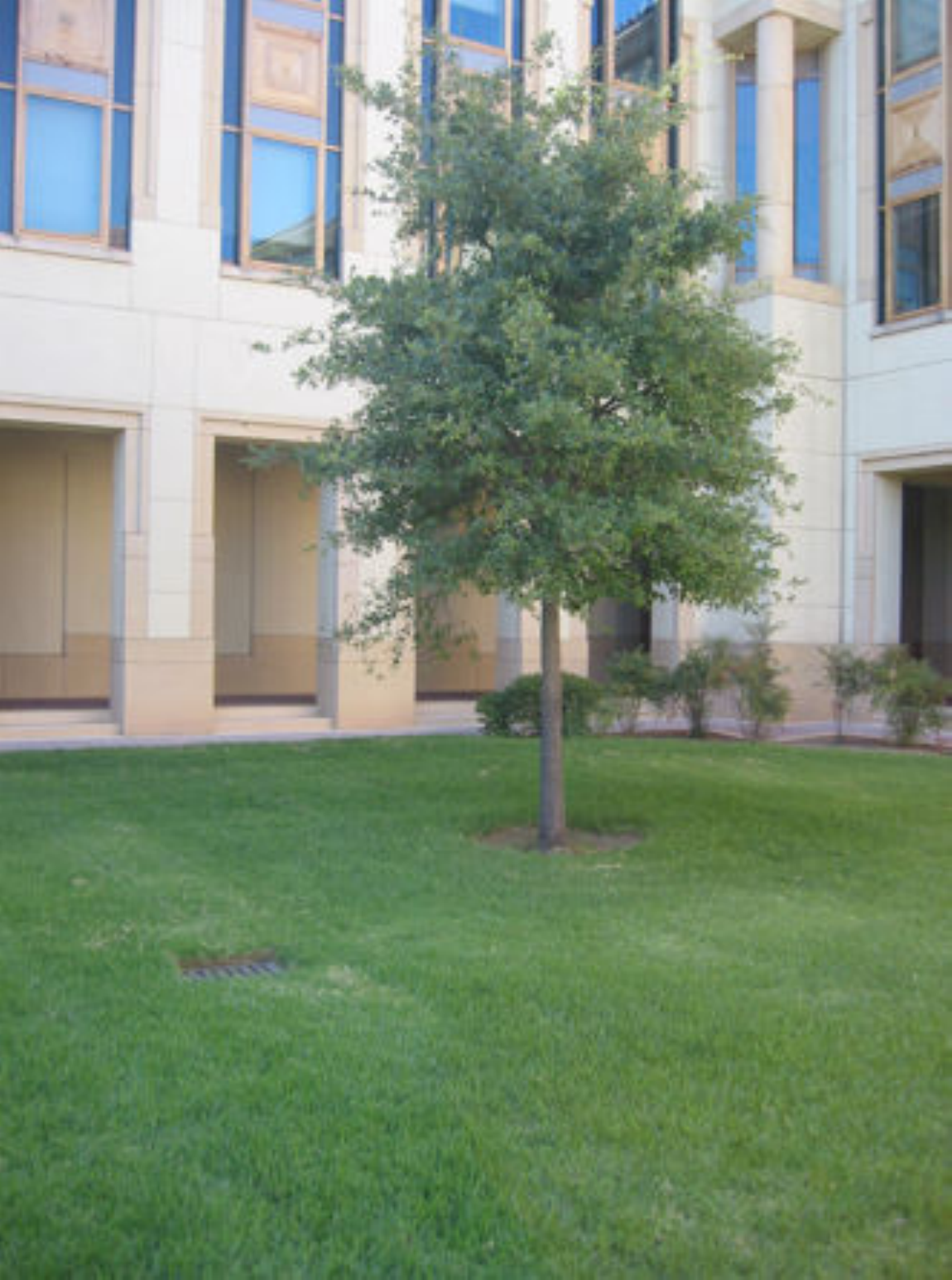}
\includegraphics[width=.133\linewidth]{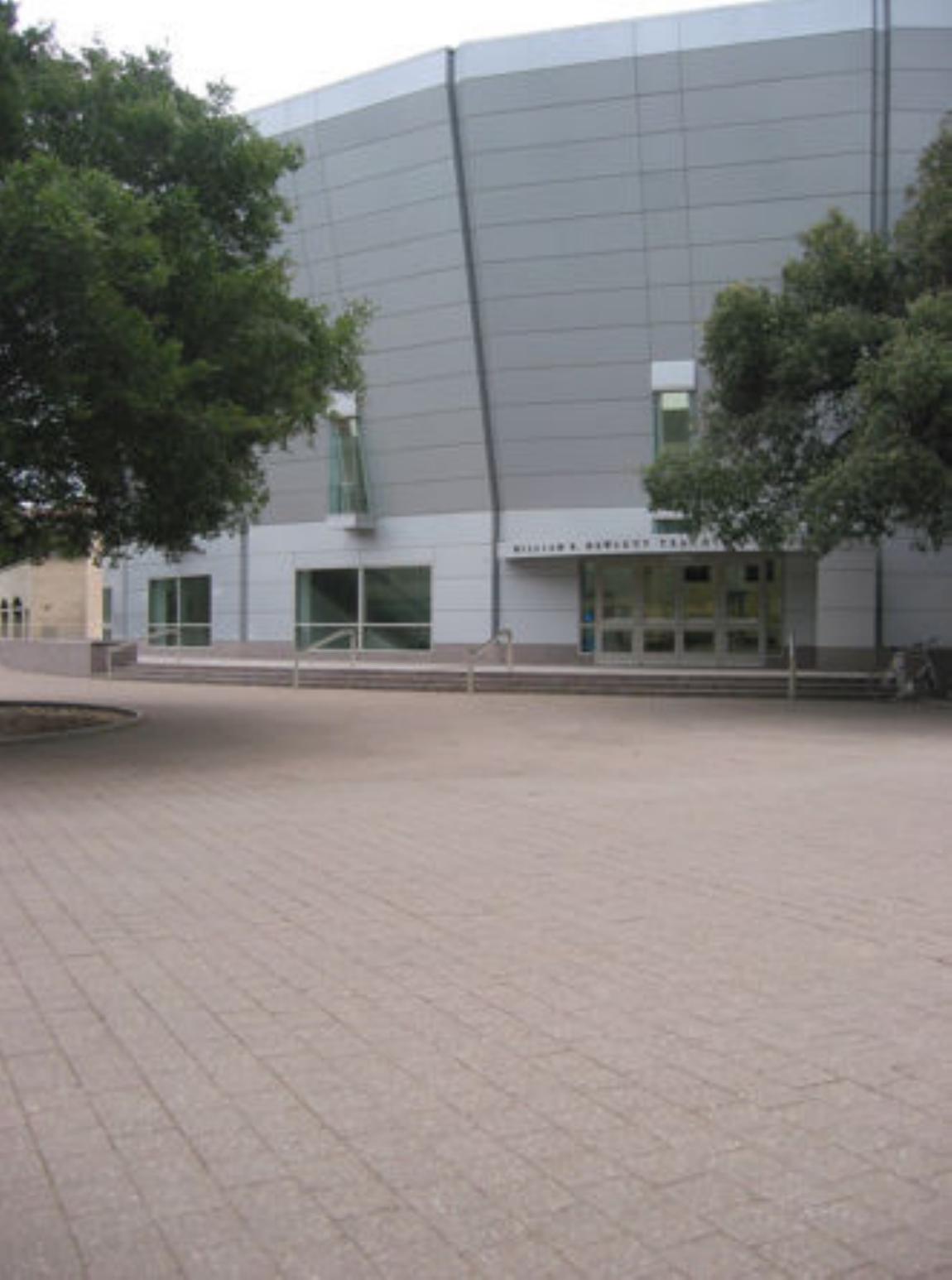}
\includegraphics[width=.133\linewidth]{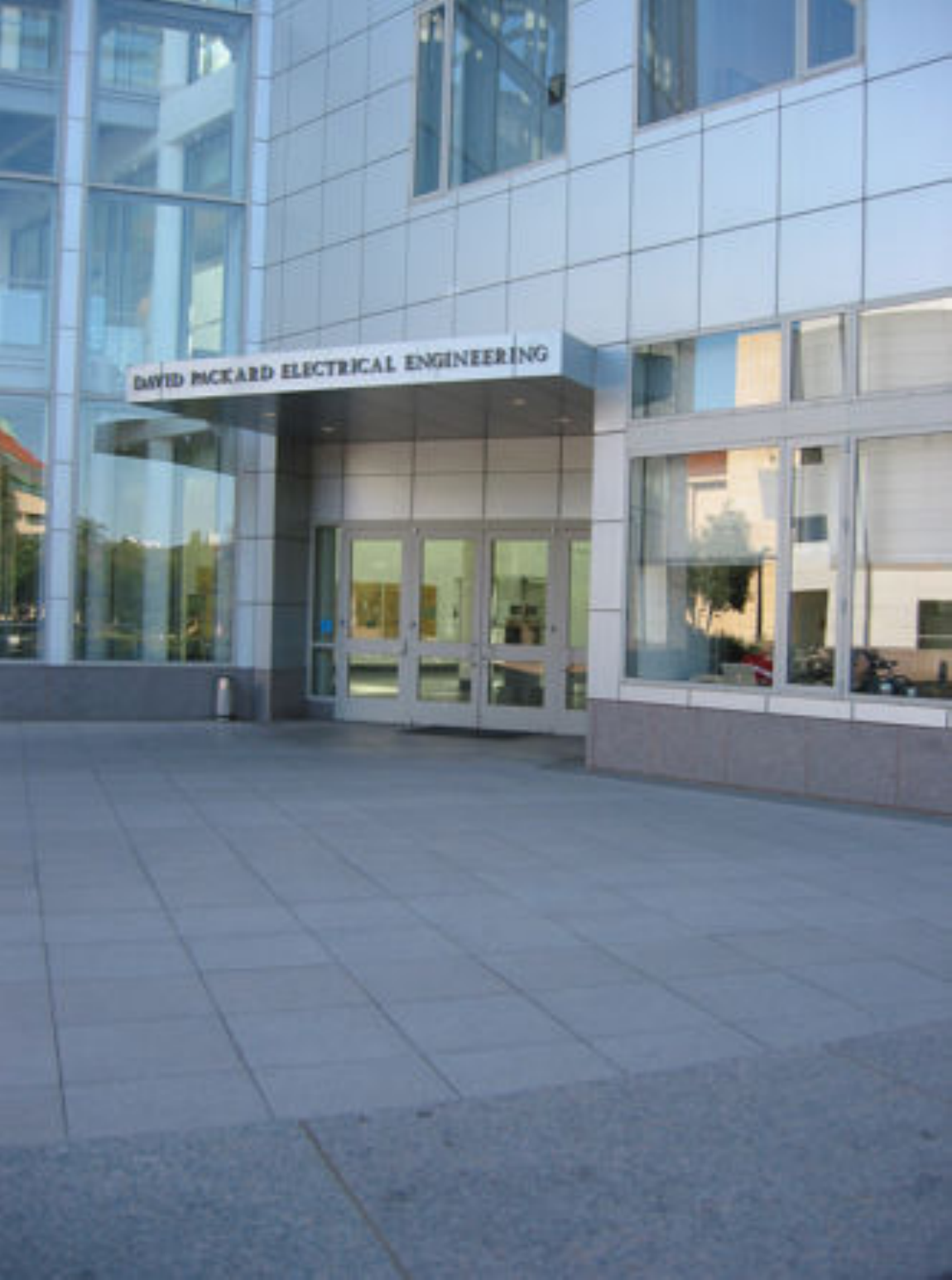}
\\
\includegraphics[width=.133\linewidth]{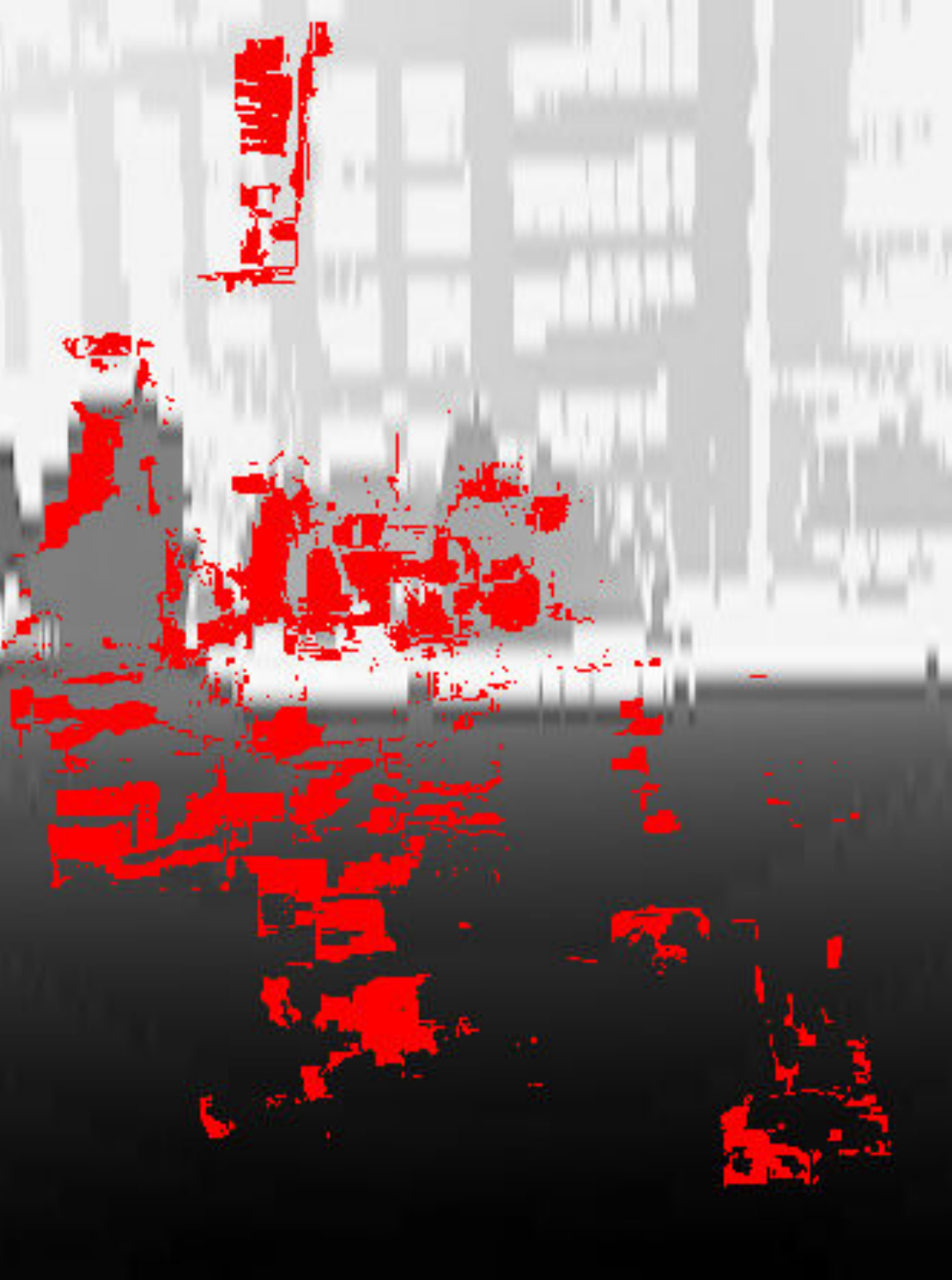}
\includegraphics[width=.133\linewidth]{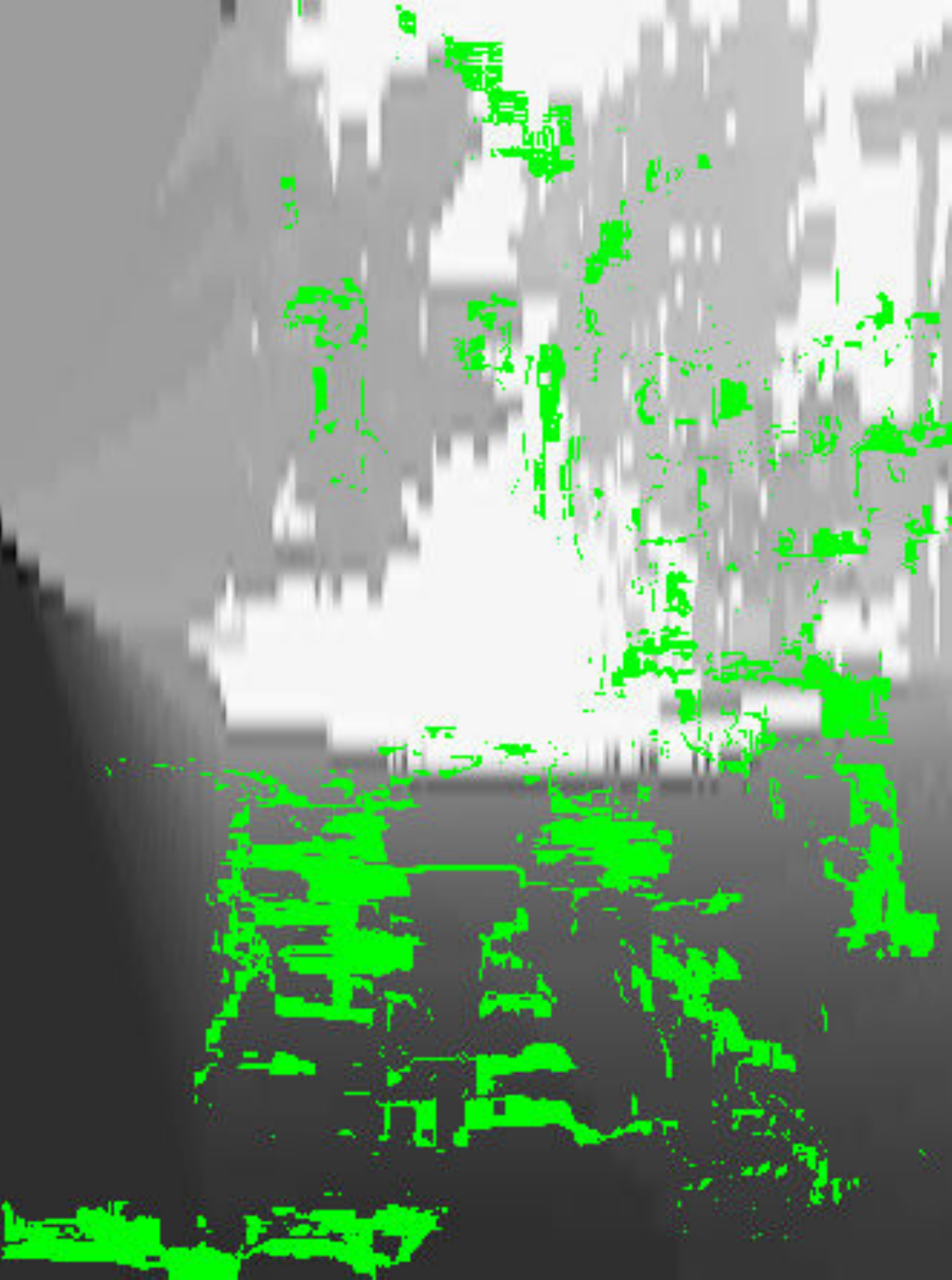}
\includegraphics[width=.133\linewidth]{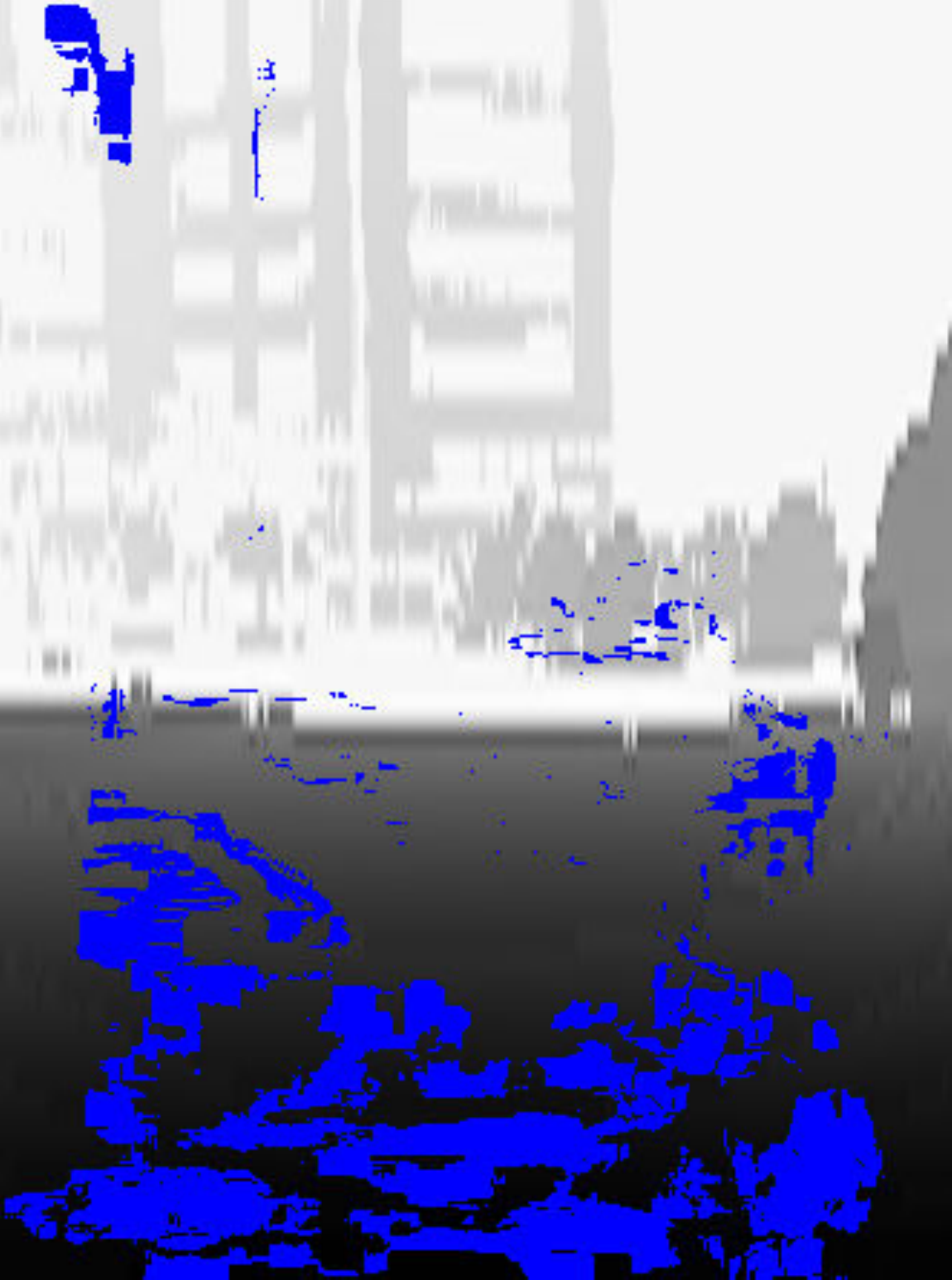}
\includegraphics[width=.133\linewidth]{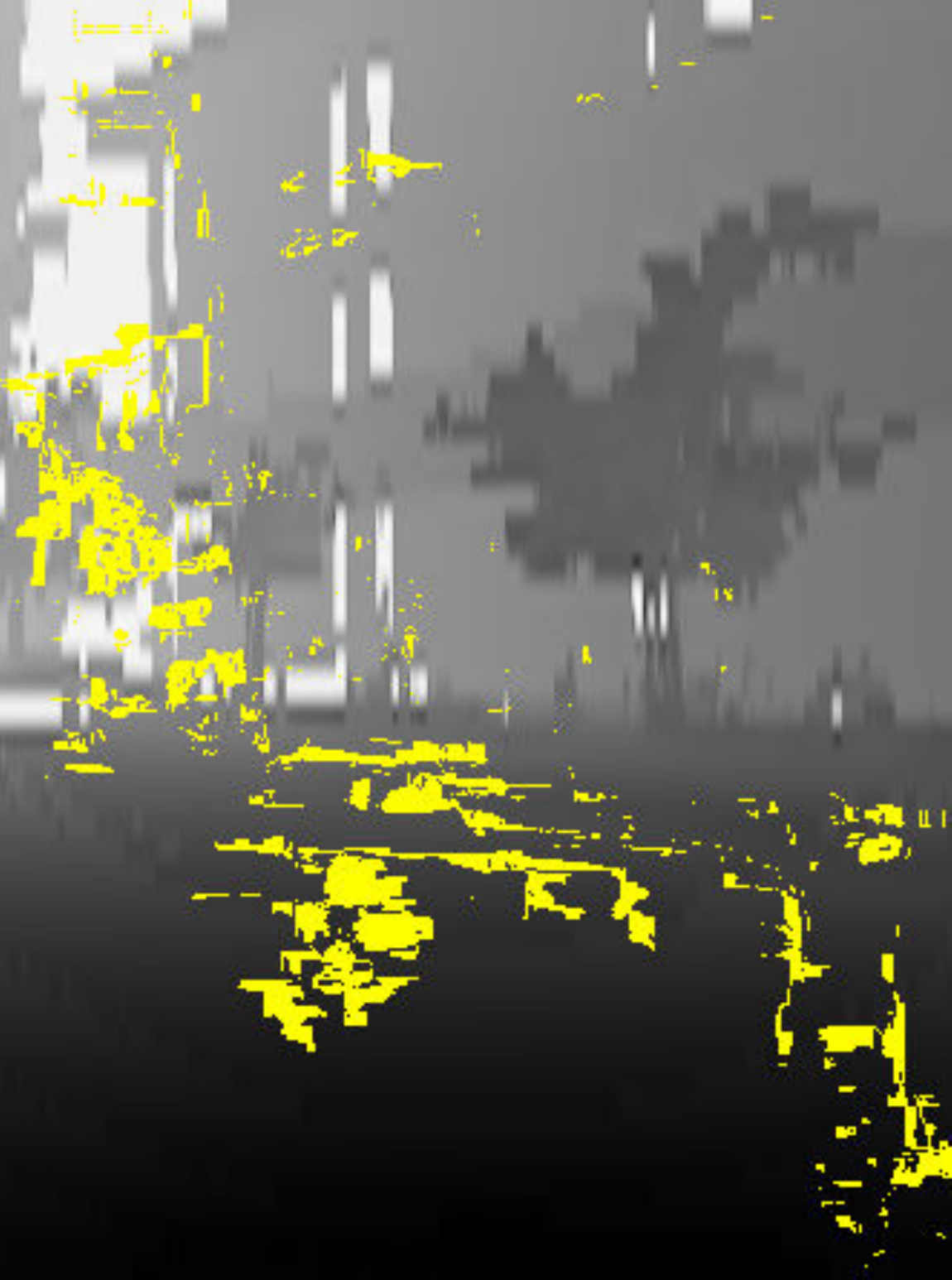}
\includegraphics[width=.133\linewidth]{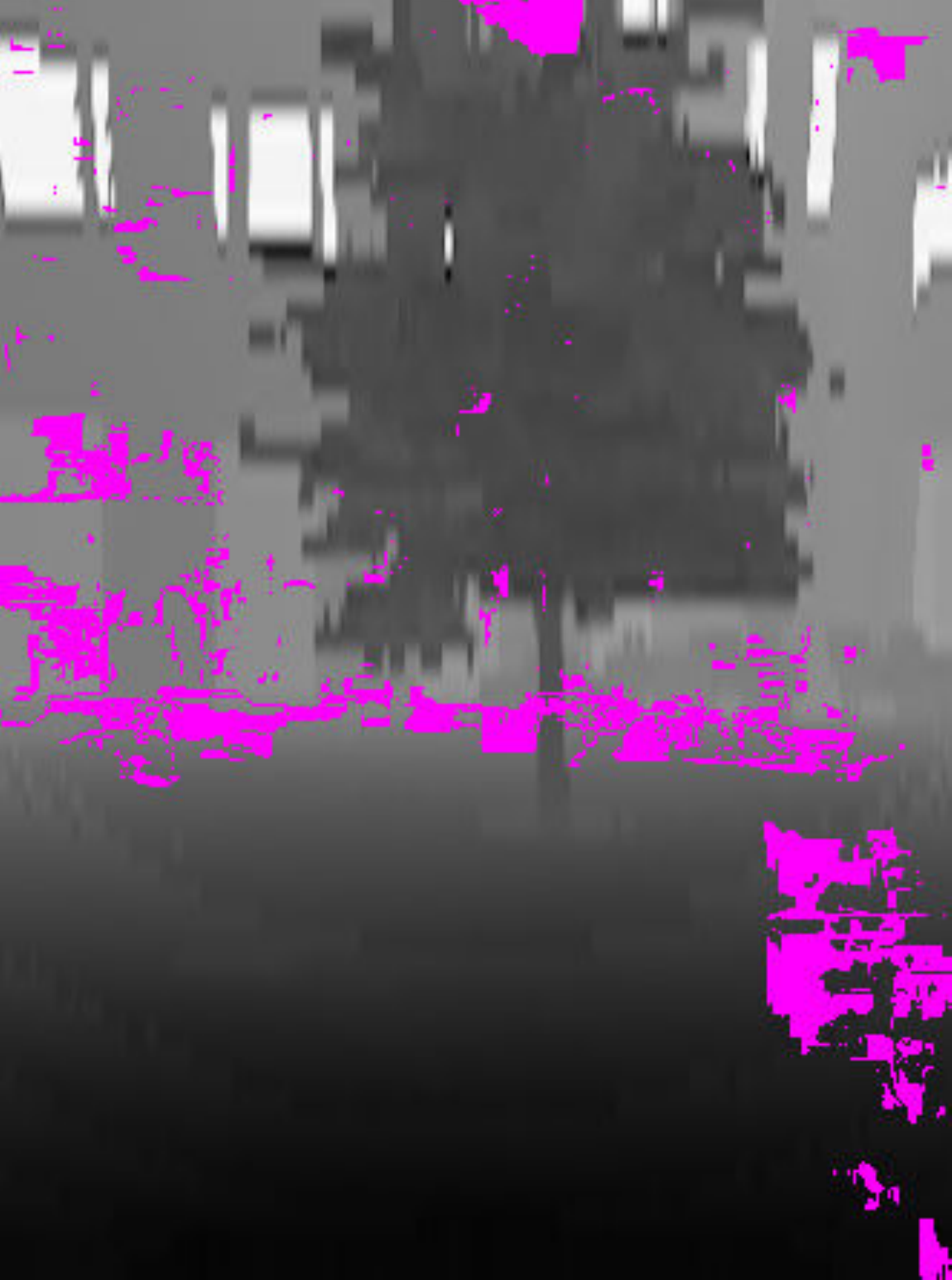}
\includegraphics[width=.133\linewidth]{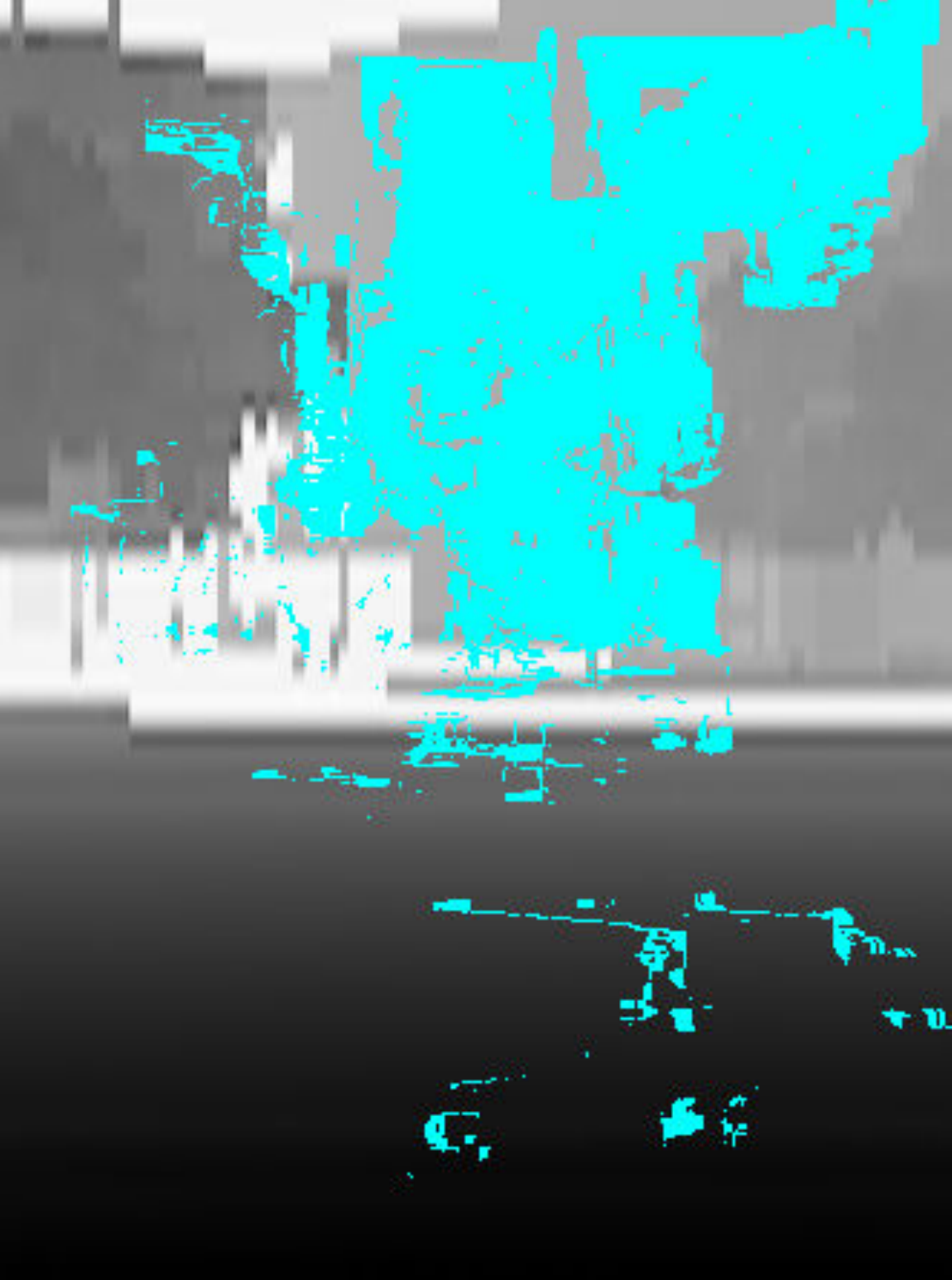}
\includegraphics[width=.133\linewidth]{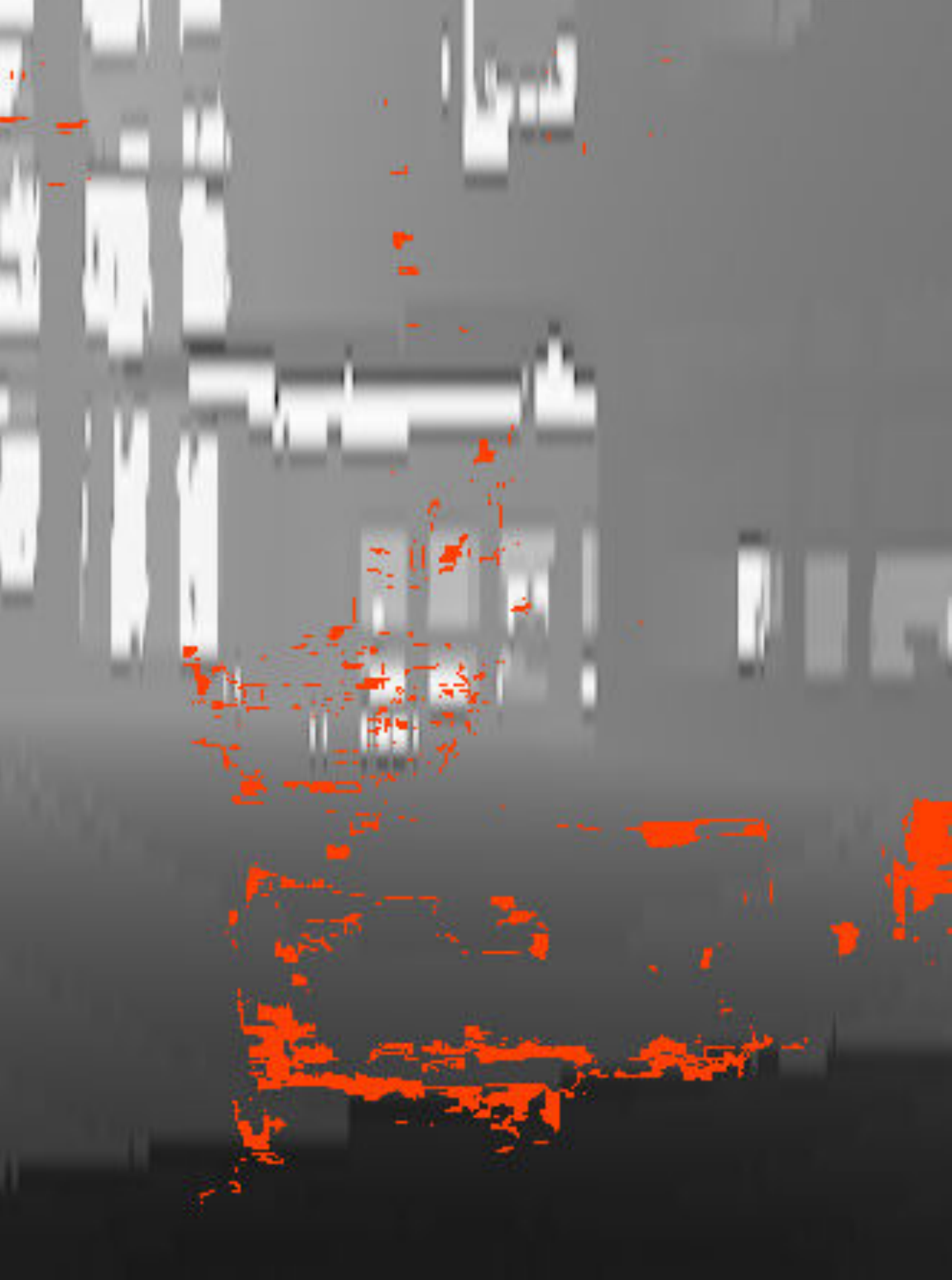}
\end{minipage}
\begin{minipage}{.146\linewidth}
\centerline{\scriptsize Contribution}
\vspace{.75mm}
\includegraphics[width=\linewidth]{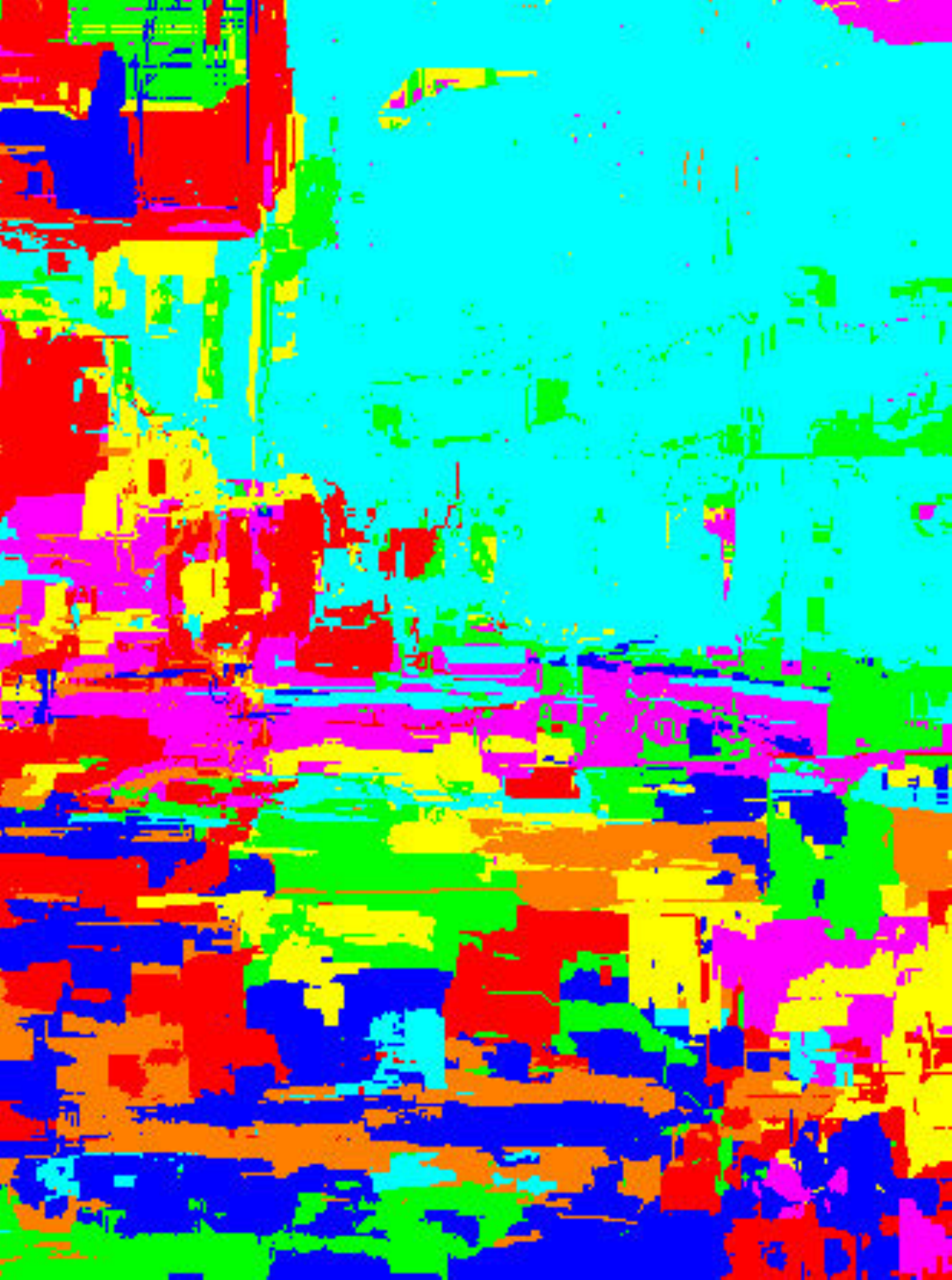}
\end{minipage}
\begin{minipage}{.146\linewidth}
\centerline{\scriptsize Inferred depth}
\includegraphics[width=\linewidth]{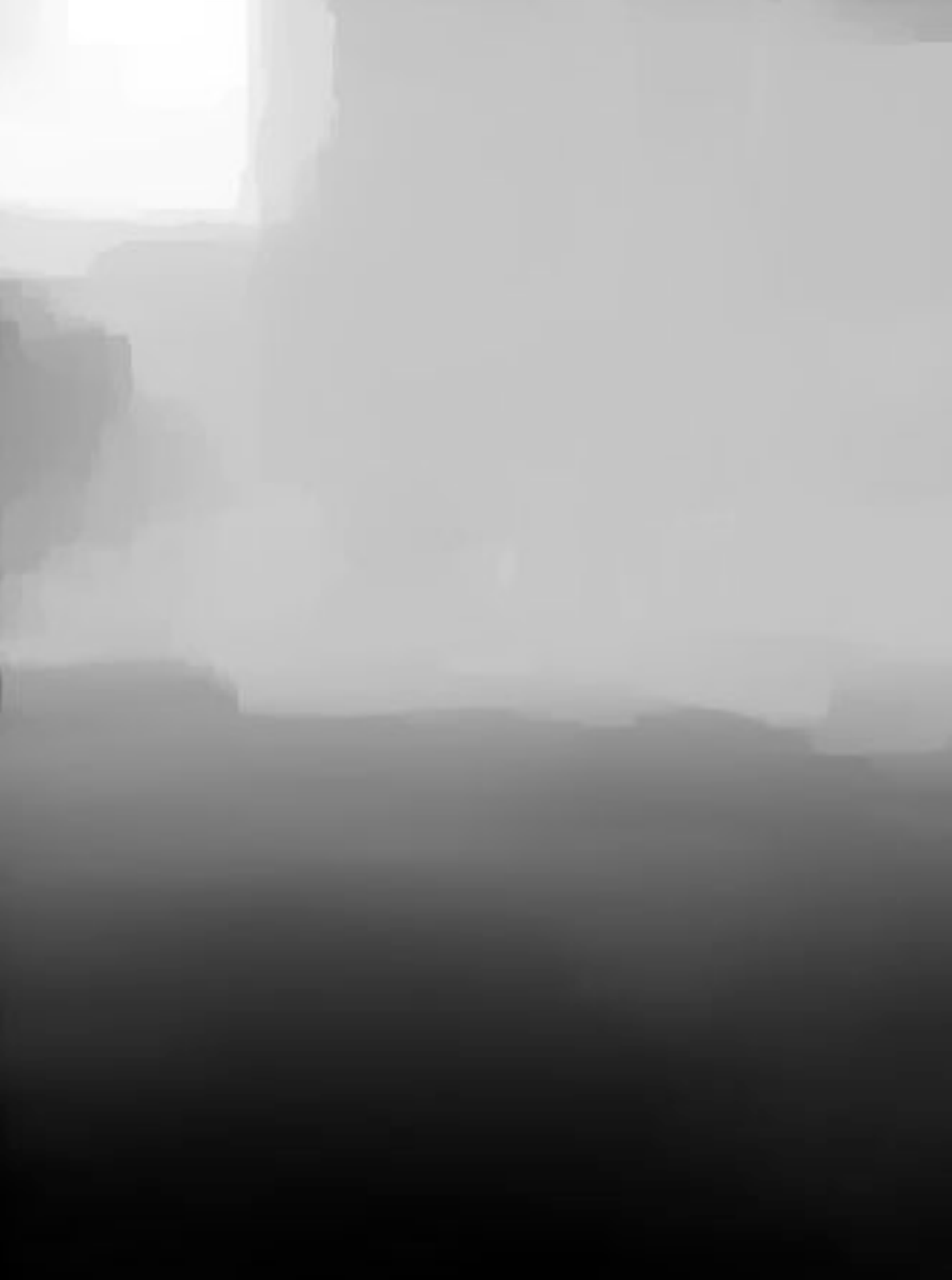}
\end{minipage}
%\vspace{-3.5mm}
\caption{Candidate contribution for depth estimation. For an input image (\emph{left}), we find top-matching candidate RGBD images (\emph{middle}), and infer depth (\emph{right}) for the input using our technique. The contribution image is color-coded to show the sources; red pixels indicate that the left-most candidate influenced the inferred depth image the most, orange indicates contribution from the right-most candidate, etc.
\vspace{-1.5mm}
}
\label{fig:candidate_contribution}\vspace{-.5em}
\end{figure*}

This robustness is likely due to the features we use when determining candidate images as well as the design of our objective function. In Fig~\ref{fig:candidate_contribution}, we show an example query image, the candidates retrieved by our algorithm, and their contribution to the inferred depth. By matching GIST features, we detect candidates that contain features consistent with the query image, such as building facades, sky, shrubbery, and similar horizon location. Notice that the depth of the building facade in the input comes mostly from another similarly oriented building facade (teal), and the ground plane and shrubbery depth come almost solely from other candidates' ground and tree depths.

%%%%%%%%%%%%%%%%%
%\begin{figure}[htp]
%\includegraphics[width=\columnwidth]{fig/viewsynthoverview-texpriors}
%\caption{For future work, we would like to bring together the two stages of our algorithm (depth estimation and view synthesis) through the use of texture priors. By incorporating this step, we could jointly optimize depth and view synthesis, allowing not only the depth to help in new view synthesis, but also allowing the synthesized views to correct depth estimates.}
%\label{fig:viewsynthoverview_texpriors}
%\end{figure}
%%%%%%%%%%%%%%%%%

%%%%%%%%%%%%%%%%%
%%%%%%%%%%%%%%%%%

%%%%%%%%%%%%%%%%%
%%%%%%%%%%%%%%%%%

\begin{figure}[t]
\includegraphics[width=.234\linewidth,clip=true,trim=0 0 50 0]{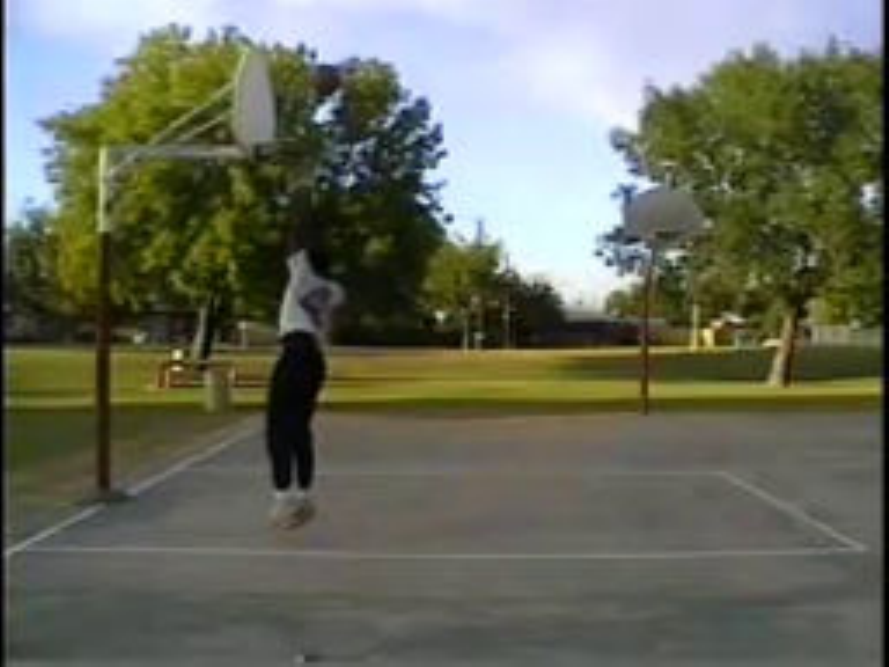}
\includegraphics[width=.234\linewidth,clip=true,trim=0 0 50 0]{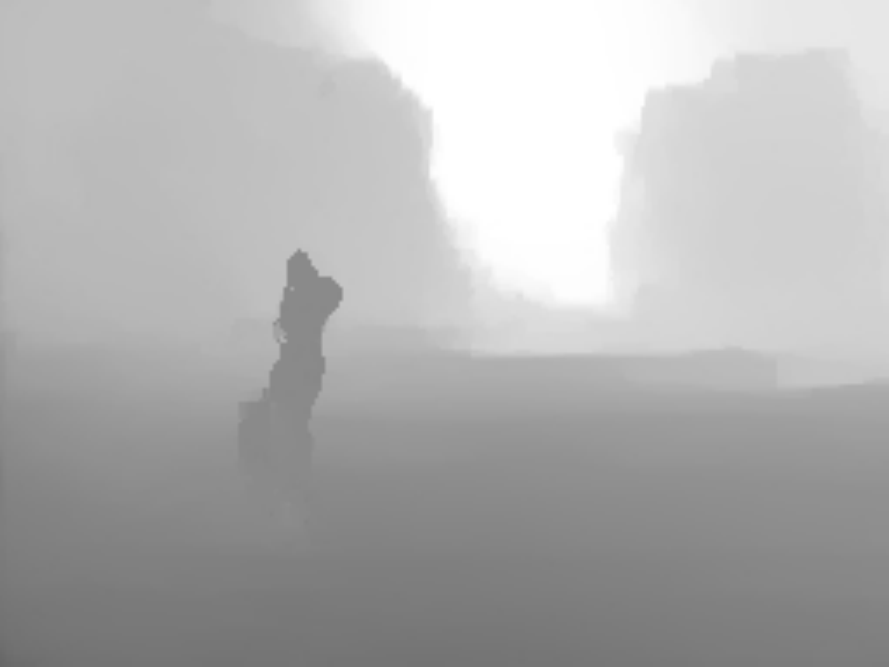} 
\hfill
\includegraphics[width=.234\linewidth,clip=true,trim=0 0 50 0]{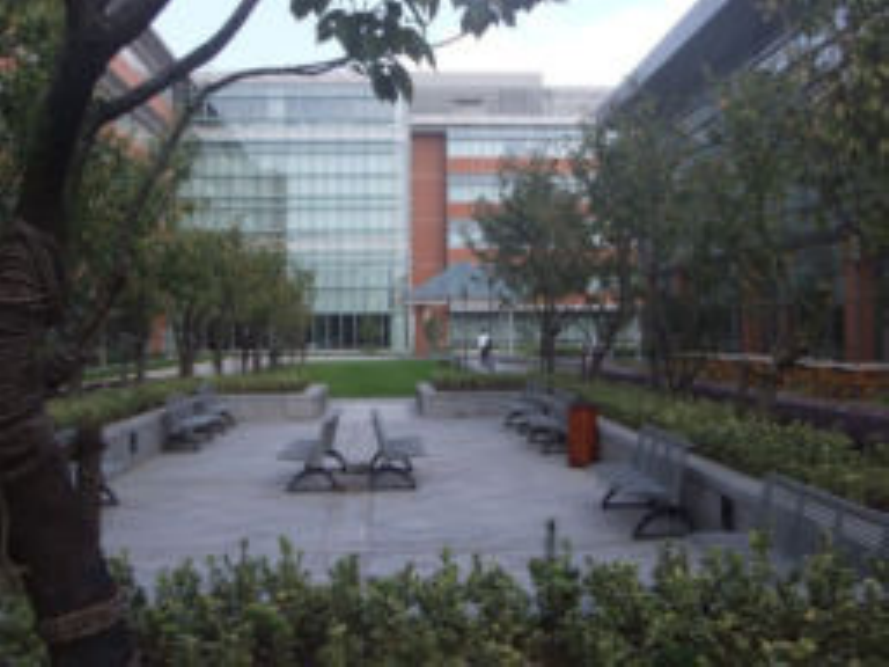}
\includegraphics[width=.234\linewidth,clip=true,trim=0 0 50 0]{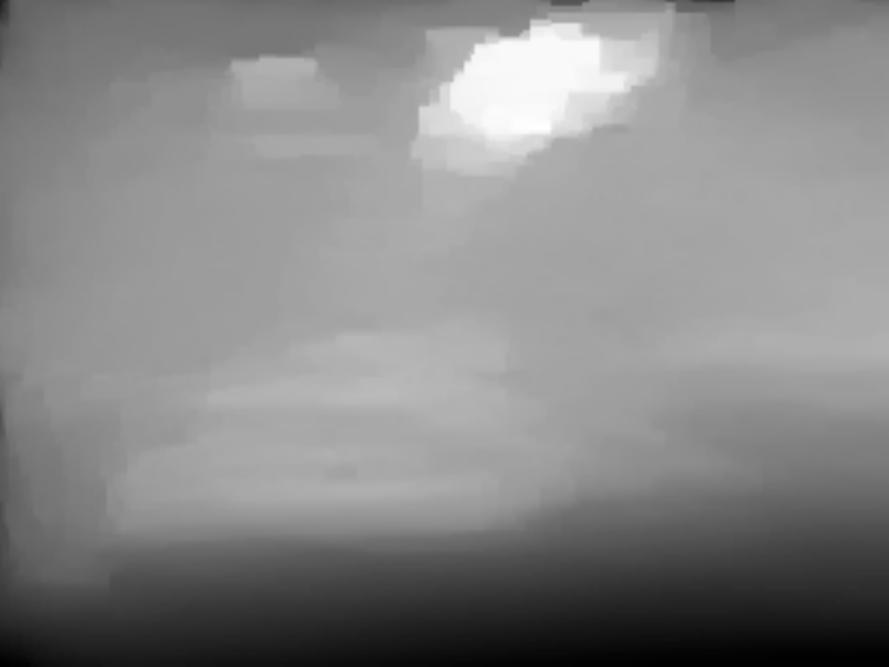}
\caption{Example failure cases. Left: thin or floating objects (pole and basketball) are ignored. Right: input image is too different from training set.
%\vspace{-2mm}
}
\label{fig:basketball}
\end{figure}

In some cases, our motion segmentation misses or falsely identifies moving pixels. This can result in inaccurate depth and 3D estimation, although our spatio-temporal regularization (Eqs.~\ref{eq:smooth},~\ref{eq:coherence}) helps to overcome this. Our algorithm also assumes that moving objects contact the ground, and thus may fail for airborne objects (see Fig~\ref{fig:basketball}).

Due to the serial nature of our method (depth estimation followed by view synthesis), our method is prone to propagating errors through the stages. For example, if an error is made during depth estimation, the result may be visually implausible. It would be ideal to use knowledge of how the synthesized views should look in order to correct issues in depth.

%%%%%%%%%%%%%%%%%%%%%%%%%%%%%%%%%%
\section{Concluding Remarks}
We have demonstrated a fully automatic technique to estimate depths for videos. Our method is applicable in cases where other methods fail, such as those based on motion parallax and structure from motion, and works even for single images and dynamics scenes. Our depth estimation technique is novel in that we use a non-parametric approach, which gives qualitatively good results, and our single-image algorithm quantitatively outperforms existing methods. Using our technique, we also show how we can generate stereoscopic videos for 3D viewing from conventional 2D videos. Specifically, we show how to generate time-coherent, visually pleasing stereo sequences using our inferred depth maps. Our method is suitable as a good starting point for converting legacy 2D feature films into 3D.

%%%%%%%%%%%%%%%%%
%%%%%%%%%%%%%%%%%

%%%%%%%%%%%%%%%%%%%%%%%%%%%%%%%%%%
\section*{Acknowledgement}
We would like to thank Tom Blank for his critical help in creating our dual-Kinect data collection system.
%This research was made in part with Government support awarded by DoD, Air Force Office of Scientific Research, National Science and Engineering Graduate (NDSEG) Fellowship, 32 CFR 168a.

%%%%%%%%%%%%%%%%%%%%%%%%%%%%%%%%%%%%%%
{%\small
\bibliographystyle{IEEEtran}
\bibliography{depthtransfer}
}

%\begin{biography}[{\includegraphics[width=1in,height=1.25in,clip,keepaspectratio]{mshell}}]{Michael Shell}

\begin{IEEEbiography}[{\includegraphics[width=1in,height=1.25in,clip,keepaspectratio]{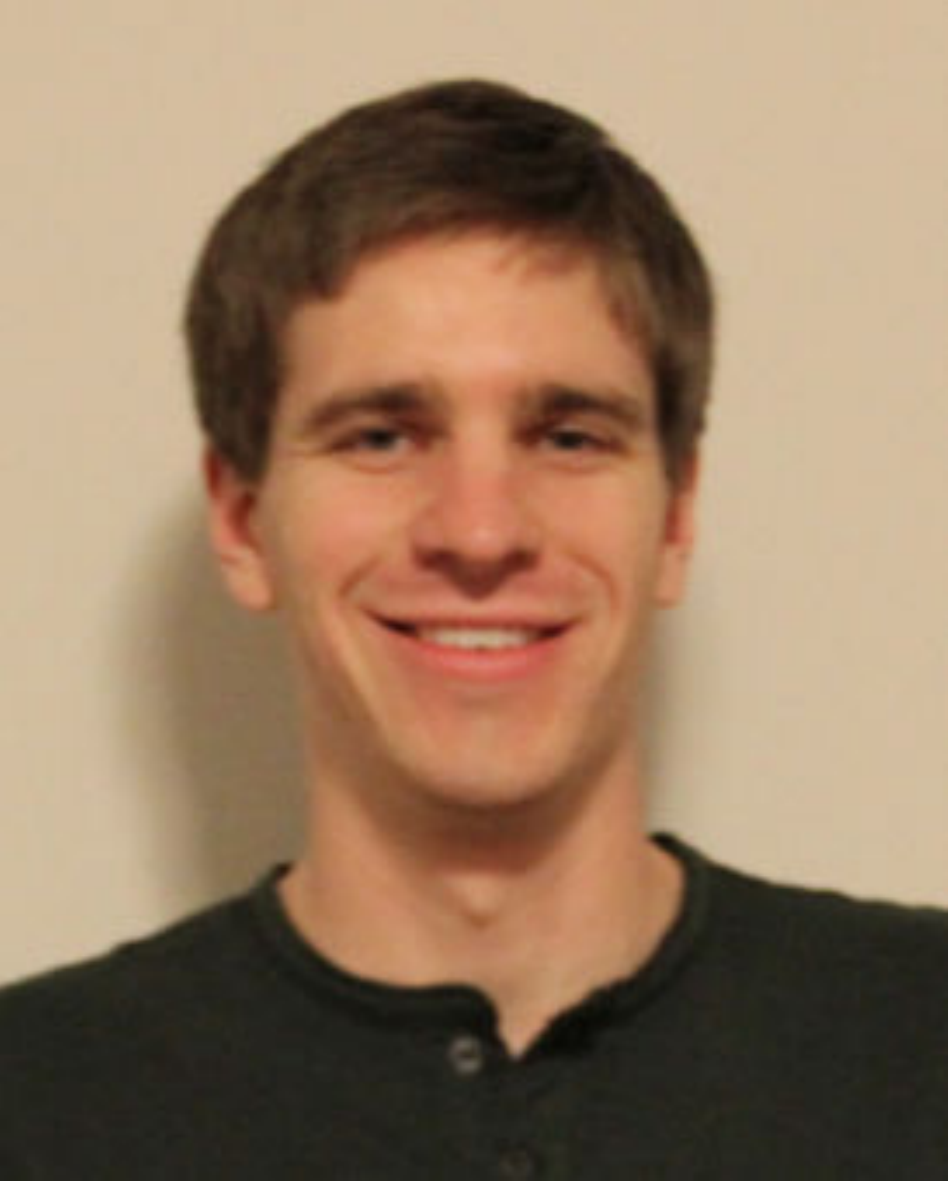}}]{Kevin Karsch} received a BS degree in computer science and mathematics from the University of Missouri in 2009, and is currently working towards his PhD in computer science at the University of Illinois. He holds the NSFGRF and NDSEG fellowship, and was recently awarded the Lemelson-MIT student prize for innovation. His research interests include computer vision, graphics, and computational photography.
\end{IEEEbiography}

\begin{IEEEbiography}[{\includegraphics[width=1in,height=1.25in,clip,keepaspectratio]{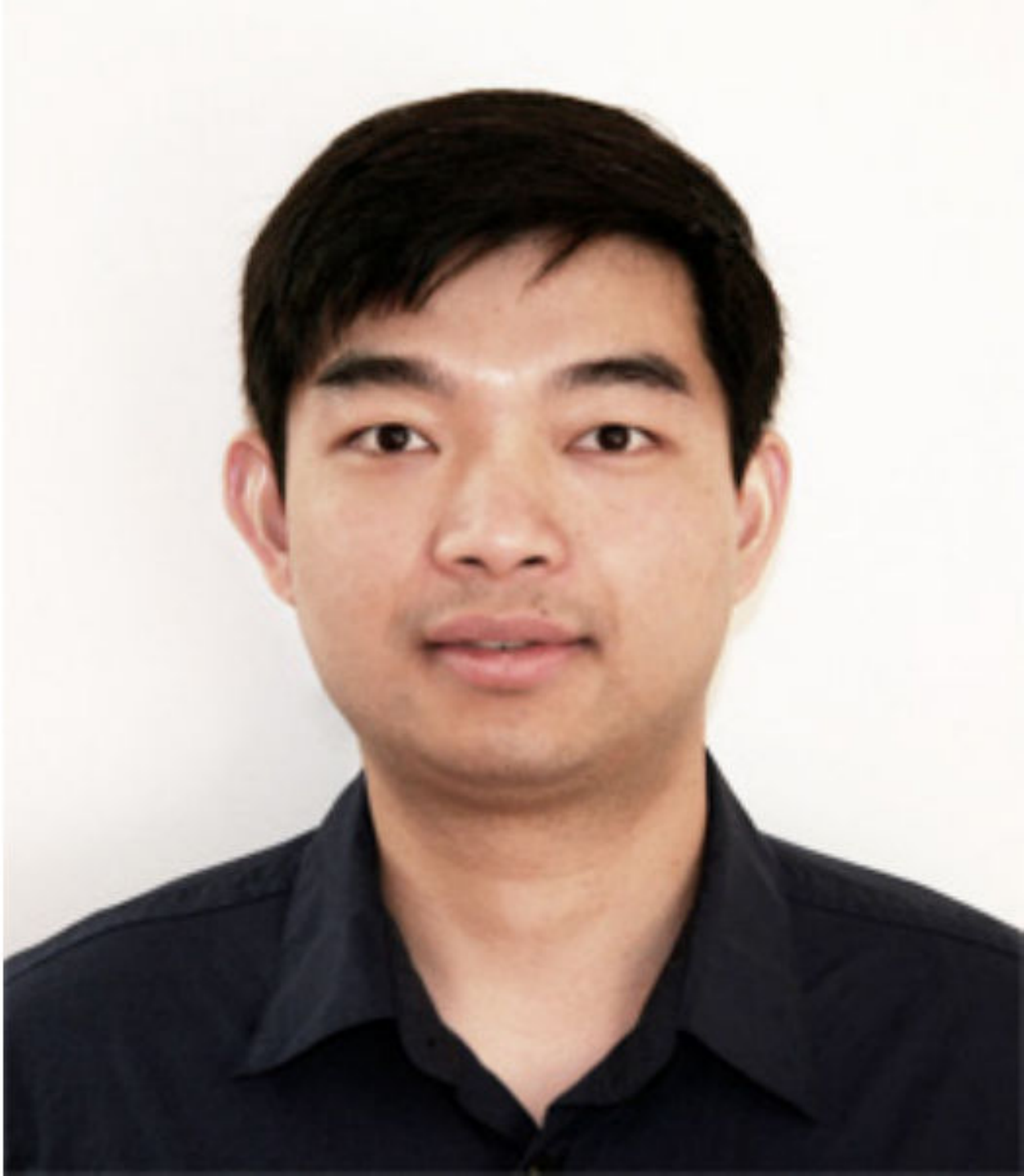}}]{Ce Liu} received the BS degree in automation and the ME degree in pattern recognition from the Department of Automation, Tsinghua University in 1999 and 2002, respectively. After receiving the PhD degree from the Massachusetts Institute of Technology in 2009, he now holds a researcher position at Microsoft Research New England. From 2002 to 2003, he worked at Microsoft Research Asia as an assistant researcher. His research interests include computer vision, computer graphics, and machine learning. He has published more than 20 papers in the top conferences and journals in these fields. He received a Microsoft Fellowship in 2005, the Outstanding Student Paper award at the Advances in Neural Information Processing Systems (NIPS) in 2006, a Xerox Fellowship in 2007, and the Best Student Paper award at the IEEE Conference on Computer Vision and Pattern Recognition (CVPR) 2009. He is a member of the IEEE.
\end{IEEEbiography}

%\newpage

\begin{IEEEbiography}[{\includegraphics[width=1in,height=1.25in,clip,keepaspectratio]{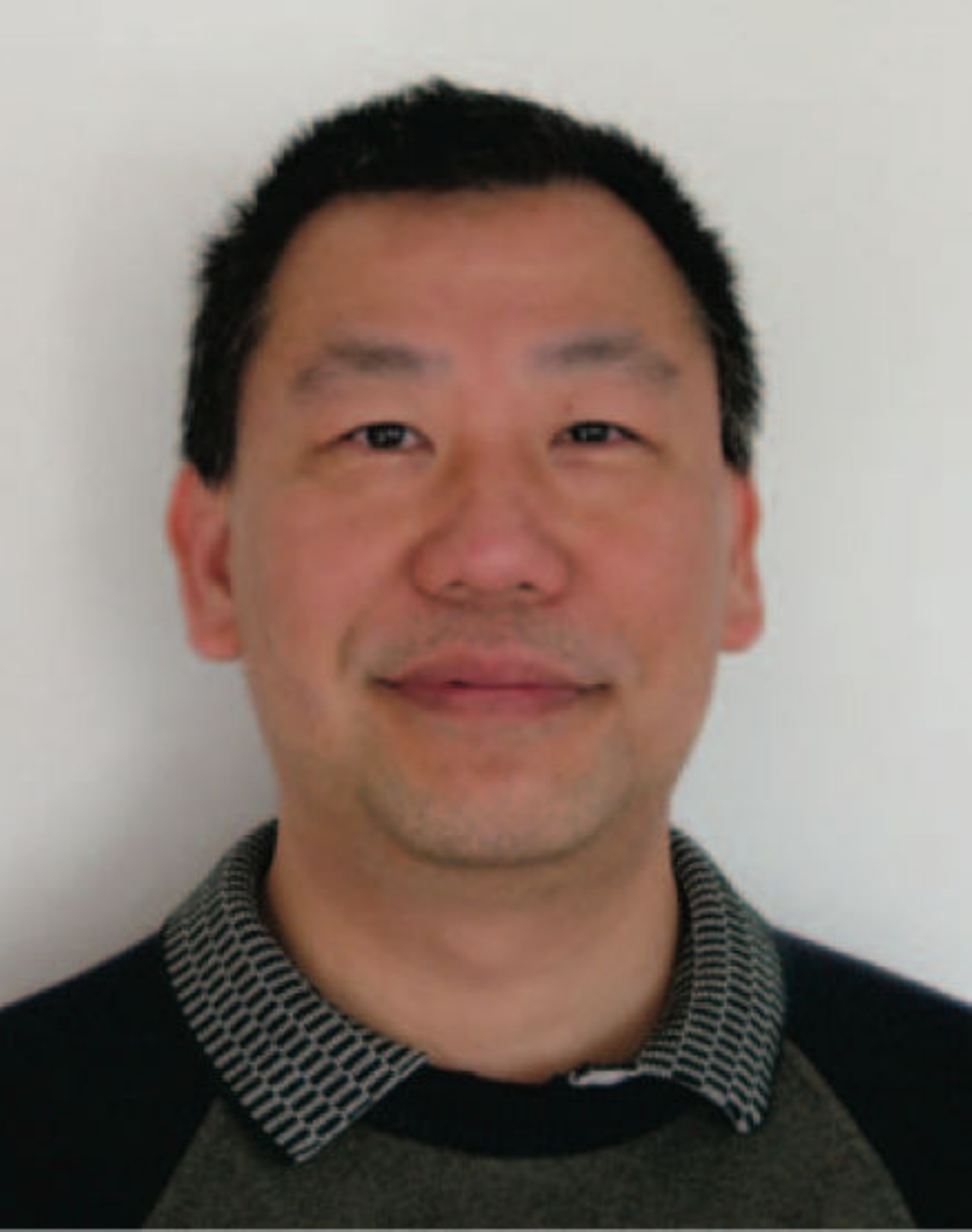}}]{Sing Bing Kang} is a principal researcher at Microsoft Research. He received his Ph.D. in robotics from Carnegie Mellon University, Pittsburgh in 1994. His areas of interest are computer vision and computer graphics, more specifically image-based modeling as well as image and video enhancement. Sing Bing has co-edited two books (``Panoramic Vision'' and ``Emerging Topics in Computer Vision'') and co-authored two books (``Image-Based Rendering'' and ``Image-Based Modeling of Plants and Trees''). He has served as area chair and member of technical committee for the major computer vision conferences (ICCV, CVPR, ECCV). In addition, he has served as papers committee member for SIGGRAPH and SIGGRAPH Asia. Sing Bing was program chair for ACCV 2007 and CVPR 2009, and is currently Associate Editor-in-Chief for IEEE Transactions on Pattern Recognition and Machine Intelligence. He is an IEEE Fellow.
\end{IEEEbiography}


\vfill


\end{document}